%% file: main.tex
\documentclass[10pt,twocolumn,letterpaper]{article}

\usepackage[pagenumbers]{cvpr} % To force page numbers, e.g. for an arXiv version

\input{preamble}

\definecolor{cvprblue}{rgb}{0.21,0.49,0.74}
\usepackage[pagebackref,breaklinks,colorlinks,allcolors=cvprblue]{hyperref}

\def\paperID{25324} % *** Enter the Paper ID here
\def\confName{CVPR}
\def\confYear{2026}

\title{Mining the Shadows: Distilling Dark Ranking Knowledge for OOD Detection}

\author{
    Dishanika Denipitiyage$^{*1}$ \quad 
    Aditya Krishna Menon$^{*2}$ \quad
    Nishavi Ranaweera$^3$ \quad
    Shaohua Zhong$^1$ \quad\\
    Suranga Seneviratne$^1$ \quad
    Sanjay Chawla$^4$ \\
    $^1$ The University of Sydney \quad 
    {\tt\small \{dishanika.denipitiyage, szho0654, suranga.seneviratne\}@sydney.edu.au}\\
    $^2$ Google \quad {\tt\small adityakmenon@google.com}\\
    $^3$ University of New South Wales \quad  {\tt\small r.tharushika\_nishavi@unsw.edu.au}\\
    $^4$ Qatar Computing Research Institute, Hamad Bin Khalifa University \quad 
{\tt\small schawla@hbku.edu.qa}
}

\begin{document}
\maketitle
\def\thefootnote{*}\footnotetext{Equal contribution.}\def\thefootnote{\arabic{footnote}}

\input{sec/0_abstract}    
\input{sec/1_intro}
\input{sec/2_background}
\input{sec/3_rankood_as_kd}
\input{sec/4_method}
\input{sec/5_theory}
\input{sec/6_experiments}

\input{sec/7_conclusion}
{
    \small
    \bibliographystyle{ieeenat_fullname}
    \bibliography{main}
}

% WARNING: do not forget to delete the supplementary pages from your submission 
\input{sec/X_suppl}

\end{document}

%% file: preamble.tex
\usepackage[T1]{fontenc}
\usepackage{amsfonts}
\usepackage{amsthm}              % proposition, assumption, ...
\usepackage[mathscr]{eucal}      % \mathscr, used by \XCal, \YCal, ...
\usepackage{stmaryrd}            % \llbracket \rrbracket, used by \indicator
\usepackage{accents}
\usepackage{upgreek}
\usepackage{nicefrac}
\usepackage{microtype}
\usepackage{float}
\usepackage[export]{adjustbox}
\usepackage{multirow}
\usepackage{colortbl}            % \cellcolor
\usepackage{makecell}
\usepackage{pifont}
\usepackage[most]{tcolorbox}     % \subtlebox, tcb theorems
\usepackage{empheq}              % boxed equations
\usepackage{soul}                % \hl, used by \hlblue

\graphicspath{{figs//}}
\allowdisplaybreaks

\DeclareRobustCommand{\hlblue}[1]{{\sethlcolor{SkyBlue!20}\hl{#1}}}

\newtcbox{\subtlebox}{nobeforeafter, math upper, tcbox raise base,
  enhanced, colframe=black!20, colback=SkyBlue!20, boxrule=0.5pt,
  arc=3pt, left=4pt, right=4pt, top=4pt, bottom=4pt}

\providecommand\numberthis{\addtocounter{equation}{1}\tag{\theequation}}

\newcommand{\defEq}{\stackrel{.}{=}}

\newcommand{\indicator}[1]{\llbracket #1 \rrbracket}

\newcommand{\argmax}{{\operatorname{argmax }}}

\renewcommand{\Pr}{\mathbb{P}}

\newcommand{\XCal}{\mathscr{X}}
\newcommand{\YCal}{\mathscr{Y}}

\newcommand{\Real}{\mathbb{R}}

\newtheorem{lemma}{Lemma}

\newtheorem{corollary}[lemma]{Corollary}

\newtheorem{proposition}[lemma]{Proposition}

\theoremstyle{definition}

\newtheorem{assumption}{Assumption}

\newtcbtheorem[number within=section]%
{remark_tcb}{Remark}{%
fonttitle=\bfseries\upshape,fontupper=\normalsize,
colframe=green!10!black,colback=green!2!white,
colbacktitle=green!20!white,coltitle=blue!75!black,
boxsep=1pt,left=2pt,right=2pt,top=1pt,bottom=1pt,
frame hidden,boxrule=1pt,
theorem style=plain}{rem}

\newtcbtheorem[number within=section]%
{assumption_tcb}{Assumption}{%
fonttitle=\bfseries\upshape,fontupper=\normalsize,
colframe=blue!5!black,colback=blue!2!white,
colbacktitle=blue!10!white,coltitle=blue!75!black,
boxsep=1pt,left=2pt,right=2pt,top=1pt,bottom=1pt,
frame hidden,boxrule=1pt,
theorem style=plain}{asm}

\newcommand{\globFeat}{\rm g}
\newcommand{\locFeat}{\rm s}
\newcommand{\nullFeat}{\rm n}
\newcommand{\sharedFeat}{\rm b}

\newcommand{\xN}{x}

\newcommand{\piCanonical}{\bar{\boldsymbol{\pi}}}

%% file: sec/0_abstract.tex
\begin{abstract}
%\emph{Out-of-distribution (OOD) detection} remains a notoriously challenging problem for deep learning classifiers. Recent work has introduced a new class of effective OOD detectors, based on a simple insight: whilst OOD samples may spuriously result in high-confidence predictions for a certain target class, their \emph{ranking} over \emph{non-target} classes does not reflect typical semantic patterns. While such \emph{ranking-based} OOD detectors have achieved impressive empirical success, they have two shortcomings. Empirically, such methods can degrade regular (in-distribution) test accuracy. Moreover, a precise formalism for \emph{why} a class ranking signal can improve OOD detection performance remains elusive. In this paper, we address both of these gaps by relating ranking-based OOD detectors to a form of \emph{knowledge distillation (KD)}. We use this perspective to design a new family of methods that maintain the strong OOD detection performance of ranking-based detectors, while \emph{boosting} the in-distribution test accuracy using the instance-level Plackett-Luce Knowledge Distillation loss (PL-KD). Further, we provide a formal justification for ranking-based OOD detectors, extending the classic notion of ``dark knowledge'' in KD to the OOD detection setting.
\emph{Out-of-distribution (OOD) detection} remains a notoriously challenging problem for deep learning classifiers. Recent work has introduced an effective new class of OOD detectors, based on a simple insight: whilst OOD samples may spuriously receive high-confidence predictions for a certain target class, their \emph{ranking} over \emph{non-target} classes may not reflect typical semantic patterns. While such \emph{ranking-based} OOD detectors have achieved impressive empirical success, they have two shortcomings: 
(i) empirically, such methods can degrade in-distribution test accuracy;
(ii) conceptually, a precise formalism for \emph{why} a class ranking signal can improve OOD detection remains elusive. In this paper, we address both of these gaps by relating ranking-based OOD detectors to a specific form of \emph{knowledge distillation (KD)}. We use this perspective to design a new 
% family of methods 
instance-level \emph{Plackett--Luce Knowledge Distillation loss} (\emph{PL-KD})
that maintains the strong OOD detection performance of ranking-based detectors, while \emph{boosting} the in-distribution test accuracy. 
% using the instance-level Plackett--Luce Knowledge Distillation loss (PL-KD). 
On CIFAR-10, CIFAR-100, and ImageNet-200, PL-KD reduces FPR95 relative to {\sc RankOOD} in every dataset and 
logit-based 
scoring-function combination we evaluate, while improving in-distribution accuracy by up to 5.3 points. Relative to standard cross-entropy training, it approximately halves both near- and far-OOD FPR95 on CIFAR-10.
% \textcolor{red}{we need to highlight some results - added these; double-check/replace with better stats?}. 
Finally, we provide a formal justification for ranking-based OOD detectors, extending the classic notion of ``dark knowledge'' in KD to the OOD detection setting.
\end{abstract}

%% file: sec/1_intro.tex
\section{Introduction}
\label{sec:intro}

Conventional supervised learning seeks a classifier capable of predicting a label for \emph{any} test sample;
however, such predictions can be unreliable when test samples do not follow the training distribution~\citep{Ben-David:2010}.
\emph{Out-of-distribution (OOD) detection} is the problem of identifying whether or not test samples follow the training distribution~\citep{Hendrycks:2017}.
% which allows for potentially abstaining from prediction on such samples.
Na\"{i}vely, one may rely on the classifier's prediction confidence 
(e.g., the softmax probability of the highest-scoring class);
unfortunately, this is sub-optimal both in theory~\citep{Hein:2019,Narasimhan:2024} and practice~\citep{Nguyen:2015}.
This has motivated a large literature on alternate OOD detectors~\citep{Lee:2018,Hendrycks:2019,Ren:2019,Huang:2021,Huang:2021b,Thulasidasan:2021,wang2022vim,Bitterwolf:2022,Katz:2022,Wei:2022,Sun:2022,Hendrycks:2022,Yang:2022,Ammar:2024,Karunanayake:2025,Karunanayake:2025b,Denipitiyage:2026,Meza:2025,YangZhu:2025,Yang:2025}.
{Despite this effort,
effective OOD detection for deep learning classifiers remains challenging: 
on the OpenOOD benchmark~\citep{Yang:2022}, 
even the best performing detectors
incorrectly flag $\sim30\%$ of inliers as being
OOD. Beyond benchmarks, such failures constrain the deployment of deep learning models in safety-critical settings, where perception systems must cope with unexpected inputs such as small animals or debris on the road~\cite{chan2021segmentmeifyoucan}. For instance, in October 2026, Tesla's CEO cited reliably detecting pets at night (``grey kittens on grey tarmac'') as the main obstacle to extending its Austin robotaxi service~\cite{lambert2026robotaxi}.}
% later into the night~\cite{lambert2026robotaxi}.}
% \noteakm{Merged with Sanjay's comments}
% The problem of OOD detection in deep learning
% systems has remained notoriously intractable despite
% a substantial body of research. On Open-OOD, a well established
% benchmark, the best ${\rm FPR}@95$ rates for near-OOD problems,
% hover around 0.3, i.e., the best OOD scoring methods
% will suppress nearly one in three inliers as being
% out of distribution. Thus it is no surprise that 
% OOD detectors are rarely used in deployed systems.}

% Recent work has introduced a class of effective new techniques~\citep{Karunanayake:2025,Karunanayake:2025b,Denipitiyage:2026}, based on a simple insight: 
% whilst OOD samples may spuriously result in high-confidence predictions for a certain class, 
% their predictions for \emph{other} classes tend to be arbitrary, and do not reflect semantic patterns inherent in the data.
% This has motivated {\sc RankOOD}~\citep{Denipitiyage:2026},
% which uses a \emph{ranking} based training objective and scoring function to achieve state-of-the-art performance across modern OOD benchmarks.
% Despite this impressive empirical success, a precise formalism for \emph{why} such a detector can perform well remains lacking.
% \todoakm{Other motivations?}
\begin{figure}
    \centering
    \includegraphics[width=0.5\textwidth]{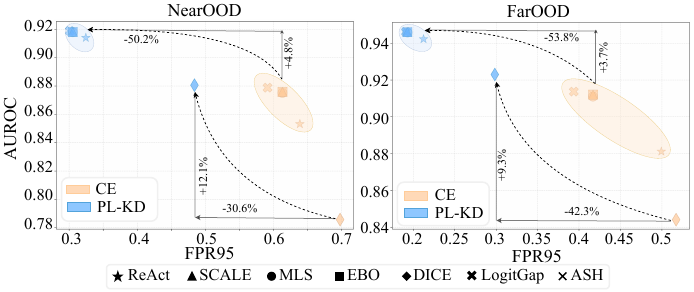}
    \caption{The OOD performance benefit of Plackett--Luce - Knowledge Distillation (PL-KD) loss over Cross Entropy (CE) - for the majority of logit-based OOD scores, PL-KD training improves both AUROC and ${\rm FPR}@95$ in both near and far OOD settings. }
    \label{fig:tldr}
    \vspace{-\baselineskip}
\end{figure}
Recent work~\citep{Karunanayake:2025,Karunanayake:2025b,Denipitiyage:2026} has introduced a new class of OOD detectors, resting on a simple insight:
whilst OOD samples may result in spuriously confident predictions for a certain target class, 
their 
% predictions 
\emph{rankings}
for \emph{non-target} classes may not reflect typical semantic structure of in-distribution samples.
% Thus, the 
% % structure 
% ranking
% of non-target class predictions can offer a distinct signature for detecting OOD samples.
% Examples of this signature include 
% a low gap in target versus average logits~\citep{Liang:2025},
% and ``unstructured'' 
% \emph{ranking} of non-target classes~\citep{Karunanayake:2025,Karunanayake:2025b,Denipitiyage:2026}. 
% Such methods have achieved strong empirical performance on modern OOD benchmarks.
% Amongst these,
% {\sc RankOOD}~\citep{Denipitiyage:2026}
% is notable for its state-of-the-art performance across modern OOD benchmarks,
% achieved via
% a {ranking} based training objective. 
% and bespoke OOD predictor.

\begin{figure*}[t!]
    \centering
    \includegraphics[width=\textwidth]{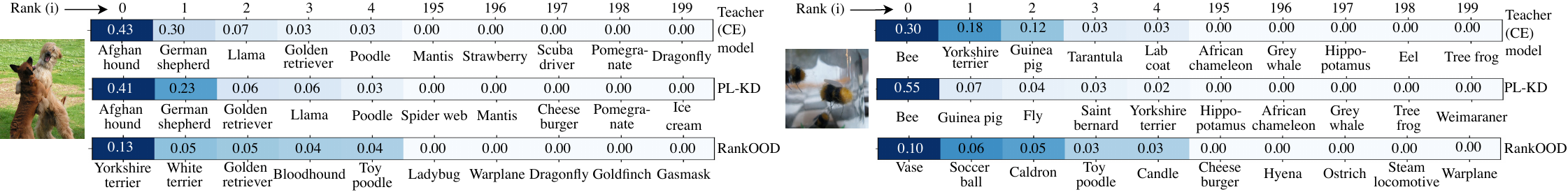}

    \caption{Two in-distribution test samples from ImageNet-200 for which {\sc RankOOD} incorrectly predicts OOD, while our PL-KD correctly identifies them as ID.
    Each row shows the top-$5$ and bottom-$5$ class probabilities for the teacher model trained with cross-entropy (CE), instance-level PL-KD, and {\sc RankOOD}.
    }
     \label{fig:illustrative-example}
    \vspace{-\baselineskip}
\end{figure*}

As demonstrated in~\citet{Karunanayake:2025b,Denipitiyage:2026}, strengthening such semantic structure via a {ranking} based training objective enables strong OOD detection performance.
Despite this impressive empirical success, 
there are two shortcomings in such methods.
Empirically, encouraging a 
strict 
% non-target signature for OOD samples 
ranking of non-target classes
can come at the expense of regular (in-distribution) test accuracy, which is an undesirable practical trade-off.
Conceptually,
a formalism for \emph{why} strengthening the structure of non-target classes can improve OOD detection remains elusive. 
% \notesuranga{Are we going too strong on RankOOD? We got SOTA in TinyImageNet and comparable results in other benchmarks.}
% \noteakm{Softened RankOOD SoTA claims}

% %

%
% % \begin{figure}[!t]
% % \centerin
% \begin{minipage}{0.4\textwidth}
%     \centering
%     \includegraphics[height=4cm]{Figures/id_vs_fpr.pdf}
%     \caption{Average ID accuracy vs average near- and far-OOD FPR@95 across CIFAR-10, CIFAR-100, and ImageNet-200. Instance-level PL-KD (ours) achieves the second-highest ID accuracy, surpassed only by CRAFT~\cite{Karunanayake:2025b}.
%     % , while consistently outperforming all post-hoc methods in FPR@95 and attaining performance comparable to ranking-based approaches.
%     }
%     \label{fig:illustrative-example}
%     \vspace{-\baselineskip}
% \end{minipage}
% \hfill
% \begin{minipage}{0.58\textwidth}
%     \centering
%     \includegraphics[width=\textwidth]{Figures/imagenet_exmples.pdf}
%     \caption{Two in-distribution test samples from ImageNet-200 for which RankOOD incorrectly predicts OOD, while our method correctly identifies them as ID. The figure shows the top-10 and bottom-10 class probabilities for the teacher model trained with cross-entropy (CE), instance-level PL-KD, and RankOOD.}
%     \label{fig:imagenet-example}
%     \vspace{-\baselineskip}
% \end{minipage}
% \end{figure}
% %
In this paper,
we provide a step towards addressing these issues by relating 
% {\sc RankOOD} 
ranking-based OOD detectors
to a form of \emph{knowledge distillation (KD)}~\citep{Bucilua:2006,Hinton:2015}.
This connection enables progress on the aforementioned shortcomings of ranking-based detectors.
Algorithmically, it offers a unified view of ranking- and KD-based approaches, resulting in a new 
\emph{PL-KD}
objective that achieves strong OOD detection performance while \emph{boosting} the in-distribution accuracy (Figure~\ref{fig:tldr}). 
Theoretically, it offers a formal grounding for why enforcing structure in non-target classes can improve OOD detection.
This
highlights the value of the ``dark knowledge'' in OOD detection,
which
complements existing analysis of its benefits in conventional classification~\citep{Lopez-Paz:2016,Mobahi:2020,Menon:2021,Nagarajan:2023}. 

% Figure~\ref{fig:illustrative-example} provides qualitative insights
% about combining ranking and instance-level teacher probabilities. For the Afghan Hound, compared to the teacher CE model,
% the PL-KD model lowers the top class confidence and correctly flips Golden Retriever before Llama; this is the impact of ranking. In the Bee example, the PL-KD is more ID-accurate than {\sc RankOOD}, and increases the top class confidence.

Figure~\ref{fig:illustrative-example} illustrates how PL-KD combines ranking
with instance-level teacher probabilities. PL-KD groups semantically similar
classes together: for the Afghan hound, it moves Golden retriever above Llama,
so that the dog breeds rank together; for the bee, it promotes Fly into the
top three and demotes Yorkshire terrier, while increasing confidence in the
correct class. In contrast, {\sc RankOOD} misclassifies both images (as Yorkshire terrier and Vase), with the remaining ranks following the class-level template of
the \emph{predicted} class rather than the image content; such
misclassifications underlie its lower ID accuracy.
Its low confidence in these predictions further leads it to flag both ID
samples as OOD.

In summary, our contributions are:
% \todoakm{add a summary figure}
\begin{enumerate}[itemsep=0pt,topsep=0pt,leftmargin=20pt,label=(\roman*)]
    \item we establish 
    that
    ranking-based OOD detectors
    are a
    specific
    instance of
    \emph{self}-distillation 
    % for specific choices of loss and supervision 
    (\S\ref{sec:rankood-as-kd});
    we then unify 
    ranking- and KD-based detectors as special cases of a more general objective (\S\ref{sec:epl}, \S\ref{sec:unified}),
    culminating in a new \emph{PL-KD} loss (\S\ref{sec:pl-kd}) that achieves
    strong 
    OOD detection performance 
    \emph{and}
    in-distribution accuracy.
    
    \item we establish that under an intuitive generative model of 
    the data
    (\S\ref{sec:theory}),
    the ``dark knowledge'' in
    KD
    encourages distinct non-target class signatures for ID versus OOD samples;
    this provides a formal grounding for the success of 
    OOD detectors based on such signatures.
    % post-hoc and training based OOD detectors.
    
    \item we empirically verify that on CIFAR and ImageNet,
    training with the new PL-KD loss
    can achieve strong OOD detection performance
    \emph{without} sacrificing ID
    test accuracy (\S\ref{sec:experiments}).
    Compared to CE and other distillation-based approaches, PL-KD reduces far- and near-OOD ${\rm FPR}@95$ by $1.5$--$54.2\%$ and $2.4$--$50.9\%$, respectively.
    % , across the three datasets.}
    % add some results numbers? let's highlight the Imagenet result here - two things to highlight, improvment over CE and other distillation methods, and the image net result. Either way some numers are required here.}
\end{enumerate}

More broadly, 
% we believe
the connection 
to KD sheds conceptual light on ranking-based OOD detectors,
and can inspire innovation leveraging insights from the KD literature. 

%% file: sec/2_background.tex
\section{Background and related work}
\label{sec:background}

We review the out-of-distribution detection problem in standard multi-class classification settings.

%
% \subsection{Multi-class classification and knowledge distillation (KD)}
\subsection{Classification and knowledge distillation (KD)}

Multi-class classification concerns learning a mapping from instances $x \in \XCal$ to labels $y \in \YCal \defEq \{ 1, 2, \ldots, L \}$ (e.g., images to semantic categories).
Conventionally, one learns a scorer $\mathbf{f} \colon \XCal \to \Real^{L}$,
where the \emph{logit} $f_y( x )$ is the predicted affinity for label $y$ on instance $x$.
Neural classifiers employ
$f_y( x ) = \mathbf{w}_y^\top \boldsymbol{\Phi}( x )$ for weights $\mathbf{w}_y \in \Real^D$ and representations $\boldsymbol{\Phi}( x ) \in \Real^D$.
These may be converted into probabilities $\mathbf{p} \colon \XCal \to \Delta( \YCal )$ via the softmax
$p_y( x ) \propto \exp( f_y( x ) )$.
% $Z( x ) \defEq \sum_{y' \in \YCal} \exp( f_{y'}( x ) )$.
This yields a \emph{classifier} $h \colon \XCal \to \YCal$ via 
$h( x ) \defEq \argmax_{y \in \YCal} \, p_y( x )$.
% , i.e., predicting the most probable class.
% In certain contexts, for convenience we write $p( y \mid x )$ to denote $p_y( x )$.

The goal in learning $h$ is to minimise the \emph{misclassification error}
$ R_{\rm err}( h ) \defEq \Pr_{\rm id}( y \neq h( x ) ) $
for test samples drawn from some 
% (unknown) 
distribution $\Pr_{\rm id}$.
To achieve this, given a labelled training sample 
% $S = \{ ( x^{(n)}, y^{(n)} ) \}_{n = 1}^{N} \sim \Pr_{\rm id}^{N}$,
$S = \{ ( x^{(n)}, y^{(n)} ) \}_{n = 1}^{N}$ 
with $( x^{(n)}, y^{(n)} ) \sim \Pr_{\rm id}$,
one minimises
$\hat{R}_{\rm ce}( \mathbf{f} ) \defEq \frac{1}{| S |} \sum_{( x, y ) \in S} \ell_{\rm ce}( y, \mathbf{f}( x ) )$
for
the \emph{softmax cross-entropy (CE)} loss
$\ell_{\rm ce}( y, \mathbf{f}( x ) ) \defEq -\log p_y( x )$.
% for $p_y( x ) \propto \exp( f_y( x ) )$,
% One then minimises the \emph{empirical surrogate risk}:
% for loss function $\ell \colon \YCal \times \Real^{N} \to \Real_+$:
% \todoakm{Should we use $(x, y) \in S$ instead?}
% \begin{equation}
%     \label{eqn:surrogate-loss}
%     % R_{\rm surr}( \mathbf{f} ) \defEq \frac{1}{N} \sum_{n = 1}^{N} \ell( y^{(n)}, \mathbf{f}( x^{(n)} ) ),
%     \hat{R}_{\rm ce}( \mathbf{f} ) \defEq \frac{1}{| S |} \sum_{( x, y ) \in S} \ell_{\rm ce}( y, \mathbf{f}( x ) ).
% \end{equation}
% where  denotes a \emph{surrogate loss function}.
% A canonical loss 
% where
% $\ell_{\rm ce}( y, \mathbf{f}( x ) ) \defEq -\log p_y( x )$
% is the \emph{softmax cross-entropy},
% for $p_y( x ) \propto \exp( f_y( x ) )$.

%
% \subsection{Knowledge distillation (KD)}

Knowledge distillation (KD)~\citep{Bucilua:2006,Hinton:2015}
is a simple but practically effective model compression technique.
Given a 
% (typically expensive) 
``teacher'' model $\mathbf{f}^{\rm t}$ 
--- 
typically trained with the regular softmax cross-entropy $\ell_{\rm ce}$
---
one trains a 
% (typically cheaper) 
``student'' model $\mathbf{f}$ with the teacher's predicted probabilities as targets:
\begin{equation}
    \label{eqn:kd}
    % \ell_{\rm kd}( \mathbf{p}^{\rm t}( x^{(n)} ), \mathbf{f}( x^{(n)} ) ) \defEq \sum_{y' \in [ L ]} p^{\rm t}_{y'}( x^{(n)} ) \cdot \ell_{\rm ce}( y', \mathbf{f}( x^{(n)} ) ).
    \ell_{\rm kd}( \mathbf{p}^{\rm t}( x ), \mathbf{f}( x ) ) \defEq \sum_{y' \in [ L ]} p^{\rm t}_{y'}( x ) \cdot \ell_{\rm ce}( y', \mathbf{f}( x ) ).
    \vspace{-2mm}
\end{equation}
In practice, one typically minimises a weighted combination of $\ell_{\rm kd}$ and $\ell_{\rm ce}$.
Note that the latter is a special case of KD using
a ``one-hot'' teacher, wherein 
% $\mathbf{p}^{\rm t}_{y'}( x^{(n)} ) = \indicator{ y' = y^{(n)} }$,
${p}^{\rm t}_{y'}( x ) = \indicator{ y' = y }$.
% Two special cases are of note.
% First, 
% the ``one-hot'' teacher, wherein 
% % $\mathbf{p}^{\rm t}_{y'}( x^{(n)} ) = \indicator{ y' = y^{(n)} }$,
% $\mathbf{p}^{\rm t}_{y'}( x ) = \indicator{ y' = y }$,
% recovers the original softmax cross-entropy.
% Second, the ``uniform'' teacher, wherein 
% % $\mathbf{p}^{\rm t}_{y'}( x^{(n)} ) = \frac{1}{L}$,
% $\mathbf{p}^{\rm t}_{y'}( x ) = \frac{1}{L}$,
% recovers label smoothing~\citep{Szegedy:2016,Muller:2019,Lukasik:2020}.
% Conventionally, distillation is applied in settings where the teacher is larger than the student.
One commonly applies \emph{temperature scaling} to the teacher (and optionally student) logits via $\mathbf{p}^{\rm t}( x ) \propto \exp( T^{-1} \cdot \mathbf{f}^{\rm t}( x ) )$ for $T > 0$, to suitably adjust the scale of ``dark knowledge''.

Interestingly, even \emph{self-distillation} --- where the teacher and student are the exact same model family --- can prove beneficial over training with $\ell_{\rm ce}$~\citep{Furlanello:2018}.
Intuitively, this can be understood 
as
% self-distillation providing a regulariser
regularising the standard CE loss~\citep{Mobahi:2020,Nagarajan:2023}.

\subsection{Out-of-distribution (OOD) detection}
\label{sec:ood-background}

In practice, one may encounter 
\emph{out-of-distribution (OOD)}
test samples, drawn from 
% a distinct distribution 
$\Pr_{\rm ood} \neq \Pr_{\rm id}$.
% A classifier's predictions are likely unreliable on such  samples,
% % ---
% % as opposed to \emph{in-distribution (ID)} samples 
% % ---
% as compared to \emph{in-distribution (ID)} samples.
% Thus, it is of interest to perform \emph{out-of-distribution (OOD) detection},
\emph{OOD detection}
is the problem of 
% identifying such samples
% by 
learning an auxiliary scorer $g \colon \XCal \to \Real$ 
% measuring the confidence that an instance is drawn from $\Pr_{\rm id}$.
% The objective for $g$ is 
to distinguish between $\Pr_{\rm id}$ and $\Pr_{\rm ood}$
---
as measured by a function of the 
{false and true positive rates},
treating ID samples as positives
% ${\rm FPR}( g ) \defEq \Pr_{\rm id}( g( x ) < 0 )$,
% ${\rm TPR}( g ) \defEq \Pr_{\rm ood}( g( x ) < 0 )$
---
% Note that 
% % in constructing $g$, 
% it 
% remains paramount for the base classifier $h$ to to minimise 
\emph{without} affecting 
the ID error $R_{\rm err}( h )$.
% By convention, higher values of $g( x )$ denote greater confidence that $x$ is drawn from $\Pr_{\rm id}$.

Broadly, there are two categories of OOD detection techniques.
\emph{Post-hoc} techniques 
% do not disturb the training algorithm for learning $\mathbf{f}$, and simply 
construct $g$ based on fixed criteria, without modifying the training algorithm~\citep{Hendrycks:2022,Liu:2020b,Liang:2025,Lee:2018b,Ren:2019,Sun:2021,wang2022vim,Morteza:2022,Ammar:2024}.
\emph{Training-based} techniques modify the training of $\mathbf{f}$, so as to better separate ID and OOD samples~\citep{Hendrycks:2019,Wei:2022,Hendrycks:2019b,Tack:2020,Isaac:2024}.
Of interest to us is \emph{two-stage} training, where the first stage model outputs refine the loss in the second stage.
% For example, 
Several works have studied various forms of (self-)KD for OOD detection~\citep{Wu:2023,Tang:2025,Yang:2025,Zhao:2025,Wang:2023,Kralik:2025}.
Recent works~\citep{Karunanayake:2025b,Denipitiyage:2026} explored a two-stage \emph{ranking} approach,
% wherein
% information from the first stage is used to refine the second stage training.
% Recently,~\citet{Karunanayake:2025b,Denipitiyage:2026} introduced a line of work that 
resting on an intriguing insight:
% whilst an OOD sample $\bar{x} \sim \Pr_{\rm ood}$ may result in a high confidence prediction in a certain target class $y$
% --- 
% i.e., $p_{y}( \bar{x} ) \gg \max_{y' \neq y} p_{y'}( \bar{x} )$ 
% ---
% the prediction on \emph{non-target} classes may not follow the expected semantic structure.
% For example, 
if an image is predicted as
the {\tt dog} class, but semantically related classes such as \{ {\tt wolf},{\tt coyote} \} do not have high scores, it is evidence that the sample may be OOD.
% The absence of such a pattern provides evidence that the sample may be OOD.
Empirically, such 
% ranking-based OOD 
detectors achieve strong OOD detection performance; 
however, they may be observed to sacrifice ID accuracy.
% on standard  benchmarks.
% Conceptually, 
% % however,
% a formalism for \emph{why} such 
% training helps is lacking.

In this work, we primarily use the \emph{LogitGap}~\citep{Liang:2025} scoring function, defined as the average gap between the largest and the remaining logits:
$g(x)$ $\defEq f_{[1]}(x) - \frac{1}{L-1} \sum_{i > 1} f_{[i]}(x)$,
where $f_{[i]}(x)$ denotes the $i^{\rm th}$-largest logit. 
We also evaluate other logit-based scoring functions, such as maximum logit (MLS) $g(x) = \max_{i} f_i(x)$, and energy (EBO) 
$g(x) = \log \sum_i \exp(f_i(x))$. 
% \textcolor{olive}{
We do not focus on 
the maximum softmax probability (MSP)
since
empirically, it tends to be outperformed by logit-based methods~\citep{Liang:2025}.
 % }
% \textcolor{red}{can we add a sentence on why probability-based is not considered?}

% The threshold $\lambda$ is selected to achieve a 95\% true positive rate (TPR) on ID samples.
% We aim to address this by establishing a relation to KD.
% further, the relationship to KD-based methods is elusive.
% However, the formal understanding of \emph{why} KD can improve OOD detection has been less explored.
% Further, as we shall shortly see, the impressive OOD detection performance can come at the price of regressing in the standard ID misclassification error.
% \todoakm{Smoother transition.}
% To make progress on this question, we now more closely inspect the mechanics of {\sc RankOOD}.

% For example,~\citet{Yang:2025} proposed to use a teacher model to identify pseudo-OOD samples, which may strengthen the training signal for the student model.
% \citet{Tang:2025} proposed to use signal from a teacher model to prune certain activations.
% \citet{Kralik:2025} proposed to use the student-teacher discrepancy as a measure of OOD likelihood.
% \citet{Wu:2023} proposed to use a richer form of distillation to boost OOD detection ability of the student.
% \citet{Wang:2023} proposed distilling an ensemble of teachers for improved OOD detection.
% \citet{Zhao:2025} proposed a distillation based data synthesis, to interpolate between ID and OOD samples during student training.
% However, .
% \todoakm{Clean up discussion}

%% file: sec/3_rankood_as_kd.tex
% \section{Ranking-based OOD detection meets knowledge distillation}
\section{Ranking- meets KD-based OOD detection}
\label{sec:rankood}

% We now explicate the relation between ranking-based OOD detection and knowledge distillation (KD).
% with a focus on the recent {\sc RankOOD} method~\citep{Denipitiyage:2026}.
% Based on this, we present a family of methods combining KD and {\sc RankOOD} (Table~\ref{tbl:summary}), 
% % two new methods that combine the strengths of 
% and examine their distinct design choices. 
% implications of design choices made in the former.
% We then critically examine the strengths and weaknesses of each approach.

%
% \subsection{A review of the {\sc RankOOD} algorithm}
% \subsection{Ranking-based OOD detection as self-distillation}
\label{sec:rankood-as-kd}

We now explicate the relation between ranking-based OOD detection and knowledge distillation (KD).
We focus on {\sc RankOOD}~\citep{Denipitiyage:2026},
% {\sc RankOOD} is 
% which is
a two-stage training method (cf.~\S\ref{sec:ood-background})
where
in the first stage,
we train an initial model $\mathbf{f}^{\rm t} \colon \XCal \to \Real^{L}$ on the training sample $S$ by minimising
% ~\eqref{eqn:surrogate-loss} with 
% the softmax cross-entropy $\ell_{\rm ce}$.
$\hat{R}_{\rm ce}( \mathbf{f}^{\rm t} )$.
For each class $y \in \YCal$,
we then compute 
a
\emph{canonical ranking} $\piCanonical^{( y )} \in \mathfrak{S}( \YCal )$ over classes
    % \todoakm{Accept $y$ as an argument?}
    % by solving a certain integer linear program 
% based on a generalised confusion matrix
(see~\citet[Section 3.1]{Denipitiyage:2026}),
which reflects the ``typical'' ranking over non-target classes.
% Intuitively, 
% for target class
% $y$  (e.g., \texttt{dog}),
% this is the ``typical'' ranking over non-target classes (e.g., \texttt{wolf} is ranked ahead of \texttt{rabbit}).
% Note that this ranking is at a \emph{class-level} rather than \emph{instance-level} (i.e., all samples with the same target class have the same canonical ranking).
% In the second stage,
% we train a \emph{new} model $\mathbf{f}$ to minimise the \emph{ListMLE} loss $\ell_{\rm pl}$~\citep{Xia:2008} 
% ---
% underpinned by the celebrated Plackett-Luce distribution over rankings~\citep{Plackett:1975,Luce:1959}
% ---
% against the canonical rankings:
% % \todoakm{Fix the notation}
% \begin{align*}
%     \hat{R}_{\rm pl}( \mathbf{f} ) &\defEq \frac{1}{| S |} \sum_{(x, y) \in S} \ell_{\rm pl}( \piCanonical^{( y )}, \mathbf{f}( x ) ) \\
%     % R_{\rm pl}( \mathbf{f} ) &\defEq \frac{1}{N} \sum_{n = 1}^{N} \ell_{\rm pl}( \piCanonical( \yN ), \mathbf{f}( \xN ) ) \\
%     \numberthis
%     \label{eqn:pl-loss}
%     \ell_{\rm pl}( \boldsymbol{\pi}, \mathbf{f}( x ) ) &\defEq \sum_{i = 1}^{L} -\log \frac{\exp( f_{\pi( i )}( x ) )}{\sum_{j = i}^{L} \exp( f_{\pi( j )}( x ) )},    
%     % \ell_{\rm pl}( \piCanonical( \yN ), \mathbf{f}( \xN ) ) &\defEq \sum_{i = 1}^{L} -\log \frac{\exp( f_{[i]}( \xN ) )}{\sum_{j = i}^{L} \exp( f_{[j]}( \xN ) )},
% \end{align*}
% where the ListMLE loss $\ell_{\rm pl}$ is defined for any ranking $\boldsymbol{\pi} \in \mathfrak{S}( \YCal )$.
In the second stage,
we train a \emph{new} model $\mathbf{f}$ 
by minimising
a ranking loss against $\boldsymbol{\bar{\pi}}^{(y)}$.
% to encourage ID samples to reflect this structure.
% to strengthen their presence in ID samples.
% \todoakm{Subscript change?}
% \begin{equation}
%     \label{eqn:rankood}
%     \hat{R}_{\rm pl}( \mathbf{f} ) \defEq \frac{1}{| S |} \sum_{(x, y) \in S} \ell_{\rm pl}( \piCanonical^{( y )}, \mathbf{f}( x ) ),
% \end{equation}
% i.e.,
Specifically, we minimise
% $ \hat{R}_{\rm pl}( \mathbf{f} ) \defEq \frac{1}{| S |} \sum_{(x, y) \in S} \ell_{\rm pl}( \piCanonical^{( y )}, \mathbf{f}( x ) ), $
% where $\ell_{\rm pl}$ denotes 
the \emph{ListMLE} loss~\citep{Xia:2008}, defined for $\boldsymbol{\pi} \in \mathfrak{S}(\YCal)$ as
\begin{equation}
    \ell_{\rm pl}( \boldsymbol{\pi}, \mathbf{f}( x ) ) \defEq \sum_{i = 1}^{L} -\log \frac{\exp( f_{\pi( i )}( x ) )}{\sum_{j = i}^{L} \exp( f_{\pi( j )}( x ) )}.
    \label{eqn:pl-loss}
\end{equation}
% underpinned by the celebrated Plackett-Luce distribution~\citep{Plackett:1975,Luce:1959}:
%\begin{align}
    % \hat{R}_{\rm pl}( \mathbf{f} ) &\defEq \frac{1}{| S |} \sum_{(x, y) \in S} \ell_{\rm pl}( \piCanonical^{( y )}, \mathbf{f}( x ) ) \\
    % R_{\rm pl}( \mathbf{f} ) &\defEq \frac{1}{N} \sum_{n = 1}^{N} \ell_{\rm pl}( \piCanonical( \yN ), \mathbf{f}( \xN ) ) \\
    %\label{eqn:pl-loss}
    %( \forall \boldsymbol{\pi} \in \mathfrak{S}( \YCal ) ) \, \ell_{\rm pl}( \boldsymbol{\pi}, \mathbf{f}( x ) ) &\defEq \sum_{i = 1}^{L} -\log \frac{\exp( f_{\pi( i )}( x ) )}{\sum_{j = i}^{L} \exp( f_{\pi( j )}( x ) )}.
    % \ell_{\rm pl}( \piCanonical( \yN ), \mathbf{f}( \xN ) ) &\defEq \sum_{i = 1}^{L} -\log \frac{\exp( f_{[i]}( \xN ) )}{\sum_{j = i}^{L} \exp( f_{[j]}( \xN ) )},
%\end{align}
% In practice, one typically minimises a weighted sum of $\ell_{\rm pl}$ and the softmax cross-entropy $\ell_{\rm ce}$.
% a slight abuse of notation,
% we 
% use 
% $f_{\pi( i )}( \xN )$ to denote the logit for the $i^{\rm th}$ 
% ranked class according to $\piCanonical( \yN )$.
% Finally, the OOD score 
% % for any $x \in \XCal$ 
% % via
% is a distance between the canonical
% and predicted ranking~\citep[Section 3.3]{Denipitiyage:2026}.

%%%
\subsection{Contrasting {\sc RankOOD} and KD}
\label{sec:contrast-rankood-kd}
\label{sec:ranking-family}
\label{sec:canonical-kd}

{\sc RankOOD} 
is evidently related to \emph{self-distillation (self-KD)}~\citep{Furlanello:2018}:
both methods
perform two-stage training,
where the second stage uses predictions from the first to modify the loss.
Compared to the canonical instantiation of self-KD~\eqref{eqn:kd},
however, {\sc RankOOD} makes two fundamental modifications.
By examining these more closely, 
we identify new methods that modify
self-KD on \emph{one} axis (cf.~Table~\ref{tbl:summary}).

% \begin{enumerate}[label=(\roman*),itemsep=\baselineskip,leftmargin=16pt,topsep=0pt]
    % \item 
    \textbf{Supervision granularity}: 
    given a training sample $(x, y)$,
    {\sc RankOOD}
    employs \emph{class-level} supervision $\piCanonical^{( y )}$.
    Importantly, 
    {every} 
    % training sample 
    $(x', y)$ with the same label $y \in \YCal$ receives the \emph{same} supervision.
By contrast, self-KD uses the 
\emph{instance-level}
teacher probabilities $\mathbf{p}^{\rm t}( x )$ as the supervision target.
    
    % rather than use 
    % \emph{instance-level} supervision via
    % the teacher's probabilities $\mathbf{p}^{\rm t}( x )$, 

%%%%%%%%%%%%
By incorporating {\sc RankOOD}'s idea of {class-level supervision} into KD,
we arrive at \textbf{class-level KD}:
for each class $y \in \YCal$, we compute the \emph{marginal teacher probabilities} $\bar{\mathbf{p}}^{(y)} \defEq \mathbb{E}_{x \mid y}[ \mathbf{p}^{\rm t}( x ) ]$,
% We now use this as supervision, 
and minimise 
$\ell_{\rm kd}( \bar{\mathbf{p}}^{(y)}, \mathbf{f}( x ) )$.
% This offers the inverse of the design choices
% in the instance-level {\sc RankOOD} described above.
%%%%%%%%%%%%

    % \item 
    \textbf{Loss function}: 
{\sc RankOOD} employs the ListMLE \emph{ranking} loss~\eqref{eqn:pl-loss}, which is an unweighted sum of $L$ loss terms.
By contrast, self-KD uses a \emph{weighted} softmax cross-entropy~\eqref{eqn:kd}.

%%%%%%%%%%%% 
By incorporating {\sc RankOOD}'s loss function into self-KD,
we arrive at
% Starting from self-KD and modifying only the loss yields 
\textbf{instance-level {\sc RankOOD}}:
% Now suppose 
% we convert this to 
we compute an \emph{instance-level label ranking}
$\boldsymbol{\pi}^{\rm t}( \xN ) \defEq {\rm argsort}( \mathbf{p}^{\rm t}( \xN ) )$ from the teacher probabilities,
and
minimise
$\ell_{\rm pl}( \boldsymbol{\pi}^{\rm t}( \xN ), \mathbf{f}( x ) )$.
% Compared to {\sc RankOOD}, this employs \emph{instance-level} supervision, which varies for different $x \in \XCal$ sharing the same target class $y$.
% Compared to self-KD, this only uses the teacher's \emph{ordering} of classes, and not its \emph{confidence} in them.  
%%%%%%%%%%%% 

% \end{enumerate}
% We now examine the implications of these modifications more closely.
% These design choices naturally suggest a \emph{family} of potential OOD detectors, which we now explore.

%
\begin{table}[!t]
    \centering
    \caption{Summary of training objectives for OOD detection, factorised based on their
    loss function (weighted softmax cross-entropy~\eqref{eqn:kd}, or ranking-based ListMLE~\eqref{eqn:pl-loss}), 
    and supervision granularity (instance- or class-level supervision).
    {\sc RankOOD}~\citep{Denipitiyage:2026} and self-KD~\citep{Furlanello:2018} are distinct on both axes. 
    By modifying self-KD along only one axis,
    we obtain two novel approaches (\hlblue{highlighted});
    by further unifying both losses in \S\ref{sec:unified}, we obtain a new 
    \emph{PL-KD}
    method that combines strengths of both KD and ranking. Going forward, our method of choice is
    the \textbf{Instance-level PL-KD.}
    % : %, which to our best knowledge are novel: 
    % KD with canonical ranking information, 
    % KD with marginal class-probabilities,
    % and {\sc RankOOD} with instance-level ranking information.
    }
    \label{tbl:summary}
    \renewcommand{\arraystretch}{1.25}
    \setlength{\tabcolsep}{8pt}     

    {\small
    \begin{tabular}{@{} l l l l @{}}
        \toprule
        & & \multicolumn{2}{c}{\textbf{Supervision granularity}} \\
        \cmidrule(l){3-4} 
        & & \multicolumn{1}{c}{Class-level (CL)} & \multicolumn{1}{c}{Instance-level (IL)} \\        \midrule
        \multirow{3}{*}{\rotatebox[origin=c]{90}{\textbf{Loss}\hspace{2pt}}} & KD      & \cellcolor{SkyBlue!20}{CL KD~(\S\ref{sec:ranking-family})} & Self-KD~\citep{Furlanello:2018} \\
        & ListMLE & {\sc RankOOD}~\citep{Denipitiyage:2026} & \cellcolor{SkyBlue!20}{IL {\sc RankOOD}~(\S\ref{sec:ranking-family})} \\
        & PL-KD & \cellcolor{SkyBlue!20}{CL PL-KD~(\S\ref{sec:method})} & \cellcolor{SkyBlue!20}{IL PL-KD~(\S\ref{sec:method})} \\
        \bottomrule
    \end{tabular}
    }

    \vspace{-\baselineskip}
\end{table}

%%%
\subsection{Bridging {\sc RankOOD} and KD}

Having arrived at four distinct methods (Table~\ref{tbl:summary}),
we now more closely examine the axes of variation. 
% in Table~\ref{tbl:summary}, namely, the supervision and loss.

%
\textbf{Supervision granularity}.
% As noted, standard~\eqref{eqn:rankood} and instance-level {\sc RankOOD}~\eqref{eqn:instance-rankood} differ in their levels of supervision granularity:
% the class-level canonical ranking $\piCanonical^{( y )}$
% the regulariser 
% The coarse class-level canonical ranking $\piCanonical^{( y )}$
% in
% standard {\sc RankOOD} %~\eqref{eqn:rankood} 
% offers the \emph{same}, aggregate ranking over non-target classes 
% for all training samples sharing the same label.
% Compared to instance-level {\sc RankOOD} (\S\ref{sec:canonical-kd}),
% on the one hand, 
% this 
Class-level supervision
offers a potential benefit compared to instance-level supervision:
the former is 
a highly \emph{robust} signal,
as aggregating class-similarities across all training samples can have a de-noising effect. 
However, this also introduces a potential risk: 
by 
enforcing the same pattern for 
\emph{every} training sample of the same class, 
one may suppress \emph{instance-specific} structure, and thus negatively impact the classification performance.
The latter is not merely hypothetical:
% in practice,
as we demonstrate in~\S\ref{sec:experiments}, 
% the impressive OOD detection performance of 
{\sc RankOOD} often 
% comes at the expense of regression on the 
degrades the
regular ID test accuracy.

\textbf{Loss function}.
% Both the KD and ListMLE losses can be understood as a certain \emph{regularised} form of softmax cross-entropy.
% For ease of comparison, let us compare standard {\sc RankOOD} and 
% % canonical rank KD.
% marginal teacher KD.
% To help distinguish the KD and ListMLE loss,
% let us compare standard {\sc RankOOD} and 
% class-level KD.
The ListMLE loss 
% in 
% standard {\sc RankOOD}
can be understood as a \emph{regularised cross-entropy}.
% By construction of the canonical ranking, 
% % the top ranked class
% $\bar{\pi}^{( y )}( 1 ) = y$.
% Thus, 
For any $(x, y) \in S$, 
by isolating the first summand of~\eqref{eqn:pl-loss},
$\ell_{\rm pl}( \piCanonical^{( y )}, \mathbf{f}( \xN ) )$ 
equals $\ell_{\rm ce}( y, \mathbf{f}( \xN ) )$
plus a \emph{regulariser}
$\Omega( \piCanonical^{( y )}, \{ f_{y'}( \xN ) \colon y' \neq y \} )$
% % is equally expressible as:
% \begin{align*}
%    \ell_{\rm pl}( \piCanonical^{( y )}, \mathbf{f}( \xN ) ) 
%    % &= \sum_{i = 1}^{L} -\log \frac{\exp( p_{\pi( i )}( x ) )}{\sum_{j = i}^{L} \exp( p_{\pi( j )}( x ) )} \\
%    &= 
%    % -\log p_{\pi( 1 )}( \xN ) 
%    \ell_{\rm ce}( y, \mathbf{f}( \xN ) )
%    + \Omega\left( \piCanonical^{( y )}, \left\{ f_{y'}( \xN ) \colon y' \neq y \right\} \right),
% \end{align*}
% % The first term is the usual softmax cross-entropy loss.
% % % , using the top-ranked class under $\piCanonical^{( y )}$ 
% % % % --- \emph{viz.} $\pi( 1 )$ in our notation --- 
% % % as the target.
% % The second term can be 
% That is, compared to the softmax cross-entropy, 
% the loss invokes an
% additional
% \emph{regulariser} 
% $\Omega( \cdot )$
on the \emph{non-target} class logits. 
% $\{ f_{y'}( \xN ) \colon y' \neq y \}$.
% intuitively, these are encouraged to respect the class ranking in $\boldsymbol{\pi}$.
% The regulariser $\Omega( \cdot )$ thus enforces additional constraints over the usual softmax cross-entropy.
% In the context of {\sc RankOOD}, we make two observations regarding $\Omega( \cdot )$.
% First,  by construction 
% and
Note that
$\Omega( \cdot )$ does \emph{not} depend on the target logit $f_{y}( \xN )$.
Thus
the ListMLE loss does {not} {explicitly} impose any further structure on this logit,
% compared to $\ell_{\rm ce}$.
% (One may expect an \emph{implicit} effect on this logit, owing to sharing of feature extraction parameters in neural network classifiers.)
% There are two implications of this.
% First, it 
% This
% suggests that {\sc RankOOD} 
and so
may not 
reduce \emph{over-confidence}~\citep{Nguyen:2015} on the target class.
% improve the detection of the MSP scorer.
% $g( x ) \defEq \max_{y \in \YCal} p_{y}( x )$.
% In particular, it may not reduce the \emph{over-confidence} of the maximum softmax probability.
% \todoakm{Perhaps needs a reference}
% Rather, to reap the benefits of {\sc RankOOD}, one must use a scorer that explicitly examines the structure of the non-top class logits.

Similarly, for the 
class-level
% canonical probability 
marginal teacher probability 
$\mathbf{\bar{p}}^{(y)}$,
% in~\S\ref{sec:canonical-kd}, 
the KD loss 
% in~\eqref{eqn:canonical-kd} 
% is expressible as
% $$ \ell_{\rm kd}( \mathbf{\bar{p}}^{(y)}, \mathbf{f}( x ) ) = \ell_{\rm ce}( y, \mathbf{f}( x ) ) + \left( f_y( x ) - \sum_{y' \in \YCal} \bar{p}^{(y)}_{y'} \cdot f_{y'}( x ) \right). $$
equals
$\ell_{\rm ce}( y, \mathbf{f}( x ) )$
plus a regulariser $f_y( x ) - \sum_{y' \in \YCal} \bar{p}^{(y)}_{y'} \cdot f_{y'}( x ) $.
The latter shrinks the target logit to the \emph{weighted mean} of other logits,
and thus potentially mitigates over-confidence in the top-scoring class.
% On the one hand, this regulariser \emph{does} affect the target logit, and thus can mitigate over-confidence in the top-scoring class.
On the other hand, ``tail'' logits receive sharply reduced signal compared to the ListMLE.
% which may prove undesirable when $L$ is large.

% Second,
% as $\Omega( \cdot )$ is instantiated with the canonical ranking $\piCanonical^{( y )}$,
% the regulariser applies to the \emph{same} set of non-target classes 
% for all training samples sharing the same label.
% On the one hand, this 
% % is expected, as it is concordance with the requirement that we respect the canonical ranking (which only depends on the label).
% offers a \emph{robust} signal,
% as aggregating class-similarities across all training samples can perform a certain de-noising. 
% On the other hand, this introduces a potential risk: 
% by forcing the logits for \emph{every} training sample belonging to a given class to follow the same pattern, one may overly suppress rich \emph{instance-dependent} structure, and thus negatively impact the classification performance.
% In fact, the latter is not a hypothetical risk:
% in practice,
% as we demonstrate in~\S\ref{sec:experiments}, 
% the impressive OOD detection performance of {\sc RankOOD} often comes at the expense of regression on the regular ID test accuracy.

% In response to the above, one natural recourse is to revert to standard self-distillation,
% which
% completely eschews both modifications in {\sc RankOOD}.
% However, one may naturally ask whether \emph{combining} insights from both methods has value.
Overall, the available design choices offer distinct trade-offs on both the supervision and loss axes.

%% file: sec/4_method.tex
\section{Unifying ranking- \& KD-based OOD detection via the PL-KD loss}
\label{sec:method}

% We now establish that the KD and ListMLE losses can be interpreted as particular approximations to a more general loss.
% This suggests a novel \emph{PL-KD loss} that combines the strengths of each.
We now unify the KD and ListMLE losses via a certain \emph{expected ranking} loss that we now describe.

\subsection{The expected Plackett-Luce (EPL) loss}
\label{sec:epl}

Fundamentally, the ListMLE~\eqref{eqn:pl-loss} and KD loss~\eqref{eqn:kd} operate on different objects:
a \emph{ranking} and \emph{distribution} over classes, respectively.
% We now provide a unified means to reason about both losses.
To bridge these,
given an instance $x \in \XCal$ and teacher distribution $\mathbf{p}^{\rm t}( x ) \in \Delta( \YCal )$ over classes,
define the \emph{induced ranking distribution} 
${p}^{\rm t}_{\rm pl}( \cdot \mid x ) \in \Delta( \mathfrak{S}( \YCal ) )$
by:
% \todoakm{Write in terms of logits?}
\begin{equation}
    \label{eqn:induced-ranking}
    ( \forall \boldsymbol{\pi} \in \mathfrak{S}( \YCal ) ) \, {p}^{\rm t}_{\rm pl}( \pi \mid x ) \defEq \prod_{i = 1}^{L} \frac{{p}^{\rm t}_{\pi( i )}( x )}{\sum_{j = i}^{L} {p}^{\rm t}_{\pi( j )}( x )}.  
\end{equation}
This is the \emph{Plackett-Luce (PL)} distribution~\citep{Plackett:1975,Luce:1959} with the class probabilities 
% ${p}^{\rm t}( \cdot \mid x )$ 
$\mathbf{p}^{\rm t}( x )$ 
as utilities.
Given student 
logits $\mathbf{f}( x )$ with probabilities $\mathbf{p}( x )$, we may similarly define the student's induced ranking distribution ${p}_{\rm pl}( \cdot \mid x )$. 
Equipped with this, define the \emph{expected PL (EPL) loss} as:
\begin{equation}
    \label{eqn:expected-ranking-loss}
    \ell_{\rm epl}( \mathbf{p}^{\rm t}( x ), \mathbf{f}( x ) ) \defEq \mathop{\mathbb{E}}_{ \boldsymbol{\pi} \sim {p}^{\rm t}_{\rm pl}( \cdot \mid x ) }\left[ -\log p_{\rm pl}( \boldsymbol{\pi} \mid x ) \right].
\end{equation}
% i.e., the average negative log-likelihood the student assigns to rankings drawn from the teacher.

In general, it is intractable to compute~\eqref{eqn:expected-ranking-loss} exactly, since ${p}^{\rm t}_{\rm pl}( \cdot \mid x )$ is defined over a set
% $\mathfrak{S}( \YCal )$
with $L!$ elements.
Interestingly, both instance-level {\sc RankOOD}~(\S\ref{sec:ranking-family}) and self-KD~\eqref{eqn:kd} can be seen as tractable (but imperfect) approximations to~\eqref{eqn:expected-ranking-loss}.

\subsection{KD and {\sc RankOOD} as EPL approximations}
\label{sec:unified}

\textbf{{\sc RankOOD} approximation}.
The simplest approximation to~\eqref{eqn:expected-ranking-loss} is to draw a \emph{single} ranking $\boldsymbol{\pi}$ from ${p}^{\rm t}_{\rm pl}( \cdot \mid x )$, and then minimise $-\log p_{\rm pl}( \boldsymbol{\pi} \mid x )$.
An elementary calculation reveals that this exactly yields the ListMLE loss~\eqref{eqn:pl-loss}.
One may further establish that the mode of ${p}^{\rm t}_{\rm pl}( \cdot \mid x )$ is exactly the ranking $\boldsymbol{\pi}^{\rm t}( x ) \defEq {\rm argsort}( \mathbf{p}^{\rm t}( x ) )$; i.e.,
one simply ranks the classes according to their associated class-probabilities.
With this choice, we exactly recover the \emph{instance-level {\sc RankOOD}} loss from~\S\ref{sec:canonical-kd}.
% \todoakm{We have to introduce this beforehand}

%
\textbf{KD approximation}.
The ListMLE loss with the ranking 
$\boldsymbol{\pi}^{\rm t}$ uses the teacher's class probabilities to guide the student;
however, unlike the KD loss, it does not explicitly consider the \emph{strength} of the teacher's class probabilities.
Interestingly, the KD loss 
$\ell_{\rm kd}$
may also be seen as a distinct approximation to $\ell_{\rm epl}$~\eqref{eqn:expected-ranking-loss}.
(See also Appendix~\ref{app:theory_extended} for a similar bound relating the KD and {\sc RankOOD} approximations.)
% In fact, we have the following two-sided bound relating the two losses.

%

\begin{proposition}
    \label{prop:sandwich-bound}
    Pick any $x \in \XCal$.
    Suppose the
    teacher distribution $\mathbf{p}^{\rm t} \colon \XCal \to \Delta( \YCal )$ satisfies 
    $ p^{\rm t}_{\rm min}( x ) \defEq \min_{y' \in \YCal} p^{\rm t}_{y'}( x ) > 0. $
    Then, for any $\mathbf{f}( x ) \in \Real^L$,
    % the KD~\eqref{eqn:kd} and EPL~\eqref{eqn:expected-ranking-loss} losses satisfy:
    % $$ \ell_{\rm kd}( \mathbf{p}^{\rm t}( x ), \mathbf{f}( x ) ) \leq \ell_{\rm epl}( \mathbf{p}^{\rm t}( x ), \mathbf{f}( x ) ) \leq ( {p^{\rm t}_{\rm min}( x )} )^{-1} \cdot \ell_{\rm kd}( \mathbf{p}^{\rm t}( x ), \mathbf{f}( x ) ). $$
    $$ \ell_{\rm kd}( \mathbf{p}^{\rm t}( x ), \mathbf{f}( x ) ) \leq \ell_{\rm epl}( \mathbf{p}^{\rm t}( x ), \mathbf{f}( x ) ) \leq \frac {\ell_{\rm kd}( \mathbf{p}^{\rm t}( x ), \mathbf{f}( x ) )}{  {p^{\rm t}_{\rm min}( x )} }. $$
\end{proposition}

% \textcolor{blue}{Suggested fix to prevent overflow}
% \begin{align*}
%     \ell_{\rm kd}( \mathbf{p}^{\rm t}( x ), \mathbf{f}( x ) )
%     &\leq \ell_{\rm epl}( \mathbf{p}^{\rm t}( x ), \mathbf{f}( x ) ) \\
%     &\leq ( p^{\rm t}_{\rm min}( x ) )^{-1} \cdot \ell_{\rm kd}( \mathbf{p}^{\rm t}( x ), \mathbf{f}( x ) ).
% \end{align*}

Proposition~\ref{prop:sandwich-bound} has two implications.
First,
the KD loss against the teacher's \emph{class} probabilities is \emph{weaker} than the EPL loss against its induced \emph{ranking} probabilities~\eqref{eqn:induced-ranking}, in that minimising the latter also minimises an upper bound on the former.
% the latter imposes an additional non-negative penalty compared to the former.
Second, 
when $p^{\rm t}_{\rm min}( x ) \gg 0$, 
a \emph{scaled} KD loss ostensibly controls the EPL loss.
However, $( p^{\rm t}_{\rm min}( x ) )^{-1} \geq L$, and so the KD loss is at least a factor of $L$ away from~\eqref{eqn:expected-ranking-loss}.
This is intuitive: 
$( p^{\rm t}_{\rm min}( x ) )^{-1} \sim L$ when the teacher's distribution is \emph{near uniform}, 
so the precise class probabilities convey roughly as much information as their {ranking}.

% as with Proposition~\ref{prop:kd-pl-bound},
% one can ostensibly control the EPL loss with a \emph{scaled} KD loss;
% however,
% when $p^{\rm t}_{\rm min}( x ) \ll 1$, the scaling factor sharply increases, further indicating KD is potentially a weak surrogate for the EPL loss.

% Notably,
% it does \emph{not} offer a lower bound on 
% $\ell_{\rm pl}( \boldsymbol{\pi}^{\rm t}( x ), \mathbf{f}( x ) )$ in terms of the KD loss.
% This is intuitive:
% since the standard ListMLE loss 
% against $\boldsymbol{\pi}^{\rm t}$
% does not account for the teacher's \emph{confidences},
% it does not incentivise mimicking them.

% Put together, the above suggests that the 
% KD loss is ``weaker'' than the
% EPL loss~\eqref{eqn:expected-ranking-loss},
% while the ListMLE loss is a high-variance estimator of it.
% % is ``stronger'' than the KD loss;
% % while also being the population analogue of the ListMLE loss.
% However, computing this loss exactly remains challenging.
% We now provide a step towards resolving this issue, which unifies aspects of both approximations.

%%%
\subsection{PL-KD: combining KD and {\sc RankOOD}}
\label{sec:pl-kd}

Both approximations above miss central aspects of the EPL loss:
the ListMLE approximation eschews the teacher confidences,
while the KD loss omits the ranking penalties.
This motivates
an alternate expression for the EPL loss. 
% which coherently combines the ranking 
% and teacher-weighting
% aspects of the ListMLE and KD loss respectively.
% which combines aspects of both the ListMLE and KD loss.

%
\begin{proposition}
    \label{prop:epl-loss}
    For any $x \in \XCal$,
    ranking $\boldsymbol{\pi} \in \mathfrak{S}( \YCal )$, 
    and $i \in [ L ]$, let $\boldsymbol{\pi}_{<i} \defEq ( \pi( 1 ), \ldots, \pi( i - 1 ) )$.
    For any teacher distribution $\mathbf{p}^{\rm t} \colon \XCal \to \Delta( \YCal )$, define the \emph{conditional ranking} 
    % distribution
{%\small
\begin{equation}
    \label{eqn:conditional-ranking}
    % ( \forall y \in \YCal ) \, 
    p^{\rm t}_{y}( x, \boldsymbol{\pi}_{< i} ) = \indicator{ y \notin \boldsymbol{\pi}_{<i} } \cdot \frac{p^{\rm t}_{y}( x )}{1 - \sum_{j = 1}^{i - 1} p^{\rm t}_{{\pi}( j )}( x )},
\end{equation}}
    and similarly for the student distribution $\mathbf{p} \colon \XCal \to \Delta( \YCal )$.
    Then, the EPL loss~\eqref{eqn:expected-ranking-loss} is equally:
% \todoakm{more digestible notation}
%\begin{empheq}[box=\subtlebox]{equation}
    %\label{eqn:refined-estimator}
    %\ell_{\rm epl}( \mathbf{p}^{\rm t}( x ), \mathbf{f}( x ) ) = \mathop{\mathbb{E}}_{ \boldsymbol{\pi} \sim {p}^{\rm t}_{\rm pl}( \cdot \mid x ) }\left[ 
%\sum_{i = 1}^{L} \ell_{\rm kd}\left( \mathbf{p}^{\rm t}( x, \boldsymbol{\pi}_{< i} ), 
%\mathbf{p}( x, \boldsymbol{\pi}_{< i} ) \right)
% \sum_{y' \notin \boldsymbol{\pi}_{<i}}   \cdot -\log \frac{p( y' \mid x )}{1 - \sum_{j = 1}^{i - 1} p( {\pi}( j ) \mid x )}
%\right].    
%\end{empheq}
%fix for overflow
\begin{empheq}[box=\subtlebox]{align}
    & \ell_{\rm epl}( \mathbf{p}^{\rm t}( x ), \mathbf{f}( x ) ) \nonumber \\
    & = \mathop{\mathbb{E}}_{ \boldsymbol{\pi} \sim p^{\rm t}_{\rm pl}( \cdot \mid x ) } \bigg[ \sum_{i = 1}^{L} \ell_{\rm kd}\big( \mathbf{p}^{\rm t}( x, \boldsymbol{\pi}_{< i} ), \mathbf{p}( x, \boldsymbol{\pi}_{< i} ) \big) \bigg].
    \label{eqn:refined-estimator}
\end{empheq}

\end{proposition}

% Compared to, the key distinction is that we now employ a series of \emph{weighted} ListMLE losses.
In words, one draws a ranking $\boldsymbol{\pi}$ as with the standard EPL loss~\eqref{eqn:expected-ranking-loss};
however, at \emph{each} position in the ranking, the corresponding loss term considers the induced conditional probability over \emph{all} possible elements for that position, not just $\pi( i )$.
In practice, we may approximate the above by drawing a single ranking $\boldsymbol{\pi}$, and minimising the sum of KD losses in~\eqref{eqn:refined-estimator}.
We term this the \emph{PL-KD} loss.
% \todoakm{consider alternate names}

PL-KD can be seen as combining the strengths of KD (i.e., the use of teacher soft confidences), and ListMLE (i.e., enforcing a ranking over the classes).
% \todoakm{explain this more}
Compared to the conventional ListMLE, each term in~\eqref{eqn:refined-estimator} employs the teacher's soft probability.
On the other hand, unlike the conventional KD loss 
---
corresponding to the first term in the summation in~\eqref{eqn:refined-estimator}
---
we consider a \emph{nested} sequence of losses, which more strongly penalises deviation from the ranking in $\boldsymbol{\pi}$.
% \todoakm{tie back to instance-level ListMLE}

As in~\S\ref{sec:ranking-family},
we term~\eqref{eqn:refined-estimator} as \emph{Instance-level PL-KD} with instance-level $\mathbf{p}^{\rm t}(x)$ 
supervision.
One may replace the \emph{instance-level} $\mathbf{p}^{\rm t}( x )$ 
% in~\eqref{eqn:refined-estimator}
with a \emph{class-level} $\bar{\mathbf{p}}^{( y )}$ (\emph{viz.} the average of $\mathbf{p}^{\rm t}( x )$ over all training samples with class $y$).
The resulting \emph{class-level PL-KD} extends the class-level KD in~\S\ref{sec:ranking-family} to more strictly account for ranking structure.
In practice, we favour instance-level PL-KD when the teacher deviations within a class are reflective of inherent uncertainty; {Appendix~\ref{app:bias-variance} formalises this, based on~\citet{Menon:2021}.}

Finally, as with standard KD, temperature scaling of teacher (and optionally student) logits is important for final performance.
In the low-temperature regime $T \to 0^+$, PL-KD reduces to instance-level {\sc RankOOD}.
For $T \gg 0$, the loss amplifies the teacher's relative ranking over classes.
Ideally, the temperature should yield the teacher best approximating the underlying data-generating distribution~\citep{Menon:2021}.

%
% \textcolor{olive}{
\textbf{PL-KD versus {\sc RankOOD} and KD}.
We may situate PL-KD on the two axes used in~\S\ref{sec:contrast-rankood-kd} to contrast {\sc RankOOD} and self-KD.
On the supervision granularity axis,
compared to class-level methods (e.g., {\sc RankOOD}),
instance-level PL-KD potentially offers benefits on examples with an \emph{atypical} non-target ranking (e.g., the Afghan Hound example in~\S\ref{sec:intro}).
On the loss function axis,
compared to the standard ListMLE loss, PL-KD 
\emph{softens} the penalty on ranking swaps between classes that are nearly identical under the teacher.
On the other hand, compared to the standard KD loss, PL-KD \emph{strengthens} the penalty on ranking swaps between 
classes with low probability under the teacher.
% classes that are distinct under the teacher.
Appendix~\ref{sec:worked-example} provides a worked example that explicates these points.
% }

%
\textbf{Relation to existing work}.
Proposition~\ref{prop:sandwich-bound} is, to our best knowledge, novel;
however, Appendix~\ref{sec:related_extended} situates this in the broader fundamental literature on \emph{class-probability estimation} and \emph{ranking}~\citep{Narasimhan:2013}.
Proposition~\ref{prop:epl-loss} mirrors expressions for \emph{sequence-} versus \emph{token-level} KL divergences~\citep{Kim:2016}, which have been established in the distinct {reinforcement learning} literature~\citep{amini2026better}.
However, to our best knowledge, its instantiation 
to label ranking settings 
--- and further application to OOD detection --- 
are novel.
In \emph{retrieval} settings,~\citet{Reddi:2021} considered a top-$k$ analogue of EPL, but not the PL-KD form we propose.
See
Appendix~\ref{sec:related_extended} for more discussion. 
% also discusses prior distinct (but broadly related) effort on combining distillation and ranking for \emph{retrieval} settings~\citep{Reddi:2021}.
% However, the latter identity 
% In Appendix~\ref{sec:related_extended}, we situate these results in the broader literature on ranking and distillation.
% \todoakm{Discuss sequence-level KD}

The idea of using self-KD is hardly new in the context of OOD detection~\citep{Wu:2023,Tang:2025,Yang:2025,Zhao:2025,Wang:2023,Kralik:2025}.
However, to our best knowledge, none of these works modify the core \emph{loss function}, unlike PL-KD which infuses KD with ranking information;
several of these existing ideas are potentially complementary, such as matching intermediate representations.
Further, these works do not offer a theoretical grounding for the value of KD, which is our next contribution.

% At this stage, we have arrived at a unification of the KD and {\sc RankOOD} approaches to OOD detection.
% As we shall see (\S\ref{sec:experiments}), the PL-KD loss performs favourably empirically in terms of both OOD detection performance and ID test accuracy.

%
% \subsection{Discussion and relation to existing work}

% One may adapt such losses in the present context, which could be an interesting direction for future work.
% \todoakm{mention the refined estimator}

%% file: sec/5_theory.tex
% \section{The blessing of dark knowledge for OOD detection}
\section{Dark knowledge can help OOD detection}
\label{sec:theory}

% Relating ranking-based OOD detection to KD has yielded an algorithmic benefit, in the form of the new PL-KD loss.
% % Having yielded a novel algorithmic 
% % We now provide a step 
% We now 
% % employ the connection
% % \todoakm{Smoother segue}
% show that this yields a conceptual benefit,
% % to make a step
% by enabling a step
% towards theoretically grounding two empirical phenomena: 
% that have been repeatedly 
% that we previously
% alluded to:
We now show that
relating ranking-based OOD detection to KD yields a conceptual benefit,
% to make a step
by enabling 
theoretical progress in grounding two empirical phenomena: 
(i)
OOD samples, 
despite having high \emph{target class} scores,
have a distinct \emph{non-target class} logit
structure;
% (as evidenced by the success of post-hoc techniques~\citep{Karunanayake:2025});
(ii) 
modifying the training loss 
% from softmax cross-entropy 
to strengthen such non-target class structure
can improve OOD detection performance.
% (as evidenced by the success of training techniques~\citep{Denipitiyage:2026}).
Formalising these requires assumptions on the data, 
which we now describe.
% on the nature of ID and OOD samples.
% We now formalise and justify assumptions on the samples' \emph{feature representations}.
% We shall make such assumptions on the \emph{feature representations} of these samples, and argue for their plausibility.

% We now establish that knowledge distillation can better detect OOD samples than standard softmax CE, by considering the structure of the {non-target} classes.
% To establish this, the basic idea is to show that 
% while standard CE and KD training can lead to a similar MSP score,
% standard CE training can lead to a higher entropy distribution over non-target classes.

%
\subsection{A generative model for ID and OOD samples}

% We consider the following OOD scoring rule, which we term {\tt EntMSP}:
% $$ s_{\rm EntMSP}( x ) = p_{[1]}( x ) - \frac{1}{\log (L - 1)} \cdot H\left( \left\{ \frac{p_{[2]}( x )}{1 - p_{[1]}( x )}, \ldots, \frac{p_{[L]}( x )}{1 - p_{[1]}( x )} \right\} \right), $$
% where $H( \cdot )$ denotes the usual entropy, normalised to live in $[0, 1]$ for a commensurate scale to $p_{[1]}( x )$.
% Intuitively, this rule deems a sample to be OOD if its MSP is low,
% \emph{and} its entropy on the non-target classes is high.

% We proceed thusly.
Recall that neural classifiers have
logits
$\mathbf{f}( x ) = \mathbf{W}^\top \boldsymbol{\Phi}( x )$,
for
weights
$\mathbf{W} = \begin{bmatrix} \mathbf{w}_1 & \ldots & \mathbf{w}_L \end{bmatrix}
\in \Real^{D \times L}$,
and feature extractor
$\boldsymbol{\Phi} \colon \XCal \to \Real^D$.
% In the following,
% let $\mathbf{W} = \begin{bmatrix} \mathbf{w}_1 & \ldots & \mathbf{w}_L \end{bmatrix}
% \in \Real^{D \times L}$ be the classification weights learned by minimising a certain training loss (e.g., softmax cross-entropy).
% Let $\boldsymbol{\Phi}( x ) \in \Real^D$ denote the feature extractor, so that the logit vector $\mathbf{f}( x ) = \mathbf{W}^\top \boldsymbol{\Phi}( x ) \in \Real^L$.
We posit a generative model
% for ID and OOD samples,
building on~\citet{Ming:2022}.
For each ID class $i \in [ L ]$, 
we assume
the mean
$\boldsymbol{\mu}^{(i)} \defEq \mathop{\mathbb{E}}_{x \sim \Pr_{\rm id}( \cdot \mid i )}[ \boldsymbol{\Phi}( x ) ]$
decomposes
into three
(suitably ``independent'')
feature types:
\emph{background},
\emph{local},
\emph{global} features 
$\boldsymbol{\mu}^{(i)}_{\sharedFeat}$,
$\boldsymbol{\mu}^{(i)}_{\locFeat}$,
$\boldsymbol{\mu}^{(i, j)}_{\globFeat}$
for each class $j \neq i$.
The latter encodes \emph{semantic relationships} between classes.

\begin{assumption}[ID feature structure]
\label{assum:id-feat}
For any 
ID class $i \in [ L ]$ and $x \sim \Pr_{\rm id}( \cdot \mid y = i )$,
% and test example
% $x \sim \Pr_{\rm id}( \cdot \mid y = i )$,
% the feature representation 
$\boldsymbol{\Phi}( x )$
% is a Gaussian, with specific structure in the mean:
% comprises
% background, local, and global features:
decomposes as follows:
for
$t_i \in \Real$,
$\{ s_{ij} \in \Real \colon i, j \in [ L ] \}$,
and 
$\sigma_{\locFeat}, \sigma_{\globFeat} \geq 0$,
$\sigma_{\nullFeat} > 0$,
{\small $$
    \boldsymbol{\Phi}( x ) = 
    t_{i}\cdot \boldsymbol{\mu}^{(i)}_{\sharedFeat} +
    \alpha^{(i)}_{\locFeat} \cdot s_{ii} \cdot \boldsymbol{\mu}^{(i)}_{\locFeat} + \sum_{j \neq i} 
    \alpha^{(i, j)}_{\globFeat} \cdot s_{ij} \cdot \boldsymbol{\mu}^{(i, j)}_{\globFeat} + 
    \boldsymbol{\epsilon}_{\nullFeat},
$$}
for
$    \alpha^{(i)}_{\locFeat} 
    \sim 
    % \sim 
    \mathscr{N}( 1, \sigma^2_{\locFeat} )$,
    % \\
    $\alpha^{(i, j)}_{\globFeat} 
    % &\sim 
    \sim 
    \mathscr{N}( 1, \sigma^2_{\globFeat} )$,
    % \\
    $\boldsymbol{\epsilon}_{\nullFeat} 
    % &\sim 
    \sim 
    \mathscr{N}( 0, \sigma_{\nullFeat}^2 \mathbf{I} ).$
% We further assume symmetry: 
Further,
$( \forall i, j ) \, s_{ij} = s_{ji}$,
$\boldsymbol{\mu}^{(i, j)}_{\globFeat} = \boldsymbol{\mu}^{(j, i)}_{\globFeat}$.
\end{assumption}

For OOD samples, we posit that
the features 
spuriously align
with the local features for a single \emph{latent} class $z$
---
chosen uniformly from all ID classes
---
but \emph{not} with the global features.
Intuitively, this means that OOD samples miss characteristics of semantically related classes.
% Further, we assume that the OOD samples have
% lower norms than ID samples~\citep{Wei:2022}.
Formally:

\begin{assumption}[OOD feature structure]
\label{assum:ood-feat}
% For OOD samples, 
% $\bar{x} \sim \Pr_{\rm ood}$, 
% The OOD distribution $\Pr_{\rm ood}$ decomposes into a mixture 
For \emph{latent} class
$z \sim {\rm Uniform}([ L ])$,
OOD samples
$\bar{x} \sim \Pr_{\rm ood}( \cdot \mid z )$ satisfy
$$\boldsymbol{\Phi}( \bar{x} ) = \alpha^{(z)}_{\locFeat} \cdot s_{zz} \cdot  \boldsymbol{\mu}^{(z)}_{\rm ood} + \boldsymbol{\epsilon}_{\nullFeat},$$
where
$\boldsymbol{\mu}^{(z)}_{\rm ood} \defEq \beta \cdot \boldsymbol{\mu}^{(z)}_{\locFeat}$
and $\beta \in ( 0, 1 )$.
% \todoakm{Clarify conditioning notation for OOD samples}
% \begin{align*}
%     \boldsymbol{\Phi}( \bar{x} ) &= \alpha^{(z)}_{\locFeat} \cdot s_{zz} \cdot  \boldsymbol{\mu}^{(z)}_{\rm ood} + \boldsymbol{\epsilon}_{\nullFeat} 
%     %\\
%     \, \text{ where } \,
%     \boldsymbol{\mu}^{(z)}_{\rm ood} \defEq \beta \cdot \boldsymbol{\mu}^{(z)}_{\locFeat}.
%     % z &\sim {\rm Uniform}([ L ]) \\
%     % \boldsymbol{\Phi}( \bar{x} ) \mid k &\sim \mathscr{N}( \boldsymbol{\mu}^{(k)}_{\rm ood}, \boldsymbol{\Sigma}^{(k)}_{\rm ood} ) \\
%     % \boldsymbol{\Phi}( \bar{x} ) \mid z &\sim \mathscr{N}( \boldsymbol{\mu}^{(z)}_{\rm ood}, \boldsymbol{\Sigma}^{(z)}_{\rm ood} ) \\
%     % \\
%     % \boldsymbol{\Sigma}^{(k)}_{\rm ood} &= \gamma \cdot \boldsymbol{\Sigma}^{(k)}_{\locFeat} + \sigma^2_{\rm ood} \mathbf{I},
%     % \\ 
%     % + \mathbf{n}_{\rm ood}( \bar{x} ).
%     % \boldsymbol{\Sigma}_{\rm ood}( \bar{x} ) &= \boldsymbol{\Sigma}_{\nullFeat}^{(i)}.
% \end{align*}
% where $\boldsymbol{\Sigma}_{\rm ood}( \bar{x} )$ lives in the nullspace of $\mathbf{W}^\top$.
% for
% noise scale
% $\sigma_{\rm ood} > 0$,
% and
% scaling factors
% $\beta, \gamma \in ( 0, 1 )$.
\end{assumption}

% %
% \begin{assumption}[Balanced feature means]
% For every $i, j \in [ L ]$,
% $ \| \boldsymbol{\mu}^{(i)} \|_2^2 = \| \boldsymbol{\mu}^{(j)} \|_2^2. $
% \end{assumption}

%
\begin{assumption}[Feature orthonormality]
\label{assum:loc-feat-orth}
The ID feature means are mutually orthogonal (orthonormal) across (within) types 
(e.g., $( \boldsymbol{\mu}_{\locFeat}^{(i)} )^\top \boldsymbol{\mu}_{\sharedFeat}^{(j)} = 0$,
$( \boldsymbol{\mu}_{\locFeat}^{(i)} )^\top \boldsymbol{\mu}_{\locFeat}^{(j)} = \indicator{ i = j }$).
% and orthonormal within types
% (e.g., $( \boldsymbol{\mu}_{\locFeat}^{(i)} )^\top \boldsymbol{\mu}_{\locFeat}^{(j)} = \indicator{ i = j }$).
% For every $i, j, k, k' \in [ L ]$,
% the features are orthonormal:
% $( \boldsymbol{\mu}_{\sharedFeat}^{(i)} )^\top \boldsymbol{\mu}_{\sharedFeat}^{(j)} = \indicator{ i = j }$,
% $( \boldsymbol{\mu}_{\locFeat}^{(i)} )^\top \boldsymbol{\mu}_{\locFeat}^{(j)} = \indicator{ i = j }$,
% and
% $( \boldsymbol{\mu}_{\globFeat}^{(i, k')} )^\top \boldsymbol{\mu}_{\globFeat}^{(j, k)} = \indicator{ (i, k') = (j, k) \lor ( i, k' ) = ( k, j ) }$;
% and mutually orthogonal:
% $( \boldsymbol{\mu}_{\locFeat}^{(i)} )^\top \boldsymbol{\mu}_{\globFeat}^{(j, k)} = 0$,
% and
% $( \boldsymbol{\mu}_{\locFeat}^{(i)} )^\top \boldsymbol{\mu}_{\sharedFeat}^{(j)} = 0$, and
% $( \boldsymbol{\mu}_{\globFeat}^{(i, j)} )^\top \boldsymbol{\mu}_{\sharedFeat}^{(k)} = 0$.
% $\| \boldsymbol{\mu}_{\locFeat}^{(i)} \|_2 = 1$ and $\| \boldsymbol{\mu}_{\globFeat}^{(j, k)} \|_2 = 1$.
% $ \mathbf{w}_i^\top \boldsymbol{\Phi}^{(j)}_{\locFeat} = 0. $
\end{assumption}

\begin{assumption}[Balanced classes]
\label{assum:balanced-class}
The ID classes are balanced, so that $( \forall i \in [ L ] ) \, \Pr_{\rm id}( y = i ) = \frac{1}{L}$.
% This setting captures many common image classification benchmarks.
% We may thus regard ID samples as drawn from a uniformly weighted mixture of class-conditionals, namely,
% $\frac{1}{L} \sum_{i = 1}^{L} \Pr_{\rm id}( \cdot \mid y = i )$.
\end{assumption}

Assumptions~\ref{assum:id-feat} -- \ref{assum:balanced-class}
aim to capture stylised characteristics that \emph{broadly} reflect practice.
We make a further simplifying assumption in the body.
% we present below a \emph{special case} of our result, which holds under .
(Appendix~\ref{app:theory_formal} avoids this, at the cost of notational burden and a first-order approximation.)

% %
% \begin{assumption}[Special case]
% \label{assum:simplify}
% The ID class self-similarities are constant 
% (i.e., 
% $t_{i} = t_0 \gg 1$,
% $s_{ii} = 1$ for every $i \in [ L ]$),
% and
% strong (i.e., $\lambda, \sigma_{\nullFeat}, s_{ij} \ll 1$ for $j \neq i$).
% Further, the number of classes $L \gg 1$.
% \end{assumption}

%
\begin{assumption}[Special case]
\label{assum:simplify}
The feature noise is isotropic ($\sigma_{\locFeat} = \sigma_{\globFeat} = 0$), and all class mean norms $\| \boldsymbol{\mu}^{(i)} \|_2$ are equal.
\end{assumption}

%%%%%%%%%
% \subsection{On the success of post-hoc non-target class signature OOD detection}
\subsection{On the viability of post-hoc OOD detection}

% We now study the viability of post-hoc OOD detection based on the non-target logit signature under the above generative model.
% That is, 
Suppose we train a classifier on ID training samples from the above distribution.
Do the logit vectors for ID and OOD samples exhibit different structure?
To study this, 
% we first pin down the structure $\mathbf{W}$.
following prior self-KD analyses~\citep{Phuong:2019,Mobahi:2020}, we 
consider \emph{regularised square loss} minimisation
with $L_2$-regularisation
strength $\lambda > 0$,
i.e.,
$R_{\rm sq}( \mathbf{W} ) \defEq \mathop{\mathbb{E}}_{( x, y ) \sim \Pr_{\rm id}}[ \ell_{\rm sq}( y, \mathbf{f}( x; \mathbf{W} ) ) ] + \lambda \cdot \| \mathbf{W} \|_F^2$
for
$\ell_{\rm sq}( y, \mathbf{f}( x ) ) = ( f_y( x ) - 1 )^2 + \sum_{y' \neq y} f_{y'}( x )^2$.

% \begin{equation}
%     \label{eqn:square-loss-risk}
%     R_{\rm sq}( \mathbf{W} ) \defEq \mathop{\mathbb{E}}_{( x, y ) \sim \Pr_{\rm id}}[ \ell_{\rm sq}( y, \mathbf{f}( x; \mathbf{W} ) ) ] + \lambda \cdot \| \mathbf{W} \|_F^2.
% \end{equation}
% Via standard results for regularised linear regression,
% the minimiser
% $\mathbf{W}^*$ of~\eqref{eqn:square-loss-risk} has an analytic form
% (see Appendix~\ref{app:helper}).
% Based on the well-established
% analytic form for 
% the minimiser
% $\mathbf{W}^*$ 
% % of~\eqref{eqn:square-loss-risk}
% ---
% see Appendix~\ref{app:helper}
% ---
% we may derive analytic expressions for the (expected) logit vectors for both ID and OOD samples.
% For clarity, we present below an \emph{informal} version of our result, which holds under additional simplifying assumptions.
% (See Appendix~\ref{app:theory_formal} for a formal statement which does not require Assumption~\ref{assum:simplify}, at the cost of greater notational burden.)
We measure how well a score $g \colon \XCal \to \Real$ separates ID from OOD samples of (latent) class $y$ by its \emph{separation}
% $$ \delta_y( g ) \defEq 
% \frac{ \underset{x \sim \Pr_{\rm id}( \cdot \mid y )}{\mathop{\mathbb{E}}}[ g( x ) ] - \underset{\bar{x} \sim \Pr_{\rm ood}( \cdot \mid y )}{\mathop{\mathbb{E}}}[ g( \bar{x} ) ] }{ \left( \underset{x \sim \Pr_{\rm id}( \cdot \mid y )}{\mathop{\mathbb{V}}}[ g( x ) ] \right)^{1/2} }. $$

\noindent\resizebox{\linewidth}{!}{$\displaystyle \delta_y( g ) \defEq 
\left( \underset{x \sim \Pr_{\rm id}( \cdot \mid y )}{\mathop{\mathbb{V}}}[ g( x ) ] \right)^{-1/2} \cdot \left[ { \underset{x \sim \Pr_{\rm id}( \cdot \mid y )}{\mathop{\mathbb{E}}}[ g( x ) ] - \underset{\bar{x} \sim \Pr_{\rm ood}( \cdot \mid y )}{\mathop{\mathbb{E}}}[ g( \bar{x} ) ] } \right]. $}

For $g( x ) = \mathbf{c}^\top \mathbf{f}( x )$ (e.g., logit differences), the ID and OOD variances coincide (Lemma~\ref{lemm:logit-cov}).
The numerator of $\delta_y( \mathbf{c}^\top \mathbf{f}( x ) )$ can be written in terms of 
% the
% \emph{teacher logits}
$ \mathbf{F} \defEq ( L \cdot \mathbf{I} + \mathbf{K} )^{-1} \mathbf{K}, $
where $ \mathbf{K} \defEq \tau^{-1} \cdot \mathbf{M}^\top \mathbf{M} $ is the (regularised) \emph{class-similarity matrix} for 
class means
$\mathbf{M} \defEq \begin{bmatrix} \boldsymbol{\mu}^{(1)} & \ldots & \boldsymbol{\mu}^{(L)} \end{bmatrix}$,
and net regularisation strength
$\tau \defEq \sigma^2_{\nullFeat} + \lambda$.
% and $N_{yy} \defEq \tau^{-1} \cdot s^2_{yy}$.
To first-order
(Proposition~\ref{corr:opt-logits-first-order}),
$F_{y,y} = \frac{K_{y,y}}{L + K_{y,y}}$ and $F_{y',y} \propto s^2_{y',y}$ for $y \neq y'$.

\begin{proposition}
\label{corr:opt-logits-first-order-informal}
Suppose $\mathbf{W}^*$ minimises 
$R_{\rm sq}$.
% the square loss population risk.
% ~\eqref{eqn:square-loss-risk}.
Under Assumptions~\ref{assum:id-feat}--\ref{assum:simplify}, 
for any $y \in [ L ]$,
the expected logits are
%fix for overflow
\begin{align*}
    \mathop{\mathbb{E}}_{x \sim \Pr_{\rm id}( \cdot \mid y )}[ \mathbf{f}( x; \mathbf{W}^* ) ]
    &= \mathbf{F} \mathbf{e}_y
    \\
    \mathop{\mathbb{E}}_{\bar{x} \sim \Pr_{\rm ood}( \cdot \mid y )}[ \mathbf{f}( \bar{x}; \mathbf{W}^* ) ]
    &= \beta \cdot {(L \cdot \tau)^{-1}} \cdot s^{2}_{yy} \cdot ( \mathbf{I} - \mathbf{F} ) \mathbf{e}_y.
\end{align*}
\end{proposition}
% %
% \begin{proposition}[informal]
% \label{corr:opt-logits-first-order-informal}
% Suppose $\mathbf{W}^*$ is the optimal square loss population risk minimiser.
% % ~\eqref{eqn:square-loss-risk}.
% Under Assumptions~\ref{assum:id-feat}--~\ref{assum:simplify}, 
% for any $y, y' \in [ L ]$,
% the expected logits for ID and OOD samples satisfy
% %fix for overflow
% \begin{align*}
%     \mathop{\mathbb{E}}_{x \sim \Pr_{\rm id}( \cdot \mid y )}[ f_{y'}( x; \mathbf{W}^* ) ]
%     &\propto 1 \cdot \indicator{y = y'} + \frac{1}{t_0^4} \cdot s^2_{y', y} \cdot \indicator{y \neq y'} \\
%     \mathop{\mathbb{E}}_{x \sim \Pr_{\rm ood}( \cdot \mid y )}[ f_{y'}( x; \mathbf{W}^* ) ]
%     &\propto \beta \cdot \bigg( \frac{1}{t_0^2} \cdot \indicator{y = y'} \\
%     &\qquad - \frac{1}{t_0^4} \cdot s^2_{y', y} \cdot \indicator{y \neq y'} \bigg).
% \end{align*}
%\begin{align*}
    %\mathop{\mathbb{E}}_{x \sim \Pr_{\rm id}( \cdot \mid y )}[ {f}_{y'}( x; \mathbf{W}^* ) ] &\propto  
    %1 \cdot \indicator{y = y'} + \frac{1}{t_0^4} \cdot s^2_{y', y} \cdot \indicator{y \neq y'}
    % \begin{cases}
    %     1 & \text{ if } y = y' \\
    %     \frac{1}{t_0^4} \cdot s^2_{y', y} & \text{ else }
    % \end{cases} 
    %\\
    %\mathop{\mathbb{E}}_{x \sim \Pr_{\rm ood}( \cdot \mid y )}[ {f}_{y'}( x; \mathbf{W}^* ) ] &\propto 
    %\beta \cdot \left( \frac{1}{t_0^2} \cdot \indicator{y = y'} -\frac{1}{t_0^4} \cdot s^2_{y', y} \cdot  \indicator{y \neq y'} \right).
    % \beta \cdot \begin{cases}
    % \frac{1}{t_0^2} \cdot 1 & \text{ if } {y = y'} \\
    % -\frac{1}{t_0^4} \cdot s^2_{y', y} & \text{ else. }
    % \end{cases}
%\end{align*}
% \end{proposition}
For ID samples, the behaviour is as expected:
the non-target logits ($y' \neq y$) 
% are \emph{positive},
are 
(to first order)
proportional to
$s^2_{y', y}$,
i.e.,
to their similarity to the target class $y$.
This encodes the ``dark knowledge'' amongst these non-target classes.

For OOD samples,
the behaviour is more interesting:
the non-target logits 
% ($y' \neq y$) 
are a \emph{negative rescaling} of their 
% corresponding logits for ID samples;
ID sample counterparts;
i.e., are (to first order) proportional to $-s^2_{y', y}$.
That is, OOD samples' non-target class ranking is on average the \emph{reverse} of that for ID samples;
thus,
\textbf{post-hoc ranking-based OOD detection can be successful}.
Intuitively, this is because OOD samples rely on local, but \emph{not} global features.

A qualifier is that the non-target logits' {scale} can be weak {relative} to the inherent noise;
i.e., $\delta_y( f_{y'} - f_{y''} )$ can be low.
Concretely, 
for any ID class $y$, 
to first order (Corollary~\ref{cor:qualifier-iso})
the gap between non-target logits $y', y''$ has standard deviation $\propto \sigma_{\nullFeat} \cdot \| \boldsymbol{\mu}^{(y')} \|_2$;
as $\boldsymbol{\mu}^{(y')}$ comprises local and background features, this can drown out the expected difference in logits which is $\propto \frac{L}{L + K_{y,y}} \cdot (s^{2}_{y', y} - s^{2}_{y'', y})$.

% the non-target logits ($y' \neq y$) 
% are \emph{negatively} proportional to
% $-s_{y', y}^2$;
% thus, classes with higher similarity to $y$ will have \emph{lower} logits, owing to the prediction solely relying on the local features.
% In both cases,
% however,
% the scale of these non-target logits will shrink 
% if the background feature $\boldsymbol{\mu}_{\sharedFeat}$ is significant ($t_0 \gg 1$).

% The above establishes that 
% in our generative model,
% Put together,
% this implies that
Overall, 
% we see that
OOD samples 
% are seen to
possess a distinctive but potentially weak
\emph{non-target class structure} under standard training.
We now show that self-KD can \emph{strengthen} this structure.

% However, the picture is less clear for \emph{confidence} based OOD detection (e.g., MaxLogit).
% % However, the stark gap to the target class logits can 
% % defeat detection based on the maximum probability and logit.
% % cause 
% % both the maximum softmax probability and logit to be spuriously high, defeating detection based on these scores. 
% % and MaxLogit detection rules to fail.
% % their \emph{maximum softmax probability} to be spuriously high.
% We now show how self-distillation can mitigate this.

% To demonstrate this, we must present expressions for the ID and OOD samples.
% we explicate an analytic form for the logit vector distributions of ID and OOD samples in the setting of .

%
% \subsection{On the success of training-based non-target class signature OOD detection}
\subsection{On the viability of KD-based OOD detection}

\begin{table*}[t]
\centering
\caption{Comparison of average near-OOD detection performance (FPR@95 $\downarrow$) across CIFAR-10, CIFAR-100, and ImageNet-200, using standard logit-based scores. All models use ResNet-18 trained for $100$ epochs.}
\label{tbl:nearood}
\renewcommand{\arraystretch}{1.25}
\resizebox{\textwidth}{!}{
\begin{tabular}{@{}lcccc|cccc|cccc@{}}
\toprule
& & \multicolumn{3}{c|}{\textbf{CIFAR-10}} & \multicolumn{4}{c|}{\textbf{CIFAR-100}} & \multicolumn{4}{c}{\textbf{ImageNet-200}} \\
% \midrule
& \textbf{\makecell{ID \\ accuracy}} & \textbf{\makecell{LogitGap \\ FPR95}} & \textbf{\makecell{MLS \\ FPR95}} & \textbf{\makecell{EBO \\ FPR95}} & 
\textbf{\makecell{ID \\ accuracy}}& \textbf{\makecell{LogitGap \\ FPR95}} & \textbf{\makecell{MLS \\ FPR95}} & \textbf{\makecell{EBO \\ FPR95}} &
\textbf{\makecell{ID \\ accuracy}} &
\textbf{\makecell{LogitGap \\ FPR95}} & \textbf{\makecell{MLS \\ FPR95}} & \textbf{\makecell{EBO \\ FPR95}} \\
\hline
\hline

CE & \textbf{0.951} & 0.613 & 0.613 & 0.613 & 0.773 & 0.547 & 0.555 & 0.555
& 0.864 & 0.548 & 0.598 & 0.602\\

{\sc RankOOD} & 0.941  & 0.364 & 0.365 & 0.365 & 0.723 &  0.595 & 0.611 & 0.662 & 0.847 & 0.542 & {0.535} & {0.535} \\
\hline

Self-KD  & 0.949  & 0.313 & 0.316 & 0.315 & 0.775 & 0.545 & 0.562 & 0.562 
& {0.866} & 0.542 & 0.566 & 0.568 \\

Instance-level {\sc RankOOD} & 0.934  & 0.355 & 0.359 & 0.363 & 0.774 & 0.544 & \textbf{0.546} &  \textbf{0.547}
& 0.873 & 0.529 & 0.552 & 0.555 \\

Instance-level PL-KD & 0.947 & \textbf{0.301} & \textbf{0.305} & \textbf{0.305} & \textbf{0.776} & \textbf{0.531} & {0.568} & 0.570 
& {\bf 0.875} & \textbf{\bf 0.516} & {\bf 0.527} & {\bf 0.529} \\

% \hline
\bottomrule
\end{tabular}
}
\vspace{-3mm}
\end{table*}

\begin{table*}[t]
\centering
\caption{Comparison of average far-OOD detection performance (FPR@95 $\downarrow$) for CIFAR-10, CIFAR-100, and ImageNet-200.}
% \todoakm{CIFAR100 CE ID accuracy slightly different to previous table?}}
\label{tbl:farood}
\renewcommand{\arraystretch}{1.25}
\resizebox{\textwidth}{!}{
\begin{tabular}{@{}lcccc|cccc|cccc@{}}
\toprule
& & \multicolumn{3}{c|}{\textbf{CIFAR-10}} & \multicolumn{4}{c|}{\textbf{CIFAR-100}} & \multicolumn{4}{c}{\textbf{ImageNet-200}} \\
% \midrule
& \textbf{\makecell{ID \\ accuracy}} & \textbf{\makecell{LogitGap \\ FPR95}} & \textbf{\makecell{MLS \\ FPR95}} & \textbf{\makecell{EBO \\ FPR95}} & 
\textbf{\makecell{ID \\ accuracy}}& \textbf{\makecell{LogitGap \\ FPR95}} & \textbf{\makecell{MLS \\ FPR95}} & \textbf{\makecell{EBO \\ FPR95}} &
\textbf{\makecell{ID \\ accuracy}} &
\textbf{\makecell{LogitGap \\ FPR95}} & \textbf{\makecell{MLS \\ FPR95}} & \textbf{\makecell{EBO \\ FPR95}} \\
\hline
\hline

CE &  \textbf{0.951} & 0.417 & 0.417 & 0.417 
& 0.773 & 0.552 & 0.549 & 0.548
& 0.864 & 0.329 & 0.340 & 0.348\\	

{\sc RankOOD} & 0.941  & 0.236 & 0.236 & 0.236
& 0.723 &  0.582 & {0.566} & {0.566} 
& 0.847 & 0.313 & {0.289} & {0.289} \\
\hline 

Self-KD  & 0.949  & 0.200 & 0.201 & 0.201
& {0.775} & 0.538 & 0.514 & 0.512
& {0.866} & {\bf 0.287} & {\bf 0.270} & {\bf 0.273} \\

Instance-level {\sc RankOOD} & 0.934  & 0.244 & 0.246 & 0.250
& 0.774 & 0.539 & 0.534 & 0.531
& 0.873 & 0.303 & 0.298 & 0.305 \\

Instance-level PL-KD & 0.947 &  \textbf{0.191 }&  \textbf{0.193} & \textbf{0.193}
& \textbf{0.776} &  \textbf{0.530} &  \textbf{0.497} &  \textbf{0.491}
& {\bf 0.875} & { 0.291} & 0.280 & { 0.287} \\

% \hline
\bottomrule
\end{tabular}
}
\vspace{-0.5mm}
\end{table*}

% \begin{table}[t]
% \centering
% \caption{Comparison of average near-ood detection performance (FPR@95 $\downarrow$) for CIFAR-10, CIFAR-100, and ImageNet-200.}
% \label{tbl:ce_ln_oe}
% \renewcommand{\arraystretch}{1.2}
% \footnotesize
% \resizebox{0.48\textwidth}{!}{
% \begin{tabular}{@{}lccc|cc}
% \toprule
% & & \multicolumn{2}{c|}{\textbf{Near}} & \multicolumn{2}{c}{\textbf{Far}} \\
% % & \multicolumn{4}{c}{\textbf{ImageNet-200}} \\
% % \midrule
% & \textbf{{ACC}} & \textbf{{FPR95}} &  \textbf{{AUROC}} & \textbf{{FPR95}} &  \textbf{{AUROC}} \\
% \hline
% \hline

% CE & {0.951} & 0.613 & 0.875 & 0.417 & 0.911\\	
% CE + PL-KD & {0.947} & 0.301 & 0.919 & 0.191 & 0.946\\	

% \hline

% LogitNorm & {0.944} & 0.298 & 0.922 & 0.137 & 0.967\\	
% LogitNorm + PL-KD & {0.949} & 0.278 & 0.926 & 0.132 & 0.967 \\	
% \hline
% OE & 0.945 & 0.201 & 0.948 & 0.136 & 0.959 \\
% OE + PL-KD & 0.945 & 0.191 & 0.950 & 0.121 & 0.962 \\

% \bottomrule
% CE & 0.772 & 0.540 & 0.812 & 0.566 & 0.792\\	
% CE + PL-KD & 0.776 & 0.531 & 0.815 & 0.530 & 0.799\\	
% LogitNorm & 0.758 & 0.548 & 0.803 & 0.547 & 0.793\\	
% LogitNorm + PL-KD & 0.777 & 0.540 & 0.811 & 0.528 & 0.806\\	
% OE & 0.766 & 0.303 & 0.890 & 0.501& 0.836\\	
% OE + PL-KD & 0.776 & 0.295 & 0.893 & 0.481 & 0.837\\	
% \bottomrule
% \end{tabular}
% }
% \vspace{-3mm}
% \end{table}

%
% We now study the viability of training-based OOD detectors, which accentuate the signal on the non-target logits.
We consider a self-distillation setup akin to~\citet{Mobahi:2020},
wherein
we regress against 
% square loss with 
the predictions under $\mathbf{W}^*$:
i.e.,
$R_{\rm sd}( \mathbf{W} ) \defEq \mathop{\mathbb{E}}_{x \sim \Pr_{\rm id}}[ \| \mathbf{f}( x; \mathbf{W}^* ) -\mathbf{f}( x; \mathbf{W} ) \|_2^2 ] + \lambda \cdot \| \mathbf{W} \|_F^2$.
This corresponds to cross-entropy based KD as $T \to +\infty$~\citep{Hinton:2015}.
% \begin{equation}
%     \label{eqn:square-loss-self-kd-risk}
%     R_{\rm sd}( \mathbf{W} ) \defEq \mathop{\mathbb{E}}_{x \sim \Pr_{\rm id}}[ \| \mathbf{f}( x; \mathbf{W}^* ) -\mathbf{f}( x; \mathbf{W} ) \|_2^2 ] + \lambda \cdot \| \mathbf{W} \|_F^2.
% \end{equation}
\citet{Mobahi:2020} established that self-distillation attenuates high-frequency components in $\mathbf{f}( x; \mathbf{W}^* )$.
% We re-state this for completeness in our setting in Lemma~\ref{lemm:opt-linear-self-kd} (Appendix):
% we evince that self-distillation scales an eigenvalue $\nu_i$ of 
% the data covariance
% by a factor of $\frac{\nu_i}{\nu_i + \lambda}$, thus strongly suppressing components wherein $\nu_i \sim 0$ (i.e., high-frequency components).
Adapting this to our setting 
% (Lemma~\ref{lemm:opt-linear-self-kd} in Appendix)
% this directly translates to 
yields
the following.
% which as before is 
% % an informal version of the precise statement 
% stated generally
% in Appendix~\ref{app:theory_formal}.
% expressions for the expected logit vectors for ID and OOD samples.
% As before, we present below an informal, simplified version; see Appendix~\ref{app:theory_formal} for a precise statement.

%
\begin{proposition}
\label{corr:opt-logits-first-order-self-kd-informal}
Suppose $\mathbf{W}^*_{\rm sd}$ minimises 
$R_{\rm sd}$.
Let 
$\mathbf{f}( x ) \defEq \mathbf{f}( x; \mathbf{W}^* )$,
$\mathbf{f}^{\rm sd}( x ) \defEq \mathbf{f}( x; \mathbf{W}^*_{\rm sd} )$,
% the \emph{self-distilled} square loss population risk.
and $\omega \defEq \lambda / \tau \in ( 0, 1 )$.
Under Assumptions~\ref{assum:id-feat}--\ref{assum:simplify},
for $y', y'' \neq y \in [ L ]$:
\begin{enumerate}[label=(\alph*),itemsep=0pt,leftmargin=16pt,topsep=0pt]
\item
$\mathbf{W}^*_{\rm sd} = \mathbf{W}^* \boldsymbol{\Gamma}$,
where
$\boldsymbol{\Gamma} \defEq ( 1 - \omega ) \cdot \mathbf{I} + \omega \cdot \mathbf{F}$.
Consequently, for \emph{every} $x \in \XCal$,
$\mathbf{f}^{\rm sd}( x ) = \boldsymbol{\Gamma} \, \mathbf{f}( x; \mathbf{W}^* )$.

\item
% Let $g_{y', y''} \defEq f_{y'}( x; \mathbf{W}^* ) - f_{y''}( x; \mathbf{W}^* )$ and
% $g^{\rm sd}_{y', y''} \defEq f_{y'}( x; \mathbf{W}^*_{\rm sd} ) - f_{y''}( x; \mathbf{W}^*_{\rm sd} )$.
Let $\gamma_0 \defEq (1 - \omega)  + \omega  \cdot F_{yy}$,
and
$b_y \defEq \beta \cdot ( L \cdot \tau  )^{-1} { s^2_{yy} }$.
Suppose $s_{y', y} \neq s_{y'', y}$.
Then, to first-order,
${\delta_{y}( f_{y} - f_{y''} )} = {\delta_{y}( f^{\rm sd}_{y} - f^{\rm sd}_{y''} )}$
and
% self-distillation multiplies
% $\delta_y( f_{y'} - f_{y''} )$
% by
\begin{align*}
    \frac{\delta_{y}( f^{\rm sd}_{y'} - f^{\rm sd}_{y''} )}{\delta_{y}( f_{y'} - f_{y''} )} &= 1 + \frac{\omega}{\gamma_0} \cdot 
% \frac{ F_{yy} - b_y \cdot ( 1 - F_{yy} ) }{ 1 + b_y } 
\left( F_{yy} - \frac{b_y}{ 1 + b_y } \right)
> 1.
% \\
% \frac{\delta_{y}( f_{y} - f_{y''} )}{\delta_{y}( f_{y} - f_{y''} )} &= 1.
\end{align*}
\end{enumerate}
\end{proposition}

Part (a) shows that under self-KD, 
a non-target logit $f^{\rm sd}_{y'}$ 
receives a target logit contribution
of $\omega \cdot F_{y', y} \cdot f_{y} \propto s^{2}_{y', y} \cdot f_y$.
In expectation, this is larger for ID than OOD samples (Proposition~\ref{corr:opt-logits-first-order-informal}).
Part (b) shows that to first-order,
self-KD {increases the separation of all non-target logit gaps}, 
while preserving the separation of target to non-target logit gaps.
Overall, \textbf{self-KD can improve ranking-based OOD detection by strengthening the non-target ranking structure}.

Proposition~\ref{corr:opt-logits-first-order-self-kd-informal} also highlights a potential limitation of self-KD:
% while the \emph{magnitude} of non-target logits is improved 
% compared to one-hot training, 
% their \emph{ranking} can be slightly degraded.
a non-target logit also receives a contribution 
% $\omega \cdot F_{y', y''} \cdot f_{y''}$ 
from other non-target siblings $y''$,
and so
non-transitive class similarities can distort the ranking.
This is suggestive of the value of leveraging \emph{explicit} ranking supervision, such as PL-KD.
Formalising this is an interesting avenue for future work.

\textbf{Relation to existing work}.
Our generative model is inspired by~\citet{Ming:2022}, who 
analysed OOD detection via a spurious feature model (see also~\citet{Zhang:2023}).
Compared to their \emph{Bayes-optimal} setup, we consider multi-class classification,
and explicate the logit behaviour under one-hot and self-KD training.
% in binary classification settings.
% Our analysis is for , and also explicates the relation to knowledge distillation.
As with~\citet{Ming:2022}, our model is reminiscent of the distinct literature on
robustness under spurious correlations~\citep{Sagawa:2019,Sagawa:2020b,Sohoni:2020}.
% However, we explicate the implications for OOD detection.
\citet{AllenZhu:2023} rigorously analysed standard KD via a certain \emph{multi-view} data model.
This has broad similarities with our generative model, while differing in details.
Further, we explicate the behaviour of self-KD for OOD detection (albeit for the special case of linear models and square loss).
Appendix~\ref{sec:related_extended} discusses additional related work.
% \textcolor{red}{Apart from the Table 1 caption, we don't seem to highlight that instance-level is the better version? What could be the better places to mention this? Probably needs Aditya's help.}

%% file: sec/6_experiments.tex
\section{Experimental results}
\label{sec:experiments}

\subsection{Setup}

\textbf{Datasets}. We present results on three standard image classification datasets: CIFAR-10, CIFAR-100, and ImageNet-200, each of which is treated as the ID dataset, with evaluation conducted against both near-OOD and far-OOD test sets following the OpenOOD protocol~\citep{Yang:2022}. For CIFAR-10, we treat CIFAR-100 and ImageNet-200 as near-OOD datasets; conversely, for CIFAR-100, CIFAR-10 and ImageNet-200 serve as near-OOD. For both CIFAR-10 and CIFAR-100, the far-OOD set includes MNIST~\citep{mnist}, SVHN~\citep{svhn}, Textures~\citep{textures}, and Places365~\citep{places}. For ImageNet-200, SSB-hard~\citep{ssb_hard} and NINCO~\citep{ninco} are used as near-OOD datasets, while iNaturalist~\citep{inaturalist}, Textures, and OpenImage-O~\citep{wang2022vim} constitute the far-OOD set. For each dataset, we use a standard ResNet~\citep{He:2015} as the underlying neural network model. 
% \textcolor{red}{In one of our OOD papers, we got a comment about testing only on ResNet. I know it's too late now, but it might be worthwhile to have the ViT results ready after submission, just in case we need them for the rebuttal.}

\textbf{Metrics}.
% We report the $\mathrm{FPR@}95$ as the primary evaluation metric.
We report ${\rm FPR}@95$, AUROC, and ID accuracy as the metrics. ${\rm FPR}@95$ denotes the false positive rate on OOD samples at 95\% TPR on ID data, while AUROC measures the ability to distinguish ID from OOD samples across all thresholds. All results are averaged over three runs. %In the main paper, we report averaged near-OOD and far-OOD ${\rm FPR}@95$. Detailed ${\rm FPR}@95$ (with standard deviation) and AUROC results are provided in Appendix~\ref{app:results_extended}.

\textbf{Baselines}. 
We evaluate Instance-level PL-KD in two settings.
First, we compare it against 
training with cross-entropy (CE), {\sc RankOOD}, and other knowledge distillation methods. For CE, we use the pre-trained checkpoints provided by OpenOOD for CIFAR-10/100 and ImageNet-200, and train {\sc RankOOD} for $100$ epochs for a fair comparison. 
% \todoakm{datasets only introduced in next para}
Second, we investigate the effectiveness of PL-KD across different training paradigms by incorporating it into LogitNorm~\cite{Wei:2022} and Outlier Exposure (OE)~\cite{Hendrycks:2019}, thereby assessing its generalisability beyond standard CE training. 
We report common \emph{logit-based} scoring functions (\S\ref{sec:ood-background}), 
% rather than probability-based alternatives, 
including 
LogitGap~\cite{Liang:2025} and Energy (EBO)~\cite{Liu:2020b}. 
% \textcolor{red}{can we refer back to 2.2 and remind the reader why we selected these?} 
Finally, we follow the OpenOOD benchmark by comparing Instance-level PL-KD against a comprehensive range of  both post-hoc methods, and training methods w/ and w/o outliers. 

\textbf{Implementation Details}. 
We train ResNet-18 on CIFAR-10, CIFAR-100 and ImageNet-200, for 100 epochs. 
All models use SGD with momentum $0.9$, an initial learning rate of $0.1$, and cosine annealing with weight decay of $5\times10^{-4}$. 
Following {\sc RankOOD}, CIFAR-10 trains on all ranks; CIFAR-100 and ImageNet-200 train on top-$10$ and bottom-$10$ ranks for ListMLE variants. 
We set ($\beta=0.9$) for CIFAR-10 and ImageNet-200, and ($\beta=0.5$) for CIFAR-100. Hyperparameters are selected via grid search on the OpenOOD ID/OOD validation sets using ${\rm FPR@95}$.
Further details, including temperature selection, are provided in Appendix~\ref{appendix:hyp_pram_selec}.

%%%
\subsection{Results}

\begin{figure}[t]
    \centering
    \includegraphics[width=0.5\textwidth]{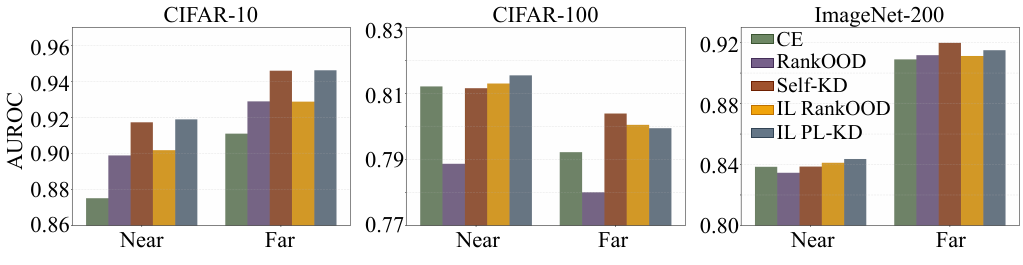}
    \caption{Near- and far-OOD detection performance (LogitGap AUROC $\uparrow$) on CIFAR-10, CIFAR-100, \& ImageNet-200.}
     \label{fig:auroc_comp}
    \vspace{-\baselineskip}
\end{figure}

Tables~\ref{tbl:nearood} and~\ref{tbl:farood} report near- and far-OOD ${\rm FPR}@95$ on
CIFAR-10, CIFAR-100 and ImageNet-200 under the 
% LogitGap, MLS and EBO scores.
logit-based scores.
Figure~\ref{fig:auroc_comp} reports LogitGap AUROC, and detailed results are in \textcolor{black}{Appendix~\ref{app:baseline_comparison}}. 
Instance-level PL-KD consistently achieves the lowest ${\rm FPR}@95$, demonstrating superior OOD detection compared to competing rank-based distillation methods, including class- and instance-level {\sc RankOOD}. 
Specifically, PL-KD reduces ${\rm FPR}@95$ by up to 19\% relative to {\sc RankOOD}, while improving ID accuracy by up to 7\%.
Compared with standard CE baselines, PL-KD improves ${\rm FPR}@95$ by up to 54.2\%, achieving the highest ID accuracy on CIFAR-100 and ImageNet-200. 
% across the three datasets.

% Furthermore, PL-KD yields substantial OOD performance gains (upto 54.2\%) over standard Cross-Entropy (CE) baselines while preserving highly competitive in-distribution (ID) accuracy, notably achieving the highest overall ID accuracy on both CIFAR-100 and ImageNet-200. 
% Finally, 
Figure~\ref{fig:auroc_comp} shows PL-KD continues to deliver competitive AUROC enhancements across nearly all ID datasets, with the only exceptions being far-OOD detection on CIFAR-100 and ImageNet-200.
Finally, as shown in Appendix Table~\ref{tbl:nearood_appendix}, PL-KD achieves $3^{\rm rd}$-best results in near-OOD detection, outperforming $35$ baselines. Notably, on ImageNet-200, PL-KD is second only to {\sc RankOOD} in OOD detection while attaining the highest ID accuracy.

%Notably, on ImageNet-200, PL-KD achieves the second best overall OOD performance, second only to {\sc RankOOD}, while attaining the highest ID accuracy.
%\vspace{-5mm}
% \todoakm{Not clear how it is "best overall" but second to RankOOD?}
% Instance-level PL-KD consistently outperforms competing methods across different OOD scores and datasets, achieving the lowest ${\rm FPR}@95$ in the majority of evaluation settings \textcolor{red}{wide range.... may not be a good wording here}. In particular, it delivers substantial improvements over {\sc RankOOD} and CE while maintaining competitive ID accuracy, achieving the highest ID accuracy on CIFAR-100 and ImageNet-200. Notably, PL-KD also improves ID accuracy over {\sc RankOOD} across all three datasets, with the exception that CE achieves the highest accuracy on CIFAR-10. As shown in Figure~\ref{fig:auroc_comp}, PL-KD continues to demonstrate competitive AUROC improvements across all ID datasets except in far-ood detection on ImageNet-200 and CIFAR-100. These findings highlight the benefit of combining knowledge distillation with instance-level logit ranking to improve OOD detection while preserving ID classification performance. \textcolor{red}{last sentence can be dropped?}
\begin{figure}
    \centering
    \includegraphics[width=0.48\textwidth]{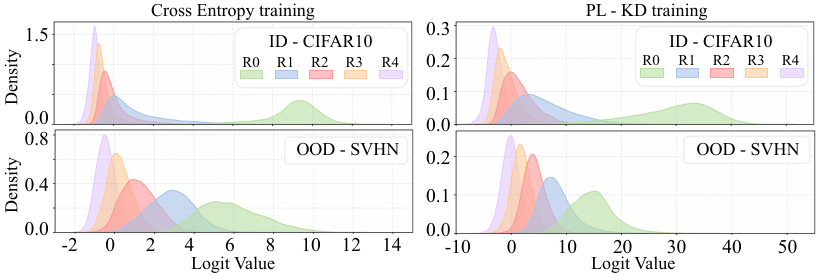}
    \caption{Ranked logit distributions of ID (CIFAR-10) and OOD (SVHN) samples under CE (left) and PL-KD (right) training.}
     \label{fig:id_ood_rank_distribution}
    \vspace{-\baselineskip}
\end{figure}

% Left: density distributions across ranking positions (R0–R4). Right: mean ranked logits with shaded standard-deviation regions.

\subsection{PL-KD across different training paradigms}
% PL-KD improves OOD detection across different baselines \textcolor{red}{here more clarity is required to differentiate from the previous results. I think what you are showing is PL-KD improves performance in combination with other baselines?}. 
Since PL-KD only modifies the second-stage supervision, it can be combined with any training objective $\ell_{\mathrm{base}}$. Given a teacher $\mathbf{f}^{\rm t}$ trained by minimising $\ell_{\mathrm{base}}$, we train a student $\mathbf{f}$ by minimising $\nu \cdot \ell_{\mathrm{base}}+(1-\nu) \cdot \ell_{\rm epl}$. In Table~\ref{tbl:ce_ln_oe}, we evaluate PL-KD with CE, LogitNorm (LN), and Outlier Exposure (OE), using their respective trained models as teachers. PL-KD consistently improves both near- and far-OOD detection over all three baselines. The improvements are more pronounced for CE,  where ${\rm FPR}@95$ decreases by $>50.1\%$  and $1.7\%$ for CIFAR-10 and CIFAR-100 respectively.
% \sanj{51\% is over CE but as a footnote we should highlight {\sc RankOOD}}.
For LogitNorm and OE, which already achieve strong OOD performance, PL-KD further reduces ${\rm FPR}@95$ by $1.5$--$11.0\%$ while achieving equal or higher ID accuracy.
These results are consistent with the analysis in \S\ref{sec:method} and \S\ref{sec:theory}.
LogitNorm constrains only the \emph{norm} of the logit vector, and OE shapes
the response to outliers. Neither imposes structure on the relative ordering
of ID \emph{non-target} logits, which is precisely what
$\ell_{\rm epl}$ supervises. PL-KD therefore provides a
complementary signal to existing OOD training objectives.
% These results demonstrate that PL-KD can enhance OOD detection across different training objectives without compromising ID classification performance. \textcolor{red}{are there any appendix tables to point to? how are these results compared to Table 2 - further better?}

% \subsection{Why PL-KD for OOD detection?} 
\subsection{PL-KD's effect on logits and OOD performance} 

% \textcolor{red}{Title: Effect of PL-KD on logits and OOD performance - current one is fine too}
Figure~\ref{fig:id_ood_rank_distribution} compares ranked logit distributions for ID (CIFAR-10) and OOD (SVHN) samples under CE and PL-KD.
CE training separates ID and OOD samples at rank $0$; however, PL-KD further amplifies this separation, with ID samples exhibiting substantially higher top-ranked logits ($\sim 30$ versus $9$). 
This behaviour follows from the soft PL-KD loss, which progressively 
matches the student and teacher distributions
after
removing the teacher's top-ranked classes. 
Unlike conventional KD, where softmax saturation can attenuate gradients for lower-ranked logits, successive renormalisation maintains supervision across ranking positions. 
While the first term
corresponds to standard KD, subsequent terms supervise lower-ranked classes. 
% Although both objectives share the same optimum (teacher logits up to a shared additive constant), PL-KD improves the optimisation of the ranking structure. 
The logit-scale separation is particularly beneficial for logit-based OOD scores, such as MaxLogit and LogitGap, which preserve differences in absolute logit values. In contrast, MSP is invariant to common logit shifts and saturates under large top-ranked logit gaps, compressing the ID-OOD separation near one.
Thus, PL-KD exposes ID-OOD differences in logit space that are partly concealed by softmax-based scores (cf. Table~\ref{tbl:appendix_fpr_cifar_10}--\ref{tbl:appendix_auroc_imagenet200}).
% Further, as shown in Appendix Table~\ref{tbl:nearood_appendix}, PL-KD achieves third-best results in near-OOD detection outperforming $35$ baselines. \textcolor{red}{highlight that near OOD imagent angle here.}
% \textcolor{red}{are there any appendix tables to point to? I think a bit more bragging is required - for example, over 30 baselines, state-of-the-art performance at least in one case. Right now, the results don't appear impactful. One way to address this is to bring some of the highlights from the appendix, like: Further, as we showed in Appendix X, PL-KD achieves .....It achieves the best performance among posthoc logit based methods ....... }

\begin{table}[t]
\centering
\caption{Near- and far-OOD detection performance of PL-KD applied to CE~\cite{Hendrycks:2017}, LogitNorm (LN)~\cite{Wei:2022}, and OE~\cite{Hendrycks:2019} baselines, using their respective trained models as teachers for knowledge distillation. The LogitGap is used as the OOD scoring function. Results are reported in accuracy, ${\rm FPR}@95$ ($\downarrow$), and AUROC ($\uparrow$).}
\label{tbl:ce_ln_oe}
\renewcommand{\arraystretch}{1.2}
\footnotesize
\resizebox{\columnwidth}{!}{
\begin{tabular}{@{}llcc|cc|cc@{}}
\toprule
% \makecell{LogitGap \\ FPR95}
& \textbf{Metric} & \textbf{CE} & \textbf{\makecell{CE + \\PL-KD}} & \textbf{LN} & \textbf{\makecell{LN + \\PL-KD}} & \textbf{OE} & \textbf{\makecell{OE + \\PL-KD}} \\
\midrule
\hline
% \multirow{3}{*}{Dataset 1}
\multirow{5}{*}{\rotatebox{90}{CIFAR-10}}
& Accuracy & {\bf 0.951} & 0.947 & 0.944 & {\bf 0.949} & {\bf 0.945} & {\bf 0.945} \\
& Near FPR95 $\downarrow$ & 0.613 & {\bf 0.301} & 0.298 & {\bf 0.278} & 0.201 & {\bf 0.191} \\
& Near AUROC $\uparrow$ & 0.875 & {\bf 0.919} & 0.922 & {\bf 0.926} & 0.948 & {\bf 0.950} \\
\cmidrule(lr){2-8}
& Far FPR95 $\downarrow$ & 0.417 & {\bf 0.191} & 0.137 & {\bf 0.132} & 0.136 & {\bf 0.121} \\
& Far AUROC $\uparrow$ & 0.911 & {\bf 0.946} & {\bf 0.967} & {\bf 0.967} & 0.959 & {\bf 0.962 }\\
\midrule
% \multirow{3}{*}{Dataset 2}
\multirow{5}{*}{\rotatebox{90}{CIFAR-100}}
& Accuracy & 0.772 & {\bf 0.776} & 0.758 & {\bf 0.777} & 0.766 & {\bf 0.776} \\
& Near FPR95 $\downarrow$ & 0.540 & {\bf 0.531} & 0.548 & {\bf 0.540} & 0.303 & {\bf 0.295} \\
& Near AUROC $\uparrow$ & 0.812 & {\bf 0.815} & 0.803 & {\bf 0.811} & 0.890 & {\bf 0.893} \\
\cmidrule(lr){2-8}
& Far FPR95 $\downarrow$ & 0.566 & {\bf 0.530} & 0.547 & {\bf 0.528} & 0.501 & {\bf 0.481} \\
& Far AUROC $\uparrow$ & 0.792 & {\bf 0.799} & 0.793 & {\bf 0.806} & 0.836 & {\bf 0.837} \\
\bottomrule
\end{tabular}
}
\vspace{-3mm}
\end{table}

%% file: sec/7_conclusion.tex
\section{Conclusion and future work}

% Recent work has introduced a class of effective new OOD detection techniques, based on a simple insight: 
% whilst OOD samples may spuriously result in high-confidence predictions for a certain class, 
% their predictions for \emph{other} classes 
% % tend to be arbitrary, and 
% do not reflect semantic patterns inherent in the data.
% Amongst these, 
% the {\sc RankOOD} method 
% --
% which measures discrepancy in the \emph{ranking of non-top classes} 
% --
% is notable for its state-of-the-art empirical performance.
% Despite this empirical success, 
% a precise formalism for \emph{why} such a detector can perform well has proven elusive.
% % a precise formalism for \emph{why} such signatures can improve OOD detection can perform well remains lacking.
% Further, encouraging a strict signature for OOD samples can come at the (undesirable) expense of regular test accuracy.
We related ranking-based OOD detectors to a form of knowledge distillation (KD), providing a lens for more holistic understanding of such methods. 
Proposed \textit{Instance-level PL-KD} achieves a better balance between ID accuracy and OOD performance, ranking among the top-$5$ methods for near-OOD detection.
% We hope the resulting algorithmic and theoretical analyses can inspire further work in the design and analysis of OOD detectors. 
%and covariate shift), 
%as well as to consider more general losses (e.g., softmax cross-entropy).
Empirically, it is of interest to combine PL-KD with complementary extant methods, such as those that regularise intermediate feature embeddings.
Another possible extension is to fuse multimodal data for OOD detection.
Theoretically, one important direction is to expand our analysis to covariate-shift generalisation.

% Specifically, we establish that {\sc RankOOD} is a form of \emph{self}-KD, where the teacher and student model belong to the same family.
% By studying the role of the loss function and supervision, we suggest two natural algorithms that combine strengths of {\sc RankOOD} and KD.
% Finally, we provide an intuitive feature model that explicates how KD can suppress spurious correlations when predicting ID samples.
% \section{LLM Usage Statement}
% % No large language models (LLMs) or other generative AI tools were used in any part of this work. Specifically, 
% No LLMs were used for 
% preparing the research idea, 
% conducting related work searches or citations,
% designing or running experiments, 
% coding assistance, 
% data generation, or 
% brainstorming. 
% %or writing and polishing the manuscript.
% \todoakm{update}

\section{LLM Usage Statement}
We used Claude and Gemini as follows.
\emph{Writing:} to edit and refine author-written text throughout.
\emph{Theory:} to verify parts of the proofs in Appendices~B--D; all proofs
were first derived and verified by the authors.
\emph{Figures:} to refine author-produced figures.
\emph{Not used:} LLMs were not used to conceive the research idea or method,
to design or run experiments, to write code, or to generate data.
The authors take full responsibility for all content.

%% file: sec/X_suppl.tex
% =====================================================================
% Supplementary material (appendix from the NeurIPS version).
% CVPR 2026 allows a single-column layout for supplementary PDFs.
% =====================================================================
\clearpage
\setcounter{page}{1}
\onecolumn
\begin{center}
    {\Large\bfseries \thetitle \par}
    \vspace{0.5em}
    {\large Supplementary Material \par}
    \vspace{1.5em}
\end{center}

\appendix

\section{Formal statements of generative model results}
\label{app:theory_formal}
\input{sec/suppl/theory_formal}
\section{Proofs of main results}
\label{app:proofs}
\input{sec/suppl/proofs}

%
\section{Additional helper results}
\label{app:helper}
\input{sec/suppl/helper}

\section{Additional theoretical results}
\label{app:theory_extended}
\input{sec/suppl/theory_extended}

\section{Additional discussion of related work}
\label{sec:related_extended}
\input{sec/suppl/related_extended}

\section{Illustration of generative model}
\label{app:illustration}
\input{sec/suppl/illustration}

\section{A worked example}
\label{sec:worked-example}
\input{sec/suppl/worked_example}

\section{Additional experimental results}
\label{app:results_extended}
\input{sec/suppl/results_appendix}

\subsection{Comparison with baselines}
\label{app:baseline_comparison}
\input{sec/suppl/results_baselines}

%% file: sec/suppl/theory_formal.tex
\subsection{Formal results: general form}

Based on the well-established
analytic form for 
the minimiser
$\mathbf{W}^*$ of the square loss
% ~\eqref{eqn:square-loss-risk}
---
see Appendix~\ref{app:helper}
---
we may derive analytic expressions for the (expected) logit vectors for both ID and OOD samples.
To express these, 
let
$\mathbf{T} \defEq \bar{\boldsymbol{\Sigma}} + \lambda \cdot \mathbf{I}$
be
the regularised
average class-covariance matrix $\bar{\boldsymbol{\Sigma}}$,
% $\mathbf{D} \defEq \mathrm{diag}( \{ (\boldsymbol{\mu}^{(y)})^\top \mathbf{T}^{-1} \boldsymbol{\mu}^{(y)} \colon y \in [ L ] \} )$,
% $\boldsymbol{\Delta} \defEq \mathbf{M}^\top \mathbf{T}^{-1} \mathbf{M} - \mathbf{D}$,
and
$\mathbf{K} \defEq \mathbf{M}^\top \mathbf{T}^{-1} \mathbf{M}$
be the whitened class similarities.
Further,
let $\mathbf{H} \defEq ( L \cdot \mathbf{I} + \mathrm{diag}( \mathbf{K} ) )^{-1} ( \mathbf{K} - \mathrm{diag}( \mathbf{K} ) )$ be the regularised off-diagonal class similarities.
In an abuse of notation, we denote by $\Pr_{\rm ood}( \cdot \mid y )$ the distribution over OOD  samples 
whose
\emph{latent} class
(per Assumption~\ref{assum:ood-feat})
is
$y \in [ L ]$.

\begin{proposition}
\label{corr:opt-logits-first-order}
Suppose $\mathbf{W}^*$ is the optimal square loss population risk minimiser.
% ~\eqref{eqn:square-loss-risk}.
Under Assumptions~\ref{assum:id-feat}--\ref{assum:balanced-class}, 
for any $y, y' \in [ L ]$,
the expected logit vectors for ID and OOD samples satisfy
\begin{align*}
    \mathop{\mathbb{E}}_{x \sim \Pr_{\rm id}( \cdot \mid y )}[ {f}_{y'}( x; \mathbf{W}^* ) ] &= 
    \frac{1}{L + K_{y, y}} \cdot
    \begin{cases}
        K_{y', y} & \text{ if } y = y' \\
        \frac{L}{L + K_{y', y'}} \cdot K_{y', y} & \text{ else }
    \end{cases} &+ \, \mathscr{O}( \| \mathbf{H} \|_F^2 ) \\     
    \mathop{\mathbb{E}}_{x \sim \Pr_{\rm ood}( \cdot \mid y )}[ {f}_{y'}( x; \mathbf{W}^* ) ] &= 
    \beta \cdot \frac{1}{L + K_{y, y}} \cdot \begin{cases}
    N_{y, y} & \text{ if } {y = y'} \\
    -\frac{N_{y, y}}{L + K_{y', y'}} \cdot K_{y', y} & \text{ else }
    \end{cases}
     &+ \, \mathscr{O}( \| \mathbf{H} \|_F^2 ),
    % \mathop{\mathbb{E}}_{x \sim \Pr_{\rm id}( \cdot \mid y )}[ {f}_{y'}( x; \mathbf{W}^* ) ] &= 
    % \frac{K_{y, y}}{L + K_{y, y}} \cdot \indicator{ y = y' } +
    % \frac{K_{y', y}}{L + K_{y', y'}} \cdot \frac{L}{L + K_{y, y}} \cdot \indicator{ y \neq y' } \\
    % \mathop{\mathbb{E}}_{x \sim \Pr_{\rm ood}( \cdot \mid y )}[ {f}_{y'}( x; \mathbf{W}^* ) ] &= 
    % \beta \cdot \frac{N_{y, y}}{L + K_{y, y}} \cdot \indicator{y = y'} - 
    % \beta \cdot \frac{K_{y', y}}{L + K_{y', y'}} \cdot \frac{N_{y, y}}{L + K_{y, y}} \cdot \indicator{ y \neq y' }.
\end{align*}
where 
$\mathbf{N} \defEq \mathrm{diag}(\{ 
{s_{yy}^2} \cdot {( \frac{\sigma^2_{\locFeat}}{L} \cdot s_{yy}^2 + \sigma^2_{\nullFeat} + \lambda )^{-1}} \colon y \in [ L ] \})$.
% and 
% $\boldsymbol{\eta}^{(y)} \defEq 
% % \mathbf{M}^\top \mathbf{T}^{-1} \boldsymbol{\mu}^{(y)}_{\locFeat}
% % =
% \mathbf{N} \mathbf{e}_y$.
\end{proposition}

\begin{proposition}
\label{corr:opt-logits-first-order-self-kd}
Suppose $\mathbf{W}^*_{\rm sd}$ is the optimal \emph{self-distilled} square loss population risk minimiser.
% ~\eqref{eqn:square-loss-self-kd-risk}.
Under Assumptions~\ref{assum:id-feat}--\ref{assum:balanced-class}, 
for any $y, y' \in [ L ]$,
the 
difference in self-distilled and regular
expected logit vectors for ID and OOD samples satisfy
\begin{align*}
    \mathop{\mathbb{E}}_{x \sim \Pr_{\rm id}( \cdot \mid y )}[ {f}_{y}( x; \mathbf{W}^*_{\rm sd} ) - {f}_{y}( x; \mathbf{W}^* ) ]  &= 
    -
    \frac{\lambda \cdot L}{L + K_{y, y}} \cdot
    \left(
    \frac{1}{L + K_{y, y}} \cdot {K^{(2)}_{y, y}} \right) + \mathscr{O}( \| \mathbf{H} \|_F^2 ) \\
    \mathop{\mathbb{E}}_{x \sim \Pr_{\rm id}( \cdot \mid y )}[ {f}_{y'}( x; \mathbf{W}^*_{\rm sd} ) - {f}_{y'}( x; \mathbf{W}^* ) ] &= 
    \frac{\lambda \cdot L}{L + K_{y, y}} \cdot \frac{1}{{L + K_{y', y'}}} \cdot
    \left(
    K_{y', y} \cdot \left[  
    \frac{K^{(2)}_{y,y}}{ Z_{y,y} }+ \frac{K^{(2)}_{y',y'}}{ Z_{y',y'} }
    \right] - K^{(2)}_{y', y} \right) + \mathscr{O}( \| \mathbf{H} \|_F^2 ) \\
    \mathop{\mathbb{E}}_{x \sim \Pr_{\rm ood}( \cdot \mid y )}[ {f}_{y}( x; \mathbf{W}^*_{\rm sd} ) - {f}_{y}( x; \mathbf{W}^* ) ] &= 
    -
    \beta \cdot \lambda \cdot \left[ \frac{N^{(2)}_{y, y}}{Z_{y, y}} - A_{y, y} \cdot N_{y, y} \right] + \mathscr{O}( \| \mathbf{H} \|_F^2 ) \\
    \mathop{\mathbb{E}}_{x \sim \Pr_{\rm ood}( \cdot \mid y )}[ {f}_{y'}( x; \mathbf{W}^*_{\rm sd} ) - {f}_{y'}( x; \mathbf{W}^* ) ] &= 
    \beta \cdot \lambda \cdot \left[ \frac{K_{y', y} \cdot N^{(2)}_{y, y}}{Z_{y', y'} \cdot Z_{y, y}} + A_{y', y} \cdot N_{y, y} \right] + \mathscr{O}( \| \mathbf{H} \|_F^2 ),
\end{align*}
where $\mathbf{K}^{(2)} \defEq \mathbf{M}^\top \mathbf{T}^{-2} \mathbf{M}$, 
$\mathbf{A} \defEq ( L \cdot \mathbf{I} + \mathbf{K} )^{-1} \mathbf{K}^{(2)} ( L \cdot \mathbf{I} + \mathbf{K} )^{-1}$,
and $\mathbf{M}^\top \mathbf{T}^{-2} \boldsymbol{\mu}_{\rm ood}^{(y)} = \frac{\beta}{s_{yy}} \cdot \mathbf{N}^{(2)} \mathbf{e}_y$ for $\mathbf{N}^{(2)} \defEq \text{diag}( \{ s^2_{yy} / (n^{(y)} + \lambda)^2 \colon y \in [ L ] \} )$.
\end{proposition}

% To intuit this, 
% as before, we consider a simplified setting 
% with
% equal class self-similarities and
% weak cross-similarities.
% In this case, 
% for $y' \neq y$,
% $K^{(2)}_{y', y} \sim \frac{K_{y', y}}{\sigma^2_{\nullFeat} + \lambda} = \frac{s^2_{y', y}}{( \sigma^2_{\nullFeat} + \lambda )^2}$ 
% and
% $A_{y', y} 
% % = \frac{K^{(2)}_{y', y}}{Z_{y, y} \cdot Z_{y', y'}} 
% \sim \frac{s^{2}_{y', y}}{( \sigma^2_{\nullFeat} + \lambda )^2 \cdot L^2}$. 
% We have
% the dominant correction term in 
% both the ID and OOD non-target logits is proportional to 
% % -\frac{K^{(2)}_{y', y}}{L + K_{y', y'}} \sim 
% % -\frac{\lambda}{L} \cdot \frac{s^2_{y', y}}{( \sigma^2_{\nullFeat} + \lambda )^2}$; 
% $-\frac{\lambda}{L} \cdot \frac{s^2_{y', y}}{( \sigma^2_{\nullFeat} + \lambda )^2}$;
% when $L$ is large, this term is negligible.
% On the other hand, for OOD samples, the dominant correction term in the 
% % non-target logits is proportional to
% target logits is proportional to
% $\frac{1}{t_0^2}$.
% Compared to the standard logits when training on one-hot labels, self-distillation 
% thus shrinks the target OOD logits towards $0$.

% $L \cdot A_{y', y} \cdot N_{y, y} \sim  \frac{K^{(2)}_{y', y}}{K_{y', y'}} \sim \frac{s^2_{y', y}}{( \sigma^2_{\nullFeat} + \lambda )^2}$; this correction is significantly larger than for ID samples.
% Compared to the standard logits when training on one-hot labels, self-distillation 
% thus shrinks the tail OOD logits towards $0$.

%%%%%%%%%%

% %
\subsection{Formal results: specialisation under Assumption~\ref{assum:simplify}}

% \textbf{Specialisation under Assumption~\ref{assum:simplify}.}
Under Assumption~\ref{assum:simplify}, 
$\alpha^{(i)}_{\locFeat} = \alpha^{(i,j)}_{\globFeat} = 1$ almost surely.
Thus, by Lemma~\ref{lemm:covariance}, every class covariance is $\boldsymbol{\Sigma}^{(i)} = \sigma^2_{\nullFeat} \mathbf{I}$.
Therefore, $\bar{\boldsymbol{\Sigma}} = \sigma^2_{\nullFeat} \mathbf{I}$ and $\mathbf{T} = \tau \mathbf{I}$ with $\tau = \sigma^2_{\nullFeat} + \lambda$.
It follows that the above objects (and those in Table~\ref{tbl:glossary}) reduce to those of \S\ref{sec:theory}:
\begin{align*}
\mathbf{K} &= \tau^{-1} \mathbf{M}^\top \mathbf{M} \\
K_{y y'} &= \indicator{y \neq y'} \cdot \tau^{-1} \cdot s^2_{y y'} + \indicator{ y = y' } \cdot \tau^{-1} \cdot \| \boldsymbol{\mu}^{(y)} \|_2^2 \\
&= \indicator{y \neq y'} \cdot \tau^{-1} \cdot s^2_{y y'} + \indicator{ y = y' } \cdot \tau^{-1} \cdot \left( t_y^2 + s^2_{yy} + \sum_{k \neq y} s^2_{yk} \right) \text{ by Lemmas~\ref{lemm:whitened-similarity-diag},~\ref{lemm:whitened-off-similarity} } \\
\mathbf{K}^{(2)} &= \tau^{-1} \mathbf{K} \\
n^{(y)} &= \sigma^2_{\nullFeat} \\
\mathbf{N} &= \tau^{-1} \mathrm{diag}( \{ s^2_{yy} \} ) \\
b_y &= L^{-1} \cdot \beta \cdot N_{yy} \\
\mathbf{D} &= \kappa \mathbf{I} \\
\mathbf{Z} &= ( L + \kappa ) \mathbf{I} \\
\mathbf{H} &= ( L + \kappa )^{-1} \cdot \boldsymbol{\Delta},
\end{align*}
where each $K_{yy} \equiv \kappa$.
Further, by Lemma~\ref{lemm:gram},
\begin{align*}
\mathbf{F} &= ( L \cdot \mathbf{I} + \mathbf{K} )^{-1} \mathbf{K} \\
&= \mathbf{I} - L ( L \cdot \mathbf{I} + \mathbf{K} )^{-1} \\
&= ( \mathbf{W}^* )^\top \mathbf{M} \\
\mathbf{A} &= ( \mathbf{W}^* )^\top \mathbf{W}^* \\
&= ({L \cdot \tau})^{-1} \mathbf{F} ( \mathbf{I} - \mathbf{F} ).
\end{align*}
Note that
$\mathbf{F}$ is symmetric, with eigenvalues $\phi_j = \kappa_j / ( L + \kappa_j ) \in [ 0, 1 )$, where $\kappa_j \geq 0$ are the eigenvalues of $\mathbf{K}$.
A result holds ``to first order'' if it includes an additive remainder that is $\mathscr{O}( \| \mathbf{H} \|^2_F )$.
Note that $\| \mathbf{H} \|_F = ( L + \kappa )^{-1} \cdot \| \boldsymbol{\Delta} \|_F$ is small when the off-diagonal similarities $\tau^{-1} \cdot s^2_{y,y'}$ are small relative to $L + \kappa$.
Finally, note that by Lemma~\ref{lemm:k-eigenvalue}, $\| \mathbf{H} \|_2 < 1$ is guaranteed, so the matrix expansions below always converge.

%
% \textbf{Specialised form of results}.
We now establish the results of \S\ref{sec:theory} under Assumption~\ref{assum:simplify}.

\begin{lemma}
\label{lemm:logit-cov}
Pick any $\mathbf{W} \in \Real^{D \times L}$,
$\mathbf{c} \in \Real^L$,
and
class $y \in [ L ]$.
Let $g( x ) \defEq \mathbf{c}^\top \mathbf{f}( x; \mathbf{W} )$.
Under Assumptions~\ref{assum:id-feat}--\ref{assum:simplify},
for $x \sim \Pr_{\rm id}( \cdot \mid y )$ and $\bar{x} \sim \Pr_{\rm ood}( \cdot \mid y )$,
\begin{align*}
	g( x ) &\sim \mathscr{N}\left( 
	\mathbf{c}^\top \mathbf{W}^\top \boldsymbol{\mu}^{(y)}, \sigma^2_{\nullFeat} \| \mathbf{W} \mathbf{c} \|_2^2
	\right) \\
	g( \bar{x} ) &\sim \mathscr{N}\left(
	\beta s_{yy} \mathbf{c}^\top \mathbf{W}^\top \boldsymbol{\mu}^{(y)}_{\locFeat}, \sigma^2_{\nullFeat} \| \mathbf{W} \mathbf{c} \|_2^2
	\right),
\end{align*}
i.e., both ID and OOD $g$'s are Gaussian with a common variance.
In particular,
\begin{align*}
\| \mathbf{W}^* \mathbf{c} \|_2^2 &= \mathbf{c}^\top \mathbf{A} \mathbf{c} \\
\| \mathbf{W}^*_{\rm sd} \mathbf{c} \|_2^2 &= \mathbf{c}^\top \boldsymbol{\Gamma} \mathbf{A} \boldsymbol{\Gamma} \mathbf{c}.
\end{align*}
\end{lemma}

\begin{proof}[Proof of Lemma~\ref{lemm:logit-cov}]
Under Assumption~\ref{assum:simplify}, 
$\boldsymbol{\Phi}( x ) = \boldsymbol{\mu}^{(y)} + \boldsymbol{\epsilon}_{\nullFeat}$ and 
$\boldsymbol{\Phi}( \bar{x} ) = \beta s_{yy} \boldsymbol{\mu}^{(y)}_{\locFeat} + \boldsymbol{\epsilon}_{\nullFeat}$, 
where $\boldsymbol{\epsilon}_{\nullFeat} \sim \mathscr{N}( \mathbf{0}, \sigma^2_{\nullFeat} \mathbf{I} )$.
Thus, $g = ( \mathbf{W} \mathbf{c} )^\top \boldsymbol{\Phi}$ is a linear transform of a Gaussian, and thus is Gaussian with the stated means and variance.
The final identities follow from Lemma~\ref{lemm:gram} and Lemma~\ref{thm:selfkd-iso}.
\end{proof}

\begin{lemma}
\label{thm:selfkd-iso}
Suppose $\mathbf{W}^*_{\rm sd}$ minimises $R_{\rm sd}$.
Under Assumptions~\ref{assum:id-feat}--\ref{assum:simplify},
$$\mathbf{W}^*_{\rm sd} = \mathbf{W}^* \boldsymbol{\Gamma},$$
so that $\mathbf{f}^{\rm sd}( x ) = \boldsymbol{\Gamma} \mathbf{f}( x )$ for every $x \in \XCal$.
\end{lemma}

\begin{proof}
By Lemma~\ref{lemm:opt-linear-self-kd},
\begin{align*}
\mathbf{W}^*_{\rm sd} &= ( \mathbf{R} + \lambda \mathbf{I} )^{-1} \mathbf{R} \mathbf{W}^* \\
&= \mathbf{W}^* - \lambda ( \mathbf{R} + \lambda \mathbf{I} )^{-1} \mathbf{W}^*,
\end{align*}
where $\mathbf{R} = \mathbb{E}_{x \sim \Pr_{\rm id}}[ \boldsymbol{\Phi}( x ) \boldsymbol{\Phi}( x )^\top ]$.
By Lemma~\ref{lemm:gram}, $( \mathbf{R} + \lambda \mathbf{I} )^{-1} \mathbf{W}^* = \tau^{-1} \mathbf{W}^* ( \mathbf{I} - \mathbf{F} )$.
Hence,
\begin{align*}
\mathbf{W}^*_{\rm sd} &= \mathbf{W}^* ( \mathbf{I} - \omega ( \mathbf{I} - \mathbf{F} ) ) \\
&= \mathbf{W}^* \boldsymbol{\Gamma}.
\end{align*}
As $\mathbf{F}$ is symmetric, so is $\boldsymbol{\Gamma}$.
Thus, $\mathbf{f}^{\rm sd}( x ) = ( \mathbf{W}^* \boldsymbol{\Gamma} )^\top \boldsymbol{\Phi}( x ) = \boldsymbol{\Gamma} \mathbf{f}( x )$.
\end{proof}

\begin{proposition}
\label{prop:first-order-iso}
Let $\mathbf{K} = \kappa \mathbf{I} + \boldsymbol{\Delta}$,
$\phi \defEq ( L + \kappa )^{-1} \cdot \kappa$, 
and $\xi \defEq ( L + \kappa )^{-2} \cdot L$. 
Under Assumptions~\ref{assum:id-feat}--\ref{assum:simplify},
\begin{align*}
	\mathbf{F} 			&= \phi \, \mathbf{I} + \xi \, \boldsymbol{\Delta} + \mathscr{O}( \| \mathbf{H} \|_F^2 ), \\
	\boldsymbol{\Gamma} &= \gamma_0' \, \mathbf{I} + \omega \xi \, \boldsymbol{\Delta} + \mathscr{O}( \| \mathbf{H} \|_F^2 ), \\
	\mathbf{A} 			&= a_0 \, \mathbf{I} + a_1 \, \boldsymbol{\Delta} + \mathscr{O}( \| \mathbf{H} \|_F^2 ),
\end{align*}
where
$\gamma_0' \defEq 1 - \omega \cdot ( 1 - \phi )$,
$a_0 = ({L \cdot \tau})^{-1} \cdot {\phi \cdot ( 1 - \phi )}$,
% = \frac{\kappa}{\tau ( L + \kappa )^2}$, 
and
$a_1 = ({L \cdot \tau})^{-1} \cdot {\xi \cdot ( 1 - 2 \phi )}$.
In particular, 
for $y' \neq y$,
\begin{align*}
    F_{yy} &= \phi + \mathscr{O}( \| \mathbf{H} \|_F^2 ) \\
    F_{y' y} &= \xi \cdot K_{y' y}  + \mathscr{O}( \| \mathbf{H} \|_F^2 ) \\
    &= \frac{L}{\tau \cdot ( L + \kappa )^2} \cdot s^2_{y' y} + \mathscr{O}( \| \mathbf{H} \|_F^2 ).
\end{align*}
% so the body's $\gamma_0 = ( 1 - \omega ) + \omega F_{yy}$ equals $\gamma_0'$ up to $\mathscr{O}( \| \mathbf{H} \|_F^2 )$, and 
\end{proposition}

\begin{proof}[Proof of Proposition~\ref{prop:first-order-iso}]
The result for $\mathbf{F}$ follows from Proposition~\ref{corr:opt-logits-first-order} with $K_{yy} = K_{y'y'} = \kappa$. 
The result for $\boldsymbol{\Gamma}$ follows since
\begin{align*}
\boldsymbol{\Gamma} &= ( 1 - \omega ) \mathbf{I} + \omega \mathbf{F} \\
&= ( 1 - \omega + \omega \cdot \phi ) \mathbf{I} + \omega \cdot \xi \cdot \boldsymbol{\Delta} + \mathscr{O}( \| \mathbf{H} \|_F^2 ).
\end{align*}
Finally,
the result for $\mathbf{A}$ follows since
\begin{align*}
(L \cdot \tau) \cdot \mathbf{A} &= \mathbf{F} ( \mathbf{I} - \mathbf{F} ) \\
&= ( \phi \mathbf{I} + \xi \boldsymbol{\Delta} ) ( ( 1 - \phi ) \mathbf{I} - \xi \boldsymbol{\Delta} ) + \mathscr{O}( \| \mathbf{H} \|_F^2 ) \\
&= \phi \cdot ( 1 - \phi ) \mathbf{I} + \xi \cdot ( 1 - 2 \phi ) \boldsymbol{\Delta} + \mathscr{O}( \| \mathbf{H} \|_F^2 ).
\end{align*}
\end{proof}

\begin{corollary}
\label{cor:qualifier-iso}
Under Assumptions~\ref{assum:id-feat}--\ref{assum:simplify}, 
for any $y \in [ L ]$ and $y', y'' \neq y$, the gap $g = f_{y'} - f_{y''}$ (with $\mathbf{W} = \mathbf{W}^*$) satisfies
\begin{align*}
\mathrm{std}( g ) &= \frac{ \sqrt{2} \, \sigma_{\nullFeat} \, \| \boldsymbol{\mu}^{(y')} \|_2 }{ \tau \cdot ( L + \kappa ) } \cdot \big( 1 + \mathscr{O}( \| \mathbf{H} \|_F ) \big) \\
\mathop{\mathbb{E}}_{x \sim \Pr_{\rm id}( \cdot \mid y )}[ g( x ) ] &= \frac{ 1 }{ \tau \cdot ( L + \kappa ) } \cdot \frac{L}{L + K_{yy}} \cdot ( s^2_{y' y} - s^2_{y'' y} ) + \mathscr{O}( \| \mathbf{H} \|_F^2 ) \\
\mathop{\mathbb{E}}_{\bar{x} \sim \Pr_{\rm ood}( \cdot \mid y )}[ g( \bar{x} ) ] &= - b_y \cdot \mathop{\mathbb{E}}_{x \sim \Pr_{\rm id}( \cdot \mid y )}[ g( x ) ].
\end{align*}
That is, the standard deviation is $\propto \sigma_{\nullFeat} \| \boldsymbol{\mu}^{(y')} \|_2$, and the expected gap is $\propto \frac{L}{L + K_{yy}} ( s^2_{y' y} - s^2_{y'' y} )$.
\end{corollary}

\begin{proof}[Proof of Corollary~\ref{cor:qualifier-iso}]
Let $\mathbf{c} = \mathbf{e}_{y'} - \mathbf{e}_{y''}$.
For the first claim, by Lemma~\ref{lemm:logit-cov} and Proposition~\ref{prop:first-order-iso}, 
\begin{align*}
	\mathrm{std}( g )^2 &= \sigma^2_{\nullFeat} \mathbf{c}^\top \mathbf{A} \mathbf{c} \mathbf{H} \\
	&= \sigma^2_{\nullFeat} ( 2 a_0 - 2 a_1 K_{y' y''} ) + \mathscr{O}( \| \mathbf{H} \|_F^2 ) \\
	&= 2 \sigma^2_{\nullFeat} a_0 ( 1 + \mathscr{O}( \| \mathbf{H} \|_F ) ).
\end{align*}
Further, the scaling factor $a_0$ is expressible as
\begin{align*}
	\sqrt{a_0} &= \frac{\sqrt{\kappa / \tau}}{L + \kappa} \\
	&= \frac{\| \boldsymbol{\mu}^{(y')} \|_2}{\tau \cdot ( L + \kappa )},
\end{align*}
where the second line is since
$\tau \cdot \kappa = \| \boldsymbol{\mu}^{(y')} \|_2^2$.

For the second claim,
by Proposition~\ref{corr:opt-logits-first-order-informal} and Proposition~\ref{prop:first-order-iso}, 
\begin{align*}
	\mathop{\mathbb{E}}_{x \sim \Pr_{\rm id}( \cdot \mid y )}[ g ] &= \mathbf{c}^\top \mathbf{F} \mathbf{e}_y \\
	&= \xi \cdot ( K_{y' y} - K_{y'' y} ) + \mathscr{O}( \| \mathbf{H} \|_F^2 ).
\end{align*}
Further, the scaling factor
$$ \frac{\xi}{\tau} = \frac{1}{\tau \cdot ( L + \kappa )} \cdot \frac{L}{L + \kappa}. $$

For the final claim,
\begin{align*}
	\mathop{\mathbb{E}}_{\bar{x} \sim \Pr_{\rm ood}( \cdot \mid y )}[ g ] &= b_y \cdot \mathbf{c}^\top ( \mathbf{I} - \mathbf{F} ) \mathbf{e}_y \\
	&= - b_y \cdot \mathbf{c}^\top \mathbf{F} \mathbf{e}_y,
\end{align*}
where the second line is since $\mathbf{c}^\top \mathbf{e}_y = 0$.
\end{proof}

\begin{proposition}
\label{thm:sep-gain}
Under Assumptions~\ref{assum:id-feat}--\ref{assum:simplify}, let $\kappa, \phi, \xi, \gamma_0'$ be as in Proposition~\ref{prop:first-order-iso}, and fix $y \in [ L ]$ and $y', y'' \neq y$ with $s_{y' y} \neq s_{y'' y}$.
For a logit gap $g_{ab} \defEq f_a - f_b$, write $g_{ab}$ and $g^{\rm sd}_{ab}$ for the scores computed with $\mathbf{W}^*$ and $\mathbf{W}^*_{\rm sd}$ respectively.
Then
\begin{align*}
    \frac{ \delta_y( g^{\rm sd}_{y' y''} ) }{ \delta_y( g_{y' y''} ) } &= \Big( 1 + \frac{\omega}{\gamma_0'} \Big( \phi - \frac{b_y}{1 + b_y} \Big) \Big) \big( 1 + \mathscr{O}( \| \mathbf{H} \|_F ) \big) \\
    \frac{ \delta_y( g^{\rm sd}_{y y'} ) }{ \delta_y( g_{y y'} ) } &= 1 + \mathscr{O}( \| \mathbf{H} \|_F^2 ) ,
\end{align*}
and the leading factor in the first ratio is strictly greater than $1$.
\end{proposition}

\begin{proof}[Proof of Proposition~\ref{thm:sep-gain}]
By Proposition~\ref{corr:opt-logits-first-order-informal},
\begin{align*}
\mathop{\mathbb{E}}_{x \sim \Pr_{\rm id}( \cdot \mid y )}[ \mathbf{f}( x ) ] &= \mathbf{F} \mathbf{e}_y \\
\mathop{\mathbb{E}}_{\bar{x} \sim \Pr_{\rm ood}( \cdot \mid y)}[ \mathbf{f}( \bar{x} ) ] &= b_y ( \mathbf{I} - \mathbf{F} ) \mathbf{e}_y.
\end{align*}
Recall that by Lemma~\ref{thm:selfkd-iso},
$\mathbf{f}^{\rm sd} = \boldsymbol{\Gamma} \mathbf{f}$.
Finally, recall that Lemma~\ref{lemm:logit-cov} provides the standard deviations of $g$.

To analyse the separation, we analyse the numerator and denominator in turn.

\emph{Mean logits.}
We use the expansions of Proposition~\ref{prop:first-order-iso}, writing $\gamma_0$ for $\gamma_0'$.
Multiplying out the expansions,
\begin{align*}
\mathbf{F} \mathbf{e}_y &= \phi \mathbf{e}_y + \xi \boldsymbol{\Delta} \mathbf{e}_y + \mathscr{O}( \| \mathbf{H} \|_F^2 ) \\
( \mathbf{I} - \mathbf{F} ) \mathbf{e}_y &= ( 1 - \phi ) \mathbf{e}_y - \xi \boldsymbol{\Delta} \mathbf{e}_y + \mathscr{O}( \| \mathbf{H} \|_F^2 ) \\
\boldsymbol{\Gamma} \mathbf{F} \mathbf{e}_y &= \gamma_0 \phi \, \mathbf{e}_y + ( \gamma_0 + \omega \phi ) \, \xi \boldsymbol{\Delta} \mathbf{e}_y + \mathscr{O}( \| \mathbf{H} \|_F^2 ) \\
\boldsymbol{\Gamma} ( \mathbf{I} - \mathbf{F} ) \mathbf{e}_y &= \gamma_0 ( 1 - \phi ) \, \mathbf{e}_y - \big( \gamma_0 - \omega ( 1 - \phi ) \big) \, \xi \boldsymbol{\Delta} \mathbf{e}_y + \mathscr{O}( \| \mathbf{H} \|_F^2 ).
\end{align*}
Thus, for the non-target gap ($\mathbf{c} = \mathbf{e}_{y'} - \mathbf{e}_{y''}$, $\mathbf{c}^\top \mathbf{e}_y = 0$), with $\Delta' \defEq K_{y' y} - K_{y'' y} \neq 0$,
\begin{align*}
\mathop{\mathbb{E}}_{x \sim \Pr_{\rm id}( \cdot \mid y )}[ g_{y' y''} ] &= \xi \Delta' + \mathscr{O}( \| \mathbf{H} \|_F^2 ) \\
\mathop{\mathbb{E}}_{x \sim \Pr_{\rm id}( \cdot \mid y )}[ g^{\rm sd}_{y' y''} ] &= ( \gamma_0 + \omega \phi ) \xi \Delta' + \mathscr{O}( \| \mathbf{H} \|_F^2 ), \\
\mathop{\mathbb{E}}_{\bar{x} \sim \Pr_{\rm ood}( \cdot \mid y)}[ g_{y' y''} ] &= - b_y \xi \Delta' + \mathscr{O}( \| \mathbf{H} \|_F^2 ) \\
\mathop{\mathbb{E}}_{\bar{x} \sim \Pr_{\rm ood}( \cdot \mid y)}[ g^{\rm sd}_{y' y''} ] &= - b_y ( \gamma_0 - \omega ( 1 - \phi ) ) \xi \Delta' + \mathscr{O}( \| \mathbf{H} \|_F^2 ).
\end{align*}

\emph{Noise.} For any $a \neq b$ and $\mathbf{c} = \mathbf{e}_a - \mathbf{e}_b$, 
$\mathbf{c}^\top \boldsymbol{\Delta} \mathbf{c} = - 2 K_{ab}$
and
$\mathbf{c}^\top \mathbf{A} \boldsymbol{\Delta} \mathbf{c} = a_0 \mathbf{c}^\top \boldsymbol{\Delta} \mathbf{c} + \mathscr{O}( \| \mathbf{H} \|_F^2 ) = - 2 a_0 K_{ab} + \mathscr{O}( \| \mathbf{H} \|_F^2 )$.
Thus, by Lemma~\ref{lemm:logit-cov} and Proposition~\ref{prop:first-order-iso},
$$ \mathrm{std}( g_{ab} )^2 = \sigma^2_{\nullFeat} \cdot \mathbf{c}^\top \mathbf{A} \mathbf{c} = \sigma^2_{\nullFeat} ( 2 a_0 - 2 a_1 K_{ab} ) + \mathscr{O}( \| \mathbf{H} \|_F^2 ). $$
Similarly,
with $\boldsymbol{\Gamma} \mathbf{c} = \gamma_0 \mathbf{c} + \omega \xi \boldsymbol{\Delta} \mathbf{c} + \mathscr{O}( \| \mathbf{H} \|_F^2 )$,
$$ \mathrm{std}( g^{\rm sd}_{ab} )^2 = \sigma^2_{\nullFeat} \, ( \boldsymbol{\Gamma} \mathbf{c} )^\top \mathbf{A} ( \boldsymbol{\Gamma} \mathbf{c} )
 = \sigma^2_{\nullFeat} \big( \gamma_0^2 ( 2 a_0 - 2 a_1 K_{ab} ) - 4 \gamma_0 \omega \xi a_0 K_{ab} \big) + \mathscr{O}( \| \mathbf{H} \|_F^2 ) . $$
Dividing these expressions, we conclude that
\begin{align*}
\numberthis
\label{eqn:sd-ratio}
\mathrm{std}( g^{\rm sd}_{ab} ) &= \big( \gamma_0 - \omega \xi K_{ab} \big) \mathrm{std}( g_{ab} ) + \mathscr{O}( \| \mathbf{H} \|_F^2 ) \cdot \mathrm{std}( g_{ab} ) \\
&= \gamma_0 \, \mathrm{std}( g_{ab} ) \big( 1 + \mathscr{O}( \| \mathbf{H} \|_F ) \big).
\end{align*}

\emph{Non-target gap.} By the analysis of the mean logits, the numerator of $\delta_y$ 
--
namely, $\mathop{\mathbb{E}}_{x \sim \Pr_{\rm id}( \cdot \mid y )}[ g_{y' y''}( x ) ] - \mathop{\mathbb{E}}_{\bar{x} \sim \Pr_{\rm ood}( \cdot \mid y )}[ g_{y' y''}( \bar{x} ) ]$
--
is $( 1 + b_y ) \xi \Delta' + \mathscr{O}( \| \mathbf{H} \|_F^2 )$ for standard one-hot training, 
and $[ ( \gamma_0 + \omega \phi ) + b_y ( \gamma_0 - \omega ( 1 - \phi ) ) ] \xi \Delta' + \mathscr{O}( \| \mathbf{H} \|_F^2 )$ for self-distillation.
By~\eqref{eqn:sd-ratio} the denominator of $\delta_y$ is multiplied by $\gamma_0 ( 1 + \mathscr{O}( \| \mathbf{H} \|_F ) )$. 
Hence,
\begin{align*}
	\frac{ \delta_y( g^{\rm sd}_{y' y''} ) }{ \delta_y( g_{y' y''} ) } &= \frac{ ( \gamma_0 + \omega \phi ) + b_y ( \gamma_0 - \omega ( 1 - \phi ) ) }{ \gamma_0 ( 1 + b_y ) } \big( 1 + \mathscr{O}( \| \mathbf{H} \|_F ) \big) \\
	&= \Big( 1 + \frac{\omega}{\gamma_0} \cdot \frac{ \phi - b_y ( 1 - \phi ) }{ 1 + b_y } \Big) \big( 1 + \mathscr{O}( \| \mathbf{H} \|_F ) \big).
\end{align*}
Note here that the remainder is $\mathscr{O}( \| \mathbf{H} \|_F )$, rather than $\mathscr{O}( \| \mathbf{H} \|_F^2 )$: 
this is because the leading term in the numerator $\xi \Delta'$ is itself $\mathscr{O}( \| \mathbf{H} \|_F )$, and non-zero if $s_{y', y} \neq s_{y'', y}$.

Finally, note that $\phi > b_y ( 1 - \phi ) \iff \phi / ( 1 - \phi ) = \kappa / L > b_y \iff \kappa > \beta s^2_{yy} / \tau$, which holds as $\kappa = K_{yy} \geq s^2_{yy} / \tau > \beta s^2_{yy} / \tau$ (i.e., the expected ID target logit exceeds the expected OOD target logit).
Therefore, the ratio is $> 1$.

\emph{Target gap.} With $\mathbf{c} = \mathbf{e}_y - \mathbf{e}_{y'}$, by the analysis of the mean logits,
\begin{align*}
	\frac{ \mathop{\mathbb{E}}_{x \sim \Pr_{\rm id}( \cdot \mid y )}[ g^{\rm sd}_{y y'}( {x} ) ] }{ \mathop{\mathbb{E}}_{x \sim \Pr_{\rm id}( \cdot \mid y )}[ g_{y y'}( {x} ) ] } &= \frac{ \gamma_0 \phi - ( \gamma_0 + \omega \phi ) \xi K_{y' y} + \mathscr{O}( \| \mathbf{H} \|_F^2 ) }{ \phi - \xi K_{y' y} + \mathscr{O}( \| \mathbf{H} \|_F^2 ) } \\
	&= \gamma_0 - \omega \xi K_{y y'} + \mathscr{O}( \| \mathbf{H} \|_F^2 ),
\end{align*}
and similarly,
\begin{align*}
\frac{\mathop{\mathbb{E}}_{\bar{x} \sim \Pr_{\rm ood}( \cdot \mid y)}[ g^{\rm sd}_{y y'}( \bar{x} ) ]}{\mathop{\mathbb{E}}_{\bar{x} \sim \Pr_{\rm ood}( \cdot \mid y)}[ g_{y y'}( \bar{x} ) ]} &= \big( \gamma_0 ( 1 - \phi ) + ( \gamma_0 - \omega ( 1 - \phi ) ) \xi K_{y' y} \big) / \big( ( 1 - \phi ) + \xi K_{y' y} \big) + \mathscr{O}( \| \mathbf{H} \|_F^2 ) \\
&= \gamma_0 - \omega \xi K_{y y'} + \mathscr{O}( \| \mathbf{H} \|_F^2 ).
\end{align*}
For the standard deviation,~\eqref{eqn:sd-ratio} with $a = y$, $b = y'$ gives the 
\emph{same} factor $\gamma_0 - \omega \xi K_{y y'} + \mathscr{O}( \| \mathbf{H} \|_F^2 )$ .
Put together,
both the numerator and denominator of $\delta_y( g_{y y'} )$ are multiplied by $c \defEq\gamma_0 - \omega \xi K_{y y'}$, up to a factor of $\mathscr{O}( \| \mathbf{H} \|_F^2 )$.
Further, the numerator $\mathop{\mathbb{E}}_{x \sim \Pr_{\rm id}( \cdot \mid y )}[ g_{y y'} ] - \mathop{\mathbb{E}}_{\bar{x} \sim \Pr_{\rm ood}( \cdot \mid y)}[ g_{y y'} ] = \phi - b_y ( 1 - \phi ) + \mathscr{O}( \| \mathbf{H} \|_F )$ does not vanish as $\| \mathbf{H} \|_F \to 0$.
Thus, the ratio $\frac{ \delta_y( g^{\rm sd}_{y y'} ) }{ \delta_y( g_{y y'} ) }$ changes by $1 + \mathscr{O}( \| \mathbf{H} \|_F^2 )$:
\begin{align*}
    \frac{ \delta_y( g^{\rm sd}_{y y'} ) }{ \delta_y( g_{y y'} ) } &= \frac{ c + \mathscr{O}( \| \mathbf{H} \|_F^2 ) }{ c + \mathscr{O}( \| \mathbf{H} \|_F^2 ) } \\
    &= 1 + \mathscr{O}( \| \mathbf{H} \|_F^2 ).
\end{align*}
\end{proof}

%% file: sec/suppl/proofs.tex
\begin{proof}[Proof of Proposition~\ref{prop:sandwich-bound}]
By definition of the induced ranking for the student,
\begin{align*}
    \ell_{\rm epl}( \mathbf{p}^{\rm t}( x ), \mathbf{f}( x ) ) &= \mathop{\mathbb{E}}_{ \boldsymbol{\pi} \sim {p}^{\rm t}_{\rm pl}( \cdot \mid x ) }\left[ -\log p_{\rm pl}( \boldsymbol{\pi} \mid x ) \right] \\
    &= \mathop{\mathbb{E}}_{ \boldsymbol{\pi} \sim {p}^{\rm t}_{\rm pl}( \cdot \mid x ) }\left[ \sum_{i = 1}^{L} -\log \frac{p_{\pi( i )}( x )}{\sum_{j = i}^{L} p_{\pi( j )}( x )} \right] \\
    &= \mathop{\mathbb{E}}_{ \boldsymbol{\pi} \sim {p}^{\rm t}_{\rm pl}( \cdot \mid x ) }\left[ \sum_{i = 1}^{L} -\log {p_{\pi( i )}( x )} + \log \left( {\sum_{j = i}^{L} p_{\pi( j )}( x )} \right) \right].
\end{align*}
For the first inequality, note that for every $i \in [ L ]$,
$$ {p_{\pi( i )}( x )} \leq {\sum_{j = i}^{L} p_{\pi( j )}( x )} \implies 
-\log \frac{p_{\pi( i )}( x )}{\sum_{j = i}^{L} p_{\pi( j )}( x )} \geq 0. $$
Thus, we may lower bound the expected ranking loss by the first term of the summation:
\begin{align*}
    \ell_{\rm epl}( \mathbf{p}^{\rm t}( x ), \mathbf{f}( x ) ) &= \mathop{\mathbb{E}}_{ \boldsymbol{\pi} \sim {p}^{\rm t}_{\rm pl}( \cdot \mid x ) }\left[ \sum_{i = 1}^{L} -\log \frac{p_{\pi( i )}( x )}{\sum_{j = i}^{L} p_{\pi( j )}( x )} \right] \\
    &\geq \mathop{\mathbb{E}}_{ \boldsymbol{\pi} \sim {p}^{\rm t}_{\rm pl}( \cdot \mid x ) }\left[ -\log \frac{p_{\pi( 1 )}( x )}{\sum_{j = 1}^{L} p_{\pi( j )}( x )} \right] \\
    &= \mathop{\mathbb{E}}_{ \boldsymbol{\pi} \sim {p}^{\rm t}_{\rm pl}( \cdot \mid x ) }\left[ -\log {p_{\pi( 1 )}( x )} \right].
\end{align*}
Now we use the following elementary fact
% \todoakm{Reference?}
about the Plackett-Luce distribution~\citep{Cao:2007}:
the distribution of the top-ranked item is simply the first term in the product~\eqref{eqn:induced-ranking}, i.e.,
$$ ( \forall \boldsymbol{\pi} \in \mathfrak{S}( \YCal ) ) \, {p}^{\rm t}_{\rm pl}( \pi( 1 ) \mid x ) = p^{\rm t}_{\pi( 1 )}( x ). $$
Thus,
$$ \ell_{\rm epl}( \mathbf{p}^{\rm t}( x ), \mathbf{f}( x ) ) \geq \mathop{\mathbb{E}}_{ y \sim {p}^{\rm t}( \cdot \mid x ) }\left[ -\log {p( y \mid x )} \right] = \ell_{\rm kd}( \mathbf{p}^{\rm t}( x ), \mathbf{f}( x ) ). $$
For the second inequality, note that for every $i \in [ L ]$, 
$$ {\sum_{j = i}^{L} p_{\pi( j )}( x )} \leq 1 \implies \log\left( {\sum_{j = i}^{L} p_{\pi( j )}( x )} \right) \leq 0. $$
Thus, we may upper bound each term of the summation:
\begin{align*}
    \ell_{\rm epl}( \mathbf{p}^{\rm t}( x ), \mathbf{f}( x ) ) &= \mathop{\mathbb{E}}_{ \boldsymbol{\pi} \sim {p}^{\rm t}_{\rm pl}( \cdot \mid x ) }\left[ \sum_{i = 1}^{L} -\log \frac{p_{\pi( i )}( x )}{\sum_{j = i}^{L} p_{\pi( j )}( x )} \right] \\
    &\leq \mathop{\mathbb{E}}_{ \boldsymbol{\pi} \sim {p}^{\rm t}_{\rm pl}( \cdot \mid x ) }\left[ \sum_{i = 1}^{L} -\log {p_{\pi( i )}( x )} \right] \\
    &= \sum_{y' \in \YCal} -\log {p_{y'}( x )} \text{ since the summation is independent of } \boldsymbol{\pi} \\
    % &= L \cdot \sum_{y' \in \YCal} \frac{1}{L} \cdot -\log {p( y' \mid x )} \\
    &\leq \frac{1}{p^{\rm t}_{\rm min}( x )} \cdot \sum_{y' \in \YCal} p^{\rm t}_{y'}( x ) \cdot -\log {p( y' \mid x )} \\
    &= \frac{1}{p^{\rm t}_{\rm min}( x )} \cdot \ell_{\rm kd}( \mathbf{p}^{\rm t}( x ), \mathbf{f}( x ) ).
\end{align*}
\end{proof}

\begin{proof}[Proof of Proposition~\ref{prop:epl-loss}]
We begin with the definition of the EPL loss~\eqref{eqn:expected-ranking-loss}, and the induced ranking distribution~\eqref{eqn:induced-ranking}:
\begin{align*}
    \ell_{\rm epl}( \mathbf{p}^{\rm t}( x ), \mathbf{f}( x ) ) &= \mathop{\mathbb{E}}_{ \boldsymbol{\pi} \sim {p}^{\rm t}_{\rm pl}( \cdot \mid x ) }\left[ -\log p_{\rm pl}( \boldsymbol{\pi} \mid x ) \right] \\
    &= \mathop{\mathbb{E}}_{ \boldsymbol{\pi} \sim {p}^{\rm t}_{\rm pl}( \cdot \mid x ) }\left[ \sum_{i = 1}^{L} -\log \frac{p_{\pi( i )}( x )}{\sum_{j = i}^{L} p_{\pi( j )}( x )} \right] \\
    &= \sum_{i = 1}^{L} \mathop{\mathbb{E}}_{ \boldsymbol{\pi} \sim {p}^{\rm t}_{\rm pl}( \cdot \mid x ) }\left[ -\log \frac{p_{\pi( i )}( x )}{\sum_{j = i}^{L} p_{\pi( j )}( x )} \right] \text{ by linearity of expectation } \\
    &= \sum_{i = 1}^{L} \mathop{\mathbb{E}}_{ \boldsymbol{\pi} \sim {p}^{\rm t}_{\rm pl}( \cdot \mid x ) }\left[ -\log \frac{p_{\pi( i )}( x )}{1 - \sum_{j = 1}^{i - 1} p_{\pi( j )}( x )} \right] \text{ since } \sum_{j = 1}^{L} p_{\pi( j )}( x ) = 1 \\
    &= \sum_{i = 1}^{L} \mathop{\mathbb{E}}_{ \boldsymbol{\pi}_{\leq i} \sim {p}^{\rm t}_{\rm pl}( \cdot \mid x ) }\left[ -\log \frac{p_{\pi( i )}( x )}{1 - \sum_{j = 1}^{i - 1} p_{\pi( j )}( x )} \right] \text{ since the summand does not depend on } \boldsymbol{\pi}_{> i} \\
    &= \sum_{i = 1}^{L} \mathop{\mathbb{E}}_{ \boldsymbol{\pi}_{< i} \sim {p}^{\rm t}_{\rm pl}( \cdot \mid x )} \mathop{\mathbb{E}}_{ {\pi}( {i} ) \sim {p}^{\rm t}_{\rm pl}( \cdot \mid x, \boldsymbol{\pi}_{<i} )} \left[ -\log \frac{p_{\pi( i )}( x )}{1 - \sum_{j = 1}^{i - 1} p_{\pi( j )}( x )} \right],
\end{align*}
where $\boldsymbol{\pi}_{<i} \defEq ( \pi( 1 ), \ldots, \pi( i - 1 ) )$.
Now observe that the innermost expectation involves the conditional distribution ${p}^{\rm t}_{\rm pl}( \cdot \mid x, \boldsymbol{\pi}_{<i} ) \in \Delta( \YCal )$, which by definition of the Plackett-Luce distribution is
$$ ( \forall y \in \YCal ) \, {p}^{\rm t}_{\rm pl}( y \mid x, \boldsymbol{\pi}_{<i} ) = \indicator{ y \notin \boldsymbol{\pi}_{<i} } \cdot \frac{p^{\rm t}_{y}( x )}{\sum_{y' \notin \boldsymbol{\pi}_{<i}} p^{\rm t}_{y'}( x )} = \indicator{ y \notin \boldsymbol{\pi}_{<i} } \cdot \frac{p^{\rm t}_{y}( x )}{1 - \sum_{j = 1}^{i - 1} p^{\rm t}_{\pi( j )}( x )}. $$
This exactly equals the induced conditional ranking distribution of~\eqref{eqn:conditional-ranking}, as does the term in the logarithm.
Thus,
\begin{align*}
    \ell_{\rm epl}( \mathbf{p}^{\rm t}( x ), \mathbf{f}( x ) ) &= \sum_{i = 1}^{L} \mathop{\mathbb{E}}_{ \boldsymbol{\pi}_{< i} \sim {p}^{\rm t}_{\rm pl}( \cdot \mid x )} \mathop{\mathbb{E}}_{ {\pi}( {i} ) \sim {p}^{\rm t}_{\rm pl}( \cdot \mid x, \boldsymbol{\pi}_{<i} )} \left[ -\log \frac{p_{\pi( i )}( x )}{1 - \sum_{j = 1}^{i - 1} p_{\pi( j )}( x )} \right] \\
    &= \sum_{i = 1}^{L} \mathop{\mathbb{E}}_{ \boldsymbol{\pi}_{< i} \sim {p}^{\rm t}_{\rm pl}( \cdot \mid x )} \mathop{\mathbb{E}}_{ y \sim \mathbf{p}^{\rm t}( x, \boldsymbol{\pi}_{<i} )} \left[ -\log p_{y}( x, \boldsymbol{\pi}_{<i} ) \right] \\
    &= \sum_{i = 1}^{L} \mathop{\mathbb{E}}_{ \boldsymbol{\pi}_{< i} \sim {p}^{\rm t}_{\rm pl}( \cdot \mid x )}\left[ \ell_{\rm kd}( \mathbf{p}^{\rm t}( x, \boldsymbol{\pi}_{<i} ), \mathbf{p}( x, \boldsymbol{\pi}_{<i} ) ) \right] \\
    &= \sum_{i = 1}^{L} \mathop{\mathbb{E}}_{ \boldsymbol{\pi} \sim {p}^{\rm t}_{\rm pl}( \cdot \mid x )}\left[ \ell_{\rm kd}( \mathbf{p}^{\rm t}( x, \boldsymbol{\pi}_{<i} ), \mathbf{p}( x, \boldsymbol{\pi}_{<i} ) ) \right] \text{ by reintroducing the successor rankings } \\
    &= \mathop{\mathbb{E}}_{ \boldsymbol{\pi} \sim {p}^{\rm t}_{\rm pl}( \cdot \mid x )}\left[ \sum_{i = 1}^{L} \ell_{\rm kd}( \mathbf{p}^{\rm t}( x, \boldsymbol{\pi}_{<i} ), \mathbf{p}( x, \boldsymbol{\pi}_{<i} ) ) \right] \text{ by linearity of expectation. }
\end{align*}
Note that in the penultimate line, we safely reintroduced the successor ranking elements $\boldsymbol{\pi}_{\geq i}$, as the summand $\ell_{\rm kd}( \mathbf{p}^{\rm t}( x, \boldsymbol{\pi}_{<i} ), \mathbf{p}( x, \boldsymbol{\pi}_{<i} ) )$ is independent of them.
\end{proof}

\begin{proof}[Proof of Proposition~\ref{corr:opt-logits-first-order-informal}]
Applying Proposition~\ref{eqn:opt-logits} with $\mathbf{T} = \tau \mathbf{I}$,
\begin{align*}
\mathop{\mathbb{E}}_{x \sim \Pr_{\rm id}( \cdot \mid y )}[ \mathbf{f}( x ) ] &= ( L \cdot \mathbf{I} + \mathbf{K} )^{-1} \mathbf{K} \mathbf{e}_y \\
&= \mathbf{F} \mathbf{e}_y \\
\mathop{\mathbb{E}}_{\bar{x} \sim \Pr_{\rm ood}( \cdot \mid y )}[ \mathbf{f}( \bar{x} ) ] &= \beta \cdot ( L \cdot \mathbf{I} + \mathbf{K} )^{-1} \mathbf{N} \mathbf{e}_y \\
&= \beta \cdot N_{yy} \cdot ( L \cdot \mathbf{I} + \mathbf{K} )^{-1} \mathbf{e}_y,
\end{align*}
where the second line follows since $\mathbf{N}$ is diagonal.
Now, observe that
\begin{align*}
( L \cdot \mathbf{I} + \mathbf{K} )^{-1} \mathbf{K} &= ( L \cdot \mathbf{I} + \mathbf{K} )^{-1} ( L \mathbf{I} + \mathbf{K} - L \mathbf{I} ) \\
&= \mathbf{I} - L \cdot ( L \cdot \mathbf{I} + \mathbf{K} )^{-1},
\end{align*}
and so
\begin{align*}
L \cdot ( L \cdot \mathbf{I} + \mathbf{K} )^{-1} &= \mathbf{I} - \mathbf{F},
\end{align*}
and $\beta \cdot L^{-1} \cdot N_{yy} = \beta \cdot ( L \cdot \tau )^{-1} \cdot s^2_{yy} = b_y$.
\end{proof}

\begin{proof}[Proof of Proposition~\ref{corr:opt-logits-first-order-self-kd-informal}]
This result follows from Lemma~\ref{thm:selfkd-iso} and Proposition~\ref{thm:sep-gain}.
For part (b), note that Proposition~\ref{thm:sep-gain} presents the results using $\phi$ and $\gamma_0' = ( 1 - \omega ) + \omega \phi$,
rather than $F_{yy}$ and $\gamma_0$.
However, since $F_{yy} = \phi + \mathscr{O}( \| \mathbf{H} \|_F^2 )$ (Proposition~\ref{prop:first-order-iso}), this results in a change of $\mathscr{O}( \| \mathbf{H} \|_F^2 )$.
Finally, note that the scaling factor $\frac{ \phi - b_y \cdot ( 1 - \phi ) }{ 1 + b_y }$ is equally expressible as $\phi - \frac{ b_y }{ 1 + b_y }$.
\end{proof}

\begin{proof}[Proof of Proposition~\ref{corr:opt-logits-first-order}]
Define 
$\boldsymbol{\Delta} \defEq \mathbf{K} - \mathbf{D}$,
where 
$\mathbf{D} \defEq \mathrm{diag}( \mathbf{K} )$.
% $\mathbf{D} \defEq \mathrm{diag}( \{ (\boldsymbol{\mu}^{(y)})^\top \mathbf{T}^{-1} \boldsymbol{\mu}^{(y)} \colon y \in [ L ] \} )$.
% \todoakm{Fix $\Delta$ notation.}
Finally, let 
$\mathbf{Z} \defEq L \cdot \mathbf{I} + \mathbf{D}$.
By Lemma~\ref{lemm:k-eigenvalue}, the 
(absolute value of the) eigenvalues of $\mathbf{Z}^{-1} \boldsymbol{\Delta}$ are less than $1$. 
Thus, we have
\begin{align*}
( L \cdot \mathbf{I} + \mathbf{K} )^{-1} &= ( \mathbf{Z} + \boldsymbol{\Delta} )^{-1} \\
&= ( \mathbf{Z} \cdot ( \mathbf{I} + \mathbf{Z}^{-1} \boldsymbol{\Delta} ) )^{-1}  \\
&= ( \mathbf{I} + \mathbf{Z}^{-1} \boldsymbol{\Delta} )^{-1} \mathbf{Z}^{-1}  \\
&= \sum_{k = 0}^{\infty} {( -1 )^k} \cdot ( {\mathbf{Z}^{-1}} \boldsymbol{\Delta} )^k \mathbf{Z}^{-1}.
\end{align*}
Note that 
$\mathbf{K} = \mathbf{D} + \boldsymbol{\Delta}$.
Thus, we have
\begin{align*}
( L \cdot \mathbf{I} + \mathbf{K} )^{-1} \cdot ( \mathbf{D} + \boldsymbol{\Delta} ) &= \sum_{k = 0}^{\infty} {( -1 )^k} \cdot 
( {\mathbf{Z}^{-1}} \boldsymbol{\Delta} )^k \mathbf{Z}^{-1}
\cdot ( \mathbf{D} + \boldsymbol{\Delta} ),
\end{align*}
and so
\begin{align*}
    \mathop{\mathbb{E}}_{x \sim \Pr_{\rm id}( \cdot \mid y )}[ \mathbf{f}( x; \mathbf{W}^* ) ] 
    &= ( L \cdot \mathbf{I} + \mathbf{K} )^{-1} ( \mathbf{D} + \boldsymbol{\Delta} ) \mathbf{e}_y \\
    &= \sum_{k = 0}^{\infty} {( -1 )^k} \cdot ( {\mathbf{Z}^{-1}} \boldsymbol{\Delta} )^k \mathbf{Z}^{-1}
    ( \mathbf{D} + \boldsymbol{\Delta} ) \mathbf{e}_y.
\end{align*}
The first-order approximation to the logits is then
\begin{align*}
    \mathop{\mathbb{E}}_{x \sim \Pr_{\rm id}( \cdot \mid y )}[ \mathbf{f}( x; \mathbf{W}^* ) ] 
    % &= ( L \cdot \mathbf{I} + \mathbf{K} )^{-1} ( \mathbf{D} + \boldsymbol{\Delta} ) \mathbf{e}_y \\
    &= ( \mathbf{I} - {\mathbf{Z}^{-1}} \boldsymbol{\Delta} ) \mathbf{Z}^{-1}
    ( \mathbf{D} + \boldsymbol{\Delta} ) \mathbf{e}_y \\
    &= \mathbf{Z}^{-1}
    ( \mathbf{D} + \boldsymbol{\Delta} ) \mathbf{e}_y
    - {\mathbf{Z}^{-1}} \boldsymbol{\Delta}
    \mathbf{Z}^{-1} \mathbf{D} \mathbf{e}_y
    - ( {\mathbf{Z}^{-1}} \boldsymbol{\Delta} )^2
    \mathbf{e}_y \\
    &= \mathbf{Z}^{-1} \mathbf{D} \mathbf{e}_y 
    + \mathbf{Z}^{-1}  \boldsymbol{\Delta} \mathbf{e}_y 
    - {\mathbf{Z}^{-1}} \boldsymbol{\Delta}
    \mathbf{Z}^{-1} \mathbf{D} \mathbf{e}_y
    + \mathscr{O}( \| \mathbf{Z}^{-1} \boldsymbol{\Delta} \|_2^2 ) \mathbf{e}_y.
\end{align*}
Thus, dropping higher-order terms,
for any $y' \in [ L ]$, 
the first-order approximation is
\begin{align*}
    \mathop{\mathbb{E}}_{x \sim \Pr_{\rm id}( \cdot \mid y )}[ {f}_{y'}( x; \mathbf{W}^* ) ] &= 
    \frac{K_{y, y}}{Z_{y, y}} \cdot \indicator{ y = y' } +
    \frac{\Delta_{y', y}}{Z_{y', y'}} \cdot \left( 1 - \frac{K_{y, y}}{Z_{y, y}} \right) \cdot \indicator{ y \neq y' } \\
    &= 
    \frac{K_{y, y}}{Z_{y, y}} \cdot \indicator{ y = y' } +
    \frac{\Delta_{y', y}}{Z_{y', y'}} \cdot \frac{L}{L + K_{y, y}} \cdot \indicator{ y \neq y' } \\
    &= 
    \frac{1}{L + K_{y, y}} \cdot
    \left(
    {K_{y, y}} \cdot \indicator{ y = y' } +
    \frac{L}{L + K_{y', y'}} \cdot K_{y', y} \cdot \indicator{ y \neq y' } \right).
\end{align*}

For OOD samples,
we similarly have
\begin{align*}
    \mathop{\mathbb{E}}_{x \sim \Pr_{\rm ood}( \cdot \mid y )}[ \mathbf{f}( x; \mathbf{W}^* ) ] 
    &= \beta \cdot s_{yy} \cdot ( L \cdot \mathbf{I} + \mathbf{K} )^{-1}
    \boldsymbol{\eta}^{(y)} \\
    &= \beta \cdot s_{yy} \cdot \sum_{k = 0}^{\infty} {( -1 )^k} \cdot ( {\mathbf{Z}^{-1}} \boldsymbol{\Delta} )^k
    \mathbf{Z}^{-1}
    \boldsymbol{\eta}^{(y)}.
    % ( \mathbf{M}^\top \mathbf{T}^{-1} \boldsymbol{\mu}^{(y)}_{\locFeat} ).
\end{align*}
The first-order approximation 
to the logits
is
\begin{align*}
\mathop{\mathbb{E}}_{x \sim \Pr_{\rm ood}( \cdot \mid y )}[ \mathbf{f}( x; \mathbf{W}^* ) ] &= \beta \cdot s_{yy} \cdot ( \mathbf{Z}^{-1} - \mathbf{Z}^{-1} \boldsymbol{\Delta} \mathbf{Z}^{-1} ) \boldsymbol{\eta}^{(y)} \\
&= \beta \cdot s_{yy} \cdot \mathbf{Z}^{-1} \boldsymbol{\eta}^{(y)} - \beta \cdot s_{yy} \cdot \mathbf{Z}^{-1} \boldsymbol{\Delta} \mathbf{Z}^{-1} \boldsymbol{\eta}^{(y)}.
\end{align*}
Thus, for any $y' \in [ L ]$,
the first-order approximation is
\begin{align*}
    \mathop{\mathbb{E}}_{x \sim \Pr_{\rm ood}( \cdot \mid y )}[ {f}_{y'}( x; \mathbf{W}^* ) ] &= 
    \beta \cdot s_{yy} \cdot \frac{{\eta}^{(y)}_{y'}}{Z_{y', y'}} - 
    \beta \cdot s_{yy} \cdot \sum_{j \in [ L ]} ( \mathbf{Z}^{-1} \boldsymbol{\Delta} \mathbf{Z}^{-1} )_{y', j} \cdot {\eta}^{(y)}_{j} \\
    &= 
    \beta \cdot s_{yy} \cdot \frac{{\eta}^{(y)}_{y'}}{Z_{y', y'}} - 
    \beta \cdot s_{yy} \cdot \sum_{j \in [ L ]} \frac{\Delta_{y', j} \cdot {\eta}^{(y)}_{j}}{Z_{y', y'} \cdot Z_{j, j}}.
\end{align*}
Recall that by Assumptions~\ref{assum:id-feat} and~\ref{assum:loc-feat-orth}, we have
$\boldsymbol{\eta}^{(y)} = \frac{1}{s_{yy}} \cdot \mathbf{N} \mathbf{e}_y$ for diagonal matrix $\mathbf{N}$.
Then,
\begin{align*}
    \mathop{\mathbb{E}}_{x \sim \Pr_{\rm ood}( \cdot \mid y )}[ {f}_{y'}( x; \mathbf{W}^* ) ] &= 
    \beta \cdot \frac{N_{y, y}}{Z_{y, y}} \cdot \indicator{y = y'} - 
    \beta \cdot \frac{\Delta_{y', y} \cdot N_{y, y}}{Z_{y', y'} \cdot Z_{y, y}}.
\end{align*}
\end{proof}

\begin{proof}[Proof of Proposition~\ref{corr:opt-logits-first-order-self-kd}]
Let $\mathbf{K}^{(2)} \defEq \mathbf{M}^\top \mathbf{T}^{-2} \mathbf{M}$, and $\mathbf{A} \defEq ( L \cdot \mathbf{I} + \mathbf{K} )^{-1} \mathbf{K}^{(2)} ( L \cdot \mathbf{I} + \mathbf{K} )^{-1}$.
By Lemma~\ref{lemm:opt-linear-self-kd},
\begin{align*}
    \mathop{\mathbb{E}}_{x \sim \Pr_{\rm id}( \cdot \mid y )}[ \mathbf{f}( x; \mathbf{W}^*_{\rm sd} ) ] &= ( \mathbf{W}^*_{\rm sd} )^\top \mathop{\mathbb{E}}_{x \sim \Pr_{\rm id}( \cdot \mid y )}[ \boldsymbol{\Phi}( x ) ] \\
    &= ( \mathbf{W}^*_{\rm sd} )^\top \mathbf{M} \mathbf{e}_y \\
    &= (\mathbf{W}^*)^\top \mathbf{M}  \mathbf{e}_y - \lambda \cdot L \cdot \mathbf{A}  \mathbf{e}_y \\
    &= \mathop{\mathbb{E}}_{x \sim \Pr_{\rm id}( \cdot \mid y )}[ \mathbf{f}( x; \mathbf{W}^* ) ] - \lambda \cdot L \cdot \mathbf{A} \mathbf{e}_y.
\end{align*}
That is, the self-distilled logits for ID samples apply a translation to the original logits, governed by $\mathbf{K}^{(2)}$.
Following Lemmas~\ref{lemm:whitened-similarity-diag} and~\ref{lemm:whitened-off-similarity},
$$ K^{(2)}_{ij} = \indicator{ i = j } \cdot \left[ 
t_j^2 \cdot \frac{1}{( l^{(j)} + \lambda )^2} +
s^2_{jj} \cdot \frac{1}{( n^{(j)} + \lambda )^2} + \sum_{k \neq j} s^{2}_{jk} \cdot \frac{1}{( m^{(j, k)} + \lambda )^2} \right] + \indicator{ i \neq j } \cdot s^{2}_{ij} \cdot \frac{1}{( m^{(i, j)} + \lambda )^2}. $$
Recall that, to a first-order approximation,
$$ ( L \cdot \mathbf{I} + \mathbf{K} )^{-1} = \mathbf{Z}^{-1} - \mathbf{Z}^{-1} \boldsymbol{\Delta} \mathbf{Z}^{-1}. $$
We
have
\begin{align*}
    \mathbf{A} &\defEq ( L \cdot \mathbf{I} + \mathbf{K} )^{-1} \mathbf{K}^{(2)} ( L \cdot \mathbf{I} + \mathbf{K} )^{-1} \\
    &= \mathbf{Z}^{-1} \mathbf{K}^{(2)} \mathbf{Z}^{-1}
    - \mathbf{Z}^{-1} \mathbf{K}^{(2)} \mathbf{Z}^{-1} \boldsymbol{\Delta} \mathbf{Z}^{-1}
    - \mathbf{Z}^{-1} \boldsymbol{\Delta} \mathbf{Z}^{-1} \mathbf{K}^{(2)} \mathbf{Z}^{-1} + 
    \mathbf{Z}^{-1} \boldsymbol{\Delta} \mathbf{Z}^{-1} \mathbf{K}^{(2)} \mathbf{Z}^{-1} \boldsymbol{\Delta} \mathbf{Z}^{-1} \\
    &= \mathbf{Z}^{-1} \mathbf{K}^{(2)} \mathbf{Z}^{-1}
    - \mathbf{Z}^{-1} \mathbf{K}^{(2)} \mathbf{Z}^{-1} \boldsymbol{\Delta} \mathbf{Z}^{-1}
    - \mathbf{Z}^{-1} \boldsymbol{\Delta} \mathbf{Z}^{-1} \mathbf{K}^{(2)} \mathbf{Z}^{-1} + 
    \mathscr{O}( \| \mathbf{H} \|_F^2 ).
\end{align*}
By expanding the individual terms, and focussing again on the first-order approximation:
$$ A_{ij} = \frac{1}{Z_{ii} \cdot Z_{jj}} \cdot \left( {K^{(2)}_{ij}}{}
    - {\Delta_{ij}}{} \cdot \left( \frac{K^{(2)}_{ii}}{ Z_{ii} }+ \frac{K^{(2)}_{jj}}{ Z_{jj} } \right) \right) + 
    \mathscr{O}( \| \mathbf{H} \|_F^2 ). $$
Note that when $i = j$,
$$ A_{ii} = \frac{1}{Z_{ii}^2} \cdot {K^{(2)}_{ii}}  +
    \mathscr{O}( \| \mathbf{H} \|_F^2 ). $$
Thus, 
for any ID class $y \in [ L ]$ and $y' \neq y$,
the first-order approximation for the expected logit for ID samples is:
\begin{align*}
    \mathop{\mathbb{E}}_{x \sim \Pr_{\rm id}( \cdot \mid y )}[ {f}_{y}( x; \mathbf{W}^*_{\rm sd} ) ]  &= 
    \frac{1}{L + K_{y, y}} \cdot
    \left(
    {K_{y, y}} - \lambda \cdot L \cdot \frac{1}{L + K_{y, y}} \cdot {K^{(2)}_{y, y}} \right) \\
    \mathop{\mathbb{E}}_{x \sim \Pr_{\rm id}( \cdot \mid y )}[ {f}_{y'}( x; \mathbf{W}^*_{\rm sd} ) ] &= 
    \frac{L}{( L + K_{y, y} ) \cdot ({L + K_{y', y'}})} \cdot
    \left(
    K_{y', y} \cdot \left( 1 + \lambda \cdot \left[  
    \frac{K^{(2)}_{y,y}}{ Z_{y,y} }+ \frac{K^{(2)}_{y',y'}}{ Z_{y',y'} }
    \right] \right) - \lambda \cdot K^{(2)}_{y', y} \right).
\end{align*}

Now consider the logits for OOD samples.
We have
\begin{align*}
    \mathop{\mathbb{E}}_{x \sim \Pr_{\rm ood}( \cdot \mid y )}[ \mathbf{f}( x; \mathbf{W}^*_{\rm sd} ) ] &= 
    (\mathbf{W}^*_{\rm sd})^\top \left( \mathop{\mathbb{E}}_{x \sim \Pr_{\rm ood}( \cdot \mid y )}[ \boldsymbol{\Phi}( x ) ] \right) \\
    &= s_{yy} \cdot (\mathbf{W}^*_{\rm sd})^\top \boldsymbol{\mu}^{(y)}_{\rm ood} \\
    &= s_{yy} \cdot (\mathbf{W}^*)^\top \boldsymbol{\mu}^{(y)}_{\rm ood} - \lambda \cdot s_{yy} \cdot ( \mathbf{W}^* )^\top ( \mathbf{R} + \lambda \cdot \mathbf{I} )^{-1} \boldsymbol{\mu}^{(y)}_{\rm ood} \\
    &= \mathop{\mathbb{E}}_{x \sim \Pr_{\rm ood}( \cdot \mid y )}[ \mathbf{f}( x; \mathbf{W}^* ) ] - \lambda \cdot s_{yy} \cdot ( \mathbf{W}^* )^\top ( \mathbf{R} + \lambda \cdot \mathbf{I} )^{-1} \boldsymbol{\mu}^{(y)}_{\rm ood}.
\end{align*}
By an application of the Woodbury identity,
\begin{align*}
    ( \mathbf{W}^* )^\top ( \mathbf{R} + \lambda \cdot \mathbf{I} )^{-1} \boldsymbol{\mu}^{(y)}_{\rm ood} &= ( L \cdot \mathbf{I} + \mathbf{K} )^{-1} \mathbf{M}^\top \mathbf{T}^{-1} ( \mathbf{R} + \lambda \cdot \mathbf{I} )^{-1} \boldsymbol{\mu}^{(y)}_{\rm ood} \\ 
    &= ( L \cdot \mathbf{I} + \mathbf{K} )^{-1} \mathbf{M}^\top \mathbf{T}^{-1} ( \mathbf{T}^{-1} - \mathbf{T}^{-1} \mathbf{M} ( L \cdot \mathbf{I} + \mathbf{K} )^{-1} \mathbf{M}^\top \mathbf{T}^{-1} ) \boldsymbol{\mu}^{(y)}_{\rm ood} \\ 
    &= ( L \cdot \mathbf{I} + \mathbf{K} )^{-1} \mathbf{M}^\top \mathbf{T}^{-2} \boldsymbol{\mu}^{(y)}_{\rm ood}
    - \mathbf{A} \mathbf{M}^\top \mathbf{T}^{-1} \boldsymbol{\mu}^{(y)}_{\rm ood}
    \\     
    &= ( L \cdot \mathbf{I} + \mathbf{K} )^{-1} \mathbf{M}^\top \mathbf{T}^{-2} \boldsymbol{\mu}^{(y)}_{\rm ood}
    - \beta \cdot \mathbf{A}  \boldsymbol{\eta}^{(y)}
    \\         
    &= ( L \cdot \mathbf{I} + \mathbf{K} )^{-1} \mathbf{M}^\top \mathbf{T}^{-2} \boldsymbol{\mu}^{(y)}_{\rm ood}
    - \frac{\beta}{s_{yy}} \cdot \mathbf{A} \mathbf{N} \mathbf{e}_ {y}.
\end{align*}
For the first term, following Lemma~\ref{lemm:mu-local-eigenvector}, we may write $\mathbf{M}^\top \mathbf{T}^{-2} \boldsymbol{\mu}^{(y)}_{\rm ood} = \frac{\beta}{s_{yy}} \cdot \mathbf{N}^{(2)} \mathbf{e}_y$, where 
$\mathbf{N}^{(2)}$ is a diagonal matrix with
$N^{(2)}_{y, y} = \frac{1}{( n^{(y)} + \lambda )^2} \cdot s^2_{yy}$.
Thus,
\begin{align*}
    ( \mathbf{W}^* )^\top ( \mathbf{R} + \lambda \cdot \mathbf{I} )^{-1} \boldsymbol{\mu}^{(y)}_{\rm ood} 
    &= \frac{\beta}{s_{yy}} \cdot ( L \cdot \mathbf{I} + \mathbf{K} )^{-1} \mathbf{N}^{(2)} \mathbf{e}_y
    - \frac{\beta}{s_{yy}} \cdot \mathbf{A} \mathbf{N} \mathbf{e}_ {y}.
\end{align*}
For the first term, using the expansion of $( L \cdot \mathbf{I} + \mathbf{K} )^{-1} $,
$$ ( L \cdot \mathbf{I} + \mathbf{K} )^{-1} \mathbf{N}^{(2)} \mathbf{e}_y = \mathbf{Z}^{-1} \mathbf{N}^{(2)} \mathbf{e}_y -  \mathbf{Z}^{-1} \boldsymbol{\Delta} \mathbf{Z}^{-1} \mathbf{N}^{(2)} \mathbf{e}_y + \mathscr{O}( \| \mathbf{H} \|_F^2 ). $$
Thus, the correction term for the  logit of class $y'$ is proportional to:
\begin{align*}
    \mathbf{e}_{y'}^\top ( \mathbf{W}^* )^\top ( \mathbf{R} + \lambda \cdot \mathbf{I} )^{-1} \boldsymbol{\mu}^{(y)}_{\rm ood} 
    &= \beta \cdot \left[ \frac{N^{(2)}_{y, y}}{Z_{y', y'}} \cdot \indicator{ y = y' } - \frac{N^{(2)}_{y, y} \cdot \Delta_{y', y}}{Z_{y, y} \cdot Z_{y', y'}} - A_{y', y} \cdot N_{y, y} \right].
\end{align*}
Thus, for any $y' \neq y$,
\begin{align*}
    \mathop{\mathbb{E}}_{x \sim \Pr_{\rm ood}( \cdot \mid y )}[ {f}_{y}( x; \mathbf{W}^*_{\rm sd} ) ] &= 
    \beta \cdot \frac{N_{y, y} - \lambda \cdot N^{(2)}_{y, y}}{Z_{y, y}} + \beta \cdot \lambda \cdot A_{y, y} \cdot N_{y, y} \\
    \mathop{\mathbb{E}}_{x \sim \Pr_{\rm ood}( \cdot \mid y )}[ {f}_{y'}( x; \mathbf{W}^*_{\rm sd} ) ] &= 
    - 
    \beta \cdot \frac{\Delta_{y', y} \cdot (N_{y, y} - \lambda \cdot N^{(2)}_{y, y})}{Z_{y', y'} \cdot Z_{y, y}} + \beta \cdot \lambda \cdot A_{y', y} \cdot N_{y, y}.
\end{align*}
\end{proof}

%%%%%%%%%%%%%

% %
% \begin{proof}[Proof of Lemma~\ref{lemm:logitgap-gm}]
% Observe that
% \todoakm{Temperature not defined in body}
% \begin{align*}
%     \log s_{\rm MSP}( \mathbf{f}( x ) ) &= \log p_{[1]}( x ) \\
%     &= \tau^{-1} \cdot f_{[1]}( x ) - \log \sum_{j = 1}^{L} \exp( \tau^{-1} \cdot f_{[j]}( x ) ).
% \end{align*}
% Now consider the mean-centered version of this score:
% \begin{align*}
%     \log s_{\rm MSP}( \mathbf{f}( x ) ) &= \log p_{[1]}( x ) - \frac{1}{L} \sum_{j=1}^{L} \log p_{[j]}( x ) \\
%     &= \tau^{-1} \cdot f_{[1]}( x ) - \frac{1}{L} \sum_{j=1}^{L} \tau^{-1} \cdot f_{[j]}( x ) \\
%     &= \tau^{-1} \cdot s_{\rm LG}( \mathbf{f}( x ) ).
% \end{align*}
% That is, the LogitGap score is simply a mean-centered version of the log softmax probability.
% \end{proof}

% %
% \begin{proof}[Proof of Lemma~\ref{lemm:kd-pl-rewrite}]
% We have
% $$ ( \forall i \in [ L - 1 ] ) \, 1 - \varrho_i = \frac{\sum_{j = i+1}^{L} p_j}{\sum_{j = i}^{L} p_j}, $$
% and $1 - \varrho_L = 0$.
% Thus,
% $$ ( \forall i \in [ L ] ) \, \prod_{j = 1}^{i - 1} ( 1 - \varrho_j ) = {\sum_{j = i}^{L} p_j}, $$
% % $$ ( \forall i \in [ L ] ) \, \prod_{j = i}^{L - 1} ( 1 - s_j ) = \frac{p_L}{\sum_{j = i}^{L} p_j}. $$
% and so
% $$ ( \forall i \in [ L ] ) \, p_i = \varrho_i \cdot \prod_{j = 1}^{i - 1} ( 1 - \varrho_j ). $$
% Equipped with this, we may rewrite the KD loss score as stated.
% \end{proof}

%% file: sec/suppl/helper.tex
\subsection{Glossary of commonly used symbols}

Table~\ref{tbl:glossary} summarises some commonly used symbols in the proofs related to the generative feature model of \S\ref{sec:theory}.

\begin{table}[!ht]
    \centering
    \renewcommand{\arraystretch}{1.25}
    \begin{tabular}{@{}ll@{}}
        \toprule
        \textbf{Symbol} & \textbf{Description} \\
        \midrule        $\boldsymbol{\mu}^{(i)}_{\locFeat}$ & Local (private) class feature mean \\
        $\boldsymbol{\mu}^{(i, j)}_{\globFeat}$ & Global (shared) feature mean \\
        $\boldsymbol{\mu}^{(i)}$ & Aggregate class feature mean \\
        \midrule        $\boldsymbol{\Sigma}^{(i)}_{\locFeat}$ & Local (private) class covariance \\
        $\boldsymbol{\Sigma}^{(i, j)}_{\globFeat}$ & Global (shared) class covariance \\
        $\boldsymbol{\Sigma}^{(i)}$ & Aggregate class covariance \\        
        $\boldsymbol{\bar{\Sigma}}$ & Average covariance across classes \\        
        \midrule
        $l^{(j)}$ & Eigenvalue corresponding to $\boldsymbol{\mu}^{(i)}_{\sharedFeat}$ (equals $\sigma^2_{\nullFeat}$) \\
        $n^{(j)}$ & Eigenvalue corresponding to $\boldsymbol{\mu}^{(i)}_{\locFeat}$ (equals $\frac{\sigma^2_{\locFeat}}{L} \cdot s_{jj}^2 
+ \sigma^2_{\nullFeat}$) \\
        $m^{(i, j)}$ & Eigenvalue corresponding to $\boldsymbol{\mu}^{(i, j)}_{\globFeat}$ (equals $\frac{2 \cdot \sigma^2_{\globFeat}}{L} \cdot s_{ij}^2 
+ \sigma^2_{\nullFeat}$) \\
        \midrule
       $\mathbf{M}$ & Class means: $\begin{bmatrix}
            \boldsymbol{\mu}^{(1)} & \ldots & \boldsymbol{\mu}^{(L)}
        \end{bmatrix}$ \\        
        $\mathbf{T}$ & Regularised average covariance: $\boldsymbol{\bar{\Sigma}} + \lambda \cdot \mathbf{I}$ \\
        $\mathbf{K}$ & Whitened class mean similarities: $\mathbf{M}^\top \mathbf{T}^{-1} \mathbf{M}$ \\
        $\mathbf{D}$ & Diagonal whitened class mean similarities: $\mathrm{diag}( \mathbf{K} )$ \\ 
        $\boldsymbol{\Delta}$ & Off-diagonal whitened class mean similarities: $\mathbf{K} - \mathbf{D}$ \\       
        $\mathbf{Z}$ & Regularised diagonal whitened class mean similarities: $L \cdot \mathbf{I} + \mathbf{D}$ \\
        $\mathbf{H}$ & Relative class mean similarities: $\mathbf{Z}^{-1} \boldsymbol{\Delta}$ \\
        $\mathbf{N}$ & $\mathrm{diag}(\{ 
( n^{(y)} + \lambda )^{-1} \cdot s_{yy}^2 \colon y \in [ L ] \})$ \\
        $\boldsymbol{\eta}^{(y)}$ & Whitened ID mean to OOD mean similarity: $\mathbf{M}^\top \mathbf{T}^{-1} \boldsymbol{\mu}_{\locFeat}^{(y)}$ \\
        \bottomrule
    \end{tabular}
    \caption{Glossary of commonly used symbols}
    \label{tbl:glossary}
\end{table}

\subsection{Properties of the generative model}

\begin{lemma}
\label{lemm:mu-similarity}
Under Assumption~\ref{assum:id-feat} and~\ref{assum:loc-feat-orth},
for each $i, j, k \in [ L ]$ with $j \neq k$,
\begin{align*}
    ( \boldsymbol{\mu}^{(i)} )^\top \boldsymbol{\mu}^{(j)}_{\sharedFeat} &= t_i \cdot \indicator{i = j} \\
    ( \boldsymbol{\mu}^{(i)} )^\top \boldsymbol{\mu}^{(j)}_{\locFeat} &= s_{ii} \cdot \indicator{i = j} \\
    ( \boldsymbol{\mu}^{(i)} )^\top \boldsymbol{\mu}^{(j, k)}_{\globFeat} &= s_{jk} \cdot \indicator{i = j \lor i = k}.
\end{align*}
\end{lemma}

\begin{proof}[Proof of Lemma~\ref{lemm:mu-similarity}]
By 
definition of the mean vector, and
Assumption~\ref{assum:loc-feat-orth}, we have
\begin{align*}
    ( \boldsymbol{\mu}^{(i)} )^\top \boldsymbol{\mu}^{(j)}_{\sharedFeat} &= t_i \cdot ( \boldsymbol{\mu}_{\sharedFeat}^{(i)} )^\top \boldsymbol{\mu}^{(j)}_{\sharedFeat} +  s_{ii} \cdot ( \boldsymbol{\mu}_{\locFeat}^{(i)} )^\top \boldsymbol{\mu}^{(j)}_{\sharedFeat} + \sum_{k' \neq i} s_{ik'} \cdot (\mu^{(i, k')}_{\globFeat})^\top \boldsymbol{\mu}^{(j)}_{\sharedFeat} \\
    &= t_i \cdot ( \boldsymbol{\mu}_{\sharedFeat}^{(i)} )^\top \boldsymbol{\mu}^{(j)}_{\sharedFeat} \\
    &= t_i \cdot \indicator{i = j} \\
    %%%
    ( \boldsymbol{\mu}^{(i)} )^\top \boldsymbol{\mu}^{(j)}_{\locFeat} &= t_i \cdot ( \boldsymbol{\mu}_{\sharedFeat}^{(i)} )^\top \boldsymbol{\mu}^{(j)}_{\locFeat} +  s_{ii} \cdot ( \boldsymbol{\mu}_{\locFeat}^{(i)} )^\top \boldsymbol{\mu}^{(j)}_{\locFeat} + \sum_{k' \neq i} s_{ik'} \cdot (\mu^{(i, k')}_{\globFeat})^\top \boldsymbol{\mu}^{(j)}_{\locFeat} \\
    &= t_i \cdot ( \boldsymbol{\mu}_{\sharedFeat}^{(i)} )^\top \boldsymbol{\mu}^{(j, k)}_{\globFeat} + s_{ii} \cdot \indicator{i = j} \\
    %%%
    ( \boldsymbol{\mu}^{(i)} )^\top \boldsymbol{\mu}^{(j, k)}_{\globFeat} &= t_i \cdot ( \boldsymbol{\mu}_{\sharedFeat}^{(i)} )^\top \boldsymbol{\mu}^{(j, k)}_{\globFeat} + s_{ii} \cdot ( \boldsymbol{\mu}_{\locFeat}^{(i)} )^\top \boldsymbol{\mu}^{(j, k)}_{\globFeat} + \sum_{k' \neq i} s_{ik'} \cdot (\mu^{(i, k')}_{\globFeat})^\top \boldsymbol{\mu}^{(j, k)}_{\globFeat} \\
    &= s_{ik} \cdot \indicator{i = j} + s_{ij} \cdot \indicator{i = k}.
\end{align*}
\end{proof}

\begin{lemma}
\label{lemm:covariance}
Under Assumption~\ref{assum:id-feat} and~\ref{assum:loc-feat-orth},
the class-conditional ID covariance matrix for any $i \in [L]$ is:
\begin{align*}
    {\boldsymbol{\Sigma}}^{(i)} &\defEq {\boldsymbol{\Sigma}}^{(i)}_{\locFeat} + {\boldsymbol{\Sigma}}^{(i)}_{\globFeat} +
    {\boldsymbol{\Sigma}}^{(i)}_{\nullFeat} \\
    {\boldsymbol{\Sigma}}^{(i)}_{\locFeat} &\defEq \sigma^2_{\locFeat} \cdot s_{ii}^2 \cdot \boldsymbol{\mu}^{(i)}_{\locFeat} (\boldsymbol{\mu}^{(i)}_{\locFeat})^\top \\
    {\boldsymbol{\Sigma}}^{(i)}_{\globFeat} &\defEq \sigma^2_{\globFeat} \cdot \sum_{k \neq i} s_{ik}^2 \cdot \boldsymbol{\mu}^{(i, k)}_{\globFeat} (\boldsymbol{\mu}^{(i, k)}_{\globFeat})^\top \\
    {\boldsymbol{\Sigma}}^{(i)}_{\nullFeat} &\defEq \sigma^2_{\nullFeat} \cdot \mathbf{I}.
\end{align*} 
\end{lemma}

\begin{proof}
Consider the class-conditional covariance of ID samples:
\begin{align*}
    {\boldsymbol{\Sigma}}^{(i)} &= \mathop{\mathbb{E}}_{x \sim \Pr_{\rm id}( \cdot \mid i )}[ ( \boldsymbol{\Phi}( x ) - \boldsymbol{\mu}^{(i)} ) ( \boldsymbol{\Phi}( x ) - \boldsymbol{\mu}^{(i)} )^\top ].
\end{align*}
Now recall that by Assumption~\ref{assum:id-feat},
$$ \boldsymbol{\Phi}( x ) - \boldsymbol{\mu}^{(i)} = ( \alpha^{(i)}_{s} - 1 ) \cdot s_{ii} \cdot \boldsymbol{\mu}^{(i)}_{s} + \sum_{k \neq i} ( \alpha^{(i, k)}_{g} - 1 ) \cdot s_{ik} \cdot \boldsymbol{\mu}^{(i, k)}_{g} + \boldsymbol{\epsilon}_{n}. $$
Further, by Assumption~\ref{assum:id-feat},
the cross-terms in the outer product are independent, and thus equate to $0$.
This leaves only the shared terms:
\begin{align*}
    {\boldsymbol{\Sigma}}^{(i)} &= \mathop{\mathbb{E}}_{x \sim \Pr_{\rm id}( \cdot \mid i )}[ ( \boldsymbol{\Phi}( x ) - \boldsymbol{\mu}^{(i)} ) ( \boldsymbol{\Phi}( x ) - \boldsymbol{\mu}^{(i)} )^\top ] \\
    &= \mathop{\mathbb{E}}_{x \sim \Pr_{\rm id}( \cdot \mid i )}[ ( \alpha^{(i)}_{s} - 1 )^2 ] \cdot s^2_{ii} \cdot \boldsymbol{\mu}^{(i)}_{s} ( \boldsymbol{\mu}^{(i)}_{s} )^\top + \sum_{k \neq i} \mathop{\mathbb{E}}_{x \sim \Pr_{\rm id}( \cdot \mid i )}[ ( \alpha^{(i, k)}_{g} - 1 )^2 ] \cdot s^2_{ik} \cdot \boldsymbol{\mu}^{(i, k)}_{g} ( \boldsymbol{\mu}^{(i, k)}_{g} )^\top + \\
    &\phantom{=} \mathop{\mathbb{E}}_{x \sim \Pr_{\rm id}( \cdot \mid i )}[ \boldsymbol{\epsilon}_{n} ( \boldsymbol{\epsilon}_{n} )^\top ] \\
    &= {\boldsymbol{\Sigma}}^{(i)}_{\locFeat} + {\boldsymbol{\Sigma}}^{(i)}_{\globFeat} +
    {\boldsymbol{\Sigma}}^{(i)}_{\nullFeat},
\end{align*}    
where
\begin{align*}
    {\boldsymbol{\Sigma}}^{(i)}_{\locFeat} &\defEq \sigma^2_{\locFeat} \cdot s_{ii}^2 \cdot \boldsymbol{\mu}^{(i)}_{\locFeat} (\boldsymbol{\mu}^{(i)}_{\locFeat})^\top \\
    {\boldsymbol{\Sigma}}^{(i)}_{\globFeat} &\defEq \sigma^2_{\globFeat} \cdot \sum_{k \neq i} s_{ik}^2 \cdot \boldsymbol{\mu}^{(i, k)}_{\globFeat} (\boldsymbol{\mu}^{(i, k)}_{\globFeat})^\top \\
    {\boldsymbol{\Sigma}}^{(i)}_{\nullFeat} &\defEq \sigma^2_{\nullFeat} \cdot \mathbf{I}.    
\end{align*}
\end{proof}

\begin{lemma}
\label{lemm:mu-shared-eigenvector}
Under Assumption~\ref{assum:id-feat} and~\ref{assum:loc-feat-orth},
for each $j \in [ L ]$,
$\boldsymbol{\mu}^{(j)}_{\sharedFeat}$ is an eigenvector of $\bar{\boldsymbol{\Sigma}}$, with eigenvalue $l^{( j )} \defEq \sigma^2_{\nullFeat}$.
\end{lemma}

\begin{proof}[Proof of Lemma~\ref{lemm:mu-shared-eigenvector}]
By Lemma~\ref{lemm:covariance} and
Assumption~\ref{assum:loc-feat-orth},
we have
\begin{align*}
    \bar{\boldsymbol{\Sigma}} \boldsymbol{\mu}^{(j)}_{\sharedFeat} &= \frac{1}{L} \sum_{i \in [ L ]} \bar{\boldsymbol{\Sigma}}^{(i)} \boldsymbol{\mu}^{(j)}_{\sharedFeat} \\
    &= 
     \sigma^2_{\nullFeat}
    \cdot \boldsymbol{\mu}^{(j)}_{\sharedFeat},
\end{align*}
i.e.,
$\boldsymbol{\mu}^{(j)}_{\sharedFeat}$ is an eigenvector of $\bar{\boldsymbol{\Sigma}}$, with eigenvalue $l^{( j )} \defEq \sigma^2_{\nullFeat}$.
\end{proof}

\begin{lemma}
\label{lemm:mu-local-eigenvector}
Under Assumption~\ref{assum:id-feat} and~\ref{assum:loc-feat-orth},
for each $j \in [ L ]$,
$\boldsymbol{\mu}^{(j)}_{\locFeat}$ is an eigenvector of $\bar{\boldsymbol{\Sigma}}$, with eigenvalue $n^{( j )} \defEq \frac{\sigma^2_{\locFeat}}{L} \cdot s_{jj}^2 
+ \sigma^2_{\nullFeat}$.
\end{lemma}

\begin{proof}[Proof of Lemma~\ref{lemm:mu-local-eigenvector}]
By Lemma~\ref{lemm:covariance} and
Assumption~\ref{assum:loc-feat-orth},
we have
\begin{align*}
    \bar{\boldsymbol{\Sigma}} \boldsymbol{\mu}^{(j)}_{\locFeat} &= \frac{1}{L} \sum_{i \in [ L ]} \bar{\boldsymbol{\Sigma}}^{(i)} \boldsymbol{\mu}^{(j)}_{\locFeat} \\
    &= \frac{1}{L} \sum_{i \in [ L ]} ( \bar{\boldsymbol{\Sigma}}^{(i)}_{\locFeat} \boldsymbol{\mu}^{(j)}_{\locFeat} + 
    \bar{\boldsymbol{\Sigma}}^{(i)}_{\globFeat} \boldsymbol{\mu}^{(j)}_{\locFeat} +
    \bar{\boldsymbol{\Sigma}}^{(i)}_{\nullFeat} \boldsymbol{\mu}^{(j)}_{\locFeat}) \\
    &=  \left( \frac{\sigma^2_{\locFeat}}{L} \cdot s_{jj}^2
    +
     \sigma^2_{\nullFeat}
    \right) \cdot \boldsymbol{\mu}^{(j)}_{\locFeat},
\end{align*}
i.e.,
$\boldsymbol{\mu}^{(j)}_{\locFeat}$ is an eigenvector of $\bar{\boldsymbol{\Sigma}}$, with eigenvalue $n^{( j )} \defEq \frac{\sigma^2_{\locFeat}}{L} \cdot s_{jj}^2 
+ \sigma^2_{\nullFeat}$.
\end{proof}

\begin{lemma}
\label{lemm:mu-global-eigenvector}
Under Assumption~\ref{assum:id-feat} and~\ref{assum:loc-feat-orth},
for each $j \neq k \in [ L ]$,
$\boldsymbol{\mu}^{(j, k)}_{\globFeat}$ is an eigenvector of $\bar{\boldsymbol{\Sigma}}$, with eigenvalue 
$m^{( j, k )} \defEq \frac{2 \cdot \sigma^2_{\globFeat}}{L} \cdot s_{jk}^2 
+ \sigma^2_{\nullFeat}$.
\end{lemma}

\begin{proof}[Proof of Lemma~\ref{lemm:mu-global-eigenvector}]
Following Lemma~\ref{lemm:mu-local-eigenvector}, and building on Assumption~\ref{assum:loc-feat-orth},
we have
\begin{align*}
    \bar{\boldsymbol{\Sigma}} \boldsymbol{\mu}^{(j, k)}_{\globFeat} &= \frac{1}{L} \sum_{i \in [ L ]} \bar{\boldsymbol{\Sigma}}^{(i)} \boldsymbol{\mu}^{(j, k)}_{\globFeat} \\
    &= \frac{1}{L} \sum_{i \in [ L ]} ( {\boldsymbol{\Sigma}}^{(i)}_{\locFeat} \boldsymbol{\mu}^{(j, k)}_{\globFeat} + 
    {\boldsymbol{\Sigma}}^{(i)}_{\globFeat} \boldsymbol{\mu}^{(j, k)}_{\globFeat} +
    {\boldsymbol{\Sigma}}^{(i)}_{\nullFeat} \boldsymbol{\mu}^{(j, k)}_{\globFeat}) \\
    &=  \left( \frac{2 \cdot \sigma^2_{\globFeat}}{L} \cdot s^2_{j, k}
    +
     \sigma^2_{\nullFeat}
    \right) \cdot \boldsymbol{\mu}^{(j, k)}_{\globFeat},
\end{align*}
i.e.,
$\boldsymbol{\mu}^{(j, k)}_{\globFeat}$ is an eigenvector of $\bar{\boldsymbol{\Sigma}}$, with eigenvalue $m^{( j, k )} \defEq \frac{2 \cdot \sigma^2_{\globFeat}}{L} \cdot s_{jk}^2 
+ \sigma^2_{\nullFeat}$.
\end{proof}

\begin{lemma}
\label{lemm:whitened-mean}
Under Assumption~\ref{assum:id-feat} and~\ref{assum:loc-feat-orth},
for each $j \in [ L ]$,
$$ \bar{\boldsymbol{\Sigma}} \boldsymbol{\mu}^{(j)} = 
t_j \cdot \sigma^2_{\nullFeat} \cdot \boldsymbol{\mu}^{(j)}_{\sharedFeat}
+
s_{jj} \cdot \left( \frac{\sigma^2_{\locFeat}}{L} \cdot s_{jj}^2
    +
     \sigma^2_{\nullFeat}
    \right) \cdot \boldsymbol{\mu}^{(j)}_{\locFeat} + \sum_{k \neq j} s_{jk} \cdot \left( \frac{2 \cdot \sigma^2_{\globFeat}}{L} \cdot s^2_{j, k}
    +
     \sigma^2_{\nullFeat}
    \right) \cdot \boldsymbol{\mu}^{(j, k)}_{\globFeat}. $$
\end{lemma}

\begin{proof}[Proof of Lemma~\ref{lemm:whitened-mean}]
By Lemmas~\ref{lemm:mu-shared-eigenvector},~\ref{lemm:mu-local-eigenvector}, and~\ref{lemm:mu-global-eigenvector}, we have
\begin{align*}
    \bar{\boldsymbol{\Sigma}} \boldsymbol{\mu}^{(j)} &= \bar{\boldsymbol{\Sigma}} \left( 
    t_j \cdot \boldsymbol{\mu}_{\sharedFeat}^{(j)} +
    s_{jj} \cdot \boldsymbol{\mu}_{\locFeat}^{(j)}  + \sum_{k \neq j} s_{jk} \cdot \boldsymbol{\mu}_{\globFeat}^{(j, k)} \right) \\
    &= 
    t_j \cdot \sigma^2_{\nullFeat} \cdot \boldsymbol{\mu}_{\sharedFeat}^{(j)} +
    s_{jj} \cdot \left( \frac{\sigma^2_{\locFeat}}{L} \cdot s_{jj}^2
    +
     \sigma^2_{\nullFeat}
    \right) \cdot \boldsymbol{\mu}^{(j)}_{\locFeat} + \sum_{k \neq j} s_{jk} \cdot \left( \frac{2 \cdot \sigma^2_{\globFeat}}{L} \cdot s^2_{j, k}
    +
     \sigma^2_{\nullFeat}
    \right) \cdot \boldsymbol{\mu}^{(j, k)}_{\globFeat}.
\end{align*}
\end{proof}

\begin{lemma}
\label{lemm:whitened-similarity-diag}
Under Assumption~\ref{assum:id-feat} and~\ref{assum:loc-feat-orth},
for each $j \in [ L ]$,
$$ ( \boldsymbol{\mu}^{(j)} )^{\top} \mathbf{T}^{-1} \boldsymbol{\mu}^{(j)} = t^2_j \cdot \frac{1}{l^{(j)} + \lambda} +
s^2_{jj} \cdot \frac{1}{n^{(j)} + \lambda} + \sum_{k \neq j} s^2_{jk} \cdot \frac{1}{m^{(j, k)} + \lambda}. $$
\end{lemma}

\begin{proof}[Proof of Lemma~\ref{lemm:whitened-similarity-diag}]
Consider the eigendecomposition $\bar{\boldsymbol{\Sigma}} = \mathbf{U} \boldsymbol{\Lambda} \mathbf{U}^\top$.
Recall that by Assumptions~\ref{assum:id-feat} and~\ref{assum:loc-feat-orth},
$\boldsymbol{\mu}^{(j)}_{\locFeat}$ is an eigenvector of $\bar{\boldsymbol{\Sigma}}$ with eigenvalue $n^{(j)}$ (Lemma~\ref{lemm:mu-local-eigenvector}),
and that
$\boldsymbol{\mu}^{(j, k)}_{\globFeat}$ is an eigenvector of $\bar{\boldsymbol{\Sigma}}$ with eigenvalue $m^{(j, k)}$ (Lemma~\ref{lemm:mu-global-eigenvector}).
Thus,
\begin{align*}
\mathbf{T}^{-1} \boldsymbol{\mu}^{(j)} &= \mathbf{U} ( \boldsymbol{\Lambda} + \lambda \cdot \mathbf{I} )^{-1} \mathbf{U}^\top \boldsymbol{\mu}^{(j)} \\
&= \mathbf{U} ( \boldsymbol{\Lambda} + \lambda \cdot \mathbf{I} )^{-1} \mathbf{U}^\top \cdot \left( 
t_j \cdot \boldsymbol{\mu}_{\sharedFeat}^{(j)} + 
s_{jj} \cdot \boldsymbol{\mu}_{\locFeat}^{(j)} + 
\sum_{k \neq j} s_{jk} \cdot \boldsymbol{\mu}_{\globFeat}^{(j,k)} \right) \\
&= t_j \cdot \frac{1}{l^{(j)} + \lambda} \boldsymbol{\mu}_{\sharedFeat}^{(j)} +
s_{jj} \cdot \frac{1}{n^{(j)} + \lambda} \cdot \boldsymbol{\mu}_{\locFeat}^{(j)} + \sum_{k \neq j} s_{jk} \cdot \frac{1}{m^{(j, k)} + \lambda} \cdot \boldsymbol{\mu}_{\globFeat}^{(j, k)}.
\end{align*}
Consequently,
by Lemma~\ref{lemm:mu-similarity},
\begin{align*}
(\boldsymbol{\mu}^{(j)})^\top \mathbf{T}^{-1} \boldsymbol{\mu}^{(j)} &= 
t_j \cdot \frac{1}{l^{(j)} + \lambda} \cdot (\boldsymbol{\mu}^{(j)})^\top \boldsymbol{\mu}_{\sharedFeat}^{(j)} +
s_{jj} \cdot \frac{1}{n^{(j)} + \lambda} \cdot (\boldsymbol{\mu}^{(j)})^\top \boldsymbol{\mu}_{\locFeat}^{(j)} + \sum_{k \neq j} s_{jk} \cdot \frac{1}{m^{(j, k)} + \lambda} \cdot (\boldsymbol{\mu}^{(j)})^\top \boldsymbol{\mu}_{\globFeat}^{(j, k)} \\
&= t^2_j \cdot \frac{1}{l^{(j)} + \lambda} + 
s^2_{jj} \cdot \frac{1}{n^{(j)} + \lambda} + \sum_{k \neq j} s^2_{jk} \cdot \frac{1}{m^{(j, k)} + \lambda}.
\end{align*}
\end{proof}

\begin{lemma}
\label{lemm:whitened-off-similarity}
Under Assumption~\ref{assum:id-feat} and~\ref{assum:loc-feat-orth},
for each $i \neq j \in [ L ]$,
$$ ( \boldsymbol{\mu}^{(i)} )^{\top} \mathbf{T}^{-1} \boldsymbol{\mu}^{(j)} = s^2_{ij} \cdot \frac{1}{m^{(i, j)} + \lambda}. $$
\end{lemma}

\begin{proof}[Proof of Lemma~\ref{lemm:whitened-off-similarity}]
The proof follows Lemma~\ref{lemm:whitened-similarity-diag},
and an application of Lemma~\ref{lemm:mu-similarity}:
\begin{align*}
(\boldsymbol{\mu}^{(i)})^\top \mathbf{T}^{-1} \boldsymbol{\mu}^{(j)} &= 
t_j \cdot \frac{1}{l^{(j)} + \lambda} \cdot (\boldsymbol{\mu}^{(i)})^\top \boldsymbol{\mu}_{\sharedFeat}^{(j)} +
s_{jj} \cdot \frac{1}{n^{(j)} + \lambda} \cdot (\boldsymbol{\mu}^{(i)})^\top \boldsymbol{\mu}_{\locFeat}^{(j)} + \sum_{k \neq j} s_{jk} \cdot \frac{1}{m^{(j, k)} + \lambda} \cdot (\boldsymbol{\mu}^{(i)})^\top \boldsymbol{\mu}_{\globFeat}^{(j, k)} \\
&= s^2_{ij} \cdot \frac{1}{m^{(i, j)} + \lambda}.
\end{align*}
\end{proof}

\begin{lemma}
\label{lemm:k-eigenvalue}
Let 
$\mathbf{K} \defEq \mathbf{M}^\top \mathbf{T}^{-1} \mathbf{M}$
be the whitened class similarities.
Then, all eigenvalues of 
$\mathbf{H} \defEq ( L \cdot \mathbf{I} + \mathrm{diag}( \mathbf{K} ) )^{-1} ( \mathbf{K} - \mathrm{diag}( \mathbf{K} ) )$ 
are less than $1$.
\end{lemma}

\begin{proof}[Proof of Lemma~\ref{lemm:k-eigenvalue}]
The elements of the matrix of interest are $H_{ij} = \indicator{ i \neq j } \cdot \frac{K_{ij}}{K_{ii} + L}$.
By a theorem of Gershgorin~\citep[Theorem 7.2.1]{Golub:2013},
the maximum eigenvalue of $\mathbf{H}$ may be bounded by
$$ \max_{i \in [ L ]} \sum_{j \neq i} \frac{K_{ij}}{K_{ii} + L}. $$
For any $i \in [L]$,
Lemmas~\ref{lemm:whitened-similarity-diag} and~\ref{lemm:whitened-off-similarity} imply that
\begin{align*}
   K_{ii} + L &= t^2_i \cdot \frac{1}{l^{(i)} + \lambda} + 
   s^2_{ii} \cdot \frac{1}{n^{(i)} + \lambda} + \sum_{j \neq i} K_{ij} + L  \\
   &> \sum_{j \neq i} K_{ij}.
\end{align*}
Thus,
$$ \max_{i \in [ L ]} \sum_{j \neq i} \frac{K_{ij}}{K_{ii} + L} < 1. $$
\end{proof}

\begin{lemma}
    \label{lemm:eta-diagonal}
Under Assumptions~\ref{assum:id-feat} and~\ref{assum:loc-feat-orth},
for every $y \in [ L ]$,
$$ \mathbf{M}^\top \mathbf{T}^{-1} \boldsymbol{\mu}^{(y)}_{\locFeat} = \frac{1}{s_{yy}} \cdot \mathbf{N} \mathbf{e}_y, $$
where 
$\mathbf{N} \defEq \mathrm{diag}(\{ 
( n^{(y)} + \lambda )^{-1} \cdot s^2_{yy} \colon y \in [ L ] \})$.
\end{lemma}

\begin{proof}[Proof of Lemma~\ref{lemm:eta-diagonal}]
Recall that by Lemma~\ref{lemm:mu-local-eigenvector},
$\boldsymbol{\mu}^{(y)}_{\locFeat}$ is an eigenvector of $\bar{\boldsymbol{\Sigma}}$ with eigenvalue $n^{(y)}$ (Lemma~\ref{lemm:mu-local-eigenvector}).
Write the eigendecomposition $\bar{\boldsymbol{\Sigma}} = \mathbf{U} \boldsymbol{\Lambda} \mathbf{U}^\top$.
Then, by Lemma~\ref{lemm:mu-similarity},
\begin{align*}
    \boldsymbol{\eta}^{(y)} &\defEq \mathbf{M}^\top \mathbf{T}^{-1} \boldsymbol{\mu}^{(y)}_{\locFeat} \\
    &= \mathbf{M}^\top ( \bar{\boldsymbol{\Sigma}} + \lambda \cdot \mathbf{I} )^{-1} \boldsymbol{\mu}^{(y)}_{\locFeat} \\
    &= \mathbf{M}^\top \mathbf{U} ( \boldsymbol{\Lambda} + \lambda \cdot \mathbf{I} )^{-1} \mathbf{U}^\top \boldsymbol{\mu}^{(y)}_{\locFeat} \\    
    &= \frac{1}{n^{(y)} + \lambda} \cdot \mathbf{M}^\top \boldsymbol{\mu}^{(y)}_{\locFeat} \\
    &= \frac{1}{n^{(y)} + \lambda} \cdot s_{yy} \cdot \mathbf{e}_y \\
    &= \frac{1}{s_{yy}} \cdot \mathbf{N} \mathbf{e}_y.
\end{align*}    
\end{proof}

\subsection{Properties of the linear regression solution}

\begin{lemma}
    \label{lemm:opt-linear}
    The optimal solution to the  square loss 
    population risk is:
    % ~\eqref{eqn:square-loss-risk} is:
    \begin{equation}
        \label{eqn:opt-lin-reg}
        \mathbf{W}^* = 
        % ( \bar{\boldsymbol{\Sigma}} + \lambda \cdot \mathbf{I} )^{-1} \cdot \mathbf{M} \cdot ( L \cdot \mathbf{I} + \mathbf{M}^\top ( \bar{\boldsymbol{\Sigma}} + \lambda \cdot \mathbf{I} )^{-1} \mathbf{M} )^{-1},
        \mathbf{T}^{-1} \mathbf{M} \cdot ( L \cdot \mathbf{I} + \mathbf{K} )^{-1},
    \end{equation}
    where $\mathbf{M} \defEq \begin{bmatrix} \boldsymbol{\mu}^{(1)} & \ldots & \boldsymbol{\mu}^{(L)} \end{bmatrix} \in \Real^{D \times L}$, and
    $\mathbf{T} \defEq  \boldsymbol{\bar{\Sigma}} + \lambda \cdot \mathbf{I}$ for
    $\boldsymbol{\bar{\Sigma}} \defEq \frac{1}{L} \sum_{y \in [L]} \boldsymbol{\Sigma}^{(y)} \in \Real^{D \times D}$.
\end{lemma}

\begin{proof}[Proof of Lemma~\ref{lemm:opt-linear}]
We may rewrite the population risk as
$$     R_{\rm sq}( \mathbf{W} ) \defEq \mathop{\mathbb{E}}_{( x, y ) \sim \Pr_{\rm id}}[ \| \mathbf{W}^\top \boldsymbol{\Phi}( x ) - \mathbf{e}_y \|_2^2 ] + \lambda \cdot \| \mathbf{W} \|_F^2, $$
where $\mathbf{e}_y \in \{ 0, 1 \}^{L}$ is a one-hot vector corresponding to class $y \in [ L ]$.
This has gradient
\begin{align*}
    G( \mathbf{W} ) &\defEq \nabla_{\mathbf{W}} R_{\rm sq}( \mathbf{W} ) \\
    &= \nabla_{\mathbf{W}}\left[ 
    \mathop{\mathbb{E}}_{( x, y ) \sim \Pr_{\rm id}}[ \| \mathbf{W}^\top \boldsymbol{\Phi}( x ) - \mathbf{e}_y \|_2^2 ] + \lambda \cdot \| \mathbf{W} \|_F^2
    \right] \\
    &= \nabla_{\mathbf{W}}\left[ 
    \mathop{\mathbb{E}}_{( x, y ) \sim \Pr_{\rm id}}[ \boldsymbol{\Phi}( x )^\top \mathbf{W} \mathbf{W}^\top \boldsymbol{\Phi}( x ) - 2 \cdot \boldsymbol{\Phi}( x )^\top \mathbf{W}  \mathbf{e}_y
    ]
    \right] + 2 \cdot \lambda \cdot \mathbf{W} \\
    &= 
    \mathop{\mathbb{E}}_{( x, y ) \sim \Pr_{\rm id}}[ 2 \cdot \boldsymbol{\Phi}( x ) \boldsymbol{\Phi}( x )^\top \mathbf{W} - 2 \cdot \boldsymbol{\Phi}( x ) \mathbf{e}_y^\top
    ]
    + 2 \cdot \lambda \cdot \mathbf{W}. 
\end{align*}
For brevity, let $\mathbf{R} \defEq \mathop{\mathbb{E}}_{( x, y ) \sim \Pr_{\rm id}}[\boldsymbol{\Phi}( x ) \boldsymbol{\Phi}( x )^\top ]$.
At optimality, the stationarity condition must hold:
\begin{align*}
    G( \mathbf{W}^* ) = 0 &\iff 
    \mathbf{R} \mathbf{W}^*
    - \mathop{\mathbb{E}}_{( x, y ) \sim \Pr_{\rm id}}[  \boldsymbol{\Phi}( x ) \mathbf{e}_y^\top
] + \lambda \cdot \mathbf{W}^* = 0 \\
&\iff \left( \mathbf{R} + \lambda \cdot \mathbf{I} \right) \mathbf{W}^* = \mathop{\mathbb{E}}_{( x, y ) \sim \Pr_{\rm id}}[ \boldsymbol{\Phi}( x ) \mathbf{e}_y^\top
] \\
&\iff \mathbf{W}^* = \left( \mathbf{R} + \lambda \cdot \mathbf{I} \right)^{-1} \mathop{\mathbb{E}}_{( x, y ) \sim 
\Pr_{\rm id}}[ \boldsymbol{\Phi}( x ) \mathbf{e}_y^\top
].
\end{align*}
Now observe that since the classes are assumed to be balanced (Assumption~\ref{assum:balanced-class}),
\begin{align*}
    \mathop{\mathbb{E}}_{( x, y ) \sim 
\Pr_{\rm id}}[ \boldsymbol{\Phi}( x ) \mathbf{e}_y^\top ] &= \frac{1}{L} \sum_{y \in [L]} \mathop{\mathbb{E}}_{x \sim 
\Pr_{\rm id}( \cdot \mid y )}[ \boldsymbol{\Phi}( x ) ] \mathbf{e}_y^\top \\
&= \frac{1}{L} \sum_{y \in [L]} \boldsymbol{\mu}^{(y)}
\mathbf{e}_y^\top \\
&= \frac{1}{L} \begin{bmatrix} \boldsymbol{\mu}^{(1)} & \ldots & \boldsymbol{\mu}^{(L)} \end{bmatrix} \\
&= \frac{1}{L} \mathbf{M},
\end{align*}
where $\mathbf{M} \defEq \begin{bmatrix} \boldsymbol{\mu}^{(1)} & \ldots & \boldsymbol{\mu}^{(L)} \end{bmatrix}$.
Similarly,
\begin{align*}
    \mathbf{R} &\defEq \mathop{\mathbb{E}}_{( x, y ) \sim \Pr_{\rm id}}[\boldsymbol{\Phi}( x ) \boldsymbol{\Phi}( x )^\top ] \\
    &= \frac{1}{L} \sum_{y \in [L]} \mathop{\mathbb{E}}_{x \sim \Pr_{\rm id}( \cdot \mid y )}[\boldsymbol{\Phi}( x ) \boldsymbol{\Phi}( x )^\top ] \\
    &= \frac{1}{L} \sum_{y \in [L]} \mathop{\mathbb{E}}_{x \sim \Pr_{\rm id}( \cdot \mid y )}[ (\boldsymbol{\Phi}( x ) - \boldsymbol{\mu}^{(y)}) (\boldsymbol{\Phi}( x ) - \boldsymbol{\mu}^{(y)})^\top + \boldsymbol{\mu}^{(y)} ( \boldsymbol{\mu}^{(y)} )^\top ] \\
    &= \frac{1}{L} \sum_{y \in [L]} ( \boldsymbol{\Sigma}^{(y)} + \boldsymbol{\mu}^{(y)} ( \boldsymbol{\mu}^{(y)} )^\top ) \\
    &= \boldsymbol{\bar{\Sigma}} + \frac{1}{L} \mathbf{M} \mathbf{M}^\top,
\end{align*}
where $\boldsymbol{\bar{\Sigma}}$ is the average of the per-class covariances.
We may now apply the Woodbury identity:
$$ ( \mathbf{A} + \mathbf{U} \mathbf{C} \mathbf{V} )^{-1} = \mathbf{A}^{-1} - \mathbf{A}^{-1} \mathbf{U} ( \mathbf{C}^{-1} + \mathbf{V} \mathbf{A}^{-1} \mathbf{U} )^{-1} \mathbf{V} \mathbf{A}^{-1}, $$
which implies
\begin{align*}
    ( \mathbf{A} + \mathbf{U} \mathbf{C} \mathbf{V} )^{-1} \mathbf{U} \mathbf{C} &= \mathbf{A}^{-1} \mathbf{U} \mathbf{C} - \mathbf{A}^{-1} \mathbf{U} ( \mathbf{C}^{-1} + \mathbf{V} \mathbf{A}^{-1} \mathbf{U} )^{-1} \mathbf{V} \mathbf{A}^{-1} \mathbf{U} \mathbf{C} \\
    &= \mathbf{A}^{-1} \mathbf{U} ( \mathbf{C} - ( \mathbf{C}^{-1} + \mathbf{V} \mathbf{A}^{-1} \mathbf{U} )^{-1} \mathbf{V} \mathbf{A}^{-1} \mathbf{U} \mathbf{C} )\\
    &= \mathbf{A}^{-1} \mathbf{U} ( \mathbf{C}^{-1} + \mathbf{V} \mathbf{A}^{-1} \mathbf{U} )^{-1},
\end{align*}
where the last line follows by another application of the Woodbury identity.
Now, applying this with the choices 
$\mathbf{A} = \bar{\boldsymbol{\Sigma}} + \lambda \cdot \mathbf{I}$,
$\mathbf{U} = \mathbf{M}$,
$\mathbf{V} = \mathbf{M}^\top$,
$\mathbf{C} = \frac{1}{L} \cdot \mathbf{I}$,
we have
\begin{align*}
    \mathbf{W}^* &= ( \bar{\boldsymbol{\Sigma}} + \lambda \cdot \mathbf{I} )^{-1} \cdot \mathbf{M} \cdot ( L \cdot \mathbf{I} + \mathbf{M}^\top ( \bar{\boldsymbol{\Sigma}} + \lambda \cdot \mathbf{I} )^{-1} \mathbf{M} )^{-1} \\
    &= \mathbf{T}^{-1} \mathbf{M} \cdot ( L \cdot \mathbf{I} + \mathbf{K} )^{-1}.
\end{align*}
\end{proof}

\begin{proposition}
\label{eqn:opt-logits}
Suppose $\mathbf{W}^*$ is the optimal square loss population risk minimiser per~\eqref{eqn:opt-lin-reg}.
Under Assumptions~\ref{assum:id-feat}--\ref{assum:balanced-class}, 
for any $y, y' \in [ L ]$,
the expected logit vectors for ID and OOD samples satisfy
\begin{align*}
    \mathop{\mathbb{E}}_{x \sim \Pr_{\rm id}( \cdot \mid y )}[ \mathbf{f}( x; \mathbf{W}^* ) ] 
    % &= 
    % \sum_{k = 0}^{\infty} {( -1 )^k} \cdot ( {\mathbf{Z}^{-1}} \mathbf{S} )^k 
    % \mathbf{Z}^{-1}
    % \mathbf{K} \mathbf{e}_y \\
    &= ( L \cdot \mathbf{I} + \mathbf{K} )^{-1} \mathbf{K} \mathbf{e}_y \\    
    \mathop{\mathbb{E}}_{x \sim \Pr_{\rm ood}( \cdot \mid y )}[ \mathbf{f}( x; \mathbf{W}^* ) ] 
    % &= 
    % \beta \cdot \sum_{k = 0}^{\infty} {( -1 )^k} \cdot ( {\mathbf{Z}^{-1}} \mathbf{S} )^k 
    % \mathbf{Z}^{-1}
    % \mathbf{N} \mathbf{e}_y \\
    &= \beta \cdot ( 
    L \cdot \mathbf{I} + \mathbf{K}
    )^{-1} \mathbf{N} \mathbf{e}_y.
    % ( \mathbf{M}^\top \mathbf{T}^{-1} \boldsymbol{\mu}^{(y)}_{\locFeat} ),
\end{align*}
% where 
% $\mathbf{S} \defEq \mathbf{K} - \mathrm{diag}( \mathbf{K} )$,
% $\mathbf{Z} \defEq L \cdot \mathbf{I} + \mathrm{diag}( \mathbf{K} )$,
% and
% $\mathbf{N} \defEq \mathrm{diag}(\{ 
% ( n^{(y)} + \lambda )^{-1} \cdot s_{yy}^2 \colon y \in [ L ] \})$.
\end{proposition}

\begin{proof}[Proof of Proposition~\ref{eqn:opt-logits}]
Let $\mathbf{T} \defEq \bar{\boldsymbol{\Sigma}} + \lambda \cdot \mathbf{I}$.
Then, by~\eqref{eqn:opt-lin-reg},
\begin{align*}
    \mathop{\mathbb{E}}_{x \sim \Pr_{\rm id}( \cdot \mid y )}[ \mathbf{f}( x; \mathbf{W}^* ) ] &= 
    (\mathbf{W}^*)^\top \left( \mathop{\mathbb{E}}_{x \sim \Pr_{\rm id}( \cdot \mid y )}[ \boldsymbol{\Phi}( x ) ] \right) \\
    &= (\mathbf{W}^*)^\top \boldsymbol{\mu}^{(y)} \\
    &= ( L \cdot \mathbf{I} + \mathbf{K} )^{-1} \mathbf{M}^\top \mathbf{T}^{-1}  \boldsymbol{\mu}^{(y)} \\
    &= ( L \cdot \mathbf{I} + \mathbf{K} )^{-1} \mathbf{K} \mathbf{e}_y.
\end{align*}

We now consider the behaviour of OOD samples.
We begin by considering the conditional distribution of OOD samples which spuriously align with a class $y \in [L]$.
By~\eqref{eqn:opt-lin-reg},
\begin{align*}
    \mathop{\mathbb{E}}_{x \sim \Pr_{\rm ood}( \cdot \mid y )}[ \mathbf{f}( x; \mathbf{W}^* ) ] &= 
    (\mathbf{W}^*)^\top \left( \mathop{\mathbb{E}}_{x \sim \Pr_{\rm ood}( \cdot \mid y )}[ \boldsymbol{\Phi}( x ) ] \right) \\
    &=  s_{yy} \cdot (\mathbf{W}^*)^\top \boldsymbol{\mu}^{(y)}_{\rm ood} \\
    &=  s_{yy} \cdot ( L \cdot \mathbf{I} + \mathbf{K} )^{-1} \mathbf{M}^\top \mathbf{T}^{-1}  \boldsymbol{\mu}^{(y)}_{\rm ood} \\
    &= \beta \cdot s_{yy} \cdot ( L \cdot \mathbf{I} + \mathbf{K} )^{-1} \boldsymbol{\eta}^{(y)},
\end{align*}
where
$\boldsymbol{\eta}^{(y)} \defEq \mathbf{M}^\top \mathbf{T}^{-1} \boldsymbol{\mu}^{(y)}_{\locFeat}$.
The result follows by Lemma~\ref{lemm:eta-diagonal}.
\end{proof}

\begin{lemma}
    \label{lemm:opt-linear-self-kd}
    The optimal solution to the  
    self-distilled square loss 
    population risk
    % ~\eqref{eqn:square-loss-self-kd-risk} 
    is:
    \begin{equation}
        \label{eqn:opt-lin-reg-self-kd}
        \mathbf{W}^*_{\rm sd} = 
        \left( \mathbf{R} + \lambda \cdot \mathbf{I} \right)^{-1} \mathbf{R} \mathbf{W}^*.
        % \frac{1}{L} \cdot \left( \mathbf{S} + \lambda \cdot \mathbf{I} \right)^{-1} \mathbf{S} \left( \mathbf{S} + \lambda \cdot \mathbf{I} \right)^{-1} \mathbf{M}.
    \end{equation}
\end{lemma}

\begin{proof}[Proof of Lemma~\ref{lemm:opt-linear-self-kd}]
Following the proof of Lemma~\ref{lemm:opt-linear}, the gradient of
the self-distilled square loss is:
% ~\eqref{eqn:square-loss-self-kd-risk} is
\begin{align*}
    G( \mathbf{W} ) &= 
    \mathop{\mathbb{E}}_{( x, y ) \sim \Pr_{\rm id}}[ 2 \cdot \boldsymbol{\Phi}( x ) \boldsymbol{\Phi}( x )^\top \mathbf{W} - 2 \cdot \boldsymbol{\Phi}( x ) \boldsymbol{\Phi}( x )^\top \mathbf{W}^*
    ]
    + 2 \cdot \lambda \cdot \mathbf{W}. 
\end{align*}
At optimality, the stationarity condition must hold:
\begin{align*}
    G( \mathbf{W}^*_{\rm sd} ) = 0 &\iff \mathbf{W}^*_{\rm sd} = \left( \mathbf{R} + \lambda \cdot \mathbf{I} \right)^{-1} \mathop{\mathbb{E}}_{( x, y ) \sim 
\Pr_{\rm id}}[ \boldsymbol{\Phi}( x ) \boldsymbol{\Phi}( x )^\top \mathbf{W}^*
] \\
&\iff \mathbf{W}^*_{\rm sd} = \left( \mathbf{R} + \lambda \cdot \mathbf{I} \right)^{-1} \mathbf{R} \mathbf{W}^*. 
% \\
% &\iff \mathbf{W}^*_{\rm sd} = \left( \mathbf{S} + \lambda \cdot \mathbf{I} \right)^{-1} \mathbf{S} \left( \mathbf{S} + \lambda \cdot \mathbf{I} \right)^{-1} \mathop{\mathbb{E}}_{( x, y ) \sim 
% \Pr_{\rm id}}[ \boldsymbol{\Phi}( x ) \mathbf{e}_y^\top ] \\
% &\iff \mathbf{W}^*_{\rm sd} = \frac{1}{L} \cdot \left( \mathbf{S} + \lambda \cdot \mathbf{I} \right)^{-1} \mathbf{S} \left( \mathbf{S} + \lambda \cdot \mathbf{I} \right)^{-1} \mathbf{M}.
\end{align*}
\end{proof}

\begin{lemma}
    \label{lemm:opt-linear-self-kd-similarity}
    Let 
    $\mathbf{W}^*_{\rm sd}$ denote
    the optimal solution to the  
    self-distilled square loss 
    population risk.
    % ~\eqref{eqn:square-loss-self-kd-risk}.
    Then,
    \begin{equation}
        \label{eqn:opt-lin-reg-self-kd-similarity}
        ( \mathbf{W}^*_{\rm sd} )^\top \mathbf{M} = (\mathbf{W}^*)^\top \mathbf{M} - \lambda \cdot L \cdot ( L \cdot \mathbf{I} + \mathbf{K} )^{-1} \mathbf{M}^\top \mathbf{T}^{-2} \mathbf{M} ( L \cdot \mathbf{I} + \mathbf{K} )^{-1}.
    \end{equation}
\end{lemma}

\begin{proof}[Proof of Lemma~\ref{lemm:opt-linear-self-kd-similarity}]
We may write
$\mathbf{R} \left( \mathbf{R} + \lambda \cdot \mathbf{I} \right)^{-1} = \mathbf{I} - \lambda \cdot \left( \mathbf{R} + \lambda \cdot \mathbf{I} \right)^{-1}$.
Thus,
\begin{align*}
( \mathbf{W}^*_{\rm sd} )^\top \mathbf{M} &= (\mathbf{W}^*)^\top \mathbf{R} \left( \mathbf{R} + \lambda \cdot \mathbf{I} \right)^{-1} \mathbf{M} \\
&= ( L \cdot \mathbf{I} + \mathbf{K} )^{-1} \mathbf{M}^\top \mathbf{T}^{-1}  \mathbf{R} \left( \mathbf{R} + \lambda \cdot \mathbf{I} \right)^{-1} \mathbf{M} \\
&= ( L \cdot \mathbf{I} + \mathbf{K} )^{-1} \mathbf{M}^\top \mathbf{T}^{-1} \mathbf{M} - \lambda \cdot ( L \cdot \mathbf{I} + \mathbf{K} )^{-1} \mathbf{M}^\top \mathbf{T}^{-1} ( \mathbf{R} + \lambda \cdot \mathbf{I} )^{-1} \mathbf{M} \\
&= (\mathbf{W}^*)^\top \mathbf{M} - \lambda \cdot L \cdot ( L \cdot \mathbf{I} + \mathbf{K} )^{-1} \mathbf{M}^\top \mathbf{T}^{-1} \mathbf{T}^{-1} \mathbf{M} ( L \cdot \mathbf{I} + \mathbf{K} )^{-1} \\
&= (\mathbf{W}^*)^\top \mathbf{M} - \lambda \cdot L \cdot ( L \cdot \mathbf{I} + \mathbf{K} )^{-1} \mathbf{M}^\top \mathbf{T}^{-2} \mathbf{M} ( L \cdot \mathbf{I} + \mathbf{K} )^{-1},
\end{align*}
where in the penultimate line we use the fact that
$\mathbf{W}^* = \frac{1}{L} \cdot ( \mathbf{R} + \lambda \cdot \mathbf{I} )^{-1} \mathbf{M}$, and the result of Lemma~\ref{lemm:opt-linear}.
\end{proof}

\begin{lemma}
\label{lemm:gram}
Let $\mathbf{W}^*$ be the optimal solution to the square loss population risk as in Lemma~\ref{lemm:opt-linear}.
Under Assumptions~\ref{assum:id-feat}--\ref{assum:balanced-class},
\begin{align*}
( \mathbf{W}^* )^\top \mathbf{M} &= ( L \cdot \mathbf{I} + \mathbf{K} )^{-1} \mathbf{K} \\
( \mathbf{W}^* )^\top \mathbf{W}^* &= ( L \cdot \mathbf{I} + \mathbf{K} )^{-1} \mathbf{M}^\top \mathbf{T}^{-2} \mathbf{M} ( L \cdot  \mathbf{I} + \mathbf{K} )^{-1} = \mathbf{A}.
\end{align*}
Additionally, under Assumption~\ref{assum:simplify},
\begin{align*}
( \mathbf{W}^* )^\top \mathbf{M} &= \mathbf{F} \\
\mathbf{A} &= (L \cdot \tau)^{-1} \cdot \mathbf{F} ( \mathbf{I} - \mathbf{F} ) \\
( \mathbf{R} + \lambda \mathbf{I} )^{-1} \mathbf{W}^* &= \tau^{-1} \mathbf{W}^* ( \mathbf{I} - \mathbf{F} ).
\end{align*}
\end{lemma}

\begin{proof}[Proof of Lemma~\ref{lemm:gram}]
By Lemma~\ref{lemm:opt-linear},
$\mathbf{W}^* = \mathbf{T}^{-1} \mathbf{M} ( L \cdot \mathbf{I} + \mathbf{K} )^{-1}$. 
Since both $\mathbf{T}$ and $( L \cdot \mathbf{I} + \mathbf{K} )$ are symmetric,
\begin{align*}
( \mathbf{W}^* )^\top \mathbf{M} &= ( L \cdot \mathbf{I} + \mathbf{K} )^{-1} \mathbf{M}^\top \mathbf{T}^{-1} \mathbf{M} \\
&= ( L \cdot \mathbf{I} + \mathbf{K} )^{-1} \mathbf{K}.
\end{align*}
Further,
\begin{align*}
( \mathbf{W}^* )^\top \mathbf{W}^* &= ( L \cdot \mathbf{I} + \mathbf{K} )^{-1} \mathbf{M}^\top \mathbf{T}^{-1} \mathbf{T}^{-1} \mathbf{M} ( L \cdot \mathbf{I} + \mathbf{K} )^{-1} \\
&= \mathbf{A},
\end{align*}
where the second line is by definition of $\mathbf{K}^{(2)} = \mathbf{M}^\top \mathbf{T}^{-2} \mathbf{M}$.

Now, under the simplifying Assumption~\ref{assum:simplify}, $\mathbf{T} = \tau \mathbf{I}$.
Thus,
$\mathbf{K}^{(2)} = \tau^{-1} \mathbf{K}$, so 
\begin{align*}
\mathbf{A} &= \tau^{-1} ( L \cdot \mathbf{I} + \mathbf{K} )^{-1} \mathbf{K} ( L \mathbf{I} + \mathbf{K} )^{-1} \\
&= (L \cdot \tau)^{-1} \mathbf{F} \cdot ( \mathbf{I} - \mathbf{F} ),
\end{align*}
where the second line is by the fact that
$( L \mathbf{I} + \mathbf{K} )^{-1} = L^{-1} \cdot ( \mathbf{I} - \mathbf{F} )$ and the commutativity of all matrices involved.

The last identity may be shown as follows.
From the proof of Lemma~\ref{lemm:opt-linear}, $( \mathbf{R} + \lambda \cdot \mathbf{I} )^{-1} \mathbf{M} = L \cdot \mathbf{W}^*$.
Under Assumption~\ref{assum:simplify}, $\mathbf{W}^* = \tau^{-1} \mathbf{M} ( L \cdot \mathbf{I} + \mathbf{K} )^{-1}$. 
Therefore,
\begin{align*}
( \mathbf{R} + \lambda \cdot \mathbf{I} )^{-1} \mathbf{W}^* &= \tau^{-1} ( \mathbf{R} + \lambda \cdot \mathbf{I} )^{-1} \mathbf{M} ( L \cdot \mathbf{I} + \mathbf{K} )^{-1} \\
& = \tau^{-1} \mathbf{W}^* \cdot L ( L \cdot \mathbf{I} + \mathbf{K} )^{-1} \\
&= \tau^{-1} \mathbf{W}^* ( \mathbf{I} - \mathbf{F} ).
\end{align*}
\end{proof}

%% file: sec/suppl/theory_extended.tex
One may relate the KD and ListMLE losses as follows.

\begin{proposition}
\label{prop:kd-pl-bound}
    Pick any $x \in \XCal$.
    Suppose the
    teacher distribution $\mathbf{p}^{\rm t} \colon \XCal \to \Delta( \YCal )$ satisfies 
    % \todoakm{Refine to use second-smallest probability}
    $$ p^{\rm t}_{[L-1]}( x ) > 0, $$
    where $p^{\rm t}_{[L-1]}( x )$ denotes the second smallest element of $\mathbf{p}^{\rm t}(x)$.
    Let $\boldsymbol{\pi}^{\rm t}$ denote the ranking according the teacher's class probabilities.
    Then, 
    there exists $b( x ) \in \Real$ such that
    for any $\mathbf{f}( x ) \in \Real^L$,
    the KD loss~\eqref{eqn:kd} and ListMLE loss~\eqref{eqn:pl-loss} are related by:
    $$ \ell_{\rm pl}( \boldsymbol{\pi}^{\rm t}( x ), \mathbf{f}( x ) ) \leq \frac{1}{p^{\rm t}_{{[L-1]}}( x )} \cdot \ell_{\rm kd}( \mathbf{p}^{\rm t}( x ), \mathbf{f}( x ) ) - b( x ). $$
\end{proposition}

Proposition~\ref{prop:kd-pl-bound} 
% has an 
% similar interpretation to the second half of Proposition~\ref{prop:sandwich-bound}:
% intuitive interpretation:
suggests
the KD loss implicitly minimises the ListMLE loss against $\boldsymbol{\pi}^{\rm t}$;
however, the bound involves a large scaling factor when $p^{\rm t}_{\rm min}( x ) \ll 1$.
This is intuitive: when the teacher's tail probabilities are small, the standard KD loss applies a much weaker penalty than the ListMLE loss.
% Interestingly, 
% unlike Proposition~\ref{prop:sandwich-bound}
% \todoakm{Perhaps move to Appendix}

%
\begin{proof}[Proof of Proposition~\ref{prop:kd-pl-bound}]
For brevity,
we omit the dependence on $x$, and consider instead a generic distribution $q \in \Delta( \YCal )$, whose elements are non-increasing; i.e.,
$q_1 \geq q_2 \geq \ldots \geq q_L$.
% \todoakm{Consistent notation.}
For predicted distribution $p \in \Delta( \YCal )$,
consider then the KD loss with $q$ as the teacher, and instance-level ListMLE loss with $q$ as the ranking signal:
\begin{align*}
    \ell_{\rm kd}( p; q ) &\defEq \sum_{i = 1}^{L} q_i \cdot -\log p_i \\
    \ell_{\rm pl}( p ) &\defEq \sum_{i = 1}^{L} -\log \frac{p_i}{\sum_{j = i}^{L} p_j}.
\end{align*}
Note that in the latter case, we implicitly use the ranking over classes implied by $q$, since the summation follows the indices of non-increasing elements in $q$.
% However, unlike the former loss, we do not use the \emph{strengths} of the individual elements in $q$.

Define the \emph{rescaled} predictions to be
$$ ( \forall i \in \YCal ) \, \varrho_i \defEq \frac{p_i}{\sum_{j = i}^{L} p_j}, $$
so that $\varrho_1 = p_1$ and $\varrho_L = 1$.
Note that $\sum_{i \in [L]} \varrho_i \neq 1$ in general.
We have
$$ ( \forall i \in [ L - 1 ] ) \, 1 - \varrho_i = \frac{\sum_{j = i+1}^{L} p_j}{\sum_{j = i}^{L} p_j}, $$
and $1 - \varrho_L = 0$.
Thus,
$$ ( \forall i \in \YCal ) \, \prod_{j = 1}^{i - 1} ( 1 - \varrho_j ) = {\sum_{j = i}^{L} p_j}, $$
% $$ ( \forall i \in \YCal ) \, \prod_{j = i}^{L - 1} ( 1 - s_j ) = \frac{p_L}{\sum_{j = i}^{L} p_j}. $$
and so
$$ ( \forall i \in \YCal ) \, p_i = \varrho_i \cdot \prod_{j = 1}^{i - 1} ( 1 - \varrho_j ). $$
Equipped with this, we may rewrite
\begin{align*}
    \ell_{\rm kd}( p; q ) &\defEq 
    % \sum_{i = 1}^{L} q_i \cdot \left[ -\log \varrho_i + \sum_{j = 1}^{i - 1} -\log (1 - s_j) \right] \\
    % &= 
    \sum_{i = 1}^{L-1} \left[ q_i \cdot -\log \varrho_i + \left( \sum_{j = i+1}^{L} q_j \right) \cdot -\log (1 - \varrho_i) \right] \\
    \ell_{\rm pl}( p ) &\defEq \sum_{i = 1}^{L-1} -\log \varrho_i.
\end{align*}

Note that each term in 
$\ell_{\rm kd}$ 
is a weighted \emph{binary} cross-entropy.
In particular,
\begin{align*}
   \ell_{\rm kd}( p; q ) &= \sum_{i = 1}^{L-1} \left( \sum_{j = i}^{L} q_j \right) \cdot \ell_{\rm BCE}( \bar{q}_{i}, \varrho_i ) \\
   \bar{q}_{i} &\defEq \frac{q_i}{\sum_{j = i}^{L} q_j}.
\end{align*}
Similarly, for any $0 < a \leq q_{L - 1}$,
\begin{align*}
   \ell_{\rm kd}( p; q ) - a \cdot \ell_{\rm pl}( p ) &= \sum_{i = 1}^{L-1} \left( \sum_{j = i}^{L} q_j - a \right) \cdot \ell_{\rm BCE}( \tilde{q}_{i}, \varrho_i ) \\
   \tilde{q}_{i} &\defEq \frac{q_i - a}{\sum_{j = i}^{L} q_j - a}.
\end{align*}
Now, the binary cross-entropy can be lower bounded by the entropy, yielding:
\begin{align*}
   \ell_{\rm kd}( p; q ) - a \cdot \ell_{\rm pl}( p ) &\geq \sum_{i = 1}^{L-1} \left( \sum_{j = i}^{L} q_j - a \right) \cdot H_{\rm bin}( \tilde{q}_{i} ),
\end{align*}
where $H_{\rm bin}$ denotes the binary Shannon entropy.
Note that the RHS does not depend on $p$, but only on $q$.
Thus, we may regard it as a constant, and conclude that for any $0 \leq a \leq q_{L - 1}$, there is a constant $b( a, q )$ such that:
$$ \ell_{\rm pl}( p ) \leq \frac{1}{a} \cdot \ell_{\rm kd}( p; q ) - b( a, q ). $$
\end{proof}

\label{app:bias-variance}
Finally, we may derive the following bias-variance decomposition in terms of the marginal teacher probabilities.
Pick any class $y \in [L]$.
Let $\mathbf{p}^{\rm t}$ be a teacher distribution,
and $\mathbf{p}^*$ be the Bayes class-probability distribution.
Further, let $\bar{\mathbf{p}}^{*, (y)}$ denote the marginal probability for the Bayes distribution.
Then,
\begin{align*}
    \mathop{\mathbb{E}}_{{x} \mid y}\left[ 
    \| \mathbf{p}^{\rm t}( x ) - \mathbf{p}^{*}( x ) \|_2^2
    \right] &= 
    \mathop{\mathbb{E}}_{{x} \mid y}\left[ 
    \| \mathbf{p}^{\rm t}( x ) - \mathbf{p}^{*}( x ) + \bar{\mathbf{p}}^{(y)} - \bar{\mathbf{p}}^{(y)} \|_2^2
    \right] \\
    &= 
    \mathop{\mathbb{E}}_{{x} \mid y}\left[ 
    \| (\mathbf{p}^{\rm t}( x ) - \bar{\mathbf{p}}^{(y)}) - (\mathbf{p}^{*}( x ) - \bar{\mathbf{p}}^{(y)}) \|_2^2
    \right] \\
    &= 
    \mathop{\mathbb{E}}_{{x} \mid y}\left[ 
    \| \mathbf{p}^{\rm t}( x ) - \bar{\mathbf{p}}^{(y)} \|_2^2 + 
    \| \mathbf{p}^{*}( x ) - \bar{\mathbf{p}}^{(y)} \|_2^2 - 
    2 \cdot (\mathbf{p}^{\rm t}( x ) - \bar{\mathbf{p}}^{(y)})^\top (\mathbf{p}^{*}( x ) - \bar{\mathbf{p}}^{(y)})
    \right] \\
    &= 
    \mathop{\mathbb{E}}_{{x} \mid y}\left[ 
    \| \mathbf{p}^{\rm t}( x ) - \bar{\mathbf{p}}^{(y)} \|_2^2 + 
    \| \mathbf{p}^{*}( x ) - \bar{\mathbf{p}}^{(y)} \|_2^2 - 
    2 \cdot (\mathbf{p}^{\rm t}( x ) - \bar{\mathbf{p}}^{(y)})^\top (\mathbf{p}^{*}( x ) - \bar{\mathbf{p}}^{*, (y)})
    \right].
\end{align*}
Therefore,
\begin{align*}
    \mathop{\mathbb{E}}_{{x} \mid y}\left[ 
    \| \mathbf{p}^{\rm t}( x ) - \mathbf{p}^{*}( x ) \|_2^2
    \right] - 
    \mathop{\mathbb{E}}_{{x} \mid y}\left[
    \| \mathbf{p}^{*}( x ) - \bar{\mathbf{p}}^{(y)} \|_2^2 
    \right]
    &= 
    \mathop{\mathbb{E}}_{{x} \mid y}\left[ 
    \| \mathbf{p}^{\rm t}( x ) - \bar{\mathbf{p}}^{(y)} \|_2^2 - 
    2 \cdot (\mathbf{p}^{\rm t}( x ) - \bar{\mathbf{p}}^{(y)})^\top (\mathbf{p}^{*}( x ) - \bar{\mathbf{p}}^{*, (y)})
    \right]
\end{align*}
The LHS is the expected difference 
in fidelity of modelling $\mathbf{p}^*$
between the instance-level teacher
and the class-level marginal teacher.
We see that the instance-level teacher has better fidelity than the class-level teacher iff
$$ \mathop{\mathbb{E}}_{{x} \mid y}\left[ 
    \| \mathbf{p}^{\rm t}( x ) - \bar{\mathbf{p}}^{(y)} \|_2^2 \right]
    <
    2 \cdot \mathop{\mathbb{E}}_{{x} \mid y}\left[
    (\mathbf{p}^{\rm t}( x ) - \bar{\mathbf{p}}^{(y)})^\top (\mathbf{p}^{*}( x ) - \bar{\mathbf{p}}^{*, (y)})
    \right], $$
i.e., the teacher's expected deviation around its class-mean cannot be ``excessive''.
Here, ``excessive'' is determined by the covariance between the teacher and Bayes probability.

%% file: sec/suppl/related_extended.tex
\subsection{Approaches to OOD detection}

Broadly, there are two categories of OOD detection techniques.
\emph{Post-hoc} techniques do not disturb the training algorithm for learning $\mathbf{f}$, and simply construct $g$ post-hoc based on certain fixed criteria.
These include the
\emph{maximum softmax probability (MSP})~\citep{Hendrycks:2017,Liang:2018}, defined as 
% the model's probability of the predicted class, i.e.,
$g_{\rm MSP}( x ) \defEq \max_{y' \in \YCal} p_{y'}( x )$,
{maximum logit (MaxLogit)}~\citep{Hendrycks:2022},
% $g_{\rm MLS}( x ) \defEq \max_{y \in \YCal} f_{y}( x )$, 
% the
{energy-based score}~\citep{Liu:2020b},
% $g_{\rm EB}( x ) \defEq -\log\left[ \sum_{y' \in \YCal} \exp( f_{y'}( x ) ) \right]$,
{LogitGap}~\citep{Liang:2025},
% \todoakm{Consistent notation}
% $g_{\rm LG}( x ) \defEq \max_{y \in \YCal} f_y( x ) - \frac{1}{L - 1} \sum_{y' \neq y} f_{y'}( x )$,
and many others~\citep{Lee:2018b,Ren:2019,Sun:2021,wang2022vim,Morteza:2022,Ammar:2024}.
% Arguably the simplest possible post-hoc rule is the \emph{maximum softmax probability (MSP})~\citep{Hendrycks:2017,Liang:2018}, defined as 
% % the model's probability of the predicted class, i.e.,
% $g_{\rm MSP}( x ) \defEq \max_{y' \in \YCal} p_{y'}( x )$.
% While intuitive,
% this rule can be easily defeated~\citep{Nguyen:2015,Hein:2019}.
% % indeed, even random inputs can lead to high predicted probabilities on certain classes~\citep{Nguyen:2015,Hein:2019}.
% This has motivated more refined rules, such as the 
% \emph{maximum logit (MaxLogit)}~\citep{Hendrycks:2022},
% % $g_{\rm MLS}( x ) \defEq \max_{y \in \YCal} f_{y}( x )$, 
% % the
% \emph{energy-based score}~\citep{Liu:2020b},
% % $g_{\rm EB}( x ) \defEq -\log\left[ \sum_{y' \in \YCal} \exp( f_{y'}( x ) ) \right]$,
% \emph{LogitGap}~\citep{Liang:2025},
% % \todoakm{Consistent notation}
% % $g_{\rm LG}( x ) \defEq \max_{y \in \YCal} f_y( x ) - \frac{1}{L - 1} \sum_{y' \neq y} f_{y'}( x )$,
% and many others~\citep{Lee:2018b,Ren:2019,Sun:2021,Wang:2022,Morteza:2022,Ammar:2024}.
\emph{Training-based} techniques seek to modify the training of $\mathbf{f}$, so as to better separate ID and OOD samples.
One possibility is to perform \emph{outlier exposure}~\citep{Hendrycks:2019}, under the 
(potentially strong)
assumption of access to a small number of known OOD samples.
% however, this assumption may be restrictive.
Without access to any OOD samples,
one may seek to encourage distinctive behaviour for ID versus OOD samples by various means of regularisation~\citep{Wei:2022,Hendrycks:2019b,Tack:2020,Isaac:2024}.
% , such as by normalising logit norms~\citep{Wei:2022}, 
% exploiting self-supervision~\citep{Hendrycks:2019b},
% contrastive learning~\citep{Tack:2020},
% encouraging reduction of feature nullspaces~\citep{Isaac:2024},
% amongst many other approaches.
Another line of work performs \emph{two-stage} training, where the first stage model outputs refine the loss in the second stage.
% These relate to KD approaches for OOD detection.
% , which we now discuss.
% \todoakm{Should we mention one- versus two-stage methods?}
For example, several works have studied various forms of (self-)KD for OOD detection~\citep{Wu:2023,Tang:2025,Yang:2025,Zhao:2025,Wang:2023,Kralik:2025}.
% , including 
% leveraging richer signals beyond logits~\citep{Wu:2023,Tang:2025}, 
% identifying or synthesising pseudo-OOD samples~\citep{Yang:2025,Zhao:2025}, 
% leveraging ensembles of teachers~\citep{Wang:2023},
% and using the student-teacher discrepancy as a scoring function~\citep{Kralik:2025}.

%
\subsection{Theoretical study of OOD detection}

While there is an extensive literature proposing various techniques for OOD detection,
the theoretical study of the problem has been more modest.
\citet{Hein:2019} established that the reliance on ReLU activations in deep networks can fundamentally be at odds with preventing over-confident predictions.
\citet{Fang:2024} showed that without further assumptions, the problem may not be PAC learnable.
\citet{Morteza:2022} established guarantees for the generalised energy model scoring rule, designed for mixture of Gaussian distributions.

% \citet{Zhang:2023} proposed a nuisance-aware training algorithm to explicitly suppress 
% the role of nuisance features that detract from OOD detection.
% \todoakm{Discuss more?}

%
\subsection{Ranking and distillation losses}

The relation between KD and ListMLE in Proposition~\ref{prop:sandwich-bound} is, to our best knowledge, novel.
However, more broadly, the formal relation between \emph{probabilistic} and \emph{ranking} losses has been previously studied.
For example, in the setting of binary classes, ~\citet{Kotlowski:2011,Agarwal:2014} established bounds relating
(excess risk under) certain \emph{proper} (e.g., log-loss) and \emph{pairwise ranking} losses,
while~\citet{Narasimhan:2013} established precise bounds relating the problem of class-probability estimation and ranking.
Roughly, these establish that accurate class-probability estimation translates to accurate ranking, while the converse does not hold without further transformations of the ranking.

Interestingly, Proposition~\ref{prop:epl-loss} can be related to results commonly employed in the \emph{reinforcement learning} literature (concerning the equivalence of various expressions of \emph{policy gradients})~\citep{GAE2016}, 
and the \emph{sequence distillation} literature (concerning the equivalence of sequence- and token-level losses)~\citep{amini2026better}.
More precisely, all these results rely on a particular instantiation of the law of total expectation.
To our best knowledge, however, the relevance of such a result in the context of label ranking (i.e., for the expected Plackett-Luce loss) has not been previously considered.

Ranking-based distillation losses 
akin to the EPL loss
have also been studied, mostly notably in~\citet{Reddi:2021}.
However, 
% unlike the regular KD loss (which we focus on),
% these natively involve computing an intractable expectation over class rankings, and thus require some form of approximation.
% Further, 
to our best knowledge, analogues of the PL-KD loss~\eqref{eqn:refined-estimator} have not been studied in this context;
rather, the proposal is akin to the standard ListMLE, with a single ranking sampled from the teacher distribution.
Further exploration of the PL-KD loss in retrieval contexts is a potentially interesting direction for future work.

%% file: sec/suppl/illustration.tex
\begin{figure}[!t]
    \centering
    \includegraphics[width=0.995\linewidth]{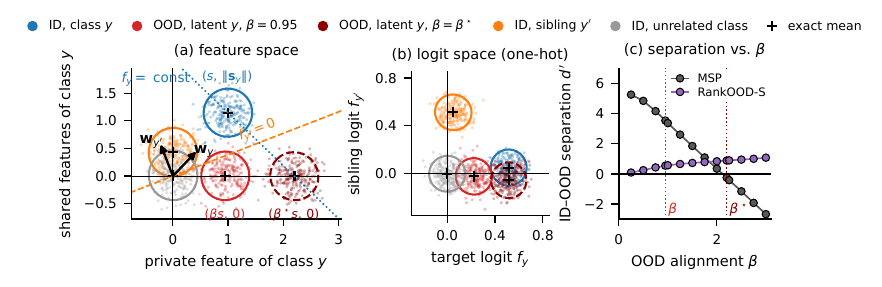}
    \caption{\textbf{Illustration of generative model (\S\ref{sec:theory}).}
In-distribution (ID) samples of class $y$ (blue) carry both $y$'s local feature, and global features shared with a related sibling class $y'$ (orange).
Out-of-distribution (OOD) samples (red) only mimic $y$'s \emph{local} feature with strength $\beta$.
When $\beta$ is large ($\beta = \beta^\star$, dark red), OOD samples match ID samples on the target logit $f_y$ and fool confidence scores (MSP).
However, the sibling weight $\mathbf{w}_{y'}$ penalises $y$'s local feature, pushing $f_{y'} < 0$ on OOD samples (Proposition~\ref{corr:opt-logits-first-order-informal}) so ranking separation ({\sc RankOOD}-S) improves with $\beta$.
}
    \label{fig:fig_paper_generative_model}
\end{figure}

Figure~\ref{fig:fig_paper_generative_model} provides an intuitive illustration of our generative model.
Here, we have three ID classes:
a target class $y$ (blue), a semantically related sibling $y'$ (orange), and an unrelated class $y''$ (gray).
In-distribution (ID) samples of class $y$ (blue) have both $y$'s local feature, and global features shared with class $y'$ (orange).
On the other hand, class $y''$ shares neither the local nor global features.
Thus, the three ID classes occupy three extremities in feature space.
Finally, we have OOD samples (red) that spuriously align with class $y$'s local feature (with strength $\beta$), but do \emph{not} possess the shared global features.

In logit space, ID samples of class $y$ get a positive sibling logit ($f_{y'} > 0$) due to the shared features.
The OOD samples appear similar to $y$ in terms of the target logit $f_y$,
but are dissimilar on non-target logits $f_{y'}$.

As $\beta$ increases, confidence-based detection (MSP) degrades and fails at $\beta^\star$, whereas ranking-based detection ({\sc RankOOD}-S) steadily improves.

%% file: sec/suppl/worked_example.tex
% ---------------------------------------------------------------------------
% Section: a worked example distinguishing the five training objectives.
% Formatted for a two-column layout (e.g. CVPR): every table and display fits
% within \columnwidth. Requires: amsmath, booktabs.
% Uses the paper's notation: \ell_{ce}, \ell_{kd} (eq. 1), \ell_{pl} (ListMLE, eq. 2),
% p^t(x), \bar\pi^{(y)}, \pi^t(x) = argsort(p^t(x)), and the PL-KD loss (eq. 6).
% Rename these labels to match the paper:
%   eq:kd -> KD loss (1)   eq:listmle -> ListMLE loss (2)   eq:plkd -> PL-KD loss (6)
%   tab:objectives -> Table 1 (summary of objectives)   sec:ranking-meets-kd -> Section 3
% ---------------------------------------------------------------------------

We illustrate how the five objectives of Table~\ref{tab:example-supervision} differ on a single training image. Consider $L = 4$ classes, $\mathcal{Y} = \{\textsf{dog}, \textsf{wolf}, \textsf{fox}, \textsf{car}\}$, and a training image $x$ with label $y = \textsf{dog}$ whose appearance is unusually fox-like. In the order (dog, wolf, fox, car), the first-stage (teacher) model assigns
\begin{equation*}
  p^t(x) = (0.70,\ 0.08,\ 0.20,\ 0.02).
\end{equation*}
Across the dog class as a whole, wolf is the usual runner-up, so the class-level canonical ranking is
\begin{equation*}
  \bar{\pi}^{(\textsf{dog})} = \textsf{dog} \succ \textsf{wolf} \succ \textsf{fox} \succ \textsf{car}.
\end{equation*}
For this particular image, however, the teacher's own ranking is
\begin{equation*}
  \pi^t(x) = \operatorname{argsort}\big(p^t(x)\big) = \textsf{dog} \succ \textsf{fox} \succ \textsf{wolf} \succ \textsf{car}.
\end{equation*}
The image is thus \emph{atypical} for its class, and this is exactly where the objectives part ways. For a typical dog image, $\bar{\pi}^{(\textsf{dog})}$ and $\pi^t(x)$ coincide, and RankOOD and its instance-level variant receive identical supervision.

\paragraph{What each objective is told.}
Table~\ref{tab:example-supervision} lists the supervision that each objective derives for $x$. The PL-KD target is the ranking distribution $p^t_{\mathrm{pl}}(\cdot \mid x)$ induced by $p^t(x)$. Its three most likely rankings are
\begin{align*}
  \textsf{dog} \succ \textsf{fox} \succ \textsf{wolf} \succ \textsf{car} &\quad (0.373), \\
  \textsf{dog} \succ \textsf{wolf} \succ \textsf{fox} \succ \textsf{car} &\quad (0.170), \\
  \textsf{fox} \succ \textsf{dog} \succ \textsf{wolf} \succ \textsf{car} &\quad (0.140),
\end{align*}
and its conditional distributions (cf. Eq.~\eqref{eqn:refined-estimator}) include, over (wolf, fox, car) and (wolf, car) respectively,
\begin{align*}
  p^t\big(x, (\textsf{dog})\big) &= (0.267,\ 0.667,\ 0.067), \\
  p^t\big(x, (\textsf{dog}, \textsf{fox})\big) &= (0.8,\ 0.2).
\end{align*}
So PL-KD is told that fox \emph{probably} follows dog, together with how confident the teacher is about each such comparison.

\begin{table}[t]
  \centering
  \caption{Supervision derived for the fox-like dog image $x$. Classes are abbreviated \textsf{d}\,(dog), \textsf{w}\,(wolf), \textsf{f}\,(fox), \textsf{c}\,(car); inst.\ = instance-level.}
  \label{tab:example-supervision}
  \small
  \setlength{\tabcolsep}{4pt}
  \begin{tabular}{@{}lll@{}}
    \toprule
    Objective & Target for $x$ & Type \\
    \midrule
    CE & $y = \textsf{d}$ & label \\
    RankOOD & $\bar{\pi}^{(\textsf{d})} = \textsf{d} \succ \textsf{w} \succ \textsf{f} \succ \textsf{c}$ & class, hard rank \\
    Self-KD & $p^t(x) = (.70, .08, .20, .02)$ & inst., soft prob. \\
    Inst.\ RankOOD & $\pi^t(x) = \textsf{d} \succ \textsf{f} \succ \textsf{w} \succ \textsf{c}$ & inst., hard rank \\
    PL-KD & $p^t_{\mathrm{pl}}(\cdot \mid x)$ & inst., soft rank \\
    \bottomrule
  \end{tabular}
\end{table}

\paragraph{Four candidate students.}
To see which outputs each objective rewards, we evaluate it on four candidate student distributions $p = \mathrm{softmax}(f(x))$, all at least as confident in the correct class as the teacher:
\begin{itemize}
  \item[\textbf{A}] matches the teacher: \\ $p = (0.70, 0.08, 0.20, 0.02)$;
  \item[\textbf{B}] follows the class-typical order, with wolf and fox swapped: \\ $p = (0.70, 0.20, 0.08, 0.02)$;
  \item[\textbf{C}] has flat non-target probabilities: \\ $p = (0.70, 0.10, 0.10, 0.10)$;
  \item[\textbf{D}] is the teacher sharpened to temperature $1/2$, $p \propto (p^t)^2$: \\ $p = (0.913, 0.012, 0.075, 0.001)$.
\end{itemize}

\begin{table}[t]
  \centering
  \caption{Loss of each candidate student (A--D) under each objective for the image $x$; bold marks the preferred (lowest-loss) student per column. The lower block gives the excess loss over student A. KD losses use the cross-entropy form of \eqref{eqn:kd}; PL-KD is the exact expectation of Eq.~\ref{eqn:refined-estimator} over all $4! = 24$ rankings.}
  \label{tab:example-losses}
  \small
  \setlength{\tabcolsep}{5pt}
  \begin{tabular}{@{}lccccc@{}}
    \toprule
    & CE & RankOOD & Self-KD & \begin{tabular}[b]{@{}c@{}}Inst.\\ RankOOD\end{tabular} & PL-KD \\
    \midrule
    A & 0.357 & 1.774 & \textbf{0.852} & 0.985 & \textbf{2.010} \\
    B & 0.357 & \textbf{0.985} & 0.962 & 1.774 & 2.464 \\
    C & 0.357 & 2.148 & 0.940 & 2.148 & 2.552 \\
    D & \textbf{0.091} & 2.091 & 1.082 & \textbf{0.309} & 2.510 \\
    \midrule
    B$-$A & 0 & $-0.789$ & 0.110 & 0.789 & 0.454 \\
    C$-$A & 0 & 0.374 & 0.089 & 1.163 & 0.543 \\
    D$-$A & $-0.266$ & 0.317 & 0.230 & $-0.676$ & 0.501 \\
    \bottomrule
  \end{tabular}
\end{table}

\paragraph{What each objective prefers.}
Table~\ref{tab:example-losses} reports the losses. The five objectives prefer four different students.

\begin{itemize}
  \item \textbf{CE} depends only on $p_{\textsf{dog}}$. It is indifferent between A, B and C, and prefers the over-confident D. It carries no information about the non-target classes.

  \item \textbf{RankOOD} prefers B. Its class-level target imposes the typical dog ordering, so it pushes this image's student towards ranking wolf above fox, against the evidence in the image. This is the risk of class-level supervision discussed in \S\ref{sec:rankood}: instance-specific structure is suppressed, which can cost ID accuracy.

  \item \textbf{Self-KD} prefers A, the correct target, but only weakly: B and C cost just $0.110$ and $0.089$ more than A. The KD loss weights the non-target shape by the teacher's non-target mass $1 - p^t_{\textsf{dog}}(x) = 0.30$. For C, the excess is exactly
  \begin{align*}
    &0.30 \times \mathrm{KL}\big(p^t(x, (\textsf{dog})) \,\big\|\, p(x, (\textsf{dog}))\big) \\
    &\quad = 0.30 \times 0.295 = 0.089.
  \end{align*}
  With a confident teacher, the ordering of the non-target classes receives little signal.

  \item \textbf{Instance-level RankOOD} has the right ordering but prefers D. The ListMLE loss rewards every widening of the margins in $\pi^t(x)$, so its infimum is approached only as the logits diverge. It uses the teacher's ordering but discards its confidence, and so treats fox $\succ$ wolf ($0.20$ vs.\ $0.08$) as being as certain as wolf $\succ$ car.

  \item \textbf{PL-KD} prefers A, and equally importantly: the excess losses of B, C and D ($0.454$, $0.543$, $0.501$) are two to six times those under Self-KD. Each position of Eq.~\ref{eqn:refined-estimator} compares renormalised conditionals at full weight, so the order of the non-target classes is supervised directly, while the soft target keeps the student calibrated rather than sharpened.
\end{itemize}

\paragraph{Summary.}
The example separates the two design axes of Table~\ref{tbl:summary}. Along the \emph{granularity} axis, class-level supervision (RankOOD) imposes an ordering that is wrong for atypical instances, which instance-level supervision avoids. Along the \emph{loss} axis, a hard ranking (instance-level RankOOD) has the right order but rewards over-confidence, while the KD loss (Self-KD) is calibrated but supervises the non-target order only weakly. PL-KD is the only objective of the five whose minimiser is the teacher's instance-level distribution \emph{and} which supervises the non-target ordering at full strength.

In practice, each second-stage objective is combined with $\ell_{ce}$, RankOOD supervises only the top and bottom ranks, and PL-KD uses a single sampled ranking per example, an unbiased estimate of Eq.~\ref{eqn:refined-estimator}. These choices do not change the qualitative picture.

%% file: sec/suppl/results_appendix.tex
Additional experimental details and extended results are provided to complement \S\ref{sec:experiments}. The main text reports mean FPR95 values for both near- and far-OOD detection across CIFAR-10, CIFAR-100, and ImageNet-200 (Table~\ref{tbl:nearood}, and \ref{tbl:farood}), averaged over three randomly initialized models. Tables~\ref{tbl:appendix_fpr_cifar_10}, \ref{tbl:appendix_fpr_cifar100}, and \ref{tbl:appendix_fpr_imagenet200} further present the corresponding means and standard deviations to capture performance variability. In addition, AUROC results for near- and far-OOD detection are reported in Tables~\ref{tbl:appendix_auroc_cifar_10}, \ref{tbl:appendix_auroc_cifar_100}, and \ref{tbl:appendix_auroc_imagenet200}. 

% \subsection{CIFAR-10}
% \label{app:results_extended_cifar10}
\setlength{\tabcolsep}{4pt} 
\begin{table}[!h]
\centering
\caption{Average FPR@95 on near-OOD and far-OOD datasets with CIFAR-10 as the ID dataset. IL denotes instance-level.}
\label{tbl:appendix_fpr_cifar_10}
\renewcommand{\arraystretch}{1.25}
\resizebox{\textwidth}{!}{
\begin{tabular}{@{}lcccccc|ccccc@{}}
\toprule
& & \multicolumn{5}{c|}{\textbf{Near-OOD}} & \multicolumn{5}{c}{\textbf{Far-OOD}} \\
\midrule
& \textbf{\makecell{ID \\ accuracy}} & \textbf{\makecell{LogitGap \\ FPR95}} & \textbf{\makecell{MSP \\ FPR95}} & \textbf{\makecell{MLS \\ FPR95}} & \textbf{\makecell{EBO \\ FPR95}} & \textbf{\makecell{RankOOD \\ FPR95}} & \textbf{\makecell{LogitGap \\ FPR95}} & \textbf{\makecell{MSP \\ FPR95}} & \textbf{\makecell{MLS \\ FPR95}} & \textbf{\makecell{EBO \\ FPR95}} & \textbf{\makecell{RankOOD \\ FPR95}} \\
\toprule
CE & \textbf{0.951$\pm$0.004} & 0.613$\pm$0.057 & 0.482$\pm$0.048 & 0.613$\pm$0.057 & 0.613$\pm$0.057 & 0.591$\pm$0.058 & 0.417$\pm$0.065 & 0.317$\pm$0.022 & 0.417$\pm$0.065 & 0.417$\pm$0.065 & 0.397$\pm$0.047\\	

RankOOD & 0.941$\pm$0.003 & {0.364$\pm$0.040} & {0.379$\pm$0.036} & {0.365$\pm$0.041} &  {0.365$\pm$0.041} & \textbf{0.372$\pm$0.038} & 0.236$\pm$0.024 & 0.255$\pm$0.029 & 0.236$\pm$0.024 & 0.236$\pm$0.025 & \textbf{0.240$\pm$0.21}\\

\hline
Self-KD  & {0.949$\pm$0.001} & 0.313$\pm$0.001 & \textbf{0.337$\pm$0.009} & 0.316$\pm$0.001 & 0.315$\pm$0.001 & 0.476$\pm$0.035 & {0.200$\pm$0.070} & \textbf{0.240$\pm$0.008} & {0.201$\pm$0.010} & {0.201$\pm$0.010} & 0.302$\pm$0.042 \\

% Class-level KD & 0.955$\pm$0.002 & 0.407$\pm$0.019 & 0.383$\pm$0.017 & 0.419$\pm$0.021 & 0.419$\pm$0.021 & 0.488$\pm$0.004 & 0.242$\pm$0.005 & 0.239$\pm$0.003 & 0.247$\pm$0.005 & 0.247$\pm$0.005 & 0.281$\pm$0.006 \\

% MT ListMLE & 0.947$\pm$0.003 & 0.476$\pm$0.068 & 0.476$\pm$0.071 & 0.478$\pm$0.067 & 0.478$\pm$0.067 & 0.446$\pm$0.047 & 0.304$\pm$0.020 & 0.309$\pm$0.021 & 0.304$\pm$0.020 & 0.304$\pm$0.020 & 0.275$\pm$0.021 \\

IL RankOOD & 0.934$\pm$0.005 & 0.355$\pm$0.017 & 0.352$\pm$0.017 & 0.359$\pm$0.016 & 0.363$\pm$0.018 & 0.435$\pm$0.058 & 0.244$\pm$0.013 & 0.250$\pm$0.010 & 0.246$\pm$0.013 & 0.250$\pm$0.013 & 0.320$\pm$0.037  \\

% TW RankOOD$_{Instance}$ & 0.957$\pm$0.001 & 0.385$\pm$0.031 & 0.361$\pm$0.023 & 0.395$\pm$0.034 & 0.396$\pm$0.034 & 0.407$\pm$0.026 & 0.223$\pm$0.008 & 0.221$\pm$0.008 & 0.226$\pm$0.009 & 0.226$\pm$0.009 & 0.244$\pm$0.021  \\

% TW RankOOD$_{Canonical}$ & 0.955$\pm$0.001 & 0.415$\pm$0.027 & 0.382$\pm$0.013 & 0.421$\pm$0.030 & 0.421$\pm$0.030 & 0.384$\pm$0.012 & 0.241$\pm$0.025 & 0.237$\pm$0.022 & 0.245$\pm$0.027 & 0.245$\pm$0.027 & 0.242$\pm$0.023 \\

IL PL-KD & 0.947$\pm$0.005 & \textbf{0.301$\pm$0.005} & 1.000$\pm$0.000 & \textbf{0.305$\pm$0.007} & \textbf{0.305$\pm$0.007} & 0.414$\pm$0.032 & \textbf{0.191$\pm$0.010} & 0.475$\pm$0.111 & \textbf{0.193$\pm$0.010} & \textbf{0.193$\pm$0.010} & 0.287$\pm$0.016 \\

% \hline
\bottomrule
\end{tabular}
}
\end{table}

\begin{table}[!h]
\centering
\caption{Average AUROC on near-OOD and far-OOD datasets with CIFAR-10 as the ID dataset. IL denotes instance-level.}
\label{tbl:appendix_auroc_cifar_10}
\renewcommand{\arraystretch}{1.25}
\resizebox{\textwidth}{!}{
\begin{tabular}{@{}lccccc|ccccc@{}}
\toprule
& & \multicolumn{4}{c|}{\textbf{Near-OOD}} & \multicolumn{5}{c}{\textbf{Far-OOD}} \\
\midrule
& \textbf{\makecell{LogitGap \\ AUROC}} & \textbf{\makecell{MSP \\ AUROC}} & \textbf{\makecell{MLS \\ AUROC}} & \textbf{\makecell{EBO \\ AUROC}} & \textbf{\makecell{RankOOD \\ AUROC}} & \textbf{\makecell{LogitGap \\ AUROC}} & \textbf{\makecell{MSP \\ AUROC}} & \textbf{\makecell{MLS \\ AUROC}} & \textbf{\makecell{EBO \\ AUROC}} & \textbf{\makecell{RankOOD \\ AUROC}} \\
\toprule
CE & 0.875$\pm$0.006 & 0.880$\pm$0.003 & 0.875$\pm$0.006 & 0.876$\pm$0.006 & 0.859$\pm$0.015 & {0.911$\pm$0.011} & {0.907$\pm$0.006} & {0.911$\pm$0.011} & {0.912$\pm$0.011} & {0.903$\pm$0.009} \\

RankOOD  & 0.899$\pm$0.006 & 0.889$\pm$0.005 & 0.899$\pm$0.006 & 0.899$\pm$0.006 & {\bf 0.895$\pm$0.005} & 0.929$\pm$0.008 & {\bf 0.918$\pm$0.008} & 0.929$\pm$0.008 & 0.929$\pm$0.008 & {\bf 0.925$\pm$0.008} \\

\hline
Self-KD   & {0.917$\pm$0.001} & { 0.891$\pm$0.002} & 0.917$\pm$0.001 & {0.917$\pm$0.001} & {0.872$\pm$0.012} & {\bf 0.946$\pm$0.004} & 0.916$\pm$0.002 & {\bf 0.946$\pm$0.004} & {\bf 0.946$\pm$0.004} & {0.919$\pm$0.014} \\

% Class-level KD  & 0.904$\pm$0.002 & 0.899$\pm$0.002 & 0.903$\pm$0.002 & 0.904$\pm$0.002 & 0.895$\pm$0.002 & 0.931$\pm$0.001 & 0.924$\pm$0.002 & 0.930$\pm$0.001 & 0.931$\pm$0.001 & 0.934$\pm$0.002 \\

% MT ListMLE & 0.892$\pm$0.006 & 0.886$\pm$0.006 & 0.892$\pm$0.006 & 0.892$\pm$0.006 & 0.895$\pm$0.004 & 0.919$\pm$0.002 & 0.912$\pm$0.004 & 0.919$\pm$0.002 & 0.920$\pm$0.002 & 0.924$\pm$0.004 \\

IL RankOOD & 0.902$\pm$0.004 & {\bf 0.899$\pm$0.004} & { 0.901$\pm$0.004} & {0.900$\pm$0.004} & 0.869$\pm$0.020 & 0.929$\pm$0.004 & 0.925$\pm$0.004 & 0.928$\pm$0.004 & 0.927$\pm$0.005 & 0.893$\pm$0.016 \\

% TW RankOOD$_{Instance}$ & 0.907$\pm$0.002 & 0.902$\pm$0.002 & 0.907$\pm$0.002 & 0.908$\pm$0.002 & 0.905$\pm$0.003 & 0.939$\pm$0.002 & 0.931$\pm$0.003 & 0.939$\pm$0.002 & 0.940$\pm$0.001 & 0.941$\pm$0.005  \\

% TW RankOOD$_{Canonical}$ & 0.905$\pm$0.004 & 0.900$\pm$0.003 & 0.904$\pm$0.004 & 0.905$\pm$0.004 & 0.908$\pm$0.001 & 0.933$\pm$0.007 & 0.924$\pm$0.008 & 0.932$\pm$0.008 & 0.934$\pm$0.008 & 0.939$\pm$0.005 \\

IL PL-KD & {\bf 0.919$\pm$0.001} & {0.876$\pm$0.001} & {\bf 0.918$\pm$0.001} & {\bf 0.918$\pm$0.001} & 0.879$\pm$0.012 & {\bf 0.946$\pm$0.004} & 0.904$\pm$0.005 &{\bf  0.946$\pm$0.004 }& {\bf 0.946$\pm$0.004} & 0.908$\pm$0.004 \\

% \hline
\bottomrule
\end{tabular}
}

\end{table}

%
% \subsection{CIFAR-100}
% \label{app:results_extended_cifar100}

\begin{table}[!h]
\centering
\caption{Average near-OOD and far-OOD FPR@95 results on CIFAR-100 (ID). IL denotes
instance-level.}
\label{tbl:appendix_fpr_cifar100}
\renewcommand{\arraystretch}{1.25}
\resizebox{\textwidth}{!}{
\begin{tabular}{@{}lcccccc|ccccc@{}}
\toprule
& & \multicolumn{5}{c|}{\textbf{Near-OOD}} & \multicolumn{5}{c}{\textbf{Far-OOD}} \\
\midrule
& \textbf{\makecell{ID \\ accuracy}} & \textbf{\makecell{LogitGap \\ FPR95}} & \textbf{\makecell{MSP \\ FPR95}} & \textbf{\makecell{MLS \\ FPR95}} & \textbf{\makecell{EBO \\ FPR95}} & \textbf{\makecell{RankOOD \\ FPR95}} & \textbf{\makecell{LogitGap \\ FPR95}} & \textbf{\makecell{MSP \\ FPR95}} & \textbf{\makecell{MLS \\ FPR95}} & \textbf{\makecell{EBO \\ FPR95}} & \textbf{\makecell{RankOOD \\ FPR95}} \\
\toprule
CE & 0.772$\pm$0.001 & 0.540$\pm$0.010 & 0.548$\pm$0.004 & 0.555$\pm$0.008 & 0.556$\pm$0.008 & 0.604$\pm$0.012 & 0.566$\pm$0.016 & {\bf 0.587$\pm$0.013} & 0.567$\pm$0.016 & 0.566$\pm$0.017 & 0.557$\pm$0.007 \\

RankOOD & 0.724$\pm$0.010 & 0.595$\pm$0.009 & 0.606$\pm$0.007 & 0.611$\pm$0.012 & 0.612$\pm$0.012 & { 0.621$\pm$0.013} & { 0.582$\pm$0.013} & 0.643$\pm$0.010 & {0.566$\pm$0.017} & {0.566$\pm$0.017} & 0.516$\pm$0.014\\

\hline
Self-KD & {0.775$\pm$0.004} & 0.545$\pm$0.010 & 0.557$\pm$0.07 & 0.562$\pm$0.009 & 0.562$\pm$0.008 & 0.583$\pm$0.007 & 0.538$\pm$0.027 & 0.579$\pm$0.026 & 0.514$\pm$0.031 & 0.512$\pm$0.032 & 0.490$\pm$0.008\\

% Class-level KD & 0.784$\pm$0.001 & 0.598$\pm$0.052 & 0.623$\pm$0.070 & 0.627$\pm$0.071 & 0.628$\pm$0.071 & 0.715$\pm$0.010 & 0.636$\pm$0.034 & 0.645$\pm$0.035 & 0.649$\pm$0.036 & 0.650$\pm$0.037 & 0.579$\pm$0.015 \\

% MT ListMLE & 0.748$\pm$0.002 & 0.547$\pm$0.002 & 0.561$\pm$0.003 & 0.569$\pm$0.008 & 0.570$\pm$0.008 & 0.534$\pm$0.015 & 0.528$\pm$0.023 & 0.584$\pm$0.032 & 0.519$\pm$0.022 & 0.518$\pm$0.023 & 0.500$\pm$0.021 \\

IL RankOOD & 0.774$\pm$0.003 & 0.544$\pm$0.005 & {\bf 0.544$\pm$0.005} & {\bf 0.546$\pm$0.004} & {\bf 0.547$\pm$0.004} & {\bf 0.578$\pm$0.010} & 0.539$\pm$0.016 & {\bf 0.552$\pm$0.015} & 0.534$\pm$0.020 & 0.531$\pm$0.022 & {\bf 0.465$\pm$0.019} \\

% TW RankOOD$_{Instance}$ & 0.773$\pm$0.004 &  0.541$\pm$0.015 & 0.546$\pm$0.014 & 0.545$\pm$0.019 & 0.547$\pm$0.020 & 0.581$\pm$0.033 & 0.564$\pm$0.011 & 0.572$\pm$0.013 & 0.569$\pm$0.018 & 0.568$\pm$0.018 & 0.532$\pm$0.012  \\

% TW RankOOD$_{Canonical}$ &  0.771$\pm$0.003 &   0.548$\pm$0.006 & 0.551$\pm$0.006 & 0.550$\pm$0.009 & 0.550$\pm$0.008 & 0.568$\pm$0.011 &  0.562$\pm$0.006 & 0.564$\pm$0.007 & 0.560$\pm$0.004 & 0.560$\pm$0.003 & 0.559$\pm$0.013 \\

IL PL-KD &  {\bf 0.778$\pm$0.003} &  {\bf 0.531$\pm$0.006} & { 0.552$\pm$0.001} & { 0.568$\pm$0.008} & { 0.570$\pm$0.008} & 0.603$\pm$0.018 & {\bf 0.530$\pm$0.020} & 0.580$\pm$0.020 & {\bf 0.497$\pm$0.017} & {\bf 0.495$\pm$0.017} & 0.507$\pm$0.021 \\

% \hline

\bottomrule
\end{tabular}
}

\end{table}

\begin{table}[!h]
\centering
\caption{Average AUROC on near-OOD and far-OOD datasets with CIFAR-100 as the ID dataset. IL denotes instance-level.}
\label{tbl:appendix_auroc_cifar_100}
\renewcommand{\arraystretch}{1.25}
\resizebox{\textwidth}{!}{
\begin{tabular}{@{}lccccc|ccccc@{}}
\toprule
& & \multicolumn{4}{c|}{\textbf{Near-OOD}} & \multicolumn{5}{c}{\textbf{Far-OOD}} \\
\midrule
& \textbf{\makecell{LogitGap \\ AUROC}} & \textbf{\makecell{MSP \\ AUROC}} & \textbf{\makecell{MLS \\ AUROC}} & \textbf{\makecell{EBO \\ AUROC}} & \textbf{\makecell{RankOOD \\ AUROC}} & \textbf{\makecell{LogitGap \\ AUROC}} & \textbf{\makecell{MSP \\ AUROC}} & \textbf{\makecell{MLS \\ AUROC}} & \textbf{\makecell{EBO \\ AUROC}} & \textbf{\makecell{RankOOD \\ AUROC}} \\
\toprule
CE & 0.812$\pm$0.001 & 0.803$\pm$0.001 & 0.811$\pm$0.001 & 0.809$\pm$0.001 & 0.784$\pm$0.006  & 0.792$\pm$0.006 & 0.778$\pm$0.005 & 0.797$\pm$0.007 & 0.798$\pm$0.007 & 0.789$\pm$0.003 \\

RankOOD  & 0.789$\pm$0.005 & 0.776$\pm$0.005 & 0.788$\pm$0.005 & 0.787$\pm$0.005 & {0.765$\pm$0.001} & { 0.780$\pm$0.004} & 0.750$\pm$0.004 &{0.794$\pm$0.005} & {0.796$\pm$0.005} & 0.820$\pm$0.013\\

\hline
Self-KD   & {0.812$\pm$0.003}& {0.797$\pm$0.002} & {0.811$\pm$0.003} & {0.811$\pm$0.003} & 0.791$\pm$0.006 & {\bf 0.804$\pm$0.010} & 0.781$\pm$0.011 & 0.818$\pm$0.010 & 0.820$\pm$0.010& 0.817$\pm$0.007\\

% Class-level KD  & 0.804$\pm$0.007 & 0.803$\pm$0.008 & 0.803$\pm$0.008 & 0.805$\pm$0.007 & 0.783$\pm$0.006 & 0.780$\pm$0.009 & 0.787$\pm$0.011 & 0.790$\pm$0.012 & 0.798$\pm$0.014 & 0.813$\pm$0.006 \\

% MT ListMLE & 0.808$\pm$0.001 & 0.795$\pm$0.002 & 0.805$\pm$0.002 & 0.805$\pm$0.002 & 0.802$\pm$0.010 & 0.808$\pm$0.006 & 0.782$\pm$0.010 & 0.826$\pm$0.006 & 0.828$\pm$0.006 & 0.820$\pm$0.007 \\

IL RankOOD & 0.813$\pm$0.001 & {\bf 0.811$\pm$0.001} & {\bf 0.815$\pm$0.001} & {\bf 0.815$\pm$0.001} &{\bf 0.796$\pm$0.004} & 0.801$\pm$0.011 & {\bf 0.795$\pm$0.011} & 0.802$\pm$0.013 & 0.802$\pm$0.015 & {\bf 0.837$\pm$0.010
}\\

% TW RankOOD$_{Instance}$  &  0.812$\pm$0.004 & 0.810$\pm$0.005 & 0.813$\pm$0.004 & 0.813$\pm$0.005 & 0.794$\pm$0.006 & 0.787$\pm$0.006 & 0.783$\pm$0.007 & 0.786$\pm$0.008 & 0.786$\pm$0.008 & 0.806$\pm$0.012  \\

% TW RankOOD$_{Canonical}$  &  0.805$\pm$0.004 & 0.805$\pm$0.004 & 0.807$\pm$0.005 & 0.807$\pm$0.005 & 0.792$\pm$0.003 &  0.786$\pm$0.005 & 0.785$\pm$0.005 & 0.787$\pm$0.005 & 0.788$\pm$0.006 & 0.789$\pm$0.010 \\

IL PL-KD &  {\bf 0.815$\pm$0.004} & {0.793$\pm$0.002} & { 0.811$\pm$0.001} & {0.809$\pm$0.002} & 0.790$\pm$0.004 &  0.799$\pm$0.010 & 0.769$\pm$0.013 & {\bf 0.821$\pm$0.006} & {\bf 0.823$\pm$0.006} & 0.803$\pm$0.010 \\

% \hline
\bottomrule
\end{tabular}
}

\end{table}

% 
% \subsection{ImageNet-200}
% \label{app:results_extended_cifar10}
\setlength{\tabcolsep}{4pt} 
\begin{table}[!h]
\centering
\caption{Average FPR@95 on near-OOD and far-OOD datasets with ImageNet-200 as the ID dataset. IL denotes instance-level.}
\label{tbl:appendix_fpr_imagenet200}
\renewcommand{\arraystretch}{1.25}

\resizebox{\textwidth}{!}{
\begin{tabular}{@{}lcccccc|ccccc@{}}
\toprule
& & \multicolumn{5}{c|}{\textbf{Near-OOD}} & \multicolumn{5}{c}{\textbf{Far-OOD}} \\
\midrule
& \textbf{\makecell{ID \\ accuracy}} & \textbf{\makecell{LogitGap \\ FPR95}} & \textbf{\makecell{MSP \\ FPR95}} & \textbf{\makecell{MLS \\ FPR95}} & \textbf{\makecell{EBO \\ FPR95}} & \textbf{\makecell{RankOOD \\ FPR95}} & \textbf{\makecell{LogitGap \\ FPR95}} & \textbf{\makecell{MSP \\ FPR95}} & \textbf{\makecell{MLS \\ FPR95}} & \textbf{\makecell{EBO \\ FPR95}} & \textbf{\makecell{RankOOD \\ FPR95}} \\
\toprule
CE & 0.864$\pm$0.001 & 0.552$\pm$0.003 & 0.548$\pm$0.004 & 0.597$\pm$0.007 & 0.602$\pm$0.007 & 0.568$\pm$0.008 &  0.326$\pm$0.009 & 0.354$\pm$0.005 & 0.340$\pm$0.015 & 0.349$\pm$0.016 & 0.363$\pm$0.006 \\

RankOOD & 0.847$\pm$0.004 & { 0.542$\pm$0.002} & 0.553$\pm$0.002 & { 0.535$\pm$0.004} & {0.536$\pm$0.005} & {\bf 0.525$\pm$0.013} & {0.313$\pm$0.013} & 0.373$\pm$0.014 & {0.289$\pm$0.013} & {0.289$\pm$0.013} & {\bf 0.297$\pm$0.009} \\
\hline

Self-KD & {0.866$\pm$0.002} & 0.542$\pm$0.003 & 0.545$\pm$0.001 & 0.566$\pm$0.004 & 0.568$\pm$0.004 & 0.597$\pm$0.008 &  {\bf 0.287$\pm$0.0086} & {0.344$\pm$0.007} & {\bf 0.270$\pm$0.005} & {\bf 0.273$\pm$0.004} & 0.342$\pm$0.017 \\

% Class-level KD & 0.864$\pm$0.004 & 0.551$\pm$0.007 & 0.553$\pm$0.008 & 0.601$\pm$0.002 & 0.605$\pm$0.002 & 0.584$\pm$0.010  
% & 0.334$\pm$0.019 & 0.359$\pm$0.018 & 0.335$\pm$0.018 & 0.341$\pm$0.016 & 0.363$\pm$0.016  \\

% Marginal-teacher ListMLE & 0.839$\pm$0.002 & 0.544$\pm$0.003 & 0.547$\pm$0.003 & 0.544$\pm$0.008 & 0.546$\pm$0.008 & 0.509$\pm$0.011 & 0.335$\pm$0.002 & 0.365$\pm$0.004 & 0.297$\pm$0.004 & 0.297$\pm$0.004 & 0.307$\pm$0.003\\

IL RankOOD & 0.873$\pm$0.000 & 0.529$\pm$0.003 & 0.531$\pm$0.004 & 0.552$\pm$0.004 & 0.555$\pm$0.004 & 0.533$\pm$0.012 & 0.303$\pm$0.002 & 0.312$\pm$0.003 & 0.298$\pm$0.002 & 0.305$\pm$0.002 & 0.320$\pm$0.002\\

% TW RankOOD$_{Instance}$ & 0.870$\pm$0.001 & 00.543$\pm$0.012 & 0.542$\pm$0.008 & 0.594$\pm$0.019 & 0.599$\pm$0.019 & 0.574$\pm$0.019 & 0.319$\pm$0.002 & 0.337$\pm$0.002 & 0.332$\pm$0.001 & 0.339$\pm$0.003 & 0.380$\pm$0.026 \\

% TW RankOOD$_{Canonical}$ & 0.866$\pm$0.003 & 0.548$\pm$0.006 & 0.549$\pm$0.010 & 0.589$\pm$0.005 & 0.594$\pm$0.004 & 0.589$\pm$0.012 & 0.339$\pm$0.014 & 0.356$\pm$0.017 & 0.363$\pm$0.008 & 0.372$\pm$0.007 & 0.414$\pm$0.026 \\

IL PL-KD & {\bf 0.875$\pm$0.003} & {\bf 0.516$\pm$0.003} & {\bf 0.522$\pm$0.004} & {\bf 0.527$\pm$0.002} & {\bf 0.529$\pm$0.002 }& 0.539$\pm$0.008 & 0.291$\pm$0.007 & {\bf 0.288$\pm$0.010} & 0.280$\pm$0.008 & 0.287$\pm$0.008 & 0.312$\pm$0.004 \\

% \hline
\bottomrule
\end{tabular}
}

\end{table}

\begin{table}[!h]
\centering
\caption{Average AUROC on near-OOD and far-OOD datasets with ImageNet-200 as the ID dataset. IL denotes instance-level.}
\label{tbl:appendix_auroc_imagenet200}
\renewcommand{\arraystretch}{1.25}
\resizebox{\textwidth}{!}{
\begin{tabular}{@{}lccccc|ccccc@{}}
\toprule
& & \multicolumn{4}{c|}{\textbf{Near-OOD}} & \multicolumn{5}{c}{\textbf{Far-OOD}} \\
\midrule
& \textbf{\makecell{LogitGap \\ AUROC}} & \textbf{\makecell{MSP \\ AUROC}} & \textbf{\makecell{MLS \\ AUROC}} & \textbf{\makecell{EBO \\ AUROC}} & \textbf{\makecell{RankOOD \\ AUROC}} & \textbf{\makecell{LogitGap \\ AUROC}} & \textbf{\makecell{MSP \\ AUROC}} & \textbf{\makecell{MLS \\ AUROC}} & \textbf{\makecell{EBO \\ AUROC}} & \textbf{\makecell{RankOOD \\ AUROC}} \\
\toprule
CE & 0.838$\pm$0.001 & 0.833$\pm$0.001 & 0.829$\pm$0.001 & 0.825$\pm$0.001 & 0.815$\pm$0.004 & 0.909$\pm$0.002 & 0.901$\pm$0.001 & 0.911$\pm$0.002 & 0.909$\pm$0.003	 & 0.882$\pm$0.006
 \\

RankOOD  & 0.835$\pm$0.003 & 0.829$\pm$0.002 & {0.838$\pm$0.003} & {\bf 0.839$\pm$0.003} & {\bf 0.842$\pm$0.006} & { 0.912$\pm$0.004} & 0.903$\pm$0.004 & {0.921$\pm$0.002}	& {0.924$\pm$0.002} &{\bf 0.910$\pm$0.005}\\

\hline
Self-KD   & {0.839$\pm$0.000} & {0.831$\pm$0.002} & 0.831$\pm$0.001 & 0.829$\pm$0.001 & 0.812$\pm$0.003 & {\bf 0.920$\pm$0.002} & { 0.905$\pm$0.002} & {\bf 0.927$\pm$0.003} & {\bf 0.927$\pm$0.004}	& 0.898$\pm$0.005 \\

IL RankOOD & 0.841$\pm$0.001 & 0.839$\pm$0.002 & 0.837$\pm$0.001 & 0.833$\pm$0.001 & 0.826$\pm$0.001 & 0.911$\pm$0.001 & 0.912$\pm$0.001 & 0.917$\pm$0.001 & 0.914$\pm$0.001 & 0.895$\pm$0.005\\

IL PL-KD & {\bf 0.844$\pm$0.001} & {\bf 0.843$\pm$0.001} & {\bf 0.844$\pm$0.001} & {\bf 0.839$\pm$0.002} & 0.822$\pm$0.002 & 0.915$\pm$0.001 & {\bf 0.921$\pm$0.002 }& 0.923$\pm$0.002 & 0.921$\pm$0.003 & 0.901$\pm$0.003 \\

% \hline
\bottomrule
\end{tabular}
}

\end{table}

%% file: sec/suppl/results_baselines.tex
For completeness, full results for all 36 baselines are reported in Table~\ref{tbl:nearood_appendix} (near-OOD) and Table~\ref{tbl:farood_appendix} (far-OOD).
Per-dataset results for each OOD benchmark are also provided. ImageNet-200 results are shown in Table~\ref{tab:fpr_tin} and Table~\ref{tab:auroc_tin}; CIFAR-100 results in Table~\ref{tab:fpr_cifar100} and Table~\ref{tab:auroc_cifar100}; and CIFAR-10 results in Table~\ref{tab:fpr_cifar10} and Table~\ref{tab:auroc_cifar10}. All per-dataset results use LogitGap as the OOD score function. \textit{Notable results of Instance-level PL-KD include making it to the top-5 for near-OOD detection in terms of FPR@95 and AUROC on all three datasets, while achieving the highest ID accuracy on ImageNet-200 and the second-highest on CIFAR-100.}

%%%%%%%%% Near OOD detection Suranga - commenting this as since we moved this to the main. %%%%%%%%%%%%%%%%

\begin{table*}[t!]
\scriptsize
\centering
\caption{ Performance comparison in \textit{near-OOD detection}. For each column, the top five methods are marked in {bold}. Note that N/A indicates missing results in OpenOOD.}
\label{tbl:nearood_appendix}
\resizebox{\textwidth}{!}{
\begin{tabular}{lccc|ccc|ccc|cc}
\toprule
Method & \multicolumn{3}{c|}{CIFAR-10} & \multicolumn{3}{c|}{CIFAR-100} & \multicolumn{3}{c|}{ImageNet-200} & \multicolumn{2}{c}{Average} \\
 & AUROC $\uparrow$ & FPR95 $\downarrow$ & ID Acc & AUROC $\uparrow$ & FPR95 $\downarrow$ & ID Acc & AUROC $\uparrow$ & FPR95 $\downarrow$ & ID Acc & AUROC $\uparrow$ & FPR95 $\downarrow$ \\
\hline
\multicolumn{12}{c}{Post-hoc inference methods} \\ 
\hline
OpenMax~\cite{openmaxbendale2016towards} & 87.62 ± 0.29 & 43.62 ± 2.27 & 95.06 ± 0.30 & 76.41 ± 0.25 & 56.58 ± 0.73 & 77.25 ± 0.10 & 80.27 ± 0.10 & 63.48 ± 0.25 & 86.37 ± 0.08 & 81.43 & 54.56  \\
MSP~\cite{Hendrycks:2017} & 88.03 ± 0.25 & 48.17 ± 3.92 & 95.06 ± 0.30 & 80.27 ± 0.11 & {54.80 ± 0.33} & 77.25 ± 0.10 & 83.34 ± 0.06 & 54.82 ± 0.35 & 86.37 ± 0.08 & 83.88 & 52.60  \\
TempScale~\cite{tempscale} & 88.09 ± 0.31 & 50.96 ± 4.32 & 95.06 ± 0.30 & 80.90 ± 0.07 & {54.49 ± 0.48} & 77.25 ± 0.10 & {\bf 83.69 ± 0.04} & 54.82 ± 0.23 & 86.37 ± 0.08 & 84.23 & 53.42  \\
ODIN~\cite{hsu:2020} & 82.87 ± 1.85 & 76.19 ± 6.08 & 95.06 ± 0.30 & 79.90 ± 0.11 & 57.91 ± 0.51 & 77.25 ± 0.10 & 80.27 ± 0.08 & 66.76 ± 0.26 & 86.37 ± 0.08 & 81.01 & 66.95  \\
MDS~\cite{Lee:2018b} & 84.20 ± 2.40 & 49.90 ± 3.98 & 95.06 ± 0.30 & 58.69 ± 0.09 & 83.53 ± 0.60 & 77.25 ± 0.10 & 61.93 ± 0.51 & 79.11 ± 0.31 & 86.37 ± 0.08 & 68.27 & 70.85  \\
MDSEns~\cite{Lee:2018b} & 60.43 ± 0.26 & 92.26 ± 0.20 & 95.06 ± 0.30 & 46.31 ± 0.24 & 95.88 ± 0.04 & 77.25 ± 0.10 & 54.32 ± 0.24 & 91.75 ± 0.10 & 86.37 ± 0.08 & 53.69 & 93.30  \\
RMDS~\cite{rmds} & 89.80 ± 0.28 & 38.89 ± 2.39 & 95.06 ± 0.30 & 80.15 ± 0.11 & 55.46 ± 0.41 & 77.25 ± 0.10 & 82.57 ± 0.25 & {\bf 54.02 ± 0.58} & 86.37 ± 0.08 & 84.17 & 49.46  \\
Gram~\cite{gram} & 58.66 ± 4.83 & 90.87 ± 1.91 & 95.06 ± 0.30 & 51.66 ± 0.77 & 92.28 ± 0.29 & 77.25 ± 0.10 & 67.67 ± 1.07 & 86.40 ± 1.21 & 86.37 ± 0.08 & 59.33 & 89.85 \\
EBO~\cite{Liu:2020b} & 87.58 ± 0.46 & 61.34 ± 4.63 & 95.06 ± 0.30 & {80.91 ± 0.08} & 55.62 ± 0.61 & 77.25 ± 0.10 & 82.50 ± 0.05 & 60.24 ± 0.57 & 86.37 ± 0.08 & 83.66 & 59.07  \\
OpenGAN~\cite{opengan} & 53.71 ± 7.68 & 94.48 ± 4.01 & 95.06 ± 0.30 & 65.98 ± 1.26 & 76.52 ± 2.59 & 77.25 ± 0.10 & 59.79 ± 3.39 & 84.15 ± 3.85 & 86.37 ± 0.08 & 59.83 & 85.05  \\
GradNorm~\cite{Huang:2021} & 54.90 ± 0.98 & 94.72 ± 0.82 & 95.06 ± 0.30 & 70.13 ± 0.47 & 85.58 ± 0.46 & 77.25 ± 0.10 & 72.75 ± 0.48 & 82.67 ± 0.30 & 86.37 ± 0.08 & 65.93 & 87.66  \\
ReAct~\cite{Sun:2021} & 87.11 ± 0.61 & 63.56 ± 7.33 & 95.06 ± 0.30 & 80.77 ± 0.05 & 56.39 ± 0.34 & 77.25 ± 0.10 & 81.87 ± 0.98 & 62.49 ± 2.19 & 86.37 ± 0.08 & 83.25 & 60.81  \\
MLS~\cite{Hendrycks:2022} & 87.52 ± 0.47 & 61.32 ± 4.62 & 95.06 ± 0.30 & \textbf{81.05 ± 0.07} & 55.47 ± 0.66 & 77.25 ± 0.10 & 82.90 ± 0.04 & 59.76 ± 0.59 & 86.37 ± 0.08 & 83.82 & 58.85  \\
KLM~\cite{Hendrycks:2022} & 79.19 ± 0.80 & 87.86 ± 6.37 & 95.06 ± 0.30 & 76.56 ± 0.25 & 77.92 ± 1.31 & 77.25 ± 0.10 & 80.76 ± 0.08 & 70.26 ± 0.64 & 86.37 ± 0.08 & 78.84 & 78.68 \\
VIM~\cite{wang2022vim} & 88.68 ± 0.28 & 44.84 ± 2.31 & 95.06 ± 0.30 & 74.98 ± 0.13 & 62.63 ± 0.27 & 77.25 ± 0.10 & 78.68 ± 0.24 & 59.19 ± 0.71 & 86.37 ± 0.08 & 80.78 & 55.55  \\
KNN~\cite{Sun:2022} & 90.64 ± 0.20 & 34.01 ± 0.38 & 95.06 ± 0.30 & 80.18 ± 0.15 & 61.22 ± 0.14 & 77.25 ± 0.10 & 81.57 ± 0.17 & 60.18 ± 0.52 & 86.37 ± 0.08 & 84.13 & 51.80  \\
DICE~\cite{dice} & 78.34 ± 0.79 & 70.04 ± 7.64 & 95.06 ± 0.30 & 79.38 ± 0.23 & 57.95 ± 0.53 & 77.25 ± 0.10 & 81.78 ± 0.14 & 61.88 ± 0.67 & 86.37 ± 0.08 & 79.83 & 63.29  \\
RankFeat~\cite{rankfeat} & 79.46 ± 2.52 & 60.88 ± 4.60 & 95.06 ± 0.30 & 61.88 ± 1.28 & 80.59 ± 1.10 & 77.25 ± 0.10 & 56.92 ± 1.59 & 92.06 ± 0.23 & 86.37 ± 0.08 & 66.09 & 77.84  \\
ASH~\cite{Djurisic:2023} & 75.27 ± 1.04 & 86.78 ± 1.82 & 95.06 ± 0.30 & 78.20 ± 0.15 & 65.71 ± 0.24 & 77.25 ± 0.10 & 82.38 ± 0.19 & 64.89 ± 0.90 & 86.37 ± 0.08 & 78.62 & 72.46  \\
SHE~\cite{she} & 81.54 ± 0.51 & 79.65 ± 3.47 & 95.06 ± 0.30 & 78.95 ± 0.18 & 59.07 ± 0.25 & 77.25 ± 0.10 & 80.18 ± 0.25 & 66.80 ± 0.74 & 86.37 ± 0.08 & 80.22 & 68.51  \\
GEN~\cite{XiXi:2023} & 88.20 ± 0.30 & 53.67 ± 3.14 & 95.06 ± 0.30 & \textbf{81.31 ± 0.08} & \textbf{54.42 ± 0.33} & 77.25 ± 0.10 & \textbf{83.68 ± 0.06} & 55.20 ± 0.20 & 86.37 ± 0.08 & 84.40 & 54.43  \\
ExCeL~\cite{Karunanayake:2025} & 86.89 ± 0.23 & 66.55 ± 0.43 & 95.06 ± 0.30 & 80.70 ± 0.06 & 55.21 ± 0.56 & 77.25 ± 0.10 & 82.40 ± 0.04 & 57.90 ± 0.40 & 86.37 ± 0.08 & 83.33 & 59.89  \\
\hline
\multicolumn{12}{c}{Training methods without outliers} \\
\hline
ConfBranch~\cite{DeVries:2018} & 89.84 ± 0.24 & {31.28 ± 0.66} & 94.88 ± 0.05 & 71.60 ± 0.62 & 70.21 ± 0.83 & 76.59 ± 0.27 & 79.10 ± 0.24 & 61.44 ± 0.34 & 85.92 ± 0.07 & 80.18 & 54.31  \\
G-ODIN~\cite{hsu:2020} & 89.12 ± 0.57 & 45.54 ± 2.52 & 94.70 ± 0.25 & 77.15 ± 0.28 & 67.58 ± 0.98 & 74.46 ± 0.04 & 77.28 ± 0.10 & 69.87 ± 0.46 & 84.56 ± 0.28 & 81.18 & 61.00 \\
CSI~\cite{Tack:2020} & 89.51 ± 0.19 & 33.66 ± 0.64 & 91.16 ± 0.14 & 71.45 ± 0.27 & 70.26 ± 0.56 & 61.60 ± 0.46 & N/A & N/A & N/A & 80.48 & 51.96  \\
ARPL~\cite{arpl} & 87.44 ± 0.15 & 40.33 ± 0.70 & 93.66 ± 0.11 & 74.94 ± 0.93 & 61.56 ± 1.81 & 70.70 ± 1.08 & 82.02 ± 0.10 & 55.74 ± 0.70 & 83.95 ± 0.32 & 81.47 & 52.54  \\
MOS~\cite{Huang:2021b} & 71.45 ± 3.09 & 78.72 ± 5.86 & 94.83 ± 0.37 & 80.40 ± 0.18 & 56.05 ± 1.01 & 76.98 ± 0.20 & 69.84 ± 0.46 & 71.60 ± 0.48 & 85.60 ± 0.20 & 73.90 & 68.79  \\
LogitNorm~\cite{Wei:2022} & \textbf{92.33 ± 0.08} & \textbf{29.34 ± 0.81} & 94.30 ± 0.25 & 78.47 ± 0.31 & 62.89 ± 0.57 & 76.34 ± 0.17 & 82.66 ± 0.15 & 56.46 ± 0.37 & 86.04 ± 0.15 &  {84.49} & {49.56}  \\
CIDER~\cite{cider} & {90.71 ± 0.16} & 32.11 ± 0.94 & N/A & 73.10 ± 0.39 & 72.02 ± 0.31 & N/A & 80.58 ± 1.75 & 60.10 ± 0.73 & N/A & 81.46 & 54.74  \\
CRAFT\cite{Karunanayake:2025b} & {91.11 ± 0.04} & 31.94 ± 1.41 & \textbf{95.80 ± 0.13} & 80.90 ± 0.33 & \textbf{53.73 ± 0.62} & \textbf{78.43 ± 0.09} & { 83.65 ± 0.41} & \textbf{ 54.62 ± 0.57} & {\bf 86.71 ± 0.08} & {\bf 85.22} & {\bf 46.76} \\
{\sc ELogitNorm}~\cite{Ding:2026} & \textbf{92.89 ± -} & \textbf{26.49 ± -} & 94.47 ± - & 79.68 ± - & {59.48 ± -} & 76.67 ± - & 83.06 ± - & {54.99 ± -} & 86.39 ± - & 85.21 & 46.99 \\
{\sc PSKD}~\cite{Yang:2025} & \textbf{91.71 ± 0.16} & {31.67 ± 0.78} & 95.14 ± 0.08 & \textbf{81.45 ± 0.44} & {54.83 ± 2.79} & 77.44 ± 0.09 & 82.84 ± 0.32 & {57.12 ± 0.63} & 86.79 ± 0.25 & \textbf{85.33} & {47.87}  \\ 
{\sc RankOOD}~\cite{Denipitiyage:2026} & 90.21 ± 0.41 & {31.72 ± 0.67} & 93.30 ± 0.003 & 80.67 ± 0.40 & \textbf{52.59 ± 0.75} & 75.30 ± 0.001 & \textbf{85.30 ± 0.18} & \textbf{50.05 ± 0.16} & 85.72 ± 0.002 & \textbf{85.39} & \textbf{44.79}  \\ 
{IL PL-KD (Ours)} & {\bf 91.89$\pm$0.10} & {\bf 30.10$\pm$0.50} & {94.68$\pm$0.50} & \textbf{81.55$\pm$0.13} & \textbf{53.09$\pm$0.59} & \textbf{77.60$\pm$0.30} & {\bf 84.36$\pm$0.07} & \textbf{51.65$\pm$0.30} & \textbf{87.53$\pm$0.26} & \textbf{85.93} & \textbf{44.95}  \\

\hline
\multicolumn{12}{c}{Training methods with outliers} \\
\hline
OE~\cite{Hendrycks:2019} & \textbf{94.82 ± 0.21} & \textbf{19.84 ± 0.95} & 94.63 ± 0.26 & \textbf{88.30 ± 0.10} & \textbf{30.73 ± 0.11} & 76.84 ± 0.42 & \textbf{84.84 ± 0.16} & \textbf{52.30 ± 0.67} & 85.82 ± 0.21 & \textbf{89.32} & \textbf{34.29} \\
MCD~\cite{mcd} & {91.03 ± 0.12} & \textbf{30.17 ± 0.06} & 94.95 ± 0.04 & 77.07 ± 0.32 & 55.88 ± 0.85 & 75.83 ± 0.04 & 83.62 ± 0.09 & {54.71 ± 0.83} & 86.12 ± 0.17 & 83.91 & \textbf{46.92}  \\
UDG~\cite{udg} & 89.91 ± 0.25 & 35.34 ± 0.95 & 92.36 ± 0.84 & 78.02 ± 0.10 & 61.42 ± 0.48 & 71.54 ± 0.64 & 74.30 ± 1.63 & 68.89 ± 1.72 & 68.11 ± 1.24 & 80.74 & 55.22  \\
MixOE~\cite{Zhang:2023b} & 88.73 ± 0.82 & 51.45 ± 7.78 & 94.55 ± 0.32 & {80.95 ± 0.20} & 55.22 ± 0.49 & 75.13 ± 0.06 & 82.62 ± 0.03 & 57.97 ± 0.40 & 85.71 ± 0.07 & 84.10 & 54.88  \\
\bottomrule
\end{tabular}

}

\end{table*}

%%%%%%%%% Far OOD detection %%%%%%%%%%%%%%%%

\begin{table*}[t!]
\scriptsize
\centering
\caption{ Performance comparison in \textit{far-OOD detection}. For each column, the top five methods are marked in {bold}. Note that N/A indicates missing results in OpenOOD.}
\label{tbl:farood_appendix}
\resizebox{\textwidth}{!}{
\begin{tabular}{lccc|ccc|ccc|ccc}
\toprule
Method & \multicolumn{3}{c|}{CIFAR-10} & \multicolumn{3}{c|}{CIFAR-100} & \multicolumn{3}{c|}{ImageNet-200} & \multicolumn{2}{c}{Average} \\
 & AUROC $\uparrow$ & FPR95 $\downarrow$ & ID Acc & AUROC $\uparrow$ & FPR95 $\downarrow$ & ID Acc & AUROC $\uparrow$ & FPR95 $\downarrow$ & ID Acc & AUROC $\uparrow$ & FPR95 $\downarrow$ \\
\hline
\multicolumn{12}{c}{Post-hoc inference methods} \\
\hline
OpenMax~\cite{openmaxbendale2016towards} & 89.62 ± 0.19 & 29.69 ± 1.21 & 95.06 ± 0.30 & 79.48 ± 0.41 & 54.50 ± 0.68 & 77.25 ± 0.10 & 90.20 ± 0.17 & 33.12 ± 0.66 & 86.37 ± 0.08 & 86.43 & 39.10  \\
MSP~\cite{Hendrycks:2017} & 90.73 ± 0.43 & 31.72 ± 1.84 & 95.06 ± 0.30 & 77.76 ± 0.44 & 58.70 ± 1.06 & 77.25 ± 0.10 & 90.13 ± 0.09 & 35.43 ± 0.38 & 86.37 ± 0.08 & 86.21 & 41.95  \\
TempScale~\cite{tempscale} & 90.97 ± 0.52 & 33.48 ± 2.39 & 95.06 ± 0.30 & 78.74 ± 0.51 & 57.94 ± 1.14 & 77.25 ± 0.10 & 90.82 ± 0.09 & 34.00 ± 0.37 & 86.37 ± 0.08 & 86.84 & 41.81  \\
ODIN~\cite{hsu:2020} & 87.96 ± 0.61 & 57.62 ± 4.24 & 95.06 ± 0.30 & 79.28 ± 0.21 & 58.86 ± 0.79 & 77.25 ± 0.10 & 91.71 ± 0.19 & 34.23 ± 1.05 & 86.37 ± 0.08 & 86.32 & 50.24  \\
MDS~\cite{Lee:2018b} & 89.72 ± 1.36 & 32.22 ± 3.40 & 95.06 ± 0.30 & 69.39 ± 1.39 & 72.26 ± 1.56 & 77.25 ± 0.10 & 74.72 ± 0.26 & 61.66 ± 0.27 & 86.37 ± 0.08 & 77.94 & 55.38  \\
MDSEns~\cite{Lee:2018b} & 73.90 ± 0.27 & 61.47 ± 0.48 & 95.06 ± 0.30 & 66.00 ± 0.69 & 66.74 ± 1.04 & 77.25 ± 0.10 & 69.27 ± 0.57 & 80.96 ± 0.38 & 86.37 ± 0.08 & 69.72 & 69.72  \\
RMDS~\cite{rmds} & 92.20 ± 0.21 & 25.35 ± 0.73 & 95.06 ± 0.30 & \textbf{82.92 ± 0.42} & 52.81 ± 0.63 & 77.25 ± 0.10 & 88.06 ± 0.34 & 32.45 ± 0.79 & 86.37 ± 0.08 & 87.73 & 36.87  \\
Gram~\cite{rmds} & 71.73 ± 3.20 & 72.34 ± 6.73 & 95.06 ± 0.30 & 73.36 ± 1.08 & 64.44 ± 2.37 & 77.25 ± 0.10 & 71.19 ± 0.24 & 84.36 ± 0.78 & 86.37 ± 0.08 & 72.09 & 73.71  \\
EBO~\cite{Liu:2020b} & 91.21 ± 0.92 & 41.69 ± 5.32 & 95.06 ± 0.30 & 79.77 ± 0.61 & 56.59 ± 1.38 & 77.25 ± 0.10 & 90.86 ± 0.21 & 34.86 ± 1.30 & 86.37 ± 0.08 & 87.28 & 44.38  \\
OpenGAN~\cite{opengan} & 54.61 ± 15.5 & 83.52 ± 11.63 & 95.06 ± 0.30 & 67.88 ± 7.16 & 70.49 ± 7.38 & 77.25 ± 0.10 & 73.15 ± 4.07 & 64.16 ± 9.33 & 86.37 ± 0.08 & 65.21 & 72.72  \\
GradNorm~\cite{Huang:2021} & 57.55 ± 3.22 & 91.90 ± 2.23 & 95.06 ± 0.30 & 69.14 ± 1.05 & 83.68 ± 1.92 & 77.25 ± 0.10 & 84.26 ± 0.87 & 66.45 ± 0.22 & 86.37 ± 0.08 & 70.32 & 80.68  \\
ReAct~\cite{Sun:2021} & 90.42 ± 1.41 & 44.90 ± 8.37 & 95.06 ± 0.30 & 80.39 ± 0.49 & 54.20 ± 1.56 & 77.25 ± 0.10 & {92.31 ± 0.56} & 28.50 ± 0.95 & 86.37 ± 0.08 & 87.71 & 42.53  \\
MLS~\cite{Hendrycks:2022} & 91.10 ± 0.89 & 41.68 ± 5.27 & 95.06 ± 0.30 & 79.67 ± 0.57 & 56.73 ± 1.33 & 77.25 ± 0.10 & 91.11 ± 0.19 & 34.03 ± 1.21 & 86.37 ± 0.08 & 87.29 & 44.15  \\
KLM~\cite{Hendrycks:2022} & 82.68 ± 0.21 & 78.31 ± 4.84 & 95.06 ± 0.30 & 76.24 ± 0.52 & 71.65 ± 2.01 & 77.25 ± 0.10 & 88.53 ± 0.11 & 40.90 ± 1.08 & 86.37 ± 0.08 & 82.48 & 63.62  \\
VIM~\cite{wang2022vim} & 93.48 ± 0.24 & 25.05 ± 0.52 & 95.06 ± 0.30 & 81.70 ± 0.62 & \textbf{50.74 ± 1.00} & 77.25 ± 0.10 & 91.26 ± 0.19 & \textbf{27.20 ± 0.30} & 86.37 ± 0.08 & {88.81} & {34.33}  \\
KNN~\cite{Sun:2022} & 92.96 ± 0.14 & 24.27 ± 0.40 & 95.06 ± 0.30 & \textbf{82.40 ± 0.17} & 53.65 ± 0.28 & 77.25 ± 0.10 & \textbf{93.16 ± 0.22} & \textbf{27.27 ± 0.75} & 86.37 ± 0.08 & \textbf{89.51} & 35.06  \\
DICE~\cite{dice} & 84.23 ± 1.89 & 51.76 ± 4.42 & 95.06 ± 0.30 & 80.01 ± 0.18 & 56.25 ± 0.60 & 77.25 ± 0.10 & 90.80 ± 0.31 & 36.51 ± 1.18 & 86.37 ± 0.08 & 85.01 & 48.17  \\
RankFeat~\cite{rankfeat} & 75.87 ± 5.06 & 57.44 ± 7.99 & 95.06 ± 0.30 & 67.10 ± 1.42 & 69.45 ± 1.01 & 77.25 ± 0.10 & 38.22 ± 3.85 & 97.72 ± 0.75 & 86.37 ± 0.08 & 60.40 & 74.87  \\
ASH~\cite{Djurisic:2023} & 78.49 ± 2.58 & 79.03 ± 4.22 & 95.06 ± 0.30 & 80.58 ± 0.66 & 59.20 ± 2.46 & 77.25 ± 0.10 & \textbf{93.90 ± 0.27} & \textbf{27.29 ± 1.12} & 86.37 ± 0.08 & 84.32 & 55.17  \\
SHE~\cite{she} & 85.32 ± 1.43 & 66.48 ± 5.98 & 95.06 ± 0.30 & 76.92 ± 1.16 & 64.12 ± 2.70 & 77.25 ± 0.10 & 89.81 ± 0.61 & 42.17 ± 1.24 & 86.37 ± 0.08 & 84.02 & 57.59  \\
GEN~\cite{XiXi:2023} & 91.35 ± 0.69 & 34.73 ± 1.58 & 95.06 ± 0.30 & 79.68 ± 0.75 & 56.71 ± 1.59 & 77.25 ± 0.10 & 91.36 ± 0.10 & 32.10 ± 0.59 & 86.37 ± 0.08 & 87.46 & 41.18  \\
ExCeL~\cite{Karunanayake:2025} & 91.69 ± 0.18 & 40.03 ± 0.84 & 95.06 ± 0.30 & {82.04 ± 0.90} & {52.24 ± 1.90} & 77.25 ± 0.10 & 91.97 ± 0.27 & 28.45 ± 0.80 & 86.37 ± 0.08 & 88.57 & 40.24  \\
\hline
\multicolumn{12}{c}{Training methods without outliers} \\
\hline
ConfBranch~\cite{DeVries:2018} & 92.85 ± 0.29 & 21.48 ± 0.94 & 94.88 ± 0.05 & 68.90 ± 1.83 & 71.82 ± 3.39 & 76.59 ± 0.27 & 90.43 ± 0.18 & 34.75 ± 0.63 & 85.92 ± 0.07 & 84.06 & 42.68 \\
G-ODIN~\cite{hsu:2020} & \textbf{95.51 ± 0.31} & 21.45 ± 1.91 & 94.70 ± 0.25 & \textbf{85.67 ± 1.58} & \textbf{42.68 ± 3.19} & 74.46 ± 0.04 & \textbf{92.33 ± 0.11} & 30.18 ± 0.49 & 84.56 ± 0.28 & \textbf{91.17} & \textbf{31.44}  \\
CSI~\cite{Tack:2020} & 92.00 ± 0.30 & 26.42 ± 0.29 & 91.16 ± 0.14 & 66.31 ± 1.21 & 76.92 ± 1.29 & 61.60 ± 0.46 & N/A & N/A & N/A & 79.16 & 51.67  \\
ARPL~\cite{arpl} & 89.31 ± 0.32 & 32.39 ± 0.74 & 93.66 ± 0.11 & 73.69 ± 1.80 & 63.14 ± 2.53 & 70.70 ± 1.08 & 89.23 ± 0.11 & 36.46 ± 0.08 & 83.95 ± 0.32 & 84.08 & 44.00  \\
MOS~\cite{Huang:2021b} & 76.41 ± 5.93 & 62.90 ± 6.62 & 94.83 ± 0.37 & 80.17 ± 1.21 & 57.28 ± 3.29 & 76.98 ± 0.20 & 80.46 ± 0.92 & 51.56 ± 0.42 & 85.60 ± 0.20 & 79.01 & 57.25  \\
LogitNorm~\cite{Wei:2022} & \textbf{96.74 ± 0.06} & \textbf{13.81 ± 0.20} & 94.30 ± 0.25 & 81.53 ± 1.26 & 53.61 ± 3.45 & 76.34 ± 0.17 & \textbf{93.04 ± 0.21} & \textbf{26.11 ± 0.52} & 86.04 ± 0.15 & \textbf{90.44} & \textbf{31.18}  \\
CIDER~\cite{cider} & \textbf{94.71 ± 0.36} & {20.72 ± 0.85} & N/A & 80.49 ± 0.68 & 54.22 ± 1.24 & N/A & 90.66 ± 1.68 & 30.17 ± 2.75 & N/A & 88.62 & 35.04  \\
CRAFT~\cite{Karunanayake:2025b} & 93.94 ± 0.20 & \textbf{19.40 ± 0.88} & \textbf{95.80 ± 0.13} & 82.03 ± 0.34 & {51.86 ± 0.49} & \textbf{78.43 ± 0.09} & 90.88 ± 0.89 & 32.67 ± 1.13 & \textbf{86.71 ± 0.08} & {88.95} & 34.64  \\
{\sc ELogitNorm}~\cite{Ding:2026} & \textbf{96.68 ± -} & \textbf{13.73 ± -} & 94.47 ± - & \textbf{84.51 ± -} & \textbf{46.86 ± -} & 76.67 ± - & \textbf{93.58 ± -} & \textbf{25.08 ± -} & 86.39 ± - & \textbf{91.59} & \textbf{28.56} \\
{\sc PSKD}~\cite{Yang:2025} & 94.56 ± 0.29 & {20.48 ± 1.30} & 95.14 ± 0.08 & {82.40 ± 0.52} & \textbf{51.56 ± 2.39} & 77.44 ± 0.09 & 91.39 ± 0.16 & {31.64 ± 0.87} & 86.79 ± 0.25 & {89.45} & {34.56}  \\ 
{\sc RankOOD}~\cite{Denipitiyage:2026} & 93.19 ± 0.84 & 20.96 ± 2.55 & 93.30 ± 0.003 & \textbf{83.63 ± 1.06} & \textbf{47.44 ± 0.80} & 75.30 ± 0.001 & 92.14 ± 0.20 & {27.73 ± 0.33} & 85.72 ± 0.002 & \textbf{89.65} & \textbf{32.04}  \\ 
{Instance-level PL-KD} & {94.62$\pm$0.41} & {\bf 19.10$\pm$0.96} & {94.68$\pm$0.46} & {79.95$\pm$1.04} & {52.99$\pm$1.95} & \textbf{77.60$\pm$0.30} & {91.49$\pm$0.08} & {29.14$\pm$0.70} & \textbf{87.53$\pm$0.26} & {88.86} & {\bf 33.74}  \\
	
\hline
\multicolumn{12}{c}{Training methods with outliers}  \\
\hline
OE~\cite{Hendrycks:2019} & \textbf{96.00 ± 0.13} & \textbf{13.13 ± 0.53} & 94.63 ± 0.26 & 81.41 ± 1.49 & 54.82 ± 2.79 & 76.84 ± 0.42 & 89.02 ± 0.18 & 34.17 ± 0.56 & 85.82 ± 0.21 & 88.81 & {34.04}  \\
MCD~\cite{mcd} & 91.00 ± 1.10 & 32.03 ± 4.21 & 94.95 ± 0.04 & 74.72 ± 0.78 & 54.39 ± 1.34 & 75.83 ± 0.04 & 88.94 ± 0.10 & 29.93 ± 0.30 & 86.12 ± 0.17 & 84.89 & 38.78  \\
UDG~\cite{udg} & {94.06 ± 0.90} & {20.35 ± 2.41} & 92.36 ± 0.84 & 79.59 ± 1.77 & 59.00 ± 3.35 & 71.54 ± 0.64 & 82.09 ± 2.78 & 62.04 ± 5.99 & 68.11 ± 1.24 & 85.25 & 47.13  \\
MixOE~\cite{Zhang:2023b} & 91.93 ± 0.69 & 33.84 ± 4.77 & 94.55 ± 0.32 & 76.40 ± 1.44 & 63.88 ± 2.48 & 75.13 ± 0.06 & 88.27 ± 0.41 & 40.93 ± 0.29 & 85.71 ± 0.07 & 85.53 & 46.22  \\
\bottomrule
\end{tabular}
}
% \captionsetup{width=0.85\linewidth}

\end{table*}

% \subsection{Detailed per-dataset OOD detection results}
% \label{sec:appendix_perdata_results}

%%%%%%%%%%%%%%%%% ImageNet-200 %%%%%%%%%%%%%%

\begin{table*}[h]
\centering
\scriptsize

\caption{FPR95 (\% $\downarrow$) of various methods for different OOD datasets when ImageNet-200 is ID. \break For each column, the top five methods are marked in \textbf{bold}. Note that N/A indicates that results are not reported in OpenOOD.}
\label{tab:fpr_tin}
\renewcommand{\arraystretch}{1.09}
\resizebox{\textwidth}{!}{
\begin{tabular}{lccc|cccc}
\toprule
Method & \multicolumn{3}{c|}{Near OOD} & \multicolumn{4}{c}{Far OOD} \\
 & SSB-hard & NINCO & Average & iNaturalist & Textures & OpenImage-O  & Average\\ 
 \hline
 \multicolumn{8}{c}{Post-hoc inference methods} \\ 
\hline
OpenMax~\cite{openmaxbendale2016towards} & 72.37 ± 0.11 & 54.59 ± 0.54 & 63.48 ± 0.25 & 24.53 ± 0.96 & 36.80 ± 0.55 & 38.03 ± 0.49 & 33.12 ± 0.66 \\
MSP~\cite{Hendrycks:2017} & 66.00 ± 0.10 & 43.65 ± 0.75 & 54.82 ± 0.35 & 26.48 ± 0.73 & 44.58 ± 0.68 & 35.23 ± 0.18 & 35.43 ± 0.38 \\
TempScale~\cite{tempscale} & 66.43 ± 0.26 & {43.21 ± 0.70} & 54.82 ± 0.23 & 24.39 ± 0.79 & 43.57 ± 0.77 & 34.04 ± 0.31 & 34.00 ± 0.37 \\
ODIN~\cite{hsu:2020} & 73.51 ± 0.38 & 60.00 ± 0.80 & 66.76 ± 0.26 & 22.39 ± 1.87 & 42.99 ± 1.56 & 37.30 ± 0.59 & 34.23 ± 1.05 \\
MDS~\cite{Lee:2018b} & 83.65 ± 0.47 & 74.57 ± 0.15 & 79.11 ± 0.31 & 58.53 ± 0.75 & 58.16 ± 0.84 & 68.29 ± 0.28 & 61.66 ± 0.27 \\
MDSEns~\cite{Lee:2018b} & 92.13 ± 0.05 & 91.36 ± 0.16 & 91.75 ± 0.10 & 83.37 ± 0.70 & 72.27 ± 0.48 & 87.26 ± 0.10 & 80.96 ± 0.38 \\
RMDS~\cite{rmds} & {65.91 ± 0.27} & \textbf{42.13 ± 1.04} & \textbf{54.02 ± 0.58} & 24.70 ± 0.90 & 37.80 ± 1.32 & 34.85 ± 0.31 & 32.45 ± 0.79 \\
Gram~\cite{gram} & 85.68 ± 0.85 & 87.13 ± 1.89 & 86.40 ± 1.21 & 85.54 ± 0.40 & 80.87 ± 1.20 & 86.66 ± 1.27 & 84.36 ± 0.78 \\
EBO~\cite{Liu:2020b} & 69.77 ± 0.32 & 50.70 ± 0.89 & 60.24 ± 0.57 & 26.41 ± 2.29 & 41.43 ± 1.85 & 36.74 ± 1.14 & 34.86 ± 1.30 \\
OpenGAN~\cite{opengan} & 88.07 ± 2.23 & 80.23 ± 5.71 & 84.15 ± 3.85 & 60.13 ± 9.79 & 66.00 ± 9.97 & 66.34 ± 8.44 & 64.16 ± 9.33 \\
GradNorm~\cite{Huang:2021} & 82.17 ± 0.62 & 83.17 ± 0.21 & 82.67 ± 0.30 & 61.31 ± 2.86 & 66.88 ± 3.59 & 71.16 ± 0.23 & 66.45 ± 0.22 \\
ReAct~\cite{Sun:2021} & 71.51 ± 1.92 & 53.47 ± 2.46 & 62.49 ± 2.19 & 22.97 ± 2.25 & \textbf{29.67 ± 1.35} & 32.86 ± 0.74 & 28.50 ± 0.95 \\
MLS~\cite{Hendrycks:2022} & 69.64 ± 0.37 & 49.87 ± 0.94 & 59.76 ± 0.59 & 25.09 ± 2.04 & 41.25 ± 1.86 & 35.76 ± 0.74 & 34.03 ± 1.21 \\
KLM~\cite{Hendrycks:2022} & 78.19 ± 2.30 & 62.33 ± 2.66 & 70.26 ± 0.64 & 26.66 ± 1.61 & 50.24 ± 1.26 & 45.81 ± 0.59 & 40.90 ± 1.08 \\
VIM~\cite{wang2022vim} & 71.28 ± 0.49 & 47.10 ± 1.10 & 59.19 ± 0.71 & 27.34 ± 0.38 & \textbf{20.39 ± 0.17} & 33.86 ± 0.63 & \textbf{27.20 ± 0.30} \\
KNN~\cite{Sun:2022} & 73.71 ± 0.31 & 46.64 ± 0.73 & 60.18 ± 0.52 & 24.46 ± 1.06 & \textbf{24.45 ± 0.29} & 32.90 ± 1.12 & \textbf{27.27 ± 0.75} \\
DICE~\cite{dice} & 70.84 ± 0.30 & 52.91 ± 1.20 & 61.88 ± 0.67 & 29.66 ± 2.62 & 40.96 ± 1.87 & 38.91 ± 1.16 & 36.51 ± 1.18 \\
RankFeat~\cite{rankfeat} & 90.79 ± 0.37 & 93.32 ± 0.11 & 92.06 ± 0.23 & 98.00 ± 0.80 & 99.40 ± 0.68 & 95.77 ± 0.85 & 97.72 ± 0.75 \\
ASH~\cite{Djurisic:2023} & 72.14 ± 0.97 & 57.63 ± 0.98 & 64.89 ± 0.90 & 22.49 ± 2.24 & \textbf{25.65 ± 0.80} & 33.72 ± 0.97 & \textbf{27.29 ± 1.12} \\
SHE~\cite{she} & 72.64 ± 0.30 & 60.96 ± 1.33 & 66.80 ± 0.74 & 34.38 ± 3.48 & 45.58 ± 2.42 & 46.54 ± 1.34 & 42.17 ± 1.24 \\
GEN~\cite{XiXi:2023} & 66.79 ± 0.26 & 43.61 ± 0.61 & 55.20 ± 0.20 & \textbf{22.03 ± 0.98} & 42.01 ± 0.92 & \textbf{32.25 ± 0.31} & 32.10 ± 0.59 \\
ExCeL~\cite{Karunanayake:2025} & 69.28 ± 0.60 & 46.51 ± 0.20 & 57.90 ± 0.40 & {22.29 ± 1.00} & 30.14 ± 0.64 & 32.91 ± 0.76 & 28.45 ± 0.80 \\
\hline
\multicolumn{8}{c}{Training methods without outliers} \\
\hline
% \rowcolor{gray!30}

ConfBranch~\cite{DeVries:2018} & 72.24 ± 0.37 & 50.63 ± 0.60 & 61.44 ± 0.34 & 23.84 ± 0.40 & 42.42 ± 2.27 & 37.99 ± 0.09 & 34.75 ± 0.63 \\
G-ODIN~\cite{hsu:2020} & 78.23 ± 0.70 & 61.52 ± 0.64 & 69.87 ± 0.46 & 26.13 ± 0.77 & \textbf{28.98 ± 1.15} & 35.43 ± 0.43 & 30.18 ± 0.49 \\
CSI~\cite{Tack:2020} & N/A & N/A & N/A & N/A & N/A & N/A & N/A \\
ARPL~\cite{arpl} & \textbf {65.73 ± 0.51} & 45.75 ± 0.89 & 55.74 ± 0.70 & 29.32 ± 0.64 & 42.87 ± 1.09 & 37.20 ± 0.69 & 36.46 ± 0.08 \\
MOS~\cite{Huang:2021b} & 74.35 ± 0.32 & 68.85 ± 0.68 & 71.60 ± 0.48 & 49.55 ± 0.73 & 51.27 ± 1.02 & 53.86 ± 0.30 & 51.56 ± 0.42 \\
LogitNorm~\cite{Wei:2022} & 67.46 ± 0.21 & 45.46 ± 0.69 & 56.46 ± 0.37 & \textbf{15.70 ± 0.61} & 32.13 ± 0.67 & \textbf{30.49 ± 0.62} & \textbf{26.11 ± 0.52} \\
CIDER~\cite{cider} & 75.50 ± 0.68 & 44.69 ± 0.88 & 60.10 ± 0.73 & 26.54 ± 2.27 & 31.51 ± 3.68 & \textbf{32.47 ± 2.40} & 30.17 ± 2.75 \\
{\sc PSKD}~\cite{Yang:2025} & 67.88 ± 0.33 & 46.36 ± 1.34 & 57.12 ± 0.63 & 24.11 ± 0.93 & 35.05 ± 1.30 & 35.77 ± 1.19 & 31.64 ± 0.87 \\
% \hline
CRAFT ~\cite{Karunanayake:2025b} & 69.07 ± 0.39 & \textbf{40.40 ± 1.13} & {54.74 ± 0.57} & 24.89 ± 1.33 & 38.73 ± 4.55 & 33.48 ± 1.74 & 32.67 ± 1.13 \\
{\sc RankOOD}~\cite{Denipitiyage:2026} & \textbf {60.68 ± 0.61} & \textbf{39.43 ± 0.99} & \textbf{50.05 ± 0.16} & \textbf{19.42 ± 0.71} & 33.81 ± 0.95 & {29.97 ± 0.56} & \textbf{27.73 ± 0.33} \\

IL PL-KD (Ours) & \textbf {64.33$\pm$0.60} & \textbf{38.96$\pm$0.26} & \textbf{51.65$\pm$0.30} & {\bf20.11$\pm$0.5}6 & 37.66$\pm$1.19 & {\bf 29.66$\pm$0.73} & 29.14$\pm$0.70 \\
\hline
\multicolumn{8}{c}{Training methods with outliers} \\
\hline
OE~\cite{Hendrycks:2019} & \textbf{64.67 ± 0.25} & \textbf{39.93 ± 1.13} & \textbf{52.30 ± 0.67} & 27.03 ± 0.47 & 41.92 ± 1.69 & 33.56 ± 0.46 & 34.17 ± 0.56 \\
MCD~\cite{mcd} & \textbf {65.69 ± 0.36} & 43.74 ± 1.32 & \textbf{54.71 ± 0.83} & \textbf{21.74 ± 0.31} & 38.11 ± 0.93 & \textbf{29.93 ± 0.19} & 29.93 ± 0.30 \\
UDG~\cite{udg} & 75.84 ± 1.86 & 61.94 ± 1.61 & 68.89 ± 1.72 & 49.26 ± 7.88 & 71.94 ± 5.25 & 64.92 ± 5.07 & 62.04 ± 5.99 \\
MixOE~\cite{Zhang:2023b} & 68.26 ± 0.48 & 47.69 ± 0.95 & 57.97 ± 0.40 & 30.84 ± 0.45 & 51.44 ± 0.61 & 40.51 ± 1.06 & 40.93 ± 0.29 \\
\bottomrule
\end{tabular}
}
% \captionsetup{width=0.8\linewidth}

\end{table*}

\begin{table*}[t]
\centering
% \scriptsize
\caption{AUROC (\% $\uparrow$) of various methods for different OOD datasets when ImageNet-200 is ID. For each column, the top five methods are marked in \textbf{bold}. Note that N/A indicates that results are not reported in OpenOOD.}
\label{tab:auroc_tin}
\renewcommand{\arraystretch}{1.09}
\resizebox{\textwidth}{!}{
\begin{tabular}{lccc|cccc}
\toprule
Method & \multicolumn{3}{c|}{Near OOD} & \multicolumn{4}{c}{Far OOD} \\
 & SSB-hard & NINCO & Average & iNaturalist & Textures & OpenImage-O  & Average\\ 
 \hline
 \multicolumn{8}{c}{Post-hoc inference methods} \\ 
\hline
OpenMax~\cite{openmaxbendale2016towards} & 77.53 ± 0.08 & 83.01 ± 0.17 & 80.27 ± 0.10 & 92.32 ± 0.32 & 90.21 ± 0.07 & 88.07 ± 0.14 & 90.20 ± 0.17 \\
MSP~\cite{Hendrycks:2017} & 80.38 ± 0.03 & 86.29 ± 0.11 & 83.34 ± 0.06 & 92.80 ± 0.25 & 88.36 ± 0.13 & 89.24 ± 0.02 & 90.13 ± 0.09 \\
TempScale~\cite{tempscale} & {80.71 ± 0.02} & {86.67 ± 0.08} & {83.69 ± 0.04} & 93.39 ± 0.25 & 89.24 ± 0.11 & 89.84 ± 0.02 & 90.82 ± 0.09 \\
ODIN~\cite{hsu:2020} & 77.19 ± 0.06 & 83.34 ± 0.12 & 80.27 ± 0.08 & \textbf{94.37 ± 0.41} & 90.65 ± 0.20 & 90.11 ± 0.15 & 91.71 ± 0.19 \\
MDS~\cite{Lee:2018b} & 58.38 ± 0.58 & 65.48 ± 0.46 & 61.93 ± 0.51 & 75.03 ± 0.76 & 79.25 ± 0.33 & 69.87 ± 0.14 & 74.72 ± 0.26 \\
MDSEns~\cite{Lee:2018b} & 50.46 ± 0.36 & 58.18 ± 0.42 & 54.32 ± 0.24 & 62.16 ± 0.73 & 80.70 ± 0.48 & 64.96 ± 0.51 & 69.27 ± 0.57 \\
RMDS~\cite{rmds} & 80.20 ± 0.23 & 84.94 ± 0.28 & 82.57 ± 0.25 & 90.64 ± 0.46 & 86.77 ± 0.38 & 86.77 ± 0.22 & 88.06 ± 0.34 \\
Gram~\cite{gram} & 65.95 ± 1.08 & 69.40 ± 1.07 & 67.67 ± 1.07 & 65.30 ± 0.20 & 80.53 ± 0.37 & 67.72 ± 0.58 & 71.19 ± 0.24 \\
EBO~\cite{Liu:2020b} & 79.83 ± 0.02 & 85.17 ± 0.11 & 82.50 ± 0.05 & 92.55 ± 0.50 & 90.79 ± 0.16 & 89.23 ± 0.26 & 90.86 ± 0.21 \\
OpenGAN~\cite{opengan} & 55.08 ± 1.84 & 64.49 ± 4.98 & 59.79 ± 3.39 & 75.32 ± 3.32 & 70.58 ± 4.66 & 73.54 ± 4.48 & 73.15 ± 4.07 \\
GradNorm~\cite{Huang:2021} & 72.12 ± 0.43 & 73.39 ± 0.63 & 72.75 ± 0.48 & 86.06 ± 1.90 & 86.07 ± 0.36 & 80.66 ± 1.09 & 84.26 ± 0.87 \\
ReAct~\cite{Sun:2021} & 78.97 ± 1.33 & 84.76 ± 0.64 & 81.87 ± 0.98 & 93.65 ± 0.88 & \textbf{92.86 ± 0.47} & \textbf{90.40 ± 0.35} & \textbf{92.31 ± 0.56} \\
MLS~\cite{Hendrycks:2022} & 80.15 ± 0.01 & 85.65 ± 0.09 & 82.90 ± 0.04 & 93.12 ± 0.45 & 90.60 ± 0.16 & 89.62 ± 0.21 & 91.11 ± 0.19 \\
KLM~\cite{Hendrycks:2022} & 77.56 ± 0.18 & 83.96 ± 0.12 & 80.76 ± 0.08 & 91.80 ± 0.21 & 86.13 ± 0.12 & 87.66 ± 0.17 & 88.53 ± 0.11 \\
VIM~\cite{wang2022vim} & 74.04 ± 0.31 & 83.32 ± 0.19 & 78.68 ± 0.24 & 90.96 ± 0.36 & \textbf{94.61 ± 0.12} & 88.20 ± 0.18 & 91.26 ± 0.19 \\
KNN~\cite{Sun:2022} & 77.03 ± 0.23 & 86.10 ± 0.12 & 81.57 ± 0.17 & \textbf{93.99 ± 0.36} & \textbf{95.29 ± 0.02} & {90.19 ± 0.32} & \textbf{93.16 ± 0.22} \\
DICE~\cite{dice} & 79.06 ± 0.05 & 84.49 ± 0.24 & 81.78 ± 0.14 & 91.81 ± 0.79 & 91.53 ± 0.21 & 89.06 ± 0.34 & 90.80 ± 0.31 \\
RankFeat~\cite{rankfeat} & 58.74 ± 0.94 & 55.10 ± 2.52 & 56.92 ± 1.59 & 33.08 ± 4.68 & 29.10 ± 2.57 & 52.48 ± 4.44 & 38.22 ± 3.85 \\
ASH~\cite{Djurisic:2023} & 79.52 ± 0.37 & 85.24 ± 0.08 & 82.38 ± 0.19 & \textbf{95.10 ± 0.47} & \textbf{94.77 ± 0.19} & \textbf{91.82 ± 0.25} & \textbf{93.90 ± 0.27} \\
SHE~\cite{she} & 78.30 ± 0.20 & 82.07 ± 0.33 & 80.18 ± 0.25 & 91.43 ± 1.28 & 90.51 ± 0.19 & 87.49 ± 0.70 & 89.81 ± 0.61 \\
GEN~\cite{XiXi:2023} & \textbf{80.75 ± 0.03} & 86.60 ± 0.08 & \textbf{83.68 ± 0.06} & 93.70 ± 0.18 & 90.25 ± 0.10 & 90.13 ± 0.06 & 91.36 ± 0.10 \\
ExCeL~\cite{Karunanayake:2025} & 79.39 ± 0.03 & 85.40 ± 0.04 & 82.40 ± 0.04 & 93.76 ± 0.43 & 92.40 ± 0.05 & 89.75 ± 0.32 & 91.97 ± 0.27 \\
\hline
\multicolumn{8}{c}{Training methods without outliers} \\
\hline
% \rowcolor{gray!30}

ConfBranch~\cite{DeVries:2018} & 75.01 ± 0.35 & 83.19 ± 0.14 & 79.10 ± 0.24 & 93.40 ± 0.09 & 89.64 ± 0.52 & 88.26 ± 0.07 & 90.43 ± 0.18 \\
G-ODIN~\cite{hsu:2020} & 72.94 ± 0.05 & 81.63 ± 0.21 & 77.28 ± 0.10 & 93.12 ± 0.21 & \textbf{93.67 ± 0.21} & 90.18 ± 0.15 & \textbf{92.33 ± 0.11} \\
CSI~\cite{Tack:2020} & N/A & N/A & N/A & N/A & N/A & N/A & N/A \\
ARPL~\cite{arpl} & 79.24 ± 0.14 & 84.81 ± 0.07 & 82.02 ± 0.10 & 91.54 ± 0.05 & 88.11 ± 0.34 & 88.04 ± 0.20 & 89.23 ± 0.11 \\
MOS~\cite{Huang:2021b} & 66.54 ± 0.49 & 73.14 ± 0.47 & 69.84 ± 0.46 & 79.69 ± 1.38 & 81.38 ± 0.75 & 80.29 ± 0.68 & 80.46 ± 0.92 \\
LogitNorm~\cite{Wei:2022} & 78.42 ± 0.23 & \textbf{86.90 ± 0.07} & 82.66 ± 0.15 & \textbf{96.26 ± 0.20} & 91.85 ± 0.21 & \textbf{91.01 ± 0.27} & \textbf{93.04 ± 0.21} \\
CIDER~\cite{cider} & 76.04 ± 2.37 & 85.13 ± 1.13 & 80.58 ± 1.75 & 90.69 ± 2.13 & 92.38 ± 1.35 & 88.92 ± 1.58 & 90.66 ± 1.68 \\

% \hline
CRAFT~\cite{Karunanayake:2025b} & 80.70 ± 0.18 & \textbf{86.74 ± 0.84} & \textbf{83.72 ± 0.41} & 92.85 ± 0.66 & 89.94 ± 0.49 & 89.85 ± 0.46 & 90.88 ± 0.89 \\
{\sc PSKD}~\cite{Yang:2025} & 79.87 ± 0.29 & 85.81 ± 0.42 & 82.84 ± 0.32 & 92.97 ± 0.19 & 91.80 ± 0.31 & 89.40 ± 0.26 & 91.39 ± 0.16 \\
{\sc RankOOD}~\cite{Denipitiyage:2026} & \textbf{82.57 ± 0.28} & \textbf{88.04 ± 0.26} & \textbf{85.30 ± 0.18} & \textbf{94.69 ± 0.17} & 90.85 ± 0.14 & \textbf{90.89 ± 0.25} & 92.14 ± 0.20 \\
IL PL-KD (Ours) & {\bf 81.12$\pm$0.06}	& {\bf 87.59$\pm$0.10}	& {\bf 84.36$\pm$0.07}	& 93.91$\pm$0.05	& 90.16$\pm$0.22	& {\bf 90.41$\pm$0.07}	& 91.49$\pm$0.08\\

% & 80.37 ± 0.00 & 86.52 ± 0.00 & 83.45 ± 0.00 & 92.93 ± 0.00 & 89.51 ± 0.00 & 89.28 ± 0.00 & 90.58 ± 0.00 \\
\hline
\multicolumn{8}{c}{Training methods with outliers} \\
\hline
OE~\cite{Hendrycks:2019} & \textbf{82.34 ± 0.16} & \textbf{87.35 ± 0.23} & \textbf{84.84 ± 0.16} & 90.30 ± 0.16 & 87.76 ± 0.32 & 89.01 ± 0.24 & 89.02 ± 0.18 \\
MCD~\cite{mcd} & \textbf{81.51 ± 0.14} & 85.74 ± 0.07 & 83.62 ± 0.09 & 90.83 ± 0.10 & 86.87 ± 0.12 & 89.12 ± 0.18 & 88.94 ± 0.10 \\
UDG~\cite{udg} & 70.73 ± 1.74 & 77.88 ± 1.56 & 74.30 ± 1.63 & 85.95 ± 2.97 & 81.79 ± 2.57 & 78.54 ± 2.98 & 82.09 ± 2.78 \\
MixOE~\cite{Zhang:2023b} & 80.23 ± 0.15 & 85.01 ± 0.10 & 82.62 ± 0.03 & 90.64 ± 0.36 & 86.80 ± 0.45 & 87.36 ± 0.49 & 88.27 ± 0.41 \\
\bottomrule
\end{tabular}}
% \captionsetup{width=0.8\linewidth}

\end{table*}

%%%%%%%%%%%%%%%%% CIFAR 100 %%%%%%%%%%%%%%

\begin{table*}[h]
\centering
% \scriptsize
\caption{ FPR95 (\% $\downarrow$) of various methods for different OOD datasets when CIFAR-100 is ID. For each column, the top five methods are marked in \textbf{bold}.}
\label{tab:fpr_cifar100}
\renewcommand{\arraystretch}{1.09}
\resizebox{\textwidth}{!}{
\begin{tabular}{lccc|ccccc}
\toprule
Method & \multicolumn{3}{c|}{Near OOD} & \multicolumn{5}{c}{Far OOD} \\
 & CIFAR-10 & TIN & Average & MNIST & SVHN & Textures & Places365 & Average\\ 
 \hline
 \multicolumn{9}{c}{Post-hoc inference methods} \\ 
\hline
OpenMax~\cite{openmaxbendale2016towards} & 60.17 ± 0.97 & 52.99 ± 0.51 & 56.58 ± 0.73 & 53.82 ± 4.74 & 53.20 ± 1.78 & 56.12 ± 1.91 & \textbf{54.85 ± 1.42} & 54.50 ± 0.68 \\
MSP~\cite{Hendrycks:2017} & \textbf{58.91 ± 0.93} & 50.70 ± 0.34 & {54.80 ± 0.33} & 57.23 ± 4.68 & 59.07 ± 2.53 & 61.88 ± 1.28 & 56.62 ± 0.87 & 58.70 ± 1.06 \\
TempScale~\cite{tempscale} & \textbf{58.72 ± 0.81} & 50.26 ± 0.16 & {54.49 ± 0.48} & 56.05 ± 4.61 & 57.71 ± 2.68 & 61.56 ± 1.43 & 56.46 ± 0.94 & 57.94 ± 1.14 \\
ODIN~\cite{hsu:2020} & 60.64 ± 0.56 & 55.19 ± 0.57 & 57.91 ± 0.51 & 45.94 ± 3.29 & 67.41 ± 3.88 & 62.37 ± 2.96 & 59.71 ± 0.92 & 58.86 ± 0.79 \\
MDS~\cite{Lee:2018b} & 88.00 ± 0.49 & 79.05 ± 1.22 & 83.53 ± 0.60 & 71.72 ± 2.94 & 67.21 ± 6.09 & 70.49 ± 2.48 & 79.61 ± 0.34 & 72.26 ± 1.56 \\
MDSEns~\cite{Lee:2018b} & 95.94 ± 0.16 & 95.82 ± 0.12 & 95.88 ± 0.04 & \textbf{2.83 ± 0.86} & 82.57 ± 2.58 & 84.94 ± 0.83 & 96.61 ± 0.17 & 66.74 ± 1.04 \\
RMDS~\cite{rmds} & 61.37 ± 0.24 & 49.56 ± 0.90 & 55.46 ± 0.41 & 52.05 ± 6.28 & 51.65 ± 3.68 & 53.99 ± 1.06 & \textbf{53.57 ± 0.43} & 52.81 ± 0.63 \\
Gram~\cite{gram} & 92.71 ± 0.64 & 91.85 ± 0.86 & 92.28 ± 0.29 & 53.53 ± 7.45 & \textbf{20.06 ± 1.96} & 89.51 ± 2.54 & 94.67 ± 0.60 & 64.44 ± 2.37 \\
EBO~\cite{Liu:2020b} & 59.21 ± 0.75 & 52.03 ± 0.50 & 55.62 ± 0.61 & 52.62 ± 3.83 & 53.62 ± 3.14 & 62.35 ± 2.06 & 57.75 ± 0.86 & 56.59 ± 1.38 \\
OpenGAN~\cite{opengan} & 78.83 ± 3.94 & 74.21 ± 1.25 & 76.52 ± 2.59 & 63.09 ± 23.3 & 70.35 ± 2.06 & 74.77 ± 1.78 & 73.75 ± 8.32 & 70.49 ± 7.38 \\
GradNorm~\cite{Huang:2021} & 84.30 ± 0.36 & 86.85 ± 0.62 & 85.58 ± 0.46 & 86.97 ± 1.44 & 69.90 ± 7.94 & 92.51 ± 0.61 & 85.32 ± 0.44 & 83.68 ± 1.92 \\
ReAct~\cite{Sun:2021} & 61.30 ± 0.43 & 51.47 ± 0.47 & 56.39 ± 0.34 & 56.04 ± 5.66 & 50.41 ± 2.02 & 55.04 ± 0.82 & {55.30 ± 0.41} & 54.20 ± 1.56 \\
MLS~\cite{Hendrycks:2022} & {59.11 ± 0.64} & 51.83 ± 0.70 & 55.47 ± 0.66 & 52.95 ± 3.82 & 53.90 ± 3.04 & 62.39 ± 2.13 & 57.68 ± 0.91 & 56.73 ± 1.33 \\
KLM~\cite{Hendrycks:2022} & 84.77 ± 2.95 & 71.07 ± 0.59 & 77.92 ± 1.31 & 73.09 ± 6.67 & 50.30 ± 7.04 & 81.80 ± 5.80 & 81.40 ± 1.58 & 71.65 ± 2.01 \\
VIM~\cite{wang2022vim} & 70.59 ± 0.43 & 54.66 ± 0.42 & 62.63 ± 0.27 & 48.32 ± 1.07 & 46.22 ± 5.46 & \textbf{46.86 ± 2.29} & 61.57 ± 0.77 & \textbf{50.74 ± 1.00} \\
KNN~\cite{Sun:2022} & 72.80 ± 0.44 & 49.65 ± 0.37 & 61.22 ± 0.14 & 48.58 ± 4.67 & 51.75 ± 3.12 & \textbf{53.56 ± 2.32} & 60.70 ± 1.03 & 53.65 ± 0.28 \\
DICE~\cite{dice} & 60.98 ± 1.10 & 54.93 ± 0.53 & 57.95 ± 0.53 & 51.79 ± 3.67 & 49.58 ± 3.32 & 64.23 ± 1.65 & 59.39 ± 1.25 & 56.25 ± 0.60 \\
RankFeat~\cite{rankfeat} & 82.78 ± 1.56 & 78.40 ± 0.95 & 80.59 ± 1.10 & 75.01 ± 5.83 & 58.49 ± 2.30 & 66.87 ± 3.80 & 77.42 ± 1.96 & 69.45 ± 1.01 \\
ASH~\cite{Djurisic:2023} & 68.06 ± 0.44 & 63.35 ± 0.90 & 65.71 ± 0.24 & 66.58 ± 3.88 & 46.00 ± 2.67 & 61.27 ± 2.74 & 62.95 ± 0.99 & 59.20 ± 2.46 \\
SHE~\cite{she} & 60.41 ± 0.51 & 57.74 ± 0.73 & 59.07 ± 0.25 & 58.78 ± 2.70 & 59.15 ± 7.61 & 73.29 ± 3.22 & 65.24 ± 0.98 & 64.12 ± 2.70 \\
GEN~\cite{XiXi:2023} & \textbf{58.87 ± 0.69} & 49.98 ± 0.05 & \textbf{54.42 ± 0.33} & 53.92 ± 5.71 & 55.45 ± 2.76 & 61.23 ± 1.40 & 56.25 ± 1.01 & 56.71 ± 1.59 \\
ExCeL~\cite{Karunanayake:2025} & 61.07 ± 0.81 & {49.35 ± 0.31} & 55.21 ± 0.56 & 54.67 ± 5.86 & {45.13 ± 0.33} & \textbf{51.14 ± 0.14} & 58.02 ± 1.28 & \textbf{52.24 ± 1.90} \\
\hline
\multicolumn{9}{c}{Training methods without outliers} \\
\hline
% \rowcolor{gray!30}

ConfBranch~\cite{DeVries:2018} & 74.56 ± 1.22 & 65.86 ± 0.56 & 70.21 ± 0.83 & 55.95 ± 6.15 & 76.01 ± 12.3 & 85.43 ± 1.17 & 69.90 ± 0.28 & 71.82 ± 3.39 \\
G-ODIN~\cite{hsu:2020} & 78.82 ± 1.86 & 56.34 ± 0.45 & 67.58 ± 0.98 & \textbf{27.19 ± 6.24} & \textbf{42.68 ± 5.74} & \textbf{35.83 ± 1.15} & 65.03 ± 1.16 & \textbf{42.68 ± 3.19} \\
CSI~\cite{Tack:2020} & 72.62 ± 0.49 & 67.90 ± 0.64 & 70.26 ± 0.56 & 80.54 ± 4.87 & 67.21 ± 3.35 & 90.51 ± 1.47 & 69.41 ± 0.58 & 76.92 ± 1.29 \\
ARPL~\cite{arpl} & 64.84 ± 1.25 & 58.27 ± 2.40 & 61.56 ± 1.81 & 59.12 ± 8.04 & 59.76 ± 1.58 & 71.66 ± 1.81 & 62.01 ± 0.89 & 63.14 ± 2.53 \\
MOS~\cite{Huang:2021b} & 60.60 ± 1.47 & 51.49 ± 0.69 & 56.05 ± 1.01 & 52.70 ± 3.81 & 56.33 ± 8.46 & 61.24 ± 2.06 & 58.86 ± 0.41 & 57.28 ± 3.29 \\
LogitNorm~\cite{Wei:2022} & 73.88 ± 1.21 & 51.89 ± 0.10 & 62.89 ± 0.57 & \textbf{34.12 ± 8.32} & 47.52 ± 8.02 & 77.38 ± 2.99 & 55.44 ± 1.45 & 53.61 ± 3.45 \\
CIDER~\cite{cider} & 82.71 ± 1.25 & 61.33 ± 0.64 & 72.02 ± 0.31 & 75.32 ± 4.21 & \textbf{17.82 ± 2.80} & 54.43 ± 2.56 & 69.30 ± 1.81 & 54.22 ± 1.24 \\
% \hline
CRAFT~\cite{Karunanayake:2025b} & 59.19 ± 0.64 & \textbf{48.26 ± 1.21} & \textbf{53.73 ± 0.62} & 48.95 ± 1.90 & 47.50 ± 5.22 & 56.97 ± 1.77 & \textbf{54.02 ± 0.30} & \textbf{51.86 ± 0.49} \\
{\sc PSKD}~\cite{Yang:2025} & 60.73 ± 4.96 & 48.93 ± 0.63 & 54.83 ± 2.79 & 42.91 ± 2.95 & 44.76 ± 4.73 & 63.23 ± 2.45 & 55.34 ± 0.60 & 51.56 ± 2.39 \\
{\sc RankOOD}~\cite{Denipitiyage:2026} & \textbf{55.71 ± 1.92} & 49.47 ± 0.80 & \textbf{52.59 ± 0.75} & \textbf{42.25 ± 2.92} & \textbf{39.21 ± 1.59} & \textbf{52.88 ± 1.89} & 55.40 ± 0.47 & \textbf{47.44 ± 0.80} \\

IL PL-KD (Ours) & 
% \textbf{58.74 ± 0.01} & \textbf{48.60 ± 0.01} & \textbf{53.67 ± 0.01} & 52.41 ± 0.02 & 61.22 ± 0.04 & 58.41 ± 0.01 & 56.05 ± 0.01 & 57.02 ± 0.02 \\
{\bf 57.21$\pm$1.27} & {\bf 48.97$\pm$0.09} & {\bf 53.09$\pm$0.59} & 49.79$\pm$1.34 & 52.59$\pm$5.74 & 56.73$\pm$1.04 & {\bf 52.84$\pm$0.46} & 52.99$\pm$1.9\\
\hline
\multicolumn{9}{c}{Training methods with outliers} \\
\hline
OE~\cite{Hendrycks:2019} & 61.26 ± 0.22 & \textbf{0.21 ± 0.01} & \textbf{30.73 ± 0.11} & 53.31 ± 9.91 & 51.84 ± 3.45 & 55.83 ± 1.82 & 58.30 ± 0.72 & 54.82 ± 2.79 \\
MCD~\cite{mcd} & 62.65 ± 0.54 & \textbf{49.10 ± 1.29} & 55.88 ± 0.85 & 62.78 ± 2.91 & \textbf{43.71 ± 3.73} & 56.89 ± 0.64 & \textbf{54.17 ± 1.13} & 54.39 ± 1.34 \\
UDG~\cite{udg} & 66.40 ± 0.51 & 56.43 ± 0.68 & 61.42 ± 0.48 & \textbf{45.14 ± 12.8} & 59.67 ± 5.62 & 71.33 ± 3.59 & 59.85 ± 0.57 & 59.00 ± 3.35 \\
MixOE~\cite{Zhang:2023b} & 61.12 ± 1.08 & \textbf{49.32 ± 0.36} & 55.22 ± 0.49 & 59.49 ± 7.74 & 73.09 ± 4.00 & 66.04 ± 0.98 & 56.93 ± 0.78 & 63.88 ± 2.48 \\
\bottomrule
\end{tabular}
}
% \captionsetup{width=0.87\linewidth}

\end{table*}

\begin{table*}[h]
\centering
% \scriptsize

\caption{ AUROC (\% $\uparrow$) of various methods for different OOD datasets when CIFAR-100 is ID. For each column, the top five methods are marked in \textbf{bold}.}
\label{tab:auroc_cifar100}
\renewcommand{\arraystretch}{1.09}
\resizebox{\textwidth}{!}{
\begin{tabular}{lccc|ccccc}
\toprule
Method & \multicolumn{3}{c|}{Near OOD} & \multicolumn{5}{c}{Far OOD} \\
 & CIFAR-10 & TIN & Average & MNIST & SVHN & Textures & Places365 & Average\\ 
 \hline
 \multicolumn{9}{c}{Post-hoc inference methods} \\ 
\hline
OpenMax~\cite{openmaxbendale2016towards} & 74.38 ± 0.37 & 78.44 ± 0.14 & 76.41 ± 0.25 & 76.01 ± 1.39 & 82.07 ± 1.53 & 80.56 ± 0.09 & 79.29 ± 0.40 & 79.48 ± 0.41 \\
MSP~\cite{Hendrycks:2017} & 78.47 ± 0.07 & 82.07 ± 0.17 & 80.27 ± 0.11 & 76.08 ± 1.86 & 78.42 ± 0.89 & 77.32 ± 0.71 & 79.22 ± 0.29 & 77.76 ± 0.44 \\
TempScale~\cite{tempscale} & {79.02 ± 0.06} & 82.79 ± 0.09 & 80.90 ± 0.07 & 77.27 ± 1.85 & 79.79 ± 1.05 & 78.11 ± 0.72 & 79.80 ± 0.25 & 78.74 ± 0.51 \\
ODIN~\cite{hsu:2020} & 78.18 ± 0.14 & 81.63 ± 0.08 & 79.90 ± 0.11 & 83.79 ± 1.31 & 74.54 ± 0.76 & 79.33 ± 1.08 & 79.45 ± 0.26 & 79.28 ± 0.21 \\
MDS~\cite{Lee:2018b} & 55.87 ± 0.22 & 61.50 ± 0.28 & 58.69 ± 0.09 & 67.47 ± 0.81 & 70.68 ± 6.40 & 76.26 ± 0.69 & 63.15 ± 0.49 & 69.39 ± 1.39 \\
MDSEns~\cite{Lee:2018b} & 43.85 ± 0.31 & 48.78 ± 0.19 & 46.31 ± 0.24 & \textbf{98.21 ± 0.78} & 53.76 ± 1.63 & 69.75 ± 1.14 & 42.27 ± 0.73 & 66.00 ± 0.69 \\
RMDS~\cite{rmds} & 77.75 ± 0.19 & 82.55 ± 0.02 & 80.15 ± 0.11 & 79.74 ± 2.49 & 84.89 ± 1.10 & \textbf{83.65 ± 0.51} & \textbf{83.40 ± 0.46} & \textbf{82.92 ± 0.42} \\
Gram~\cite{gram} & 49.41 ± 0.58 & 53.91 ± 1.58 & 51.66 ± 0.77 & 80.71 ± 4.15 & \textbf{95.55 ± 0.60} & 70.79 ± 1.32 & 46.38 ± 1.21 & 73.36 ± 1.08 \\
EBO~\cite{Liu:2020b} & \textbf{79.05 ± 0.11} & 82.76 ± 0.08 & {80.91 ± 0.08} & 79.18 ± 1.37 & 82.03 ± 1.74 & 78.35 ± 0.83 & 79.52 ± 0.23 & 79.77 ± 0.61 \\
OpenGAN~\cite{opengan} & 63.23 ± 2.44 & 68.74 ± 2.29 & 65.98 ± 1.26 & 68.14 ± 18.8 & 68.40 ± 2.15 & 65.84 ± 3.43 & 69.13 ± 7.08 & 67.88 ± 7.16 \\
GradNorm~\cite{Huang:2021} & 70.32 ± 0.20 & 69.95 ± 0.79 & 70.13 ± 0.47 & 65.35 ± 1.12 & 76.95 ± 4.73 & 64.58 ± 0.13 & 69.69 ± 0.17 & 69.14 ± 1.05 \\
ReAct~\cite{Sun:2021} & 78.65 ± 0.05 & 82.88 ± 0.08 & 80.77 ± 0.05 & 78.37 ± 1.59 & 83.01 ± 0.97 & 80.15 ± 0.46 & {80.03 ± 0.11} & 80.39 ± 0.49 \\
MLS~\cite{Hendrycks:2022} & \textbf{79.21 ± 0.10} & 82.90 ± 0.05 & \textbf{81.05 ± 0.07} & 78.91 ± 1.47 & 81.65 ± 1.49 & 78.39 ± 0.84 & 79.75 ± 0.24 & 79.67 ± 0.57 \\
KLM~\cite{Hendrycks:2022} & 73.91 ± 0.25 & 79.22 ± 0.28 & 76.56 ± 0.25 & 74.15 ± 2.59 & 79.34 ± 0.44 & 75.77 ± 0.45 & 75.70 ± 0.24 & 76.24 ± 0.52 \\
VIM~\cite{wang2022vim} & 72.21 ± 0.41 & 77.76 ± 0.16 & 74.98 ± 0.13 & 81.89 ± 1.02 & 83.14 ± 3.71 & \textbf{85.91 ± 0.78} & 75.85 ± 0.37 & 81.70 ± 0.62 \\
KNN~\cite{Sun:2022} & 77.02 ± 0.25 & \textbf{83.34 ± 0.16} & 80.18 ± 0.15 & 82.36 ± 1.52 & 84.15 ± 1.09 & \textbf{83.66 ± 0.83} & 79.43 ± 0.47 & \textbf{82.40 ± 0.17} \\
DICE~\cite{dice} & 78.04 ± 0.32 & 80.72 ± 0.30 & 79.38 ± 0.23 & 79.86 ± 1.89 & 84.22 ± 2.00 & 77.63 ± 0.34 & 78.33 ± 0.66 & 80.01 ± 0.18 \\
RankFeat~\cite{rankfeat} & 58.04 ± 2.36 & 65.72 ± 0.22 & 61.88 ± 1.28 & 63.03 ± 3.86 & 72.14 ± 1.39 & 69.40 ± 3.08 & 63.82 ± 1.83 & 67.10 ± 1.42 \\
ASH~\cite{Djurisic:2023} & 76.48 ± 0.30 & 79.92 ± 0.20 & 78.20 ± 0.15 & 77.23 ± 0.46 & {85.60 ± 1.40} & 80.72 ± 0.70 & 78.76 ± 0.16 & 80.58 ± 0.66 \\
SHE~\cite{she} & 78.15 ± 0.03 & 79.74 ± 0.36 & 78.95 ± 0.18 & 76.76 ± 1.07 & 80.97 ± 3.98 & 73.64 ± 1.28 & 76.30 ± 0.51 & 76.92 ± 1.16 \\
GEN~\cite{XiXi:2023} & \textbf{79.38 ± 0.04} & { 83.25 ± 0.13} & \textbf{81.31 ± 0.08} & 78.29 ± 2.05 & 81.41 ± 1.50 & 78.74 ± 0.81 & {80.28 ± 0.27} & 79.68 ± 0.75 \\
ExCeL~\cite{Karunanayake:2025} & 78.14 ± 0.09 & \textbf{83.26 ± 0.03} & 80.70 ± 0.06 & 78.99 ± 1.73 & \textbf{85.91 ± 0.73} & \textbf{83.28 ± 0.58} & 79.98 ± 0.57 & {82.04 ± 0.90} \\
\hline
\multicolumn{9}{c}{Training methods without outliers} \\
\hline
% \rowcolor{gray!30}
ConfBranch~\cite{DeVries:2018} & 68.80 ± 0.73 & 74.41 ± 0.54 & 71.60 ± 0.62 & 74.29 ± 4.44 & 65.51 ± 8.07 & 65.39 ± 0.16 & 70.42 ± 0.26 & 68.90 ± 1.83 \\
G-ODIN~\cite{hsu:2020} & 73.04 ± 0.39 & 81.26 ± 0.29 & 77.15 ± 0.28 & \textbf{91.15 ± 2.86} & 83.74 ± 3.10 & \textbf{89.62 ± 0.36} & 78.17 ± 0.62 & \textbf{85.67 ± 1.58} \\
CSI~\cite{Tack:2020} & 69.50 ± 0.43 & 73.40 ± 0.13 & 71.45 ± 0.27 & 51.79 ± 6.77 & 80.24 ± 1.80 & 62.22 ± 0.98 & 70.99 ± 0.54 & 66.31 ± 1.21 \\
ARPL~\cite{arpl} & 73.38 ± 0.78 & 76.50 ± 1.11 & 74.94 ± 0.93 & 73.77 ± 5.89 & 76.45 ± 1.00 & 69.93 ± 1.33 & 74.62 ± 0.57 & 73.69 ± 1.80 \\
MOS~\cite{Huang:2021b} & 78.54 ± 0.13 & 82.26 ± 0.25 & 80.40 ± 0.18 & 80.68 ± 1.65 & 81.59 ± 3.81 & 79.92 ± 0.57 & 78.50 ± 0.34 & 80.17 ± 1.21 \\
LogitNorm~\cite{Wei:2022} & 74.57 ± 0.39 & 82.37 ± 0.24 & 78.47 ± 0.31 & \textbf{90.69 ± 1.38} & 82.80 ± 4.57 & 72.37 ± 0.67 & {80.25 ± 0.61} & 81.53 ± 1.26 \\
CIDER~\cite{cider} & 67.55 ± 0.60 & 78.65 ± 0.35 & 73.10 ± 0.39 & 68.14 ± 3.98 & \textbf{97.17 ± 0.34} & 82.21 ± 1.93 & 74.43 ± 0.64 & 80.49 ± 0.68 \\

% \hline
CRAFT~\cite{Karunanayake:2025b} & {78.67 ± 0.21} & 83.14 ± 0.73 & 80.90 ± 0.33 & 80.34 ± 1.84 & 85.16 ± 1.15 & 80.91 ± 0.45 & \textbf{81.71 ± 0.12} & 82.03 ± 0.34 \\
{\sc PSKD}~\cite{Yang:2025} & \textbf{79.16 ± 0.93} & \textbf{83.74 ± 0.18} & \textbf{81.45 ± 0.44} & \textbf{84.19 ± 0.78} & \textbf{86.20 ± 1.32} & 78.64 ± 0.85 & \textbf{80.59 ± 0.67} & \textbf{82.40 ± 0.52} \\
{\sc RankOOD}~\cite{Denipitiyage:2026} & 78.84 ± 0.79 & 82.50 ± 0.39 & 80.67 ± 0.40 & \textbf{84.00 ± 2.15} & \textbf{87.75 ± 1.25} & 82.04 ± 0.70 & \textbf{80.73 ± 0.46} & \textbf{83.63 ± 1.06} \\
IL PL-KD (Ours) & 
% \textbf{79.43 ± 0.00} & \textbf{83.36 ± 0.00} & \textbf{81.40 ± 0.00} & 77.14 ± 0.01 & 77.00 ± 0.02 & 78.48 ± 0.00 & 79.91 ± 0.01 & 78.13 ± 0.00 \\
{\bf 79.90$\pm$0.28} & 83.20$\pm$0.06 & {\bf 81.55$\pm$0.13} & 78.59$\pm$0.53 & 80.55$\pm$3.17 & 79.53$\pm$0.37 & {\bf 81.10$\pm$0.25} & 79.95$\pm$1.04 \\
\hline
\multicolumn{9}{c}{Training methods with outliers} \\
\hline
OE~\cite{Hendrycks:2019} & 76.70 ± 0.19 & \textbf{99.89 ± 0.02} & \textbf{88.30 ± 0.10} & 80.68 ± 5.82 & 84.37 ± 1.34 & 82.18 ± 0.68 & 78.39 ± 0.41 & 81.41 ± 1.49 \\
MCD~\cite{mcd} & 75.40 ± 0.46 & 78.75 ± 0.21 & 77.07 ± 0.32 & 68.25 ± 1.99 & 75.92 ± 0.37 & 77.07 ± 0.76 & 77.65 ± 0.09 & 74.72 ± 0.78 \\
UDG~\cite{udg} & 75.15 ± 0.15 & 80.90 ± 0.21 & 78.02 ± 0.10 & {83.88 ± 5.98} & 79.80 ± 1.61 & 75.57 ± 0.80 & 79.11 ± 0.17 & 79.59 ± 1.77 \\
MixOE~\cite{Zhang:2023b} & 78.17 ± 0.29 & \textbf{83.73 ± 0.12} & {80.95 ± 0.20} & 76.06 ± 5.52 & 72.28 ± 0.81 & 77.34 ± 0.91 & 79.92 ± 0.30 & 76.40 ± 1.44 \\
\bottomrule
\end{tabular}
}
% \captionsetup{width=0.87\linewidth}

\end{table*}

%%%%%%%%%%%%%%%%% CIFAR 10 %%%%%%%%%%%%%%

\begin{table*}[h]
\centering
% \scriptsize
\caption{FPR95 (\% $\downarrow$) of various methods for different OOD datasets when CIFAR-10 is ID. For each column, the top five methods are marked in \textbf{bold}.}
\label{tab:fpr_cifar10}
\renewcommand{\arraystretch}{1.09}
\resizebox{\textwidth}{!}{
\begin{tabular}{lccc|ccccc}
\toprule
Method & \multicolumn{3}{c|}{Near OOD} & \multicolumn{5}{c}{Far OOD} \\
 & CIFAR-100 & TIN & Average & MNIST & SVHN & Textures & Places365 & Average\\ 
 \hline
\multicolumn{9}{c}{Post-hoc inference methods} \\ 
\hline
OpenMax~\cite{openmaxbendale2016towards} & 48.06 ± 3.25 & 39.18 ± 1.44 & 43.62 ± 2.27 & 23.33 ± 4.67 & 25.40 ± 1.47 & 31.50 ± 4.05 & 38.52 ± 2.27 & 29.69 ± 1.21 \\ 
MSP~\cite{Hendrycks:2017} & 53.08 ± 4.86 & 43.27 ± 3.00 & 48.17 ± 3.92 & 23.64 ± 5.81 & 25.82 ± 1.64 & 34.96 ± 4.64 & 42.47 ± 3.81 & 31.72 ± 1.84 \\ 
TempScale~\cite{tempscale} & 55.81 ± 5.07 & 46.11 ± 3.63 & 50.96 ± 4.32 & 23.53 ± 7.05 & 26.97 ± 2.65 & 38.16 ± 5.89 & 45.27 ± 4.50 & 33.48 ± 2.39 \\ 
ODIN~\cite{hsu:2020} & 77.00 ± 5.74 & 75.38 ± 6.42 & 76.19 ± 6.08 & 23.83 ± 12.3 & 68.61 ± 0.52 & 67.70 ± 11.1 & 70.36 ± 6.96 & 57.62 ± 4.24 \\ 
MDS~\cite{Lee:2018b} & 52.81 ± 3.62 & 46.99 ± 4.36 & 49.90 ± 3.98 & 27.30 ± 3.55 & 25.96 ± 2.52 & 27.94 ± 4.20 & 47.67 ± 4.54 & 32.22 ± 3.40 \\ 
MDSEns~\cite{Lee:2018b} & 91.87 ± 0.10 & 92.66 ± 0.42 & 92.26 ± 0.20 & \textbf{1.30 ± 0.51} & 74.34 ± 1.04 & 76.07 ± 0.17 & 94.16 ± 0.33 & 61.47 ± 0.48 \\ 
RMDS~\cite{rmds} & 43.86 ± 3.49 & 33.91 ± 1.39 & 38.89 ± 2.39 & 21.49 ± 2.32 & 23.46 ± 1.48 & 25.25 ± 0.53 & 31.20 ± 0.28 & 25.35 ± 0.73 \\ 
Gram~\cite{gram} & 91.68 ± 2.24 & 90.06 ± 1.59 & 90.87 ± 1.91 & 70.30 ± 8.96 & 33.91 ± 17.4 & 94.64 ± 2.71 & 90.49 ± 1.93 & 72.34 ± 6.73 \\ 
EBO~\cite{Liu:2020b} & 66.60 ± 4.46 & 56.08 ± 4.83 & 61.34 ± 4.63 & 24.99 ± 12.9 & 35.12 ± 6.11 & 51.82 ± 6.11 & 54.85 ± 6.52 & 41.69 ± 5.32 \\ 
OpenGAN~\cite{opengan} & 94.84 ± 3.83 & 94.11 ± 4.21 & 94.48 ± 4.01 & 79.54 ± 19.7 & 75.27 ± 26.9 & 83.95 ± 14.9 & 95.32 ± 4.45 & 83.52 ± 11.6 \\ 
GradNorm~\cite{Huang:2021} & 94.54 ± 1.11 & 94.89 ± 0.60 & 94.72 ± 0.82 & 85.41 ± 4.85 & 91.65 ± 2.42 & 98.09 ± 0.49 & 92.46 ± 2.28 & 91.90 ± 2.23 \\ 
ReAct~\cite{Sun:2021} & 67.40 ± 7.34 & 59.71 ± 7.31 & 63.56 ± 7.33 & 33.77 ± 18.0 & 50.23 ± 15.9 & 51.42 ± 11.4 & 44.20 ± 3.35 & 44.90 ± 8.37 \\ 
MLS~\cite{Hendrycks:2022} & 66.59 ± 4.44 & 56.06 ± 4.82 & 61.32 ± 4.62 & 25.06 ± 12.9 & 35.09 ± 6.09 & 51.73 ± 6.13 & 54.84 ± 6.51 & 41.68 ± 5.27 \\ 
KLM~\cite{Hendrycks:2022} & 90.55 ± 5.83 & 85.18 ± 7.60 & 87.86 ± 6.37 & 76.22 ± 12.1 & 59.47 ± 7.06 & 81.95 ± 9.95 & 95.58 ± 2.12 & 78.31 ± 4.84 \\ 
VIM~\cite{wang2022vim} & 49.19 ± 3.15 & 40.49 ± 1.55 & 44.84 ± 2.31 & 18.36 ± 1.42 & 19.29 ± 0.41 & \textbf{21.14 ± 1.83} & 41.43 ± 2.17 & 25.05 ± 0.52 \\ 
KNN~\cite{Sun:2022} & 37.64 ± 0.31 & 30.37 ± 0.65 & 34.01 ± 0.38 & 20.05 ± 1.36 & 22.60 ± 1.26 & 24.06 ± 0.55 & 30.38 ± 0.63 & 24.27 ± 0.40 \\ 
DICE~\cite{dice} & 73.71 ± 7.67 & 66.37 ± 7.68 & 70.04 ± 7.64 & 30.83 ± 10.5 & 36.61 ± 4.74 & 62.42 ± 4.79 & 77.19 ± 12.6 & 51.76 ± 4.42 \\ 
RankFeat~\cite{rankfeat} & 65.32 ± 3.48 & 56.44 ± 5.76 & 60.88 ± 4.60 & 61.86 ± 12.8 & 64.49 ± 7.38 & 59.71 ± 9.79 & 43.70 ± 7.39 & 57.44 ± 7.99 \\ 
ASH~\cite{Djurisic:2023} & 87.31 ± 2.06 & 86.25 ± 1.58 & 86.78 ± 1.82 & 70.00 ± 10.6 & 83.64 ± 6.48 & 84.59 ± 1.74 & 77.89 ± 7.28 & 79.03 ± 4.22 \\ 
SHE~\cite{she} & 81.00 ± 3.42 & 78.30 ± 3.52 & 79.65 ± 3.47 & 42.22 ± 20.6 & 62.74 ± 4.01 & 84.60 ± 5.30 & 76.36 ± 5.32 & 66.48 ± 5.98 \\ 
GEN~\cite{XiXi:2023} & 58.75 ± 3.97 & 48.59 ± 2.34 & 53.67 ± 3.14 & 23.00 ± 7.75 & 28.14 ± 2.59 & 40.74 ± 6.61 & 47.03 ± 3.22 & 34.73 ± 1.58 \\ 
ExCeL~\cite{Karunanayake:2025} & 71.16 ± 1.34 & 61.42 ± 0.26 & 66.29 ± 0.43 & {15.46 ± 1.89} & 31.78 ± 3.65 & 53.67 ± 2.19 & 55.09 ± 1.12 & 39.00 ± 0.84 \\ 
\hline
\multicolumn{9}{c}{Training methods without outliers} \\
\hline
% \rowcolor{gray!30}

ConfBranch~\cite{DeVries:2018} & \textbf{34.44 ± 0.81} & {28.11 ± 0.61} & \textbf{31.28 ± 0.66} & {15.79 ± 2.00} & \textbf{14.06 ± 0.84} & 27.24 ± 1.32 & 28.85 ± 1.03 & 21.48 ± 0.94 \\ 
G-ODIN~\cite{hsu:2020} & 48.86 ± 2.91 & 42.21 ± 2.18 & 45.54 ± 2.52 & \textbf{4.53 ± 2.08} & \textbf{10.72 ± 0.88} & 27.27 ± 6.73 & 43.30 ± 3.57 & 21.45 ± 1.91 \\ 
CSI~\cite{Tack:2020} & 37.57 ± 0.89 & 29.74 ± 0.42 & 33.66 ± 0.64 & 24.41 ± 1.57 & 17.56 ± 0.12 & 28.95 ± 1.33 & 34.76 ± 1.52 & 26.42 ± 0.29 \\ 
ARPL~\cite{arpl} & 43.38 ± 0.37 & 37.28 ± 1.21 & 40.33 ± 0.70 & 21.49 ± 2.03 & 35.68 ± 3.48 & 35.19 ± 1.79 & 37.21 ± 0.80 & 32.39 ± 0.74 \\ 
MOS~\cite{Huang:2021b} & 79.38 ± 5.06 & 78.05 ± 6.69 & 78.72 ± 5.86 & 65.95 ± 17.5 & 57.79 ± 5.79 & 76.78 ± 3.86 & 51.09 ± 1.33 & 62.90 ± 6.62 \\ 
LogitNorm~\cite{Wei:2022} & \textbf{34.37 ± 1.30} & \textbf{24.30 ± 0.54} & \textbf{29.34 ± 0.81} & \textbf{3.93 ± 1.99} & \textbf{8.33 ± 1.78} & {21.94 ± 0.85} & \textbf{21.04 ± 0.71} & \textbf{13.81 ± 0.20} \\ 
CIDER~\cite{cider} & {35.60 ± 0.78} & 28.61 ± 1.10 & 32.11 ± 0.94 & 24.76 ± 2.82 & \textbf{8.04 ± 0.43} & 25.05 ± 3.29 & \textbf{25.03 ± 1.36} & {20.72 ± 0.85} \\ 
% \hline
CRAFT~\cite{Karunanayake:2025b} & 36.61 ± 2.93 & \textbf{27.28 ± 0.09} & 31.94 ± 1.41 & 17.13 ± 0.99 & 14.58 ± 4.62 & \textbf{20.78 ± 0.04} & \textbf{25.12 ± 0.15} & \textbf{19.40 ± 0.88} \\ 
{\sc PSKD}~\cite{Yang:2025} & \textbf{35.34 ± 0.94} & 28.01 ± 0.73 & 31.67 ± 0.78 & \textbf{11.41 ± 0.50} & 17.60 ± 4.85 & 23.56 ± 1.48 & 29.34 ± 2.41 & 20.48 ± 1.30 \\
{\sc RankOOD}~\cite{Denipitiyage:2026} & {35.42 ± 0.65} & {28.01 ± 1.41} & {31.72 ± 0.67} & 17.18 ± 4.09 & 15.75 ± 3.20 & 22.66 ± 5.39 & 28.27 ± 0.83 & 20.96 ± 2.55 \\
IL PL-KD (Ours) & 
{\bf 34.17$\pm$0.74} & {\bf 26.02$\pm$0.24} & {\bf 30.10$\pm$0.49} & {\bf 11.52$\pm$0.26} & 14.70$\pm$2.01 & {\bf 21.68$\pm$0.76} & 28.49$\pm$1.37 & {\bf 19.10$\pm$0.96}\\
% 40.75 ± 0.01 & 32.19 ± 0.02 & 36.47 ± 0.01 & 19.26 ± 0.03 & 18.50 ± 0.02 & 32.60 ± 0.08 & 29.59 ± 0.03 & 24.99 ± 0.03\\
\hline
\multicolumn{9}{c}{Training methods with outliers} \\
\hline
OE~\cite{Hendrycks:2019} & 36.71 ± 2.06 & \textbf{2.97 ± 1.17} & \textbf{19.84 ± 0.95} & 24.67 ± 2.55 & \textbf{1.25 ± 0.36} & \textbf{12.07 ± 2.14} & \textbf{14.53 ± 2.80} & \textbf{13.13 ± 0.53} \\ 
MCD~\cite{mcd} & \textbf{34.36 ± 0.37} & \textbf{25.98 ± 0.44} & \textbf{30.17 ± 0.06} & 62.11 ± 11.8 & 19.43 ± 5.93 & 22.51 ± 5.16 & \textbf{24.10 ± 1.58} & 32.03 ± 4.21 \\ 
UDG~\cite{udg} & 40.75 ± 0.69 & 29.93 ± 1.27 & 35.34 ± 0.95 & 16.61 ± 5.14 & 17.39 ± 7.87 & \textbf{19.70 ± 1.89} & 27.70 ± 1.80 & \textbf{20.35 ± 2.41} \\ 
MixOE~\cite{Zhang:2023b} & 58.29 ± 8.25 & 44.62 ± 7.57 & 51.45 ± 7.78 & 38.28 ± 13.4 & 20.36 ± 3.99 & 33.19 ± 4.28 & 43.54 ± 4.95 & 33.84 ± 4.77 \\ 
\bottomrule
\end{tabular}
}
% \captionsetup{width=0.87\linewidth}

\end{table*}

\begin{table*}[h]
\centering
% \scriptsize
\caption{AUROC (\% $\uparrow$) of various methods for different OOD datasets when CIFAR-10 is ID. For each column, the top five methods are marked in \textbf{bold}.}
\label{tab:auroc_cifar10}
\renewcommand{\arraystretch}{1.09}
\resizebox{\textwidth}{!}{
\begin{tabular}{lccc|ccccc}
\toprule
Method & \multicolumn{3}{c|}{Near OOD} & \multicolumn{5}{c}{Far OOD} \\
 & CIFAR-100 & TIN & Average & MNIST & SVHN & Textures & Places365 & Average\\ 
 \hline
\multicolumn{9}{c}{Post-hoc inference methods} \\ 
\hline
OpenMax~\cite{openmaxbendale2016towards} & 86.91 ± 0.31 & 88.32 ± 0.28 & 87.62 ± 0.29 & 90.50 ± 0.44 & 89.77 ± 0.45 & 89.58 ± 0.60 & 88.63 ± 0.28 & 89.62 ± 0.19 \\
MSP~\cite{Hendrycks:2017} & 87.19 ± 0.33 & 88.87 ± 0.19 & 88.03 ± 0.25 & 92.63 ± 1.57 & 91.46 ± 0.40 & 89.89 ± 0.71 & 88.92 ± 0.47 & 90.73 ± 0.43 \\
TempScale~\cite{tempscale} & 87.17 ± 0.40 & 89.00 ± 0.23 & 88.09 ± 0.31 & 93.11 ± 1.77 & 91.66 ± 0.52 & 90.01 ± 0.74 & 89.11 ± 0.52 & 90.97 ± 0.52 \\
ODIN~\cite{hsu:2020} & 82.18 ± 1.87 & 83.55 ± 1.84 & 82.87 ± 1.85 & 95.24 ± 1.96 & 84.58 ± 0.77 & 86.94 ± 2.26 & 85.07 ± 1.24 & 87.96 ± 0.61 \\
MDS~\cite{Lee:2018b} & 83.59 ± 2.27 & 84.81 ± 2.53 & 84.20 ± 2.40 & 90.10 ± 2.41 & 91.18 ± 0.47 & 92.69 ± 1.06 & 84.90 ± 2.54 & 89.72 ± 1.36 \\
MDSEns~\cite{Lee:2018b} & 61.29 ± 0.23 & 59.57 ± 0.53 & 60.43 ± 0.26 & \textbf{99.17 ± 0.41} & 66.56 ± 0.58 & 77.40 ± 0.28 & 52.47 ± 0.15 & 73.90 ± 0.27 \\
RMDS~\cite{rmds} & 88.83 ± 0.35 & 90.76 ± 0.27 & 89.80 ± 0.28 & 93.22 ± 0.80 & 91.84 ± 0.26 & 92.23 ± 0.23 & 91.51 ± 0.11 & 92.20 ± 0.21 \\
Gram~\cite{gram} & 58.33 ± 4.49 & 58.98 ± 5.19 & 58.66 ± 4.83 & 72.64 ± 2.34 & 91.52 ± 4.45 & 62.34 ± 8.27 & 60.44 ± 3.41 & 71.73 ± 3.20 \\
EBO~\cite{Liu:2020b} & 86.36 ± 0.58 & 88.80 ± 0.36 & 87.58 ± 0.46 & 94.32 ± 2.53 & 91.79 ± 0.98 & 89.47 ± 0.70 & 89.25 ± 0.78 & 91.21 ± 0.92 \\
OpenGAN~\cite{opengan} & 52.81 ± 7.69 & 54.62 ± 7.68 & 53.71 ± 7.68 & 56.14 ± 24.1 & 52.81 ± 27.6 & 56.14 ± 18.3 & 53.34 ± 5.79 & 54.61 ± 15.5 \\
GradNorm~\cite{Huang:2021} & 54.43 ± 1.59 & 55.37 ± 0.41 & 54.90 ± 0.98 & 63.72 ± 7.37 & 53.91 ± 6.36 & 52.07 ± 4.09 & 60.50 ± 5.33 & 57.55 ± 3.22 \\
ReAct~\cite{Sun:2021} & 85.93 ± 0.83 & 88.29 ± 0.44 & 87.11 ± 0.61 & 92.81 ± 3.03 & 89.12 ± 3.19 & 89.38 ± 1.49 & 90.35 ± 0.78 & 90.42 ± 1.41 \\
MLS~\cite{Hendrycks:2022} & 86.31 ± 0.59 & 88.72 ± 0.36 & 87.52 ± 0.47 & 94.15 ± 2.48 & 91.69 ± 0.94 & 89.41 ± 0.71 & 89.14 ± 0.76 & 91.10 ± 0.89 \\
KLM~\cite{Hendrycks:2022} & 77.89 ± 0.75 & 80.49 ± 0.85 & 79.19 ± 0.80 & 85.00 ± 2.04 & 84.99 ± 1.18 & 82.35 ± 0.33 & 78.37 ± 0.33 & 82.68 ± 0.21 \\
VIM~\cite{wang2022vim} & 87.75 ± 0.28 & 89.62 ± 0.33 & 88.68 ± 0.28 & 94.76 ± 0.38 & 94.50 ± 0.48 & \textbf{95.15 ± 0.34} & 89.49 ± 0.39 & 93.48 ± 0.24 \\
KNN~\cite{Sun:2022} & {89.73 ± 0.14} & 91.56 ± 0.26 & 90.64 ± 0.20 & 94.26 ± 0.38 & 92.67 ± 0.30 & 93.16 ± 0.24 & {91.77 ± 0.23} & 92.96 ± 0.14 \\
DICE~\cite{dice} & 77.01 ± 0.88 & 79.67 ± 0.87 & 78.34 ± 0.79 & 90.37 ± 5.97 & 90.02 ± 1.77 & 81.86 ± 2.35 & 74.67 ± 4.98 & 84.23 ± 1.89 \\
RankFeat~\cite{rankfeat} & 77.98 ± 2.24 & 80.94 ± 2.80 & 79.46 ± 2.52 & 75.87 ± 5.22 & 68.15 ± 7.44 & 73.46 ± 6.49 & 85.99 ± 3.04 & 75.87 ± 5.06 \\
ASH~\cite{Djurisic:2023} & 74.11 ± 1.55 & 76.44 ± 0.61 & 75.27 ± 1.04 & 83.16 ± 4.66 & 73.46 ± 6.41 & 77.45 ± 2.39 & 79.89 ± 3.69 & 78.49 ± 2.58 \\
SHE~\cite{she} & 80.31 ± 0.69 & 82.76 ± 0.43 & 81.54 ± 0.51 & 90.43 ± 4.76 & 86.38 ± 1.32 & 81.57 ± 1.21 & 82.89 ± 1.22 & 85.32 ± 1.43 \\
GEN~\cite{XiXi:2023} & 87.21 ± 0.36 & 89.20 ± 0.25 & 88.20 ± 0.30 & 93.83 ± 2.14 & 91.97 ± 0.66 & 90.14 ± 0.76 & 89.46 ± 0.65 & 91.35 ± 0.69 \\
ExCeL~\cite{Karunanayake:2025} & 85.31 ± 0.26 & 88.48 ± 0.19 & 86.89 ± 0.23 & {95.87 ± 0.45} & 91.40 ± 1.43 & 89.66 ± 0.64 & 89.84 ± 0.41 & 91.69 ± 0.18 \\
\hline
\multicolumn{9}{c}{Training methods without outliers} \\
\hline
% \rowcolor{gray!30}

ConfBranch~\cite{DeVries:2018} & 88.91 ± 0.25 & 90.77 ± 0.25 & 89.84 ± 0.24 & 94.49 ± 0.77 & \textbf{95.42 ± 0.35} & 91.10 ± 0.41 & 90.39 ± 0.40 & 92.85 ± 0.29 \\
G-ODIN~\cite{hsu:2020} & 88.14 ± 0.60 & 90.09 ± 0.54 & 89.12 ± 0.57 & \textbf{98.95 ± 0.53} & \textbf{97.76 ± 0.14} & \textbf{95.02 ± 1.10} & 90.31 ± 0.65 & \textbf{95.51 ± 0.31} \\
CSI~\cite{Tack:2020} & 88.16 ± 0.16 & 90.87 ± 0.23 & 89.51 ± 0.19 & 92.55 ± 1.15 & 95.18 ± 0.45 & 90.71 ± 0.44 & 89.56 ± 0.51 & 92.00 ± 0.30 \\
ARPL~\cite{arpl} & 86.76 ± 0.16 & 88.12 ± 0.14 & 87.44 ± 0.15 & 92.62 ± 0.88 & 87.69 ± 0.97 & 88.57 ± 0.43 & 88.39 ± 0.16 & 89.31 ± 0.32 \\
MOS~\cite{Huang:2021b} & 70.57 ± 3.04 & 72.34 ± 3.16 & 71.45 ± 3.09 & 74.81 ± 10.1 & 73.66 ± 9.14 & 70.35 ± 3.11 & 86.81 ± 1.85 & 76.41 ± 5.93 \\
LogitNorm~\cite{Wei:2022} & \textbf{90.95 ± 0.22} & \textbf{93.70 ± 0.06} & \textbf{92.33 ± 0.08} & \textbf{99.14 ± 0.45} & \textbf{98.25 ± 0.41} & \textbf{94.77 ± 0.43} & \textbf{94.79 ± 0.16} & \textbf{96.74 ± 0.06} \\
CIDER~\cite{cider} & 89.47 ± 0.19 & {91.94 ± 0.19} & {90.71 ± 0.16} & 93.30 ± 1.08 & \textbf{98.06 ± 0.07} & 93.71 ± 0.39 & \textbf{93.77 ± 0.68} & \textbf{94.71 ± 0.36} \\
% \hline
CRAFT~\cite{Karunanayake:2025b} & \textbf{90.18 ± 0.14} & {92.04 ± 0.06} & \textbf{91.11 ± 0.04} & 94.59 ± 0.02 & 94.94 ± 1.10 & 93.46 ± 0.29 & \textbf{92.77 ± 0.10} & 93.94 ± 0.20 \\
{\sc PSKD}~\cite{Yang:2025} & \textbf{90.62 ± 0.16} & \textbf{92.80 ± 0.16} & \textbf{91.71 ± 0.16} & \textbf{97.20 ± 0.30} & 94.98 ± 1.57 & 93.38 ± 0.37 & \textbf{92.69 ± 0.40} & 94.56 ± 0.29 \\
{\sc RankOOD}~\cite{Denipitiyage:2026} & 89.11 ± 0.24 & 91.32 ± 0.58 & 90.21 ± 0.41 & 93.95 ± 2.23 & 94.70 ± 1.03 & 92.63 ± 1.53 & 91.51 ± 0.36 & 93.19 ± 0.84 \\
IL PL-KD (Ours) & 
{\bf 90.77$\pm$0.08} & {\bf 93.01$\pm$0.06} & {\bf 91.89$\pm$0.07} & {\bf 96.78$\pm$0.14} & 95.38$\pm$1.08 & 93.63$\pm$0.04 &{ 92.68$\pm$0.39} & \textbf{94.62$\pm$0.41} \\
% 89.52 ± 0.00 & 91.21 ± 0.00 & 90.37 ± 0.00 & 94.01 ± 0.01 & 94.00 ± 0.01 & 91.88 ± 0.01 & 91.74 ± 0.00 & 92.90 ± 0.01\\
\hline
\multicolumn{9}{c}{Training methods with outliers} \\
\hline
OE~\cite{Hendrycks:2019} & \textbf{90.54 ± 0.53} & \textbf{99.11 ± 0.34} & \textbf{94.82 ± 0.21} & 90.22 ± 1.31 & \textbf{99.60 ± 0.14} & \textbf{97.58 ± 0.27} & \textbf{96.58 ± 0.70} & \textbf{96.00 ± 0.13} \\
MCD~\cite{mcd} & {89.88 ± 0.07} & \textbf{92.18 ± 0.18} & {91.03 ± 0.12} & 84.22 ± 2.10 & 93.76 ± 2.30 & 93.35 ± 1.30 & {92.66 ± 0.36} & 91.00 ± 1.10 \\
UDG~\cite{udg} & 88.62 ± 0.32 & 91.20 ± 0.20 & 89.91 ± 0.25 & {95.81 ± 1.52} & 94.55 ± 2.27 & \textbf{93.92 ± 0.44} & 91.97 ± 0.41 & {94.06 ± 0.90} \\
MixOE~\cite{Zhang:2023b} & 87.47 ± 0.97 & 90.00 ± 0.73 & 88.73 ± 0.82 & 91.66 ± 2.21 & 93.82 ± 1.27 & 91.84 ± 0.51 & 90.38 ± 0.55 & 91.93 ± 0.69 \\
\bottomrule
\end{tabular}
}
% \captionsetup{width=0.87\linewidth}

\end{table*}

\subsection{Computational cost}
\label{app:gpu_time}
Table~\ref{tbl:gpu_time} reports per-epoch GPU training time in seconds. CIFAR-10/100 models are trained using 2x Nvidia RTX 4090 GPUs with a combined VRAM of 48GB for 100 epochs and Imagenet-200 models are trained using a single NVIDIA H200 GPU for 100 epochs. The proposed distillation variants are also computationally more efficient than RankOOD.

\begin{table}[b]
\centering
\footnotesize
\caption{Per-epoch GPU time (sec.) across different datasets.}

\label{tbl:gpu_time}
\renewcommand{\arraystretch}{0.9}
\setlength{\tabcolsep}{4pt}
\begin{tabular}{lcrc}
\toprule 
\textbf{Method} & \textbf{CIFAR-10} & \textbf{CIFAR-100} & \textbf{ImageNet-200}  \\
\midrule

CE     & 6.02 & 6.14 & 252.67 \\
RankOOD   & 13.80  & 14.34 & 1402.12 \\
\midrule
Self-KD	& 7.45	& 8.77	& 796.74\\
Class-level KD & 6.26 & 7.12 & 771.27\\
IL RankOOD & 18.8 & 14.64 & 849.18 \\
IL PL-KD & 8.59 & 10.48 & 716.42\\
\bottomrule
\vspace{-7mm}
\end{tabular}
\end{table}

\subsection{Detailed experimental setup}

We present details on relevant hyper-parameters for Instance-level PL-KD training in Tab~\ref{tbl:training_settings}.  For all image datasets (CIFAR-10, CIFAR-100, ImageNet-200), we use ResNet-18 model. For all training, we employ standard augmentations as per OpenOOD~\cite{Yang:2022}.

\begin{table}[t]
\centering
\caption{Summary of training settings on image datasets.}
\label{tbl:training_settings}
\begin{tabular}{lccc}
\toprule
\textbf{Hyperparameter} & \textbf{CIFAR-10} & \textbf{CIFAR-100} & \textbf{ImageNet-200} \\
\midrule
Batch size & 128 & 256 & 256 \\
Epochs & 100 & 100 & 100 \\
PL-KD loss weight & 0.5 & 0.5 & 0.5 \\
Image size & 32 & 32 & 224 \\
Teacher temperature & 1.0 & 5.0 & 0.5 \\
Student temperature & 9.0 & 10.0 & 0.3 \\
Peak learning rate & 0.1 & 0.1 & 0.1 \\
Momentum & 0.9 & 0.9 & 0.9 \\
Learning rate decay factor & $5 \cdot 10^{-4}$ & $5 \cdot 10^{-4}$ & $5 \cdot 10^{-4}$ \\
% Teacher temperature & 1.0 & 5.0 & 0.5 \\

\bottomrule
\end{tabular}
\end{table}

\subsection{Hyper-parameter selection}
\begin{figure*}
    \centering
    \includegraphics[width=0.98\textwidth]{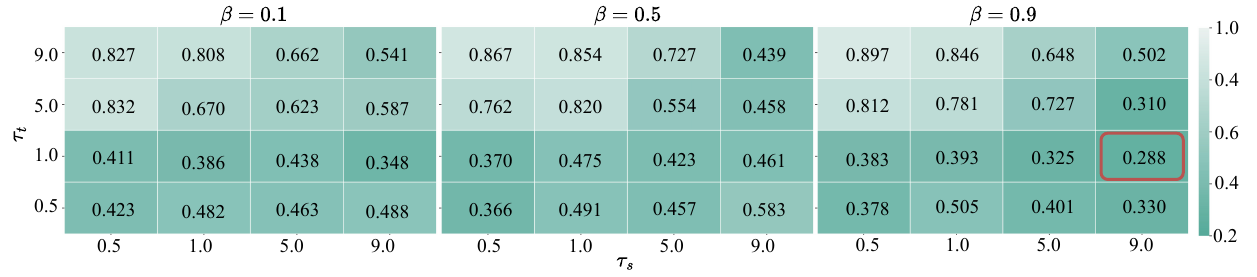}

    \caption{Effect of the teacher temperature \(\tau_t\), student temperature \(\tau_s\), and loss sensitivity \(\beta\) on the FPR95 (\(\downarrow\)) for OOD detection. The model is trained on CIFAR-10 as the in-distribution (ID) dataset and evaluated on the OOD validation set. Each heatmap reports the FPR95 for different combinations of teacher and student temperatures at a fixed \(\beta\). The best configuration is achieved with \(\beta=0.9\), \(\tau_t=1.0\), and \(\tau_s=9.0\), yielding an FPR95 of 0.288.}
     \label{fig:hp_cifar10}
    \vspace{-\baselineskip}
\end{figure*}

\begin{figure*}
    \centering
    \includegraphics[width=0.98\textwidth]{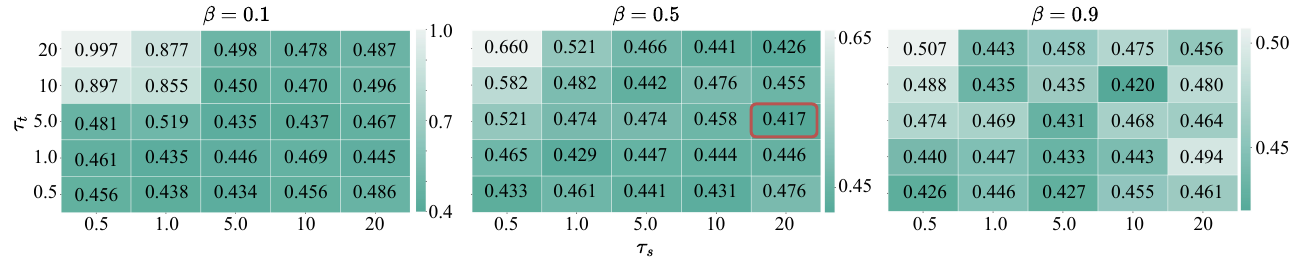}

    \caption{Effect of the teacher temperature \(\tau_t\), student temperature \(\tau_s\), and loss sensitivity \(\beta\) on the FPR95 (\(\downarrow\)) for OOD detection. The model is trained on CIFAR-100 as the in-distribution (ID) dataset and evaluated on the OOD validation set. Each heatmap reports the FPR95 for different combinations of teacher and student temperatures at a fixed \(\beta\). The best configuration is achieved with \(\beta=0.5\), \(\tau_t=5.0\), and \(\tau_s=20.0\).}
     \label{fig:hp_cifar100}
    \vspace{-\baselineskip}
\end{figure*}

\begin{figure*}
    \centering
    \includegraphics[width=0.98\textwidth]{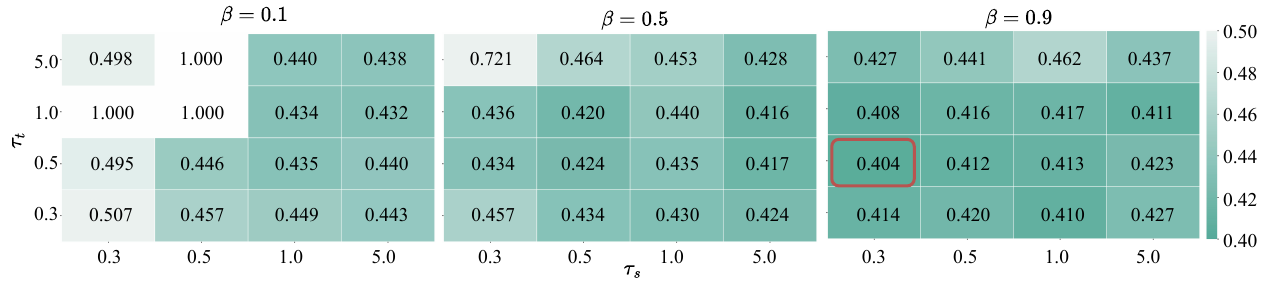}

    \caption{Effect of the teacher temperature \(\tau_t\), student temperature \(\tau_s\), and loss sensitivity \(\beta\) on the FPR95 (\(\downarrow\)) for OOD detection. The model is trained on ImageNet-200 as the in-distribution (ID) dataset and evaluated on the OOD validation set. Each heatmap reports the FPR95 for different combinations of teacher and student temperatures at a fixed \(\beta\). The best configuration is achieved with \(\beta=0.9\), \(\tau_t=0.5\), and \(\tau_s=0.3\).}
     \label{fig:hp_imagenet}
    \vspace{-\baselineskip}
\end{figure*}

We perform hyperparameter optimization and model selection based on the validation FPR@95 and classification accuracy, using the validation splits defined by the OpenOOD benchmark for three datasets: CIFAR-10, CIFAR-100, and ImageNet-200. Specifically, $\beta$ denotes the cross-entropy loss weight, while $\tau_t$ and $\tau_s$ represent the temperature parameters of the teacher and student models, respectively. Figures~\ref{fig:hp_cifar10},\ref {fig:hp_cifar100} and \ref {fig:hp_imagenet} present heatmaps of the FPR@95 scores obtained across the evaluated hyperparameter configurations.
\label{appendix:hyp_pram_selec}

